\pdfoutput=1
\documentclass{article}

\PassOptionsToPackage{numbers, compress}{natbib}

\usepackage[preprint]{neurips_2021}

\usepackage[utf8]{inputenc} 
\usepackage[T1]{fontenc}    
\usepackage{xcolor}         
\usepackage{url}            
\usepackage{booktabs}       
\usepackage{amsfonts}       
\usepackage{nicefrac}       
\usepackage{microtype}      

\usepackage{mathtools} 
\usepackage{lipsum}
\usepackage{graphicx}
\usepackage[]{cleveref}
\usepackage{subcaption}
\usepackage{tikz}

\usepackage{setspace}
\usepackage{pdflscape}
\usepackage{wrapfig}

\usepackage{multirow}
\usepackage{makecell}
\usepackage{paralist}
\usepackage{amssymb}
\usepackage{amsmath}
\usepackage{amsthm}
\DeclareMathOperator*{\argmin}{arg\,min}
\DeclareMathOperator*{\argmax}{arg\,max}
\renewcommand{\vec}[1]{{#1}}
\usepackage{xcolor}

\usepackage{bbm}

\usepackage{fancyhdr}
\usepackage{color}
\usepackage{algorithm}
\usepackage{algorithmic}
\usepackage{array}
\newcolumntype{H}{>{\setbox0=\hbox\bgroup}c<{\egroup}@{}}

\usepackage{pifont}
\newcommand{\cmark}{\ding{51}}%
\newcommand{\xmark}{\ding{55}}%

\usepackage{array}
\newcolumntype{?}{!{\vrule width 1pt}}

\usepackage{setspace}
\usepackage{pdflscape}
\usepackage{titlesec}
\usepackage{amsthm}
\usepackage{thmtools, thm-restate}
\usepackage{enumitem}

\usepackage{sidecap}

\titlespacing*{\section}{0pt}{0.2ex plus .05ex minus .05ex}{0.1ex plus .05ex minus .05ex}
\titlespacing*{\subsection}{0pt}{0.1ex plus .05ex minus .05ex}{0.05ex plus .05ex minus .05ex}
\setlist[itemize]{leftmargin=*,itemsep=0em}
\setlist[enumerate]{leftmargin=*,itemsep=0em}

\crefname{figure}{Fig.}{Figs.}
\crefname{definition}{Defn.}{Defns.}
\crefname{corollary}{Corollary}{Corollaries}
\crefname{proposition}{Prop.}{Props.}
\crefname{theorem}{Thm.}{Thms.}
\crefname{remark}{Remark}{Remarks}
\crefname{principle}{Principle}{Principles}
\crefname{lemma}{Lemma}{Lemmas}
\crefname{claim}{Claim}{Claims}
\crefname{table}{Tab.}{Tabs.}
\crefname{section}{\S}{\S\S}
\crefname{subsection}{\S}{\S\S}
\crefname{subsubsection}{\S}{\S\S}

\title{DIVINE: Diverse Influential Training Points for Data Visualization and Model Refinement}

\author{
Umang Bhatt\\
University of Cambridge\\
\texttt{usb20@cam.ac.uk}
\And
Isabel Chien\\
University of Cambridge\\
\texttt{ic390@cam.ac.uk}
\And
Muhammad Bilal Zafar\\
Amazon AWS AI\\
\texttt{zafamuh@amazon.com}
\And
Adrian Weller\\
University of Cambridge \& The Alan Turing Institute\\
\texttt{aw665@cam.ac.uk}
}

\begin{document}
\maketitle

\begin{abstract}
As the complexity of machine learning (ML) models increases, resulting in a lack of prediction explainability, several methods have been developed to explain a model's behavior in terms of the training data points that most influence the model. However, these methods tend to mark outliers as highly influential points, limiting the insights that practitioners can draw from points that are not representative of the training data. In this work, we take a step towards finding influential training points that also represent the training data well. We first review methods for assigning importance scores to training points. Given importance scores, we propose a method to select a set of DIVerse INfluEntial (DIVINE) training points as a useful explanation of model behavior. As practitioners might not only be interested in finding data points influential with respect to model accuracy, but also with respect to other important metrics, we show how to evaluate training data points on the basis of group fairness. Our method can identify unfairness-inducing training points, which can be removed to improve fairness outcomes. Our quantitative experiments and user studies show that visualizing DIVINE points helps practitioners understand and explain model behavior better than earlier approaches. 
\end{abstract}

\section{Introduction}
\label{intro}
Training point importance is a useful form of explainability for practitioners when reasoning about a machine learning (ML) model's behavior~\citep{jeyakumar2020can}. 
This form of explanation identifies which training points are most important to a ML model 
~\citep{ghorbani2019data,koh2017understanding,yeh2018representer}. 
To compute training point importance, popular methods include calculating Data Shapley values~\citep{ghorbani2019data,kwon2020efficient} or using 
influence functions 
to estimate the impact to the model of dropping one or more points from training data~\citep{koh2017understanding,koh2019accuracy}.
However, the top-$m$ most important points returned by these methods are often redundant, in the sense that several may be very similar, limiting the extent of explanation provided~\citep{barshan2020relatif,bhatt2020explainable}. 
To address this shortcoming, we devise an approach for selecting a set of DIVINE (DIVerse INfluEntial) training points. 

\begin{figure}[t]
\centering
    \begin{subfigure}[b]{0.3\linewidth}
    \centering
    \includegraphics[width=0.95\linewidth]{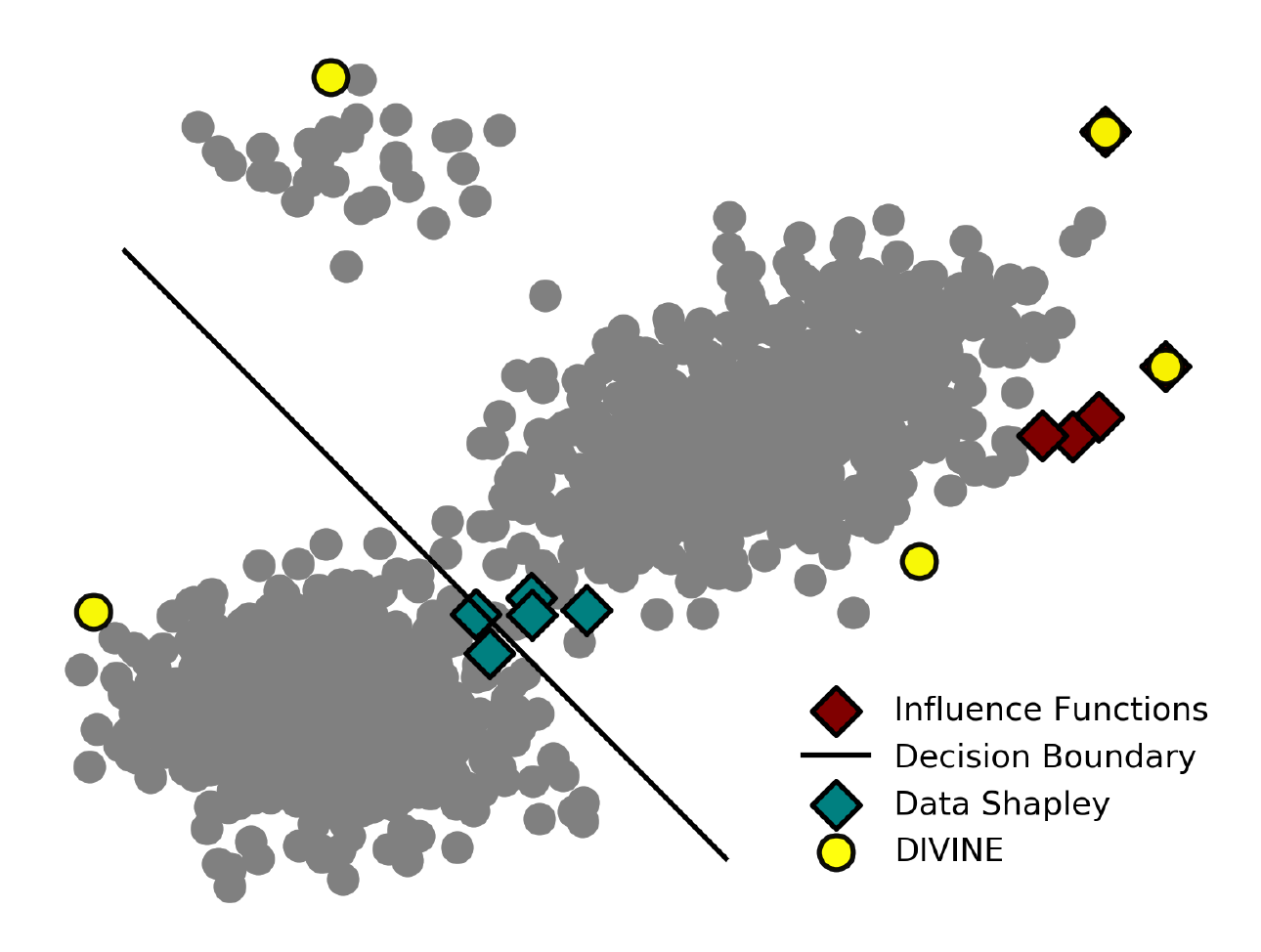}
            \caption{Synthetic Data}
            \label{fig:exisiting}
    \end{subfigure}
    \begin{subfigure}[b]{0.3\linewidth}
    \centering
    \includegraphics[width=0.95\linewidth]{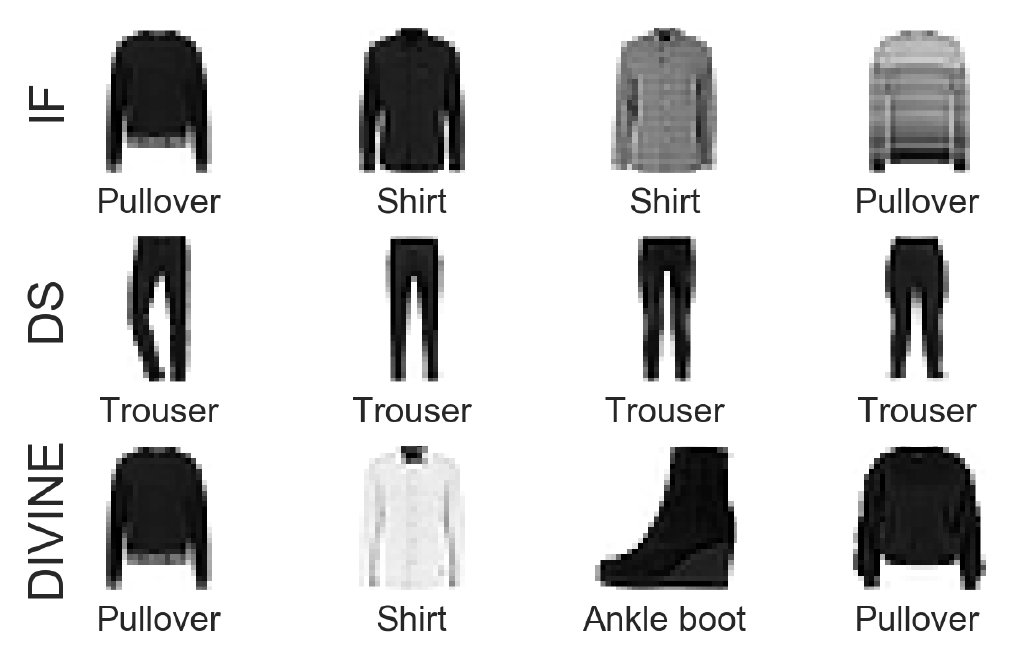}
            \caption{Global: FashionMNIST}
            \label{fig:global_fmnist}
    \end{subfigure}
    \begin{subfigure}[b]{0.3\linewidth}
            \centering
          \includegraphics[width=0.99\linewidth]{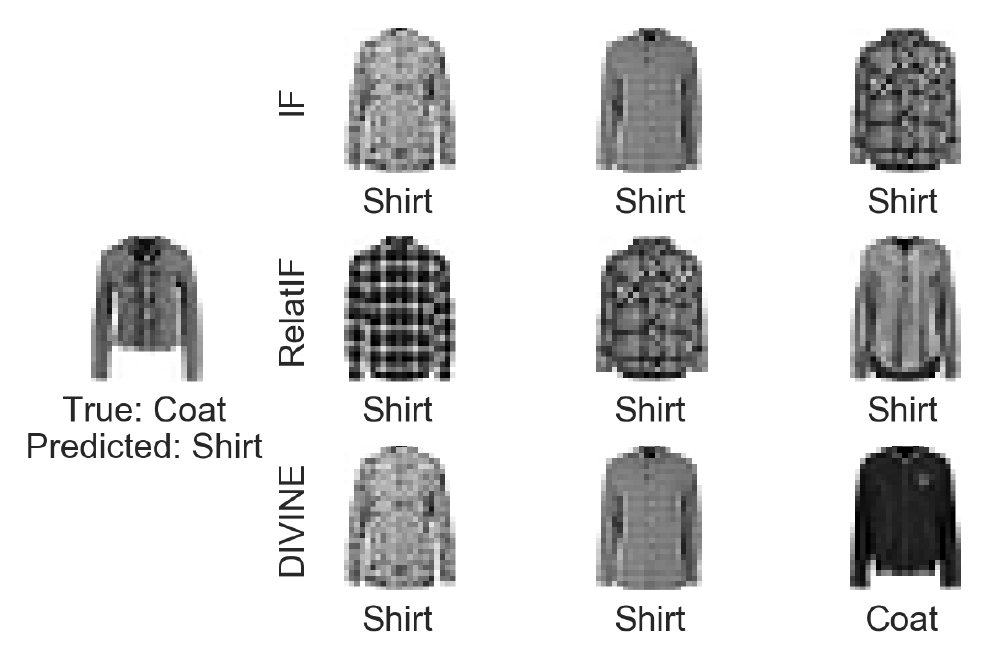}
            \caption{Local: FashionMNIST}
            \label{fig:local_fmnist}
    \end{subfigure}
    \caption{\small In~\ref{fig:exisiting}, we show that our method selects DIVINE points (yellow circles) that are spread across the feature space. This contrasts IF (red diamonds) and Data Shapley (blue diamonds), which select points located in one region. Note the overlap between IF and DIVINE points in the top right. In~\ref{fig:global_fmnist} and~\ref{fig:local_fmnist}, we show that DIVINE points (third row) are more diverse than ones selected by IF (first row) or other methods (second row). DIVINE is calculated by trading off IF and $\mathcal{R}_{SR}$. The predicted label is listed under each point.}\label{fig:one}
    \vspace{-0.5cm}
\end{figure}

Figure~\ref{fig:exisiting} shows that the top-$5$ influential points with respect to the approximate leave-one-out estimate of~\citet{koh2017understanding} and with respect to Data Shapley~\citep{ghorbani2019data} are all located in a small vicinity (red and blue diamonds respectively). 
Due to this lack of diversity, practitioners may miss key insights from underrepresented data points.
Some regions, such as the cluster of points in the top left corner, are ignored by both methods.
In Figure~\ref{fig:exisiting}, our DIVINE points, denoted by yellow circles, not only lie in regions of high influence but also across the feature space. 
Our method provides the flexibility, under a common assumption, to operate on top of training point importance scores from a wide range of methods, including Data Shapley (DS)~\citep{ghorbani2019data} and influence functions (IF)~\citep{koh2017understanding}.
Beyond the synthetic setting of Figure~\ref{fig:exisiting}, consider a misclassified test point, as in Figure~\ref{fig:local_fmnist}. The influential training points according to two competitive methods (Influence~\citep{koh2017understanding} and RelatIF~\citep{barshan2020relatif}) are very similar to the test point. The resulting explanation contains redundant information. A explanation containing diverse points is more useful: notice the coat that appears in the DIVINE points but not in the others. 
Our user studies (Section~\ref{user_study}) show that this additional diversity allows practitioners to be more accurate in simulating model behavior and enhances trustworthiness in the model. 

Moreover, existing methods for training point importance focus on the impact of a data point on loss or accuracy, but practitioners may also want to value data points with respect to other important metrics, such as fairness. 
To bridge this gap, we develop an efficient method for computing importance scores with respect to group fairness metrics~\citep{hardt2016equality}. Practitioners can use these scores to visualize data points that harm model fairness. 
We describe how to refine a model and improve fairness outcomes by removing \textit{unfairness-inducing points}, while minimizing impact on other metrics.

Our main contributions are:
\begin{enumerate}
    \item We devise a method for finding a diverse set of training points that are influential to a model.
    Our top $m$ DIVINE points, when trading off influence with diversity objectives, can provide a more comprehensive overview of model behavior (Section~\ref{submod}). 
    \item We discuss how to value a training point's influence on group fairness metrics (Section~\ref{fair}). 
    \item Experiments on synthetic and real-world datasets show that DIVINE can help explore diverse, influential regions of the feature space and can also help improve fairness in the model outcomes by identifying most unfairness inducing points (Section~\ref{experiments}).
    \item Extensive user studies show that DIVINE leads to enhanced user trust and better task simulatability as compared to existing approaches (Section~\ref{user_study}).
\end{enumerate}

\section{Background: Assigning Training Point Importance}
\label{valuation}
We start by reviewing earlier methods for obtaining training point importance scores.
Consider a ML model parameterized by $\theta \in \Theta$. Given training data $\mathcal{D} = \{(x_i,y_i)\}_{i=1}^{n}$  and a loss function $l(x,y;\theta)$, weighted empirical risk minimization estimates $\hat \theta = \argmin_{\theta \in \Theta}\sum_{i=1}^{n} w_i l(x_i,y_i; \theta)$, 
where $w_i$ is the weight given to training point $i$.
Usually, each training point has equal weight, e.g., $w_i = \frac{1}{n}$.
The leave-one-out (LOO) model obtained as a result of dropping the $i$-th training point (i.e., $w_i = 0$) has LOO parameters denoted by $\hat \theta^{\text{LOO}}_i$. Dropping a set of training points $\mathcal{U}$ can be done by setting all respective $w_i$'s in the set to zero. The resulting model is denoted by  $\hat \theta^{\text{LOO}}_\mathcal{U}$.
Within the LOO framework, the importance of the $i$-th training point can be written as $I_i^{\text{LOO}} = f(\hat \theta^{\text{LOO}}_i) - f(\hat \theta)$
where 
$f: \Theta \rightarrow \mathbb{R}$ measures a quantity of interest  (e.g., loss). We call $f$ the evaluation function. Instead of assigning importance by a difference between the value of $f$ with LOO and original parameters, we can also take an absolute difference, squared difference, sigmoid, etc. depending on the application.
In this paper, we focus consideration on cases where $f$ is the loss $f_{\text{loss}}(\theta) = \sum_{i=1}^{n} l(x_i,y_i; \theta)$ on $n$ data points, or is a group fairness metric, like equal accuracy (Section~\ref{fair}).%

Unless otherwise specified, we assume that $f(\theta)$is a non-negative scalar and that lower $f(\theta)$ is desirable. Let $\theta_{\text{new}}$ denote the new parameters (e.g., $\theta_{\text{new}} = \hat \theta^{\text{LOO}}_i$) and let $\theta_{\text{old}}$ denote the old ERM parameters (e.g., $\theta_{\text{old}} = \hat \theta$).
Thus, a positive $I_i =  f(\theta_{\text{new}}) - f(\theta_{\text{old}})$  implies that including the $i$-th point is \textit{helpful} for lowering $f$ when learning $\theta$: upon removing the $i$-th point, the value of $f$ at the new parameters increased, which is undesirable. A negative $I_i$ implies that including the $i$-th point is \textit{harmful} for lowering $f$. A large absolute magnitude of $I_i$ implies that a point $i$ is influential. 

\subsection{Influence Functions}
Retraining for different weight configurations can be computationally expensive. \citet{koh2017understanding} develop algorithms to approximate the effect of removing a training point on the loss at a test point by re-weighting its contribution. Suppose we modify the weight of $x_i$ from $w_i = \frac{1}{n}$ to $w_i = \frac{1}{n} + \epsilon_i$. Let $\hat \theta_{\epsilon_i}$ be the parameters obtained upon re-weighting. If we let $\epsilon_i = -\frac{1}{n}$, this amounts to dropping $x_i$ from the training data. Influence functions (IF) from robust statistics can be used to approximate $\hat \theta_{\epsilon_i}$~\citep{influence,hampel1974influence}. Assuming the loss $l$ is twice differentiable and convex in $\theta$, we can linearly approximate the parameters upon dropping $x_i$ as $\hat \theta^{\text{IF}}_i \approx \hat \theta -  H_{\hat \theta}^{-1} \nabla_{\theta} \; l\left(x_i,y_i; \hat \theta \right)\epsilon_i$,
where $H_{\theta} = \frac{1}{n}\sum_{i=1}^{n} \nabla_{\theta}^2 l(x_i,y_i; \theta)$ is the Hessian of the loss $l$. 
For details, see \citep{koh2017understanding}. We let $I_i^{\text{IF}} = f(\hat \theta^{\text{IF}}_i) - f(\hat \theta)$ be the importance score of the $i$-th training point according to IF. 
Per \citep{koh2019accuracy}, we can estimate the influence of dropping the $i$-th training point on any evaluation function $f$ as:
\begin{equation}
    I_i^{\text{IF}} \approx I_i := \nabla_{\theta} \; f(\hat \theta)^\intercal H_{\hat \theta}^{-1} \nabla_{\theta} \; l(x_i,y_i; \hat \theta) .
    \label{generic}
\end{equation}
When $f$ is \textit{loss}, \citet{koh2019accuracy} note that influence is additive, which means that importance scores are additive. This implies that the importance of training points in $\mathcal{U}$ is given by: $I_\mathcal{U}^{\text{IF}} = \sum_{i \in \mathcal{U}} I_i^{\text{IF}}$.

\subsection{Data Shapley}
Instead of computing parameters to obtain an importance score with respect to an evaluation function $f$, techniques like Data Shapley (DS) aim to  directly compute importance scores~\citep{ghorbani2019data}.
Shapley values are a game-theoretic way to attribute value to \textit{players} in game. \citet{ghorbani2019data} apply Shapley values to training point importance. 
They propose to compute the importance of $x_i$ as $ I_i^{\text{DS}} = C^{-1}\sum_{\mathcal{S} \subseteq \mathcal{D}\setminus \{i\}} f(\hat \theta^{\text{LOO}}_\mathcal{U}) - f(\hat \theta^{\text{LOO}}_{\mathcal{U} \cup \{i\}})$,
where $C = \binom{n-1}{|\mathcal{S}|}$, $\mathcal{S}$ is a subset of the training data and $\mathcal{U} = \mathcal{D} \setminus \mathcal{S}$.  
Most works regarding DS take $f$ to be loss, accuracy, or AUC.
We can efficiently approximate DS using Monte Carlo Sampling~\citep{ghorbani2019data,kwon2020efficient}.

\subsection{Other Methods}
\label{cfp}
\citet{khanna2019interpreting} use Fisher kernels 
to select influential training points efficiently. Their method recovers IF if $l(x,y;\theta)$ is negative log-likelihood.
Specific to neural networks, \citet{yeh2018representer} decompose the (pre-activation) prediction for a test point into a linear combination of activations for training points, using a modified representer theorem.
The resulting training point weights correspond to a point's importance with respect to $f_\text{loss}$.
\citet{bhatt2020counterfactual} assign training point importance based on how much loss is incurred when finding a set of alternative parameters $\theta^\prime$ that classify a specific point $x$ differently from $\hat \theta$ but that perform similarly on the training data. Their setup is a special case of IF~\citep{koh2017understanding}.
The methods discussed thus far are model-specific and depend on a specified $f$.
Another line of work has searched for prototypes, which are representative points that summarize a dataset independent of model parameters~\citep{bien2011prototype,gurumoorthy2019efficient}. 
\citet{kim2016MMD} use maximum mean discrepancy (MMD) to find prototypes but do not assign importance scores to the selected points.
A large MMD implies that the samples are likely from different distributions~\citep{gretton2012kernel}. Since prototypes are model-agnostic, we omit them from our evaluation, as we are interested in finding diverse training points important to a model.

\section{Selecting Diverse Samples}
\label{submod}
As shown in Figure~\ref{fig:exisiting}, the top-$m$ influential points based on importance scores can result in a set of points that are similar to each other. We desire $m$ points that are simultaneously influential (high importance) and diverse across the feature space to serve as an explanation of model behavior. 
To achieve these desiderata, we propose the following objective:
\begin{align}
    \max_{\mathcal{S} \in \mathcal{D},|\mathcal{S}| = m} \mathcal{I}(\mathcal{S}) + \gamma \mathcal{R}(\mathcal{S}) ,
    \label{opt}
\end{align}
where $\mathcal{S}$ is a subset of $m$ important points from the dataset $\mathcal{D}$, $\mathcal{I}(\mathcal{S})$ is a normalized function ($\mathcal{I}(\emptyset) = 0$) that captures the importance of the points in $\mathcal{S}$, $\mathcal{R}(\mathcal{S})$ is a  function that captures the diversity of the points in $\mathcal{S}$, and $\gamma$ controls the trade-off between the two terms.
Solving the optimization problem in Equation \ref{opt} yields a set $\mathcal{S}$ of $m$ DIVerse and INfluEntial points which we call DIVINE points.
Setting $\gamma=0$ recovers the traditional setup of selecting $m$ points with the highest importance.
Our setup is reminiscent of combining loss functions (e.g., one to penalize training error and one to regularize for sparsity, smoothness, etc.): we effectively regularize for diversity in the $m$ influential points we select. Our formulation in Equation~\ref{opt} is similar to that of \citet{lin-bilmes-2011-class}, who select relevant yet diverse sentences to summarize a document, and that of \citet{prasad2014submodular}, who scale diverse set selection to exponentially large datasets.

 
To this end, we take $\mathcal{I}(\mathcal{S}) = \sum_{i \in \mathcal{S}} I_i$ to be the sum of the importance scores of points in $\mathcal{S}$. We propose three submodular $\mathcal{R}(\mathcal{S})$: 
\begin{inparaenum}[A.]
    \item Sum-Redundancy: $\mathcal{R}_{\text{SR}}(\mathcal{S}) = \kappa - \sum_{u,v \in \mathcal{S}} \phi(u,v)$~\citep{libbrecht2018choosing};
    \item Facility-Location: $\mathcal{R}_{\text{FL}}(\mathcal{S}) = \sum_{u \in \mathcal{D}} \max_{v \in \mathcal{S}} \phi(u,v)$~\citep{krause2014submodular}; and
    \item MMD: $\mathcal{R}_\text{MMD}(\mathcal{S}) = c_1\sum_{u \in \mathcal{D}, v \in \mathcal{S}} \phi\left(u, v\right) - c_2\sum_{u, v \in \mathcal{S}} \phi\left(u, v\right)$~\citep{kim2016MMD},
\end{inparaenum}
where $\phi(u,v)$ is the similarity between two points, $\kappa = \sum_{u,v \in \mathcal{D}} \phi(u,v)$, $c_1 = \frac{2}{n|\mathcal{S}|}$, and $c_2 = \frac{1}{|\mathcal{S}|^{2}}$. We let $\phi$ be the radial basis function kernel. 

\begin{wrapfigure}{R}{0.5\textwidth}
\vspace{-1cm}
  \begin{minipage}{0.5\textwidth}
    \centering
\begin{algorithm}[H]
   \caption{Greedy DIVINE selection}
   \label{alg:example}
\begin{algorithmic}[1]
   \STATE {\bfseries Input:} Dataset $\mathcal{D}$, Trade-off parameter $\gamma$, number of  diverse influential points $m$
   \FORALL{$x_i \in \mathcal{D}$}
   \STATE $I_i \gets$ influence($x_i$)
   \ENDFOR
   \STATE $\mathcal{S} \gets \emptyset$; $I_\mathcal{S} \gets 0$
   \WHILE{$|\mathcal{S}| < m$}
   \STATE \small{$\mathcal{S} \gets \mathcal{S} \cup \argmax_{x_i \in \mathcal{D} \setminus \mathcal{S}}   \left[ I_\mathcal{S} + I_i + \gamma \mathcal{R}(\mathcal{S})\right]$}
   \STATE $I_\mathcal{S} \gets \sum_{ i \in \mathcal{S}} I_i$
   \ENDWHILE
   \STATE {\bfseries Output:} Set of $m$ DIVINE points $\mathcal{S}$
\end{algorithmic}
\end{algorithm}
  \end{minipage}
 \vspace{-0.1cm}
\end{wrapfigure}

While $\mathcal{R}_{\text{SR}}$ encourages us to find $m$ influential points that are diverse from each other, both $\mathcal{R}_{\text{MMD}}$ and $\mathcal{R}_{\text{FL}}$ encourage our influential points to be representative of the training data. $\mathcal{R}_{\text{SR}}$ is known as penalty-based diversity and penalizes similarity between points in $\mathcal{S}$~\citep{lin-bilmes-2011-class,tschiatschek2014learning}.
$\mathcal{R}_{\text{FL}}(\mathcal{S})$ maximizes the average similarity between a training point and its most similar point in $\mathcal{S}$. 
$\mathcal{R}_\text{MMD}$ ensures the $m$ selected influential points are similar to the training data while being different from each other.
We use $\mathcal{R}_{\text{SR}}$ in the main text, but practitioners can select $\mathcal{R}$ that is appropriate for their use case. 

\subsection{Optimization Procedure}

It is well-known that a modular function $\mathcal{I}(\mathcal{S})$ plus a submodular function $\mathcal{R}(\mathcal{S})$ is submodular, rendering the overall objective submodular~\citep{bach2011learning}. As such, we take a greedy approach to performing the optimization in Equation~\ref{opt}, as outlined in Algorithm~\ref{alg:example}. Greedy selection returns a set that typically performs very well, and has a guarantee of at worst  $(1 - \frac{1}{e})$ 
performance of the optimal set $\mathcal{S}^*$~\citep{nemhauser1978analysis}.
We can also take a stochastic greedy approach  per~\citet{mirzasoleiman2015lazier}. Instead of using the entire dataset to find the element with maximum marginal gain, we would randomly sample $s$ points at each iteration, calculate the marginal gain for each of the $s$ points, and add the one with the highest gain to $\mathcal{S}$ until $|\mathcal{S}| = m$. In practice, stochastic greedy is preferred on large datasets, where the computational cost of full greedy alone can be high.
Moreover, some may find our additivity assumption, which lets $\mathcal{I}$ be modular, too restrictive. However, note that, by construction, DS satisfies linearity, which implies modularity. For IF, we find in Appendix~\ref{app:setup} that modularity holds for various $f$ as long as $m$ is not too large. 
Furthermore, instead of calculating $I_i$ for the entire dataset and then performing greedy selection, we could select the first point greedily, recalculate $I_i$ for the remaining points, greedily select the next point, and repeat until we have $m$ points. Appendix~\ref{add_experiments} shows that this works similarly to our approach. Future work can develop other selection methods.


\subsection{Local Explanations for Individual Points}
In addition to obtaining a global diverse set of influential training points that explains a model's behavior, our framework is amenable to obtaining local explanations: a diverse set of points that explains a model's prediction for a specific point, $x_i$. To accomplish this, we can let $f$ be the loss at $x_i$: $f_i(\theta) = l\left(x_i,y_i; \theta \right)$. For IF, the impact of the $j$-th training point on $x_i$ can be approximated as: $I_i = \nabla_{\theta} \; l\left(x_i,y_i; \hat \theta \right) H_{\hat \theta}^{-1} \nabla_{\theta} \; l\left(x_j,y_j; \hat \theta \right)$~\citep{koh2017understanding}. 
\citet{barshan2020relatif} notice that the top-$m$ influential points selected by Equation~\ref{generic} tend to be outliers, and add locality constraints to this objective. However, they solve a slightly different problem to us: their method, RelatIF, is concerned with data points being atypical, whereas DIVINE focuses on providing a diverse explanation by ensuring that the same region does not get marked as influential repeatedly.
Our method would select a diverse set of outliers (if those are indeed influential) whereas the constraints of \citet{barshan2020relatif} would not permit it. We compare our local DIVINE explanations to RelatIF in Section~\ref{examples}.


\section{Fairness Valuation}
\label{fair}
Existing approaches to valuing data points usually take $f$ to be a model's training loss~\citep{barshan2020relatif,koh2017understanding}, accuracy, or AUC~\citep{ghorbani2019data}.
We propose that $f$ can also be any group fairness criteria. 
This allows us to value training data based on their helpfulness or harmfulness for achieving various notions of fairness.\footnote{This problem can be seen as the inverse of fair data augmentation or fair active learning~\citep{anahideh2020fair}, approaches which identify additional data to collect.} In the feature importance literature, \citet{datta2016algorithmic} use quantitative input influence to evaluate the effect of removing a feature on a model's fairness. \citet{lundberg2020explaining} use Shapley values to decompose demographic parity in terms of feature contributions.
Our work can be seen as an influential point analog. While they extend feature importance to identify features that contribute to unfairness, we extend training point importance to identify which points contribute to unfairness.

Let our model have parameters $\theta$ and a binary predicted outcome $\hat y \in \{-1,1\}$ for some input $x$. Let $y$ be the actual outcome for $x$. Let $A = \{a,b\}$ be a binary sensitive attribute that is contained explicitly in or encoded implicitly in $x$. When we refer to subgroups, we mean partitions of $\mathcal{D}$ based on $A$. Let the training points in Group 1 be given by $\mathcal{D}_{a} = \mathcal{D}_{-a} \cup \mathcal{D}_{+a}$, where $\mathcal{D}_{+a} = \{x \in \mathcal{D} | A = a, y = +1 \}$ are positives in Group 1, and the training points in Group 2 be given by $\mathcal{D}_{b} = \mathcal{D}_{-b} \cup \mathcal{D}_{+b}$. We can further partition each set by predicted outcome, e.g. $\mathcal{D}_{+b} = \mathcal{D}_{+b+} \cup \mathcal{D}_{+b-}$ where $\mathcal{D}_{+b-} = \{x \in \mathcal{D} | A = b, y = +1, \hat y = -1  \}$ captures false negatives in $\mathcal{D}_{+b}$. 
In this work, we define unfairness as the difference in accuracy between subgroups: this is sometimes referred to as (un)equal accuracy.\footnote{Extension to other group fairness notions is straightforward.}
\citet{berk2018fairness} says $\theta$ is fair (with respect to equal accuracy) if the following is close to $0$: 
\begin{equation}
\small
f_{\text{unf}}(\theta) = \sum_{j \in \{-1,1\}} |P(\hat y = j |A = a, y = j) - P(\hat y = j |A = b, y = j)| ,
\label{eq:ea}
\end{equation}
where $P(\hat y = +1 |A = a, y = +1)$ is the true positive rate for Group 1 under $\theta$. We take the sum of the absolute difference in true positive rates between subgroups and the absolute difference in true negative rates between subgroups as a measure of unfairness, per Equation~\ref{eq:ea}; the larger its magnitude, the more unfair. Note $f_{\text{unf}}(\theta) \in [0,2]$.
Practitioners can calculate $f_{\text{unf}}$ on training, validation, or test data.
They may leverage importance scores with respect to $f_{\text{unf}}$ to identify points hurting their model's fairness. We refer to the points harmful to $f_{\text{unf}}$ as \textit{unfairness-inducing points}. 
By removing such points from their datasets, practitioners can potentially improve model fairness and accuracy.

\begin{wrapfigure}{R}{0.4\textwidth}
\centering
\vspace{-0.6cm}
\includegraphics[width=0.92\linewidth]{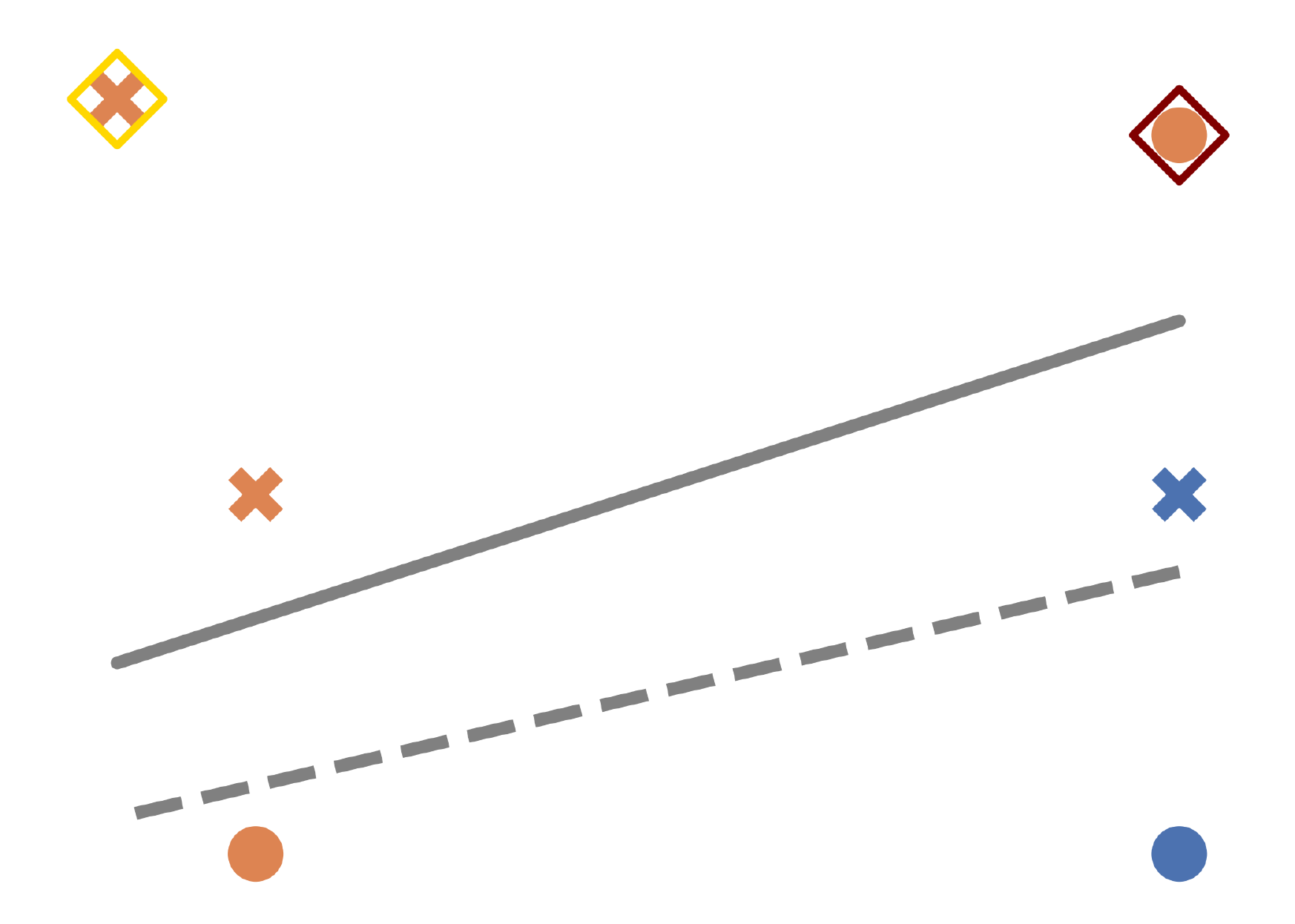}
            \caption{\small Toy Example}
            \label{fig:sanity}
\vspace{-0.3cm}
\end{wrapfigure}

To check if our method can detect erroneous, unfairness-inducing points, we construct a toy example. In Figure~\ref{fig:sanity}, a set of four $2$-dimensional data points are drawn on the four corners of a square. The top two points have label $-1$ ($\times$), and the bottom two points have label $1$ ($\bullet$). We assign sensitive attributes by saying the left side is from Group 1 (orange) and the right side is from Group 2 (blue). A fifth point is added farther away in the top left (orange $\times$). A logistic regression model is fit to these five points (dashed line), obtaining perfect accuracy ($100\%$) and ideal unfairness of $0$. We then inject a poisonous point into the dataset (top right orange $\bullet$). Note that, with respect to the original model, this point is incorrectly classified and has an inconsistent sensitive attribute. A logistic regression model is fit to all six points (solid line). This poisoned model gets $\frac{5}{6}$ points correct, but has an unfairness of $1$. We find importance scores for all six points with respect $f_\text{loss}$ and $f_\text{unf}$. The most influential point with respect to $f_\text{loss}$ is the correctly classified outlier (yellow diamond); however, the most influential point with respect to $f_\text{unf}$ is the poisonous point (red diamond).
This demonstrates that $f_\text{unf}$ can detect unfairness-inducing points and does not simply find outliers.

\section{Experiments}
\label{experiments}
We evaluate our approach on multiple datasets and identify DIVINE points with respect to multiple $f$ measures. We first visualize the set of DIVINE points found by running Algorithm~\ref{alg:example}. We compare importance scores found with respect to $f_\text{loss}$ and $f_\text{unf}$. We then learn fairer models by removing low-value points, those with high $f_\text{unf}$. We qualitatively review the unfairness-inducing points.

To validate our method, we run experiments on the following datasets: synthetic data, LSAT~\citep{kusner2017counterfactual}, Bank Marketing~\citep{Dua:2019}, COMPAS~\citep{compas}, Adult~\citep{Dua:2019}, a two-class variant of MNIST~\citep{lecun1998mnist}, and FashionMNIST~\citep{xiao2017fashion}.\footnote{Code for our experiments can be found at 
\url{https://github.com/umangsbhatt/divine-release}.
}  
We primarily consider logistic loss, $l(x, y; \theta) = \log(1 + \exp(y\;\theta^\intercal x))$, such that the logistic likelihood is given by $p(y|x) = \sigma(y\;\theta^\intercal x)$ where $\sigma(a) = \frac{1}{1 + \exp(a)}$. For image datasets, we train multilayer perceptons and convolutional neural networks, both with cross-entropy loss.

For synthetic experiments, we follow 
\citet{zafar2017fairness} generating a synthetic dataset. First, we generate $2,000$ binary labels uniformly at random. We then assign a feature-vector to each label by sampling from two Gaussian distributions: $p(x|y=1) = \mathcal{N}(x; \mu=[3; 3], \Sigma=[2, 1; 1, 2])$ and 
$p(x|y=-1) = \mathcal{N}(x; \mu=[-2; -2], \Sigma=[1, 0; 0, 1])$. We then sample $50$ points with label $y=1$ from $p(x|y=1) = \mathcal{N}(x; \mu=[-2; 8], \Sigma=[1, 0; 0, 1])$. We draw a sensitive attribute for all samples from a Bernoulli distribution: $p(A = a) = \frac{p(x^\prime | y = 1)}{p(x^\prime | y = 1) + p(x^\prime | y = -1)}$, where $x^\prime$ is a rotated version of $x$.

\begin{figure*}[t]
\centering
    \begin{subfigure}[b]{0.245\linewidth}        
            \includegraphics[width=\textwidth]{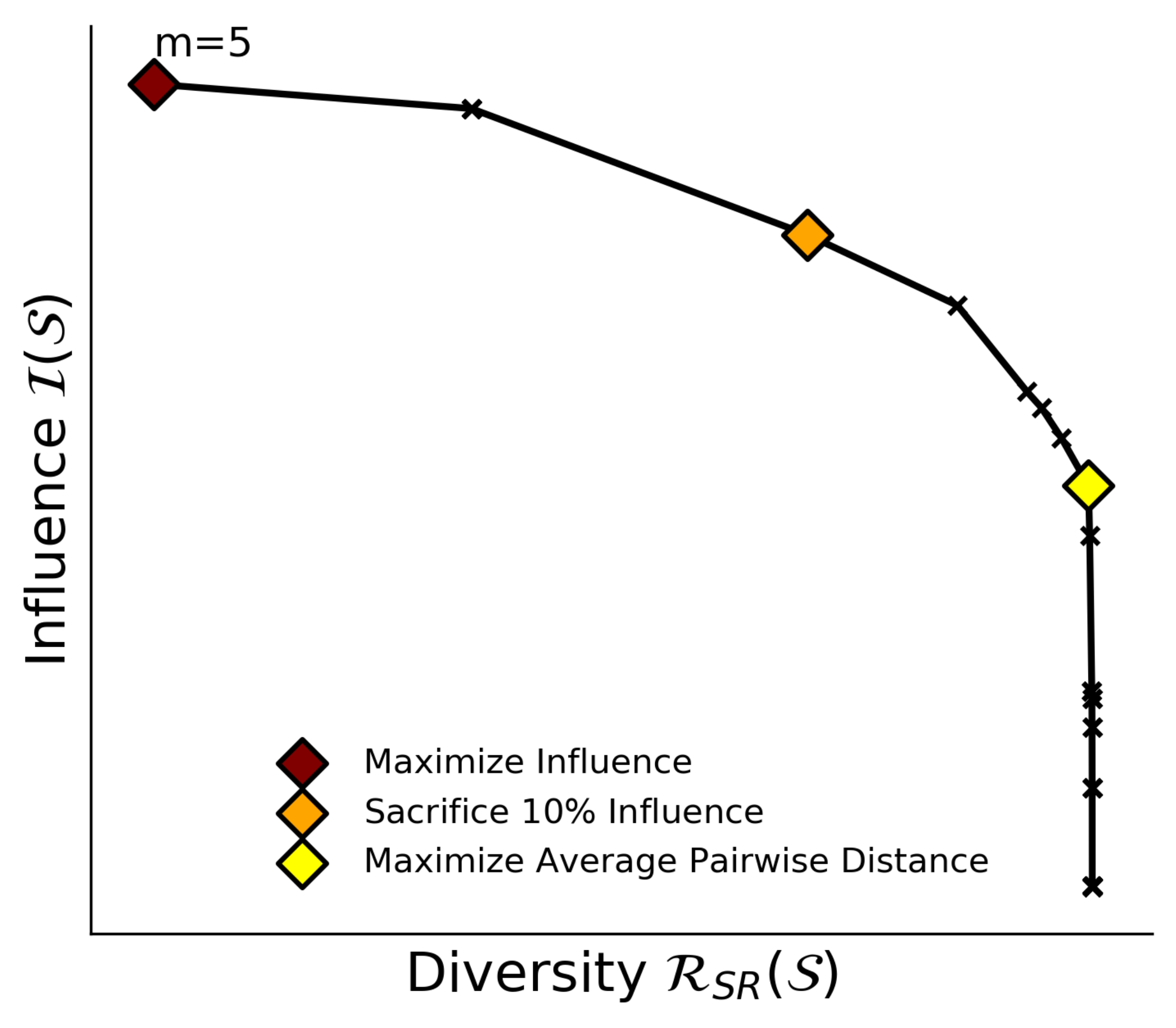}
            \caption{$\mathcal{I}$-$\mathcal{R}$ Tradeoff}
            \label{trade}
    \end{subfigure}
    \begin{subfigure}[b]{0.245\linewidth}
        \includegraphics[width=\textwidth]{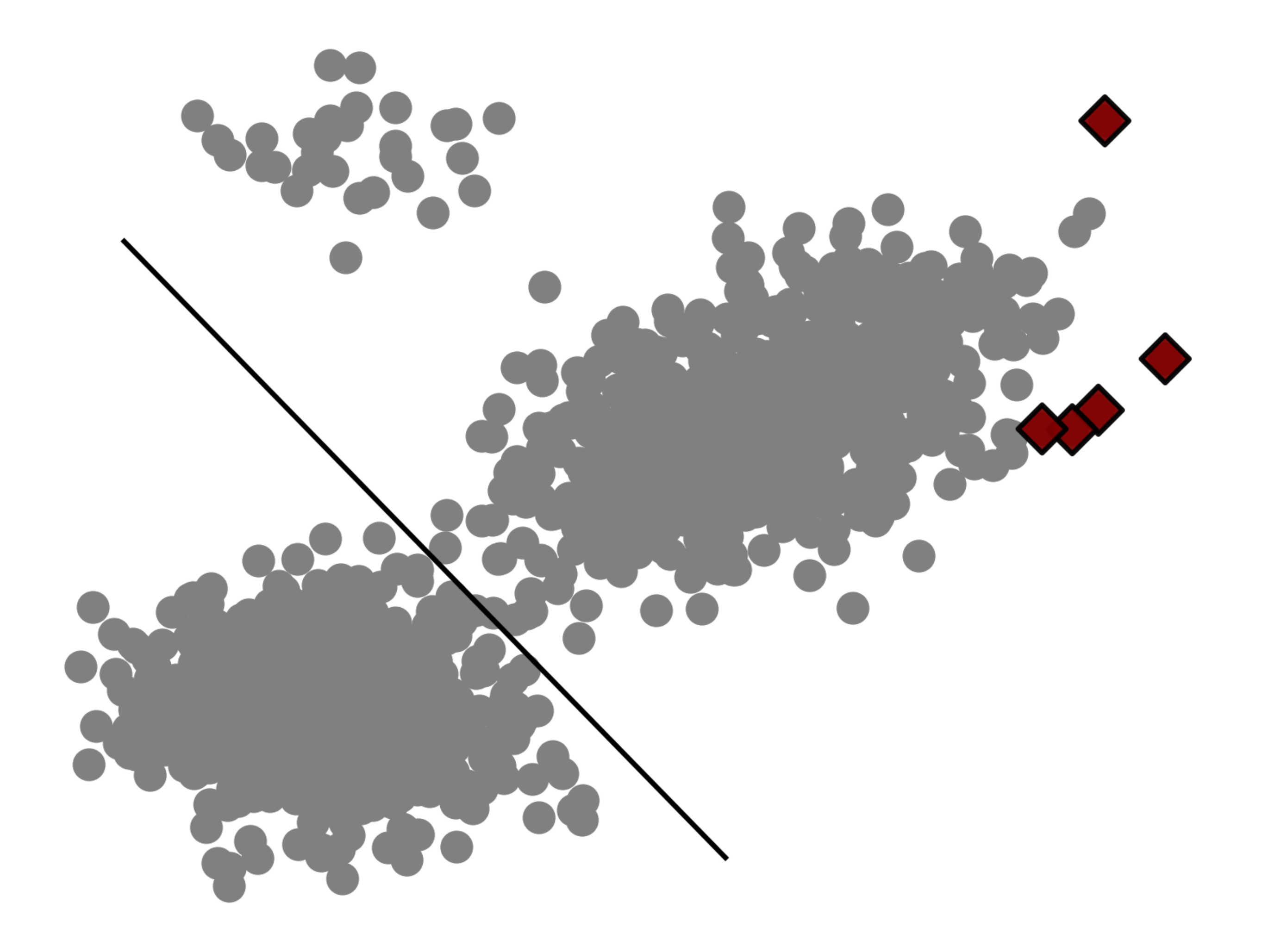}
        \caption{DIVINE $\gamma = 0$}
        \label{if_d}
    \end{subfigure}
    \begin{subfigure}[b]{0.245\linewidth}    
        \includegraphics[width=\textwidth]{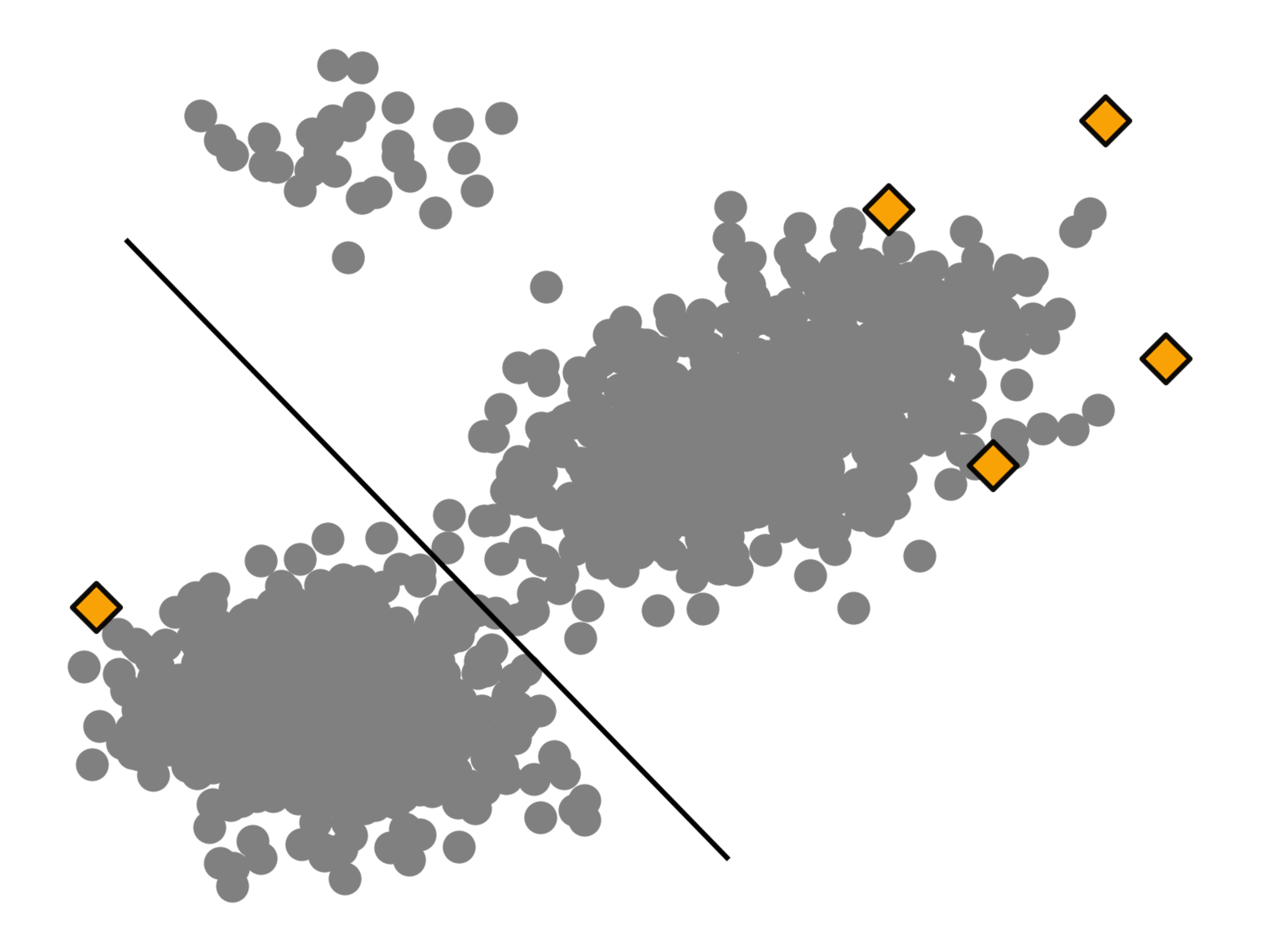}
        \caption{DIVINE $\gamma = 161$}
        \label{twenty}
    \end{subfigure}
    \begin{subfigure}[b]{0.245\linewidth}
        \includegraphics[width=\textwidth]{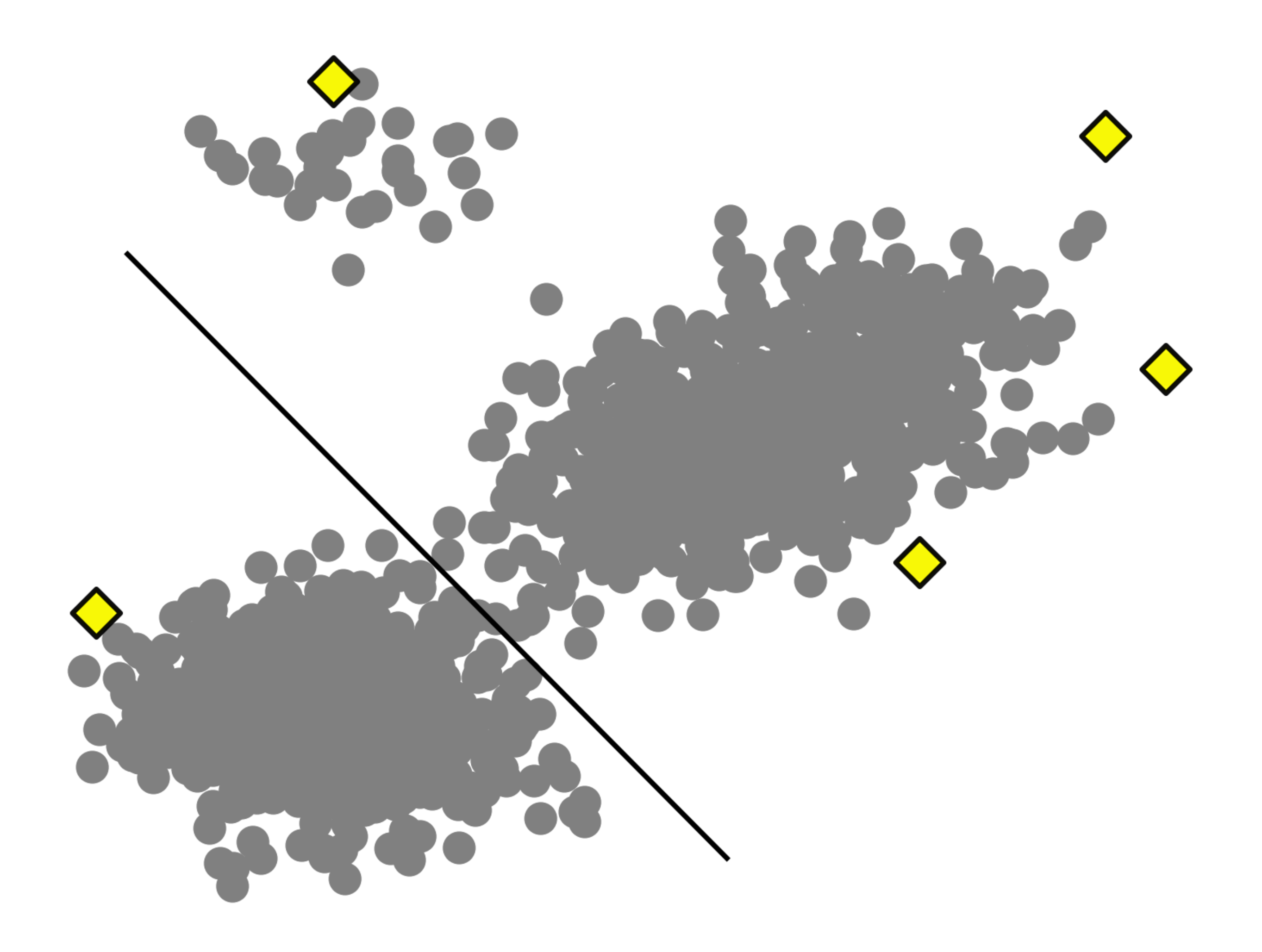}
            \caption{DIVINE $\gamma = 434$}
            \label{apd}
    \end{subfigure}%
    \caption{\small We characterize the trade-off between influence and diversity in~\ref{trade} by varying $\gamma$. We also visualize the top-5 DIVINE points for select values of $\gamma$. The red diamond in~\ref{trade} is $\gamma = 0$, which recovers the top points from IF alone plotted in~\ref{if_d}. The orange diamond in~\ref{trade} is for the $\gamma$ we find such that our DIVINE points have $10\%$ less influence than IF points; these are visualized in~\ref{twenty}. The yellow diamond in~\ref{trade} is the $\gamma$ we find such that we maximize the average pairwise distance between our DIVINE points, seen in~\ref{apd}.  
     }\label{fig:all_subplots}
    \vspace{-0.6cm}
\end{figure*}

\subsection{Selecting Diverse Influential Points}
\label{div_inf}
\begin{wrapfigure}{R}{0.34\textwidth}
\centering
\vspace{-0.3cm}
\includegraphics[width=\linewidth]{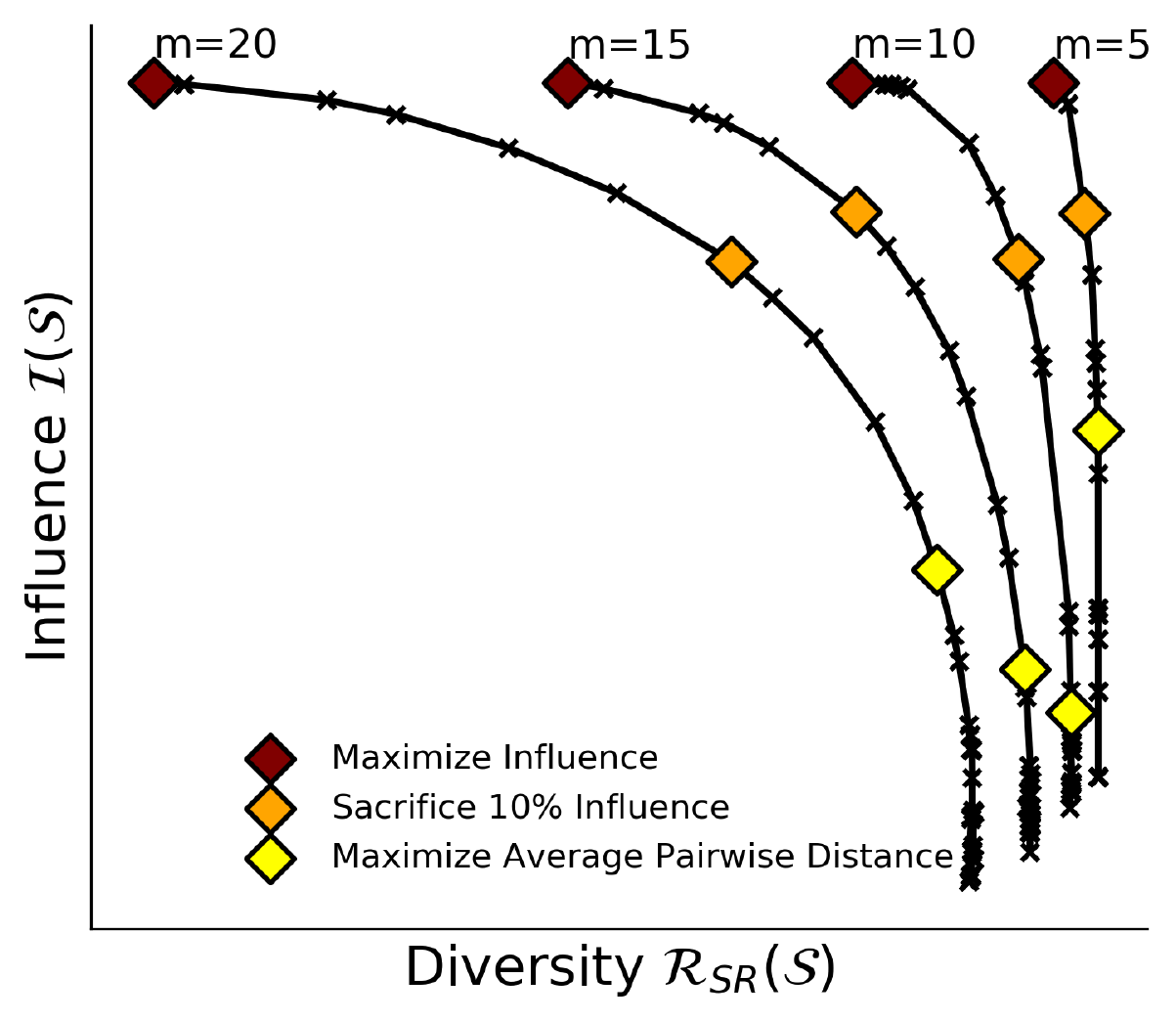}
        \caption{\small Tradeoff for various $m$}
        \label{fig:tradem}
        \vspace{-0.3cm}
\end{wrapfigure}
First, we validate our approach to greedily select DIVINE points as a global explanation for a logistic regression model trained on our synthetic data. In Figure~\ref{fig:all_subplots}, we show how DIVINE values data points using $\mathcal{R}_\text{SR}$ and IF with $f_{\text{loss}}$ on the training data.
In Figure~\ref{trade}, we characterize the trade-off between influence and diversity. We obtain the black line by varying $\gamma$. We normalize influence such that we consider how much less influence DIVINE points contain than the top IF points on the y-axis.
The red diamond represents $\gamma = 0$, which maximizes influence, i.e., top IF points. We suggest two ways to select $\gamma$. One option is to specify a specific amount of influence to sacrifice. In Figure~\ref{twenty}, we find $\gamma$ by specifying that we want our top-$5$ DIVINE points to have $10\%$ less influence than the top-$5$ IF points; we indicate the corresponding point in orange on the trade-off curve in Figure~\ref{trade}. Another option is to find the $\gamma$ maximizes the average pairwise distance between points in $\mathcal{S}$: $\sum_{u,v \in \mathcal{S}} d(u,v)$, which is depicted by the yellow diamonds in Figure~\ref{apd}.
In both selection mechanisms, we select $\gamma$ by running a log sweep over $\gamma \in [1e{-4}, 1e5]$.
Our mechanisms for selecting $\gamma$ ensure our set of points have high diversity at the expense of little influence.
In Figure~\ref{fig:tradem}, we show how our trade-off curves vary as we add more DIVINE points ($m = \{5,10,15,20\}$). Each trade-off curve has the same shape as Figure~\ref{trade}, but due to scaling might appear linear when compared to curves for larger $m$; for example, the rightmost curve in Figure~\ref{fig:tradem} is the same as the curve shown in Figure~\ref{trade}.
As we increase $m$, the diversity ($\mathcal{R}_\text{SR}$) of the IF points (red) decreases, implying redundancy in the selected points. This confirms the findings of~\citet{barshan2020relatif}. Practitioners can select any place along the black curve to identify positions that trade-off influence and diversity. Our suggested $\gamma$ selection strategies are shown as yellow and orange diamonds. 
In Appendix~\ref{add_experiments}, we show similar curves to Figure~\ref{fig:tradem} when valuing points with other methods like DS, when using other diversity measures like $\mathcal{R}_\text{FL}$ and $\mathcal{R}_\text{MMD}$, and when using other datasets.
\textit{We find that when optimizing Equation~\ref{opt} and varying $\gamma$, we maintain high influence while achieving the desired diversity}.

\subsection{Modes within Unfairness Inducing Points}
\label{learning}
\begin{figure}[tb]
  \begin{minipage}{\textwidth}
  \begin{minipage}[b]{0.4\textwidth}
    \centering
    \resizebox{\linewidth}{!}{%
    \begin{tabular}{l|l|l|l?l|l|l|l}
\toprule
\multicolumn{4}{c?}{$\mathcal{I}(\mathcal{S}) + \gamma\mathcal{R}_\text{SR}(\mathcal{S})$} & \multicolumn{4}{c}{$\mathcal{I}(\mathcal{S})$} \\ \toprule
LSAT    & GPA  & $A$  & FYA   & LSAT    & GPA  & $A$ & FYA   \\ \toprule
    14    &     2.9  & F  &    Pass    &   20      &   2.8  & M   &   Pass     \\ \hline
     25   &    3.6  &  M   &    Pass    &    25     &   3.6   & M   &     Pass   \\ \hline
     20   &     2.8 & M    &    Pass    &    20     &   3.2   & M   &    Pass   \\ \hline
     41   &      3.8 & F   & Fail    &    14     &    2.9   & F  &    Pass    \\ \hline
     33   &      4.0 & F   & Fail    &    21     &    3.1   & M  &    Pass    \\ \hline
     45   &      3.9 & F   & Fail    &    23     &    2.8   & M  &    Pass    \\ \hline
     29   &      3.7 & F   & Fail    &    22     &    2.9   & M  &    Pass    \\ \hline
     37   &      3.8 & F   & Fail    &    26     &    3.7   & M  &    Pass    \\ \bottomrule
\end{tabular}
}
      \captionof{table}{\small LSAT Unfairness Inducing Points}
      \label{example_removal}
    \end{minipage}
      \hfill
    \begin{minipage}[b]{0.24\textwidth}
    \centering
            \includegraphics[width=\linewidth]{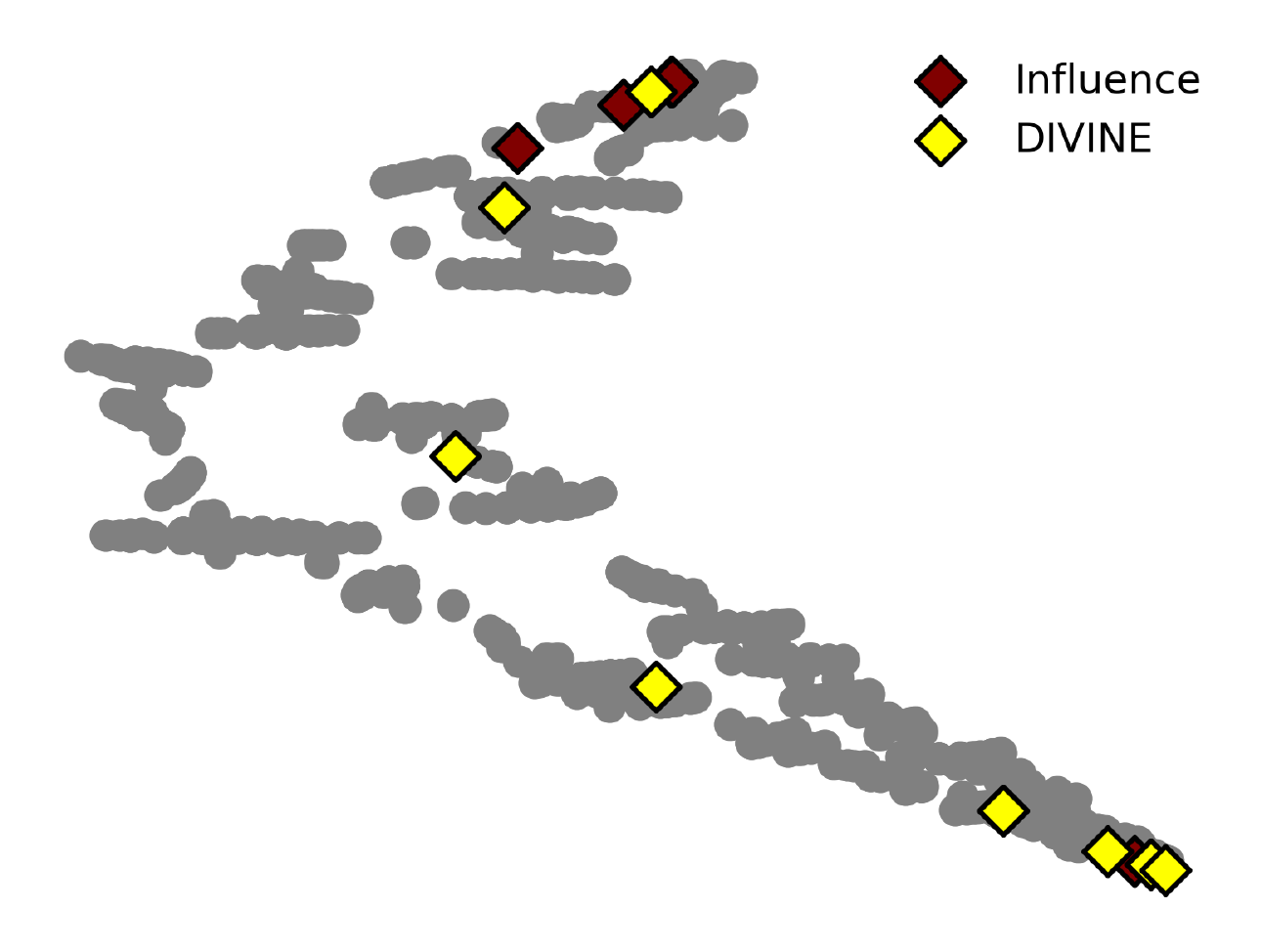}
    \captionof{figure}{\small TSNE}
    \label{fig:tsne}
  \end{minipage}
  \hfill
    \begin{minipage}[b]{0.34\textwidth}
    \centering
            \includegraphics[width=0.85\linewidth]{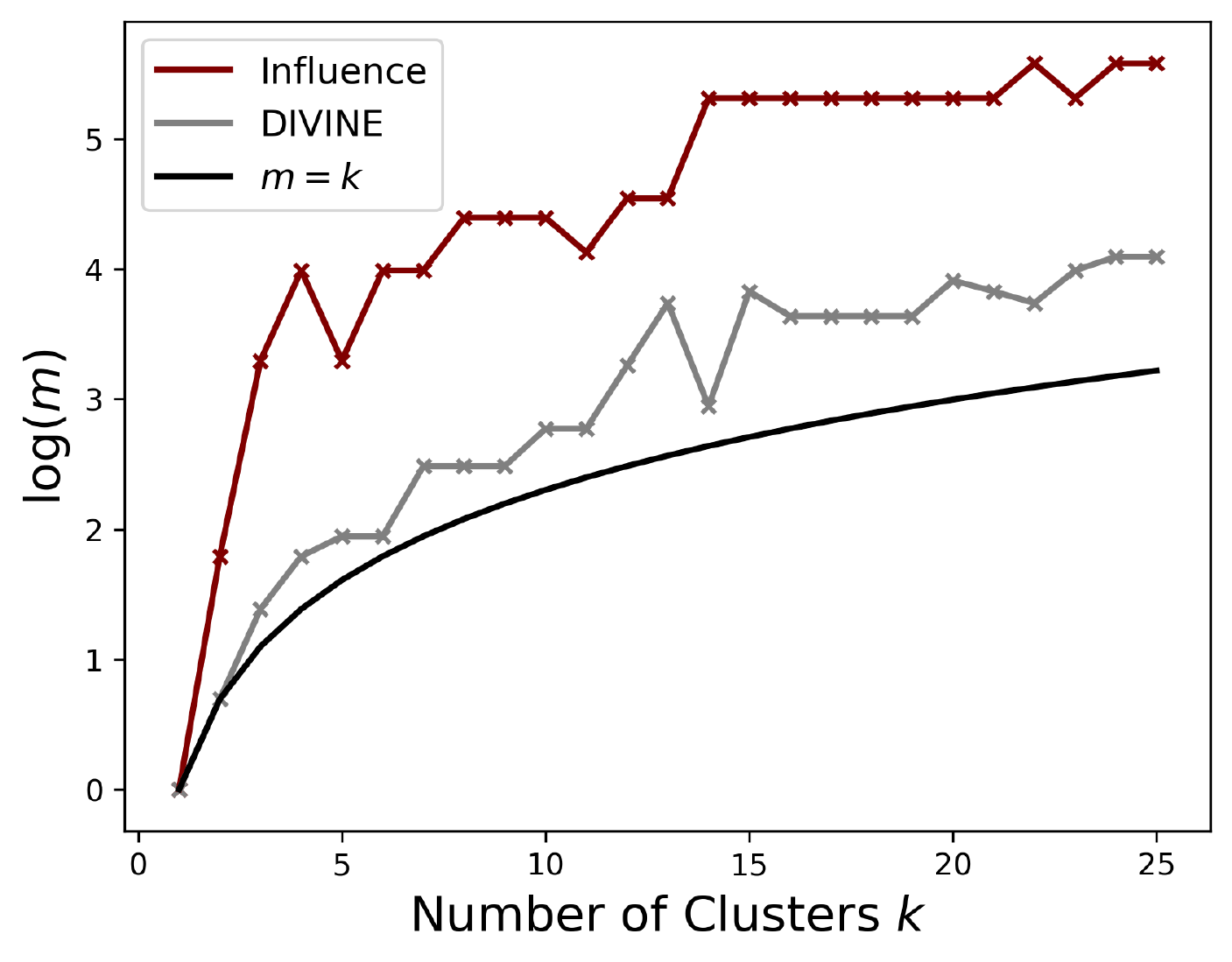}
    \captionof{figure}{\small LSAT Clustering}
    \label{fig:clu}
  \end{minipage}
  \end{minipage}
  \vspace{-0.5cm}
 \end{figure}

Studying the diverse modes of data that contribute to canonical model behavior, herein with respect to unfairness, can help practitioners analyze their models.
In Table~\ref{example_removal} for LSAT, we qualitatively compare the top-$8$ diverse unfairness-inducing points found by Equation~\ref{opt} (left) and the top-$8$ most influential points (right).  
We use IF with respect to $f_\text{unf}$ for $\mathcal{I}$ and $\mathcal{R}_\text{SR}$.
$A$ is the sensitive attribute: male (M) or female (F). FYA is first-year average, which is binarized to pass/fail. The maximum possible LSAT score is 48, and maximum GPA is 4.0. Notice the lack of diversity in the points on the right: most points are males with poor LSAT test scores and low GPA grades yet pass their course. 
DIVINE points include not only points with poor LSAT scores and low GPAs that pass but also points with high LSAT test scores and high GPAs (which are mostly female) yet fail. 
By trading off influence and diversity (left), we identify an unfairness-inducing ``mode'' of the dataset---female participants with high LSAT scores and GPAs, but fail---which is not identified by influence alone (right). We visualize this diversity in low dimensions via TSNE~\citep{van2014accelerating} in Figure~\ref{fig:tsne}: notice how all the IF points are clustered. This ability to detect modes missed by the top IF points highlights the utility of DIVINE. With DIVINE, we find multiple modes that lead to unfairness in our model.
Quantitatively, we show that DIVINE does a better job of covering clusters of data in input space.
In Figure~\ref{fig:clu}, we cluster the full dataset using KMeans into $k$ clusters and then find the $m$ such that one point from each cluster is in the top-$m$ points of IF and DIVINE. The black line is a lower-bound, $m=k$. DIVINE points requires a smaller $m$ than IF to represent all $k$ clusters of the data in the top points. The redundancy of the top IF points make it difficult to get a holistic picture of model behavior, as top IF points lie in clusters where we already have important points identified.
\textit{Unlike IF points, DIVINE points allow us to identify various modes of data that contribute to a model's unfairness. }

\subsection{Removing Unfairness Inducing Points}
\begin{figure*}[tb]
\centering
\includegraphics[width=0.8\linewidth]{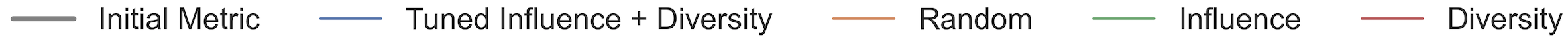}
    \begin{subfigure}[b]{0.24\linewidth}            
            \includegraphics[width=\textwidth]{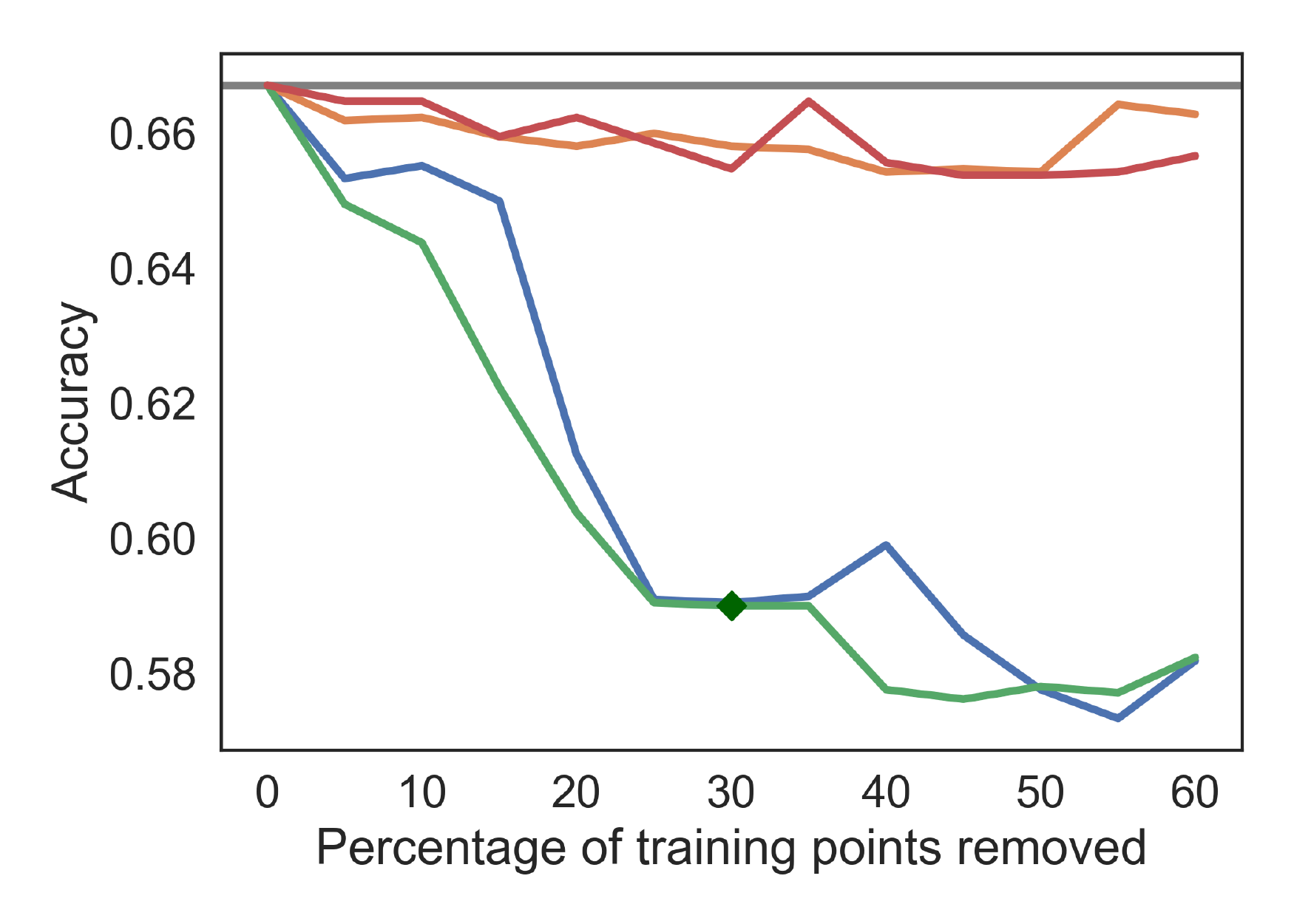}
    \end{subfigure}%
    \begin{subfigure}[b]{0.24\linewidth}
            \centering
            \includegraphics[width=\textwidth]{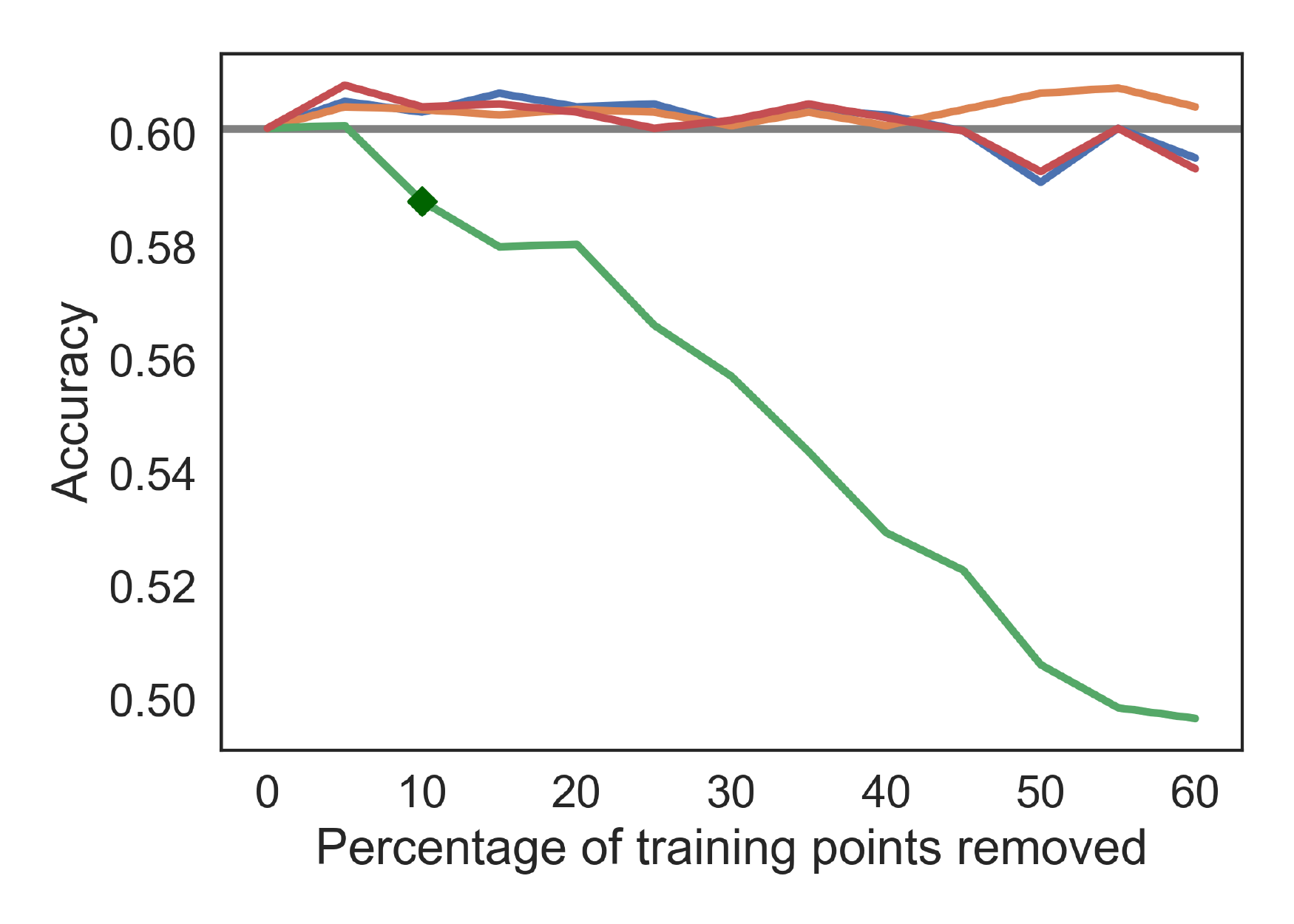}
    \end{subfigure}
    \begin{subfigure}[b]{0.24\linewidth}    
            \includegraphics[width=\textwidth]{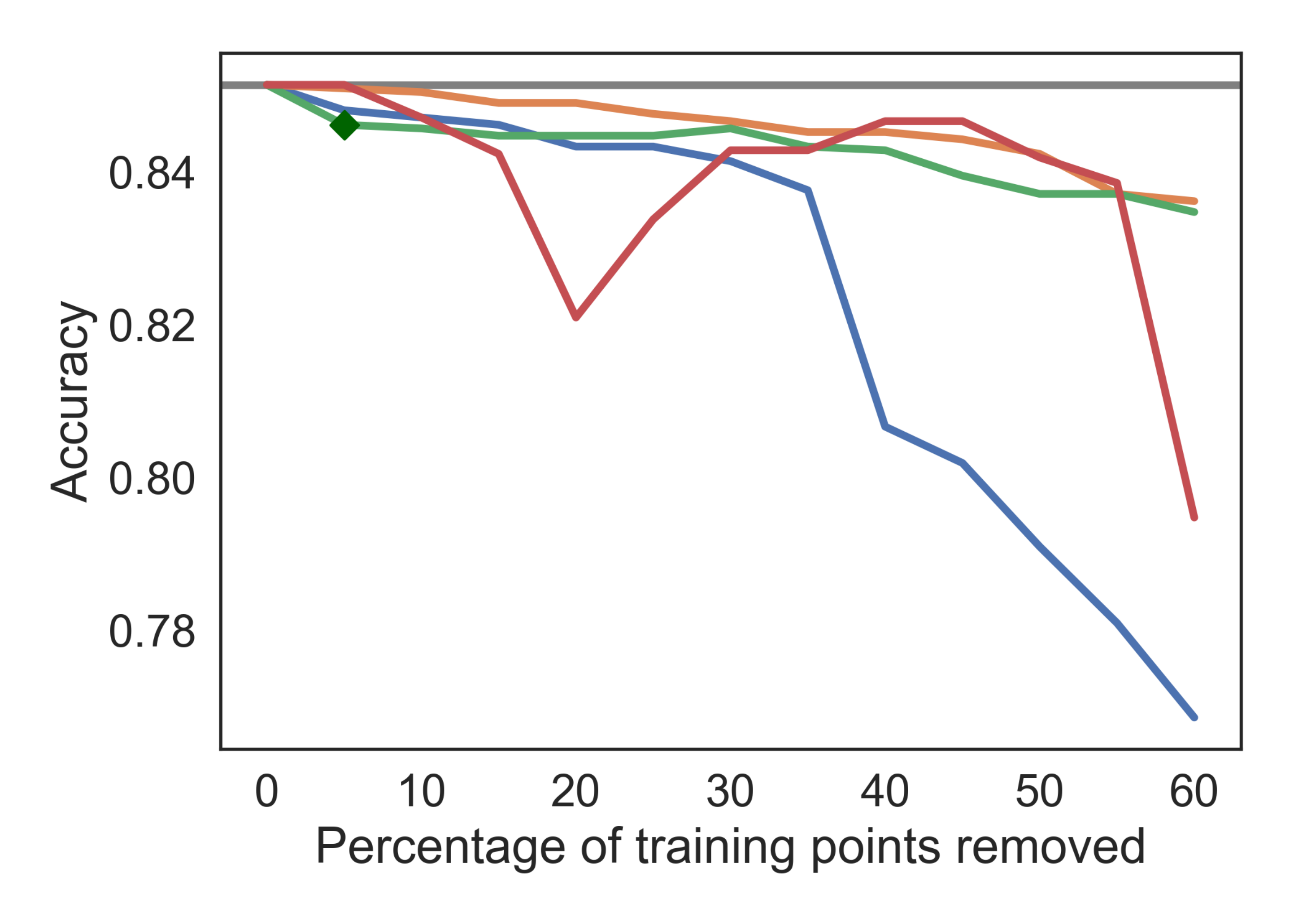}
    \end{subfigure}%
    \begin{subfigure}[b]{0.24\linewidth}
            \centering
            \includegraphics[width=\textwidth]{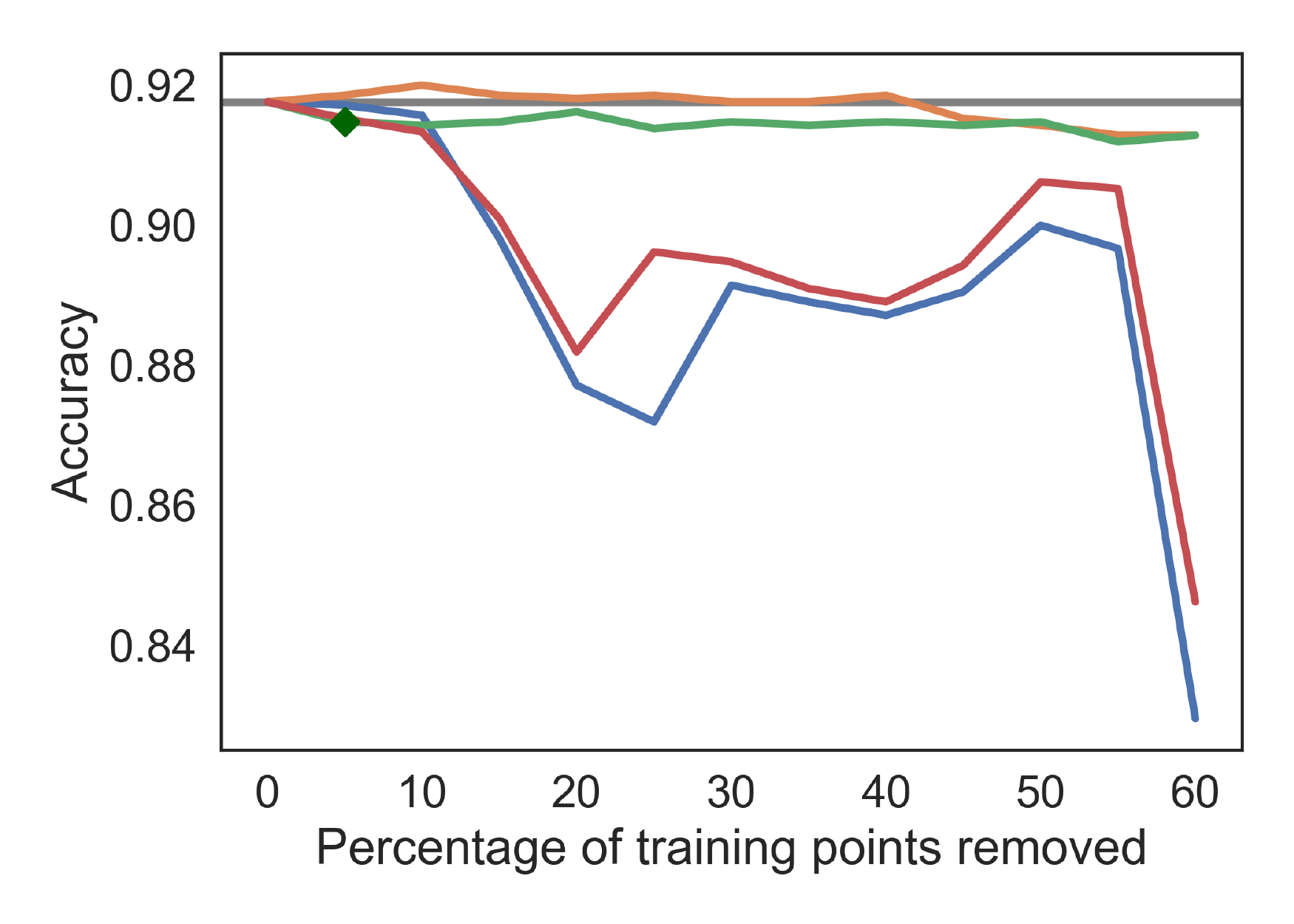}
    \end{subfigure}
     \begin{subfigure}[b]{0.24\linewidth}            
            \includegraphics[width=\textwidth]{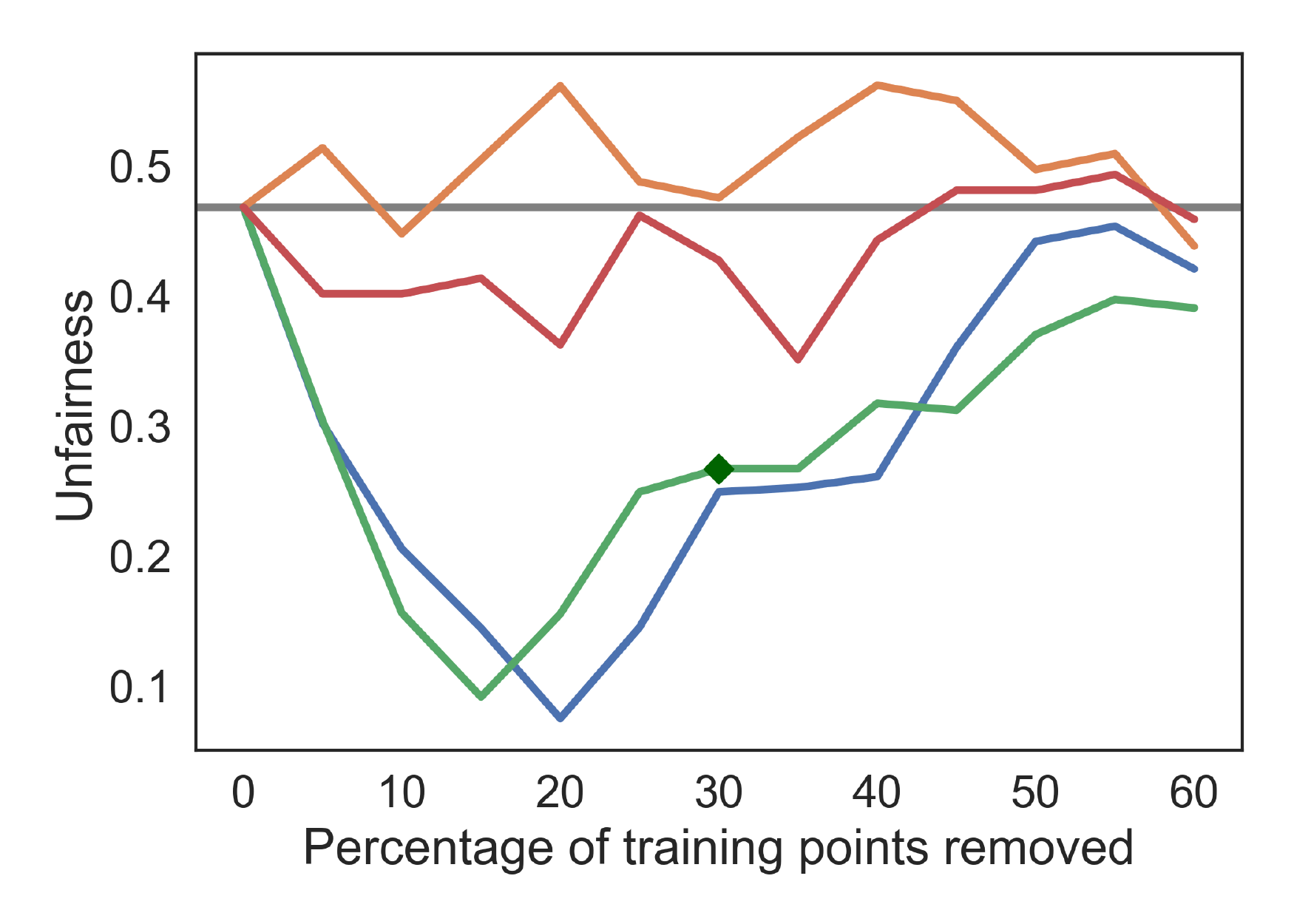}
            \caption{COMPAS}
    \end{subfigure}%
    \begin{subfigure}[b]{0.24\linewidth}
            \centering
            \includegraphics[width=\textwidth]{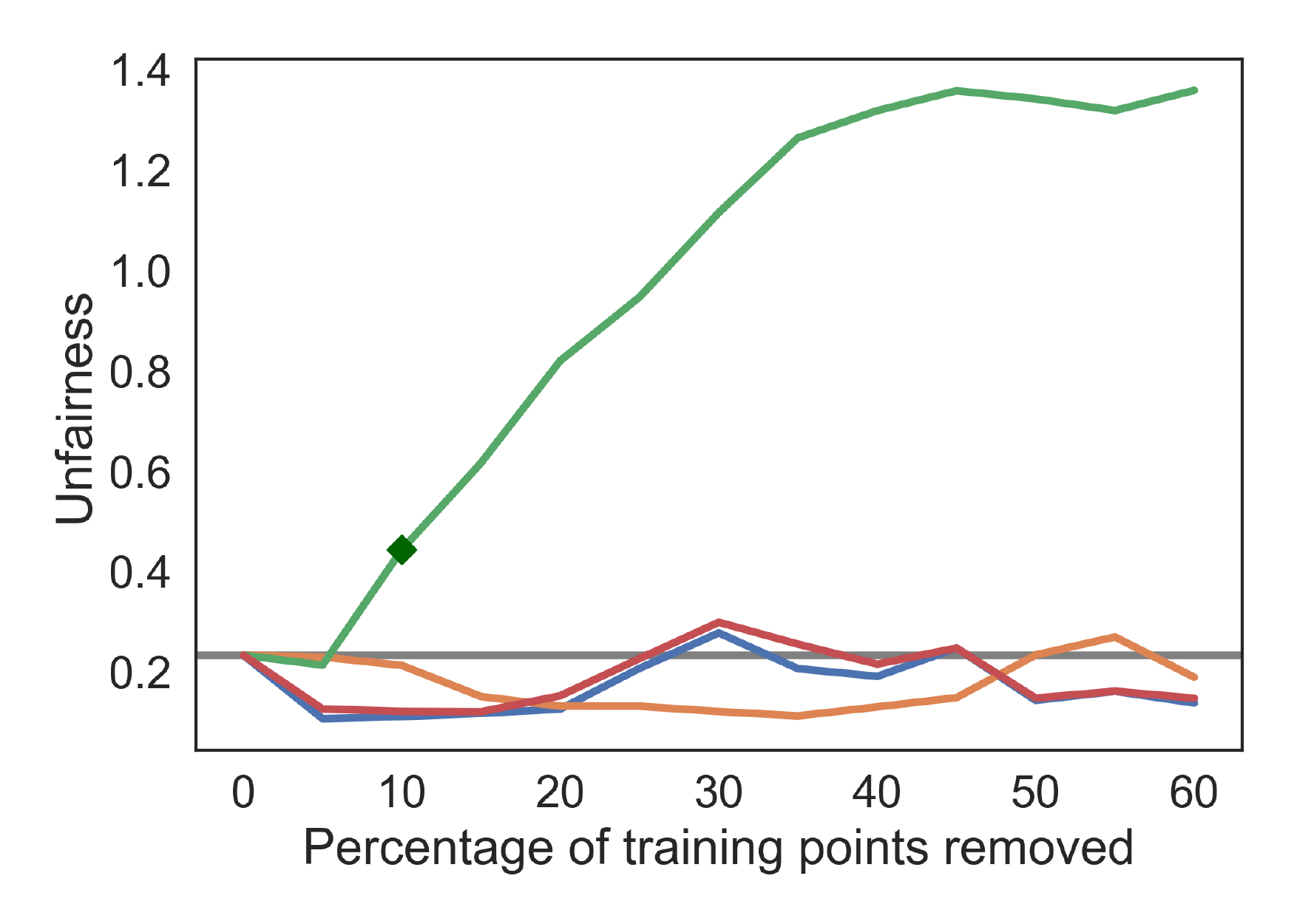}
            \caption{LSAT}
    \end{subfigure}
    \begin{subfigure}[b]{0.24\linewidth}            
            \includegraphics[width=\textwidth]{figures/pdfs/adult_LOOScorer_redundancy_0.0001_accuracy_train_selection_plot.pdf}
            \caption{Adult}
    \end{subfigure}%
    \begin{subfigure}[b]{0.24\linewidth}
            \centering
            \includegraphics[width=\textwidth]{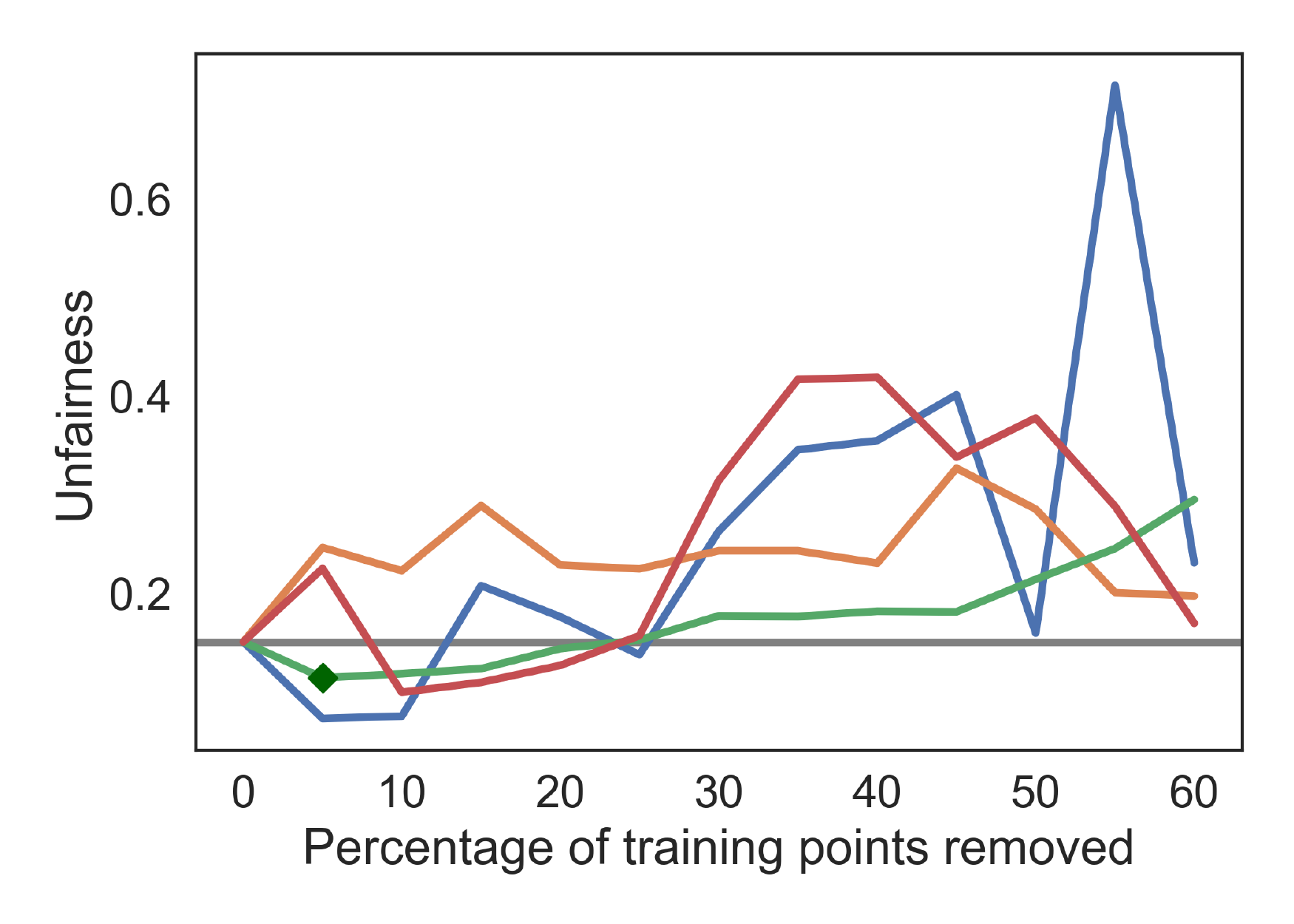}
            \caption{Bank}
    \end{subfigure}
    \caption{\small Training data performance on four datasets after removal of unfairness-inducing points. First row shows accuracy. Second row shows unfairness. Methods for selecting points to remove denoted by line color. Blue selects DIVINE points per Equation \ref{opt} with IF, $\mathcal{R}_\text{SR}$, and $\gamma$ \textit{tuned} via pairwise distance. Orange denotes randomly selected points. Green uses IF to select points ($\gamma = 0$). Red chooses points by maximizing $\mathcal{R}_\text{SR}$ alone. Grey indicates original model's performance. Diamond indicates the point after which all unfairness-inducing points have been removed.
    After this point, we expect unfairness to increase, as we start removing low importance (but helpful) points.
    With both IF and DIVINE, valuing data with respect to $f_\text{unf}$ identifies harmful data to remove; after removal, fairness outcomes improve greatly, though accuracy may drop slightly.
    }
    \label{fig:removal}
    \vspace{-0.5cm}
\end{figure*}

Once we have detected unfairness-inducing points, we may hope to improve our models fairness outcomes. We now consider removing unfairness-inducing points identified with DIVINE. We first calculate importance scores $I_i$ for each training point with respect to $f_\text{unf}$.
To use Algorithm~\ref{alg:example} to find points to remove, we negate each importance score (harmful points now have positive importance), allowing us to perform submodular \textit{maximization} via Equation~\ref{opt}.
We iteratively select sets of $m$ unfairness-inducing points to remove based on Equation~\ref{opt}. We let $m$ be equal to $5\%$ of the training data size. After removing the selected points, we retrain. 
In Figure~\ref{fig:removal}, we plot accuracy and unfairness ($f_\text{unf}$) after removing up to $60\%$ of the training data.
For all $4$ tabular datasets, $f_\text{unf}$ remains stable or decreases until a large fraction of the dataset has been removed. The corresponding drop in accuracy is minor. Consider IF on COMPAS: dropping first $10\%$ of the training points reduces unfairness by almost $70\%$ while only incurring a ~$2\%$ drop in accuracy.
This implies that in many cases, a \textit{significant} drop in unfairness can be achieved by dropping a small fraction of the training data points.

In Figure~\ref{fig:removal}, we remove sets of $m$ points based on importance scores calculated with respect to the original model. Instead, we can recalculate importance scores after every set of $m$ points is dropped. Results for removal with recalculation are in Appendix~\ref{add_experiments}. In Figure~\ref{fig:removal}, we report performance metrics on the training data after removing $m$ unfairness-inducing points. However, we can calculate performance metrics on the test data to report the effects of removal on accuracy and unfairness in generalization as well. We can also calculate $f_\text{unf}$ on held-out validation data (instead of on training data as in Figure~\ref{fig:removal}) when scoring training points and can even report performance metrics on test data. In Appendix~\ref{add_experiments}, we conduct experiments with validation and test data.


The benefit of removal of DIVINE points is reflected in the performance of LSAT in Figure~\ref{fig:removal}. In the first $5\%$ of points removed via IF with respect to $f_\text{unf}$ (green), $80\%$ of those points are females and $26\%$ are correctly classified by the original model. The model performs poorly after these points are removed. In contrast, in the first $5\%$ of points removed with DIVINE (blue), $50\%$ of points are female and $60\%$ are correctly classified points. DIVINE points more closely resemble the training data, which is $45\%$ female and correctly classifies $60\%$ of points. \textit{Thus, we find that we can value data points with respect to unfairness and then remove harmful points to improve fairness outcomes.}



\section{User Studies}
\label{user_study}
We conduct user studies to validate how useful DIVINE points are for explanation and for simulatability. Details about all user studies can be found in Appendix~\ref{hse}.  For all experiments in this section, we take DIVINE to  be IF with $\mathcal{R}_\text{SR}$ and find $\gamma$ by maximizing average pairwise distance.

\subsection{Manually Examining DIVINE Explanations}
\label{examples}
We consider how DIVINE points can be used as global and local training point-based explanations. 
We first assess if DIVINE points provide sufficient diversity and are not dominated by outliers like IF points~\citep{barshan2020relatif}.
While Section~\ref{experiments} primarily focuses on DIVINE points in tabular data and convex models, here we use image data and non-convex models. 
We train a convolutional neural network on FashionMNIST.
The top $4$ images obtained by IF and DS are similar to each other, as shown in Figure~\ref{fig:global_fmnist}, while DIVINE points are more varied. Notably, the amount of influence contained in the top-$4$ points from IF and DIVINE are roughly similar: for $10\%$ less influence, we obtain $50\%$ more diversity in $\mathcal{R}_\text{SR}$. Similar results hold for a multi-layer perceptron trained on a binary subset of MNIST ($1$ vs. $7$) in Appendix~\ref{add_experiments}: DIVINE points consider both $7$s and $1$s of various modes (differing thickness or style) to be influential.
In Figure~\ref{fig:local_fmnist}, when explaining a misclassified test point, DIVINE finds visually different training points to serve as an explanation. 
The average pairwise distance between the top DIVINE points is nearly double the average pairwise distance between the top IF points (Appendix~\ref{add_experiments}). As demonstrated on MNIST and FashionMNIST, DIVINE points have more diversity in feature space, which can help practitioners see which input regions influence their model.


Next, we conduct a user study to validate the utility of displaying DIVINE points as an explanation for practitioners. We asked $20$ participants with computer science experience to rank the diversity of the top-$m$ influential points from various methods (details in Appendix~\ref{hse}). When shown the top-$5$ FashionMNIST points from IF and DIVINE, $100\%$ of participants said DIVINE was more diverse than IF. When shown the top-$10$ FashionMNIST points from IF and DIVINE, $80\%$ of participants said DIVINE was more diverse. One participant noted that ``[DIVINE] seemed to be more distinct and varied,'' while another said that ``with [DIVINE] a more representative selection was used.'' 

In practice, influential training points can be useful for displaying the trustworthiness of a model~\citep{cai2019human}. \citet{zhou2019effects} display the top IF points and ask participants to rate the trustworthiness of the resultant model. Similarly, we displayed $10$ FashionMNIST points and asked participants to rate the trustworthiness of the resultant model. We show one set of points from DIVINE and another from IF.  They provided a trustworthy rating from 1 (Least) to 5 (Most) for each set of points.
DIVINE, $4.1 \pm 0.88$, was deemed more trustworthy then IF, $2.6 \pm 1.26$, ($p=0.003$, t-test). A participant said ``[DIVINE] has more variety so it is more trustworthy.''   Participants were then asked to decide if IF, RelatIF, or DIVINE provided a more useful local explanation for a misclassification on FashionMNIST. After seeing the misclassification and the top-$3$ points from each of the three methods, $50\%$ of participants preferred DIVINE, $30\%$ preferred RelatIF, and $20\%$ preferred IF. One participant said that ``the mistake made by the model is made obvious since the shape between shirt and coat are shown in [DIVINE],'' which suggests the modes detected by DIVINE (per Section~\ref{learning}) are useful in practice. \textit{This confirms that not only are DIVINE points quantitatively more diverse than IF points, but are also qualitatively perceived to be more diverse, trustworthy, and useful.}





\subsection{DIVINE for Simulatability}
\label{linedrawing}
\begin{wrapfigure}{R}{0.27\textwidth}
\vspace{-0.3cm}
\centering
\includegraphics[width=\linewidth]{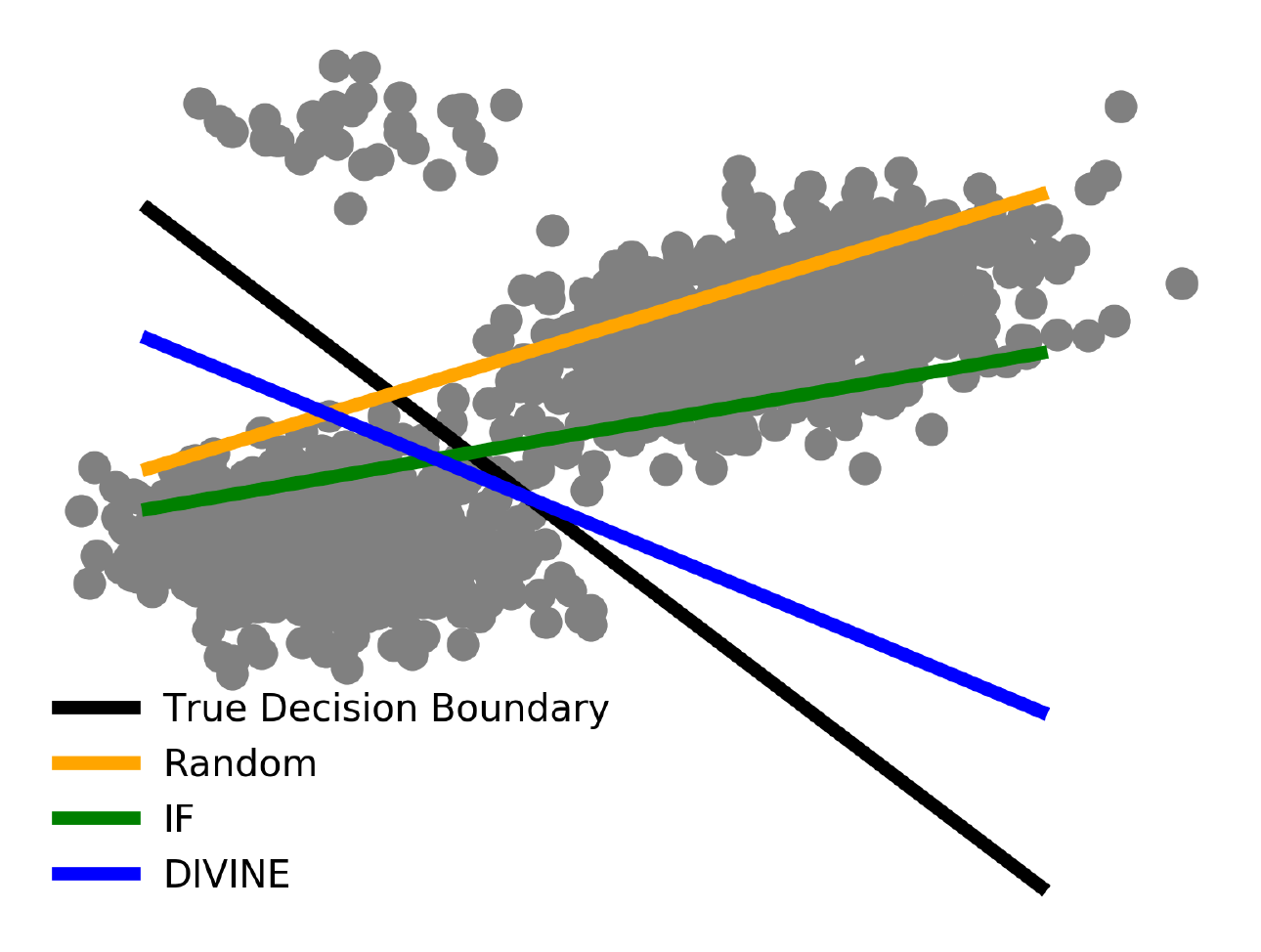}
        \caption{\small User Drawn\\ Decision Boundaries}
        \label{fig:boundary}
\vspace{-0.3cm}
\end{wrapfigure}
Many user studies for explainability consider how users can perform forward simulation, i.e. where a user uses an explanation to predict the model's behavior on an unseen test point~\citep{doshi2017towards,hase-bansal-2020-evaluating}. Another important consideration of explanations is \textbf{simulatability}~\citep{hoffman2018metrics,lipton2018mythos}, which measures how well a user can reason about an entire model given an explanation. 
We posit that diversity in influential samples will help practitioners with simulatability.
To test the simulatability of DIVINE points, we ask  practitioners to reconstruct a model decision boundary given a set of points from our synthetic data. Survey details can be found in Appendix~\ref{hse}. Our goal is to measure the similarity between the user-drawn decision boundary and the true decision boundary.
To a new set of $20$ participants, we display $m$ points on a grid: each point is colored by its predicted class. $10$ participants see $m=5$ points. The rest see $m=10$. We then ask the participant to draw a decision boundary that separates the two classes. We ask them to draw a decision boundary after seeing the top-$m$ IF points, top-$m$ DIVINE points, or $m$ random points. When shown $5$ points, we find that the cosine similarity between the user drawn boundary and the true boundary is $0.91 \pm 0.25$ with DIVINE, $0.59 \pm 0.78$ with IF, and $0.48 \pm 0.81$ with Random.
When shown $10$ points, we find that the cosine similarity between the user drawn boundary and the true boundary is $0.99 \pm 0.00$ with DIVINE, is $0.39 \pm 0.88$ with IF, and $0.33 \pm 0.70$ with Random. We show the average user drawn decision boundary (after observing $10$ points) in Figure~\ref{fig:boundary}. Notice that the average decision boundary drawn after observing DIVINE points is closer to the true decision boundary. We find that DIVINE points were considerably more helpful that IF points to participants (p= $0.011$, t-test).
While DIVINE points are not optimized for decision boundary reconstruction and might be misleading in some cases, it is reassuring to know that they provide sufficiently diverse explanations such that users can reconstruct the model decision boundary.
Though our study was performed on 2D synthetic data with linear decision boundaries, our findings are promising.  We hope in future work to extend this study to higher dimensions.
\textit{We conclude that practitioners find DIVINE points helpful for simulatability}.

\section{Conclusions, Limitations, and Future Work}
\label{conc}
\begin{wrapfigure}{R}{0.45\textwidth}
\vspace{-0.3cm}
\small
\resizebox{\linewidth}{!}{%
\begin{tabular}{l|l|l|l}
\toprule
             Method    & $\theta$ & $\mathcal{R}(\mathcal{S})$ & $f(\theta)$ \\ 
\midrule
Prototypes~\citep{kim2016MMD} & \xmark & \cmark  &  N/A \\
\cline{1-4} 
IF~\citep{koh2017understanding} & \cmark & \xmark  &  Loss \\ 
\cline{1-4} 
DS~\citep{ghorbani2019data} & \cmark & \xmark  &  Accuracy and AUC \\ 
\cline{1-4} 
DIVINE (Ours) & \cmark & \cmark  &  Loss, Unfairness, etc. \\ 
\bottomrule
\end{tabular}
}
\captionof{table}{Comparison}
\label{tab:practitioner_smallea}
\vspace{-0.3cm}
\end{wrapfigure}
In this work, we propose an approach for finding DIVerse INfluEntial (DIVINE) training points. 
We note that existing training point importance methods tend to assign high importance to similar points; hence, we propose a method to select a diverse and influential subset of the data using submodular optimization. In Table~\ref{tab:practitioner_smallea}, we summarize how DIVINE compares to existing training point importance methods. Our method enables practitioners to inject diversity into explanations of model behavior, derived from training point importance scores.
Additionally, previous work has mainly investigated influential points with respect to a model's loss. We go further by considering valuation of data points with respect to model fairness. We then examine how unfairness can be reduced. We use DIVINE to detect and remove unfairness-inducing points, leading to improvements in model fairness.
Our experiments on synthetic and real-world datasets demonstrate that, using DIVINE, practitioners can visualize a diverse summary of influential training points and thus understand the possible modes of data that contribute to their model's behavior.
In our user studies, we find that practitioners perceive DIVINE points to be more diverse, more useful, and more trustworthy. Practitioners also find DIVINE helpful for model simulatability.

We acknowledge that practitioners may not want diversity in their influential training points, if they do not desire a complete picture of model behavior. For local explanations, 
one participant in our user study (Section~\ref{examples}) noted that DIVINE points ``may include conflicting and uncomparable [sic] items.'' As such, practitioners may want to clarify the goals for using training point importance and leverage diversity accordingly.
In Appendix~\ref{guide}, we provide a detailed guide for how practitioners can leverage DIVINE and our codebase. 
Specific to removing points, we note that removing unfairness-inducing points might be too harsh. Future work might learn to down-weight those points with weight $w_i \in \left (-\frac{1}{n},\frac{1}{n}\right ]$ as opposed to dropping them with weight $w_i = -\frac{1}{n}$.
Future work could also value data using alternative metrics, such as robustness or privacy.
Nonetheless, practitioners can leverage DIVINE to value training points based on their effect on model-specific evaluation metrics and to summarize model behavior either locally or globally. 
We hope DIVINE is a helpful intervention for practitioners to generate data visualizations and refine their models.


\clearpage

\section*{Acknowledgements}
The authors thank Matthew Ashman, Krishna Gummadi, Weiyang Liu, Aditya Nori, Elre Oldewage, Richard E. Turner, and Timothy Ye for their comments on this manuscript. UB acknowledges support from the Mozilla Foundation and from DeepMind and the Leverhulme Trust via the Leverhulme Centre for the Future of Intelligence (CFI). IC acknowledges support from Microsoft Research.
AW acknowledges support from a Turing AI Fellowship under grant EP/V025379/1, The Alan Turing Institute under EPSRC grant EP/N510129/1 \& TU/B/000074, and the Leverhulme Trust via CFI. 

\bibliography{example_paper}

\begin{thebibliography}{54}
\providecommand{\natexlab}[1]{#1}
\providecommand{\url}[1]{\texttt{#1}}
\expandafter\ifx\csname urlstyle\endcsname\relax
  \providecommand{\doi}[1]{doi: #1}\else
  \providecommand{\doi}{doi: \begingroup \urlstyle{rm}\Url}\fi

\bibitem[Anahideh et~al.(2020)Anahideh, Asudeh, and
  Thirumuruganathan]{anahideh2020fair}
Hadis Anahideh, Abolfazl Asudeh, and Saravanan Thirumuruganathan.
\newblock Fair active learning.
\newblock \emph{arXiv preprint arXiv:2001.01796}, 2020.

\bibitem[Angwin et~al.(2016)Angwin, Larson, Mattu, Kirchner, and
  ProPublica]{compas}
Julia Angwin, Jeff Larson, Surya Mattu, Lauren Kirchner, and ProPublica.
\newblock Machine bias, 2016.

\bibitem[Bach(2011)]{bach2011learning}
Francis Bach.
\newblock Learning with submodular functions: A convex optimization
  perspective.
\newblock \emph{arXiv preprint arXiv:1111.6453}, 2011.

\bibitem[Barshan et~al.(2020)Barshan, Brunet, and
  Dziugaite]{barshan2020relatif}
Elnaz Barshan, Marc-Etienne Brunet, and Gintare~Karolina Dziugaite.
\newblock Relatif: Identifying explanatory training examples via relative
  influence.
\newblock \emph{arXiv preprint arXiv:2003.11630}, 2020.

\bibitem[Berk et~al.(2017)Berk, Heidari, Jabbari, Kearns, and
  Roth]{berk2018fairness}
Richard Berk, Hoda Heidari, Shahin Jabbari, Michael Kearns, and Aaron Roth.
\newblock Fairness in criminal justice risk assessments: The state of the art.
\newblock \emph{Sociological Methods \& Research}, page 0049124118782533, 2017.

\bibitem[Bhatt et~al.(2020{\natexlab{a}})Bhatt, Xiang, Sharma, Weller, Taly,
  Jia, Ghosh, Puri, Moura, and Eckersley]{bhatt2020explainable}
Umang Bhatt, Alice Xiang, Shubham Sharma, Adrian Weller, Ankur Taly, Yunhan
  Jia, Joydeep Ghosh, Ruchir Puri, Jos{\'e}~MF Moura, and Peter Eckersley.
\newblock Explainable machine learning in deployment.
\newblock In \emph{Proceedings of the 2020 Conference on Fairness,
  Accountability, and Transparency}, pages 648--657, 2020{\natexlab{a}}.

\bibitem[Bhatt et~al.(2020{\natexlab{b}})Bhatt, Zafar, Gummadi, and
  Weller]{bhatt2020counterfactual}
Umang Bhatt, Muhammad~Bilal Zafar, Krishna Gummadi, and Adrian Weller.
\newblock Counterfactul accuracies for alternative models.
\newblock \emph{ICLR Workshop on Machine Learning in Real Life (ML-IRL)},
  2020{\natexlab{b}}.

\bibitem[Bien and Tibshirani(2011)]{bien2011prototype}
Jacob Bien and Robert Tibshirani.
\newblock Prototype selection for interpretable classification.
\newblock \emph{The Annals of Applied Statistics}, pages 2403--2424, 2011.

\bibitem[Breiman(2001)]{breiman2001statistical}
Leo Breiman.
\newblock Statistical modeling: The two cultures.
\newblock \emph{Statistical Science}, 16\penalty0 (3):\penalty0 199--231, 2001.

\bibitem[Cai et~al.(2019)Cai, Reif, Hegde, Hipp, Kim, Smilkov, Wattenberg,
  Viegas, Corrado, Stumpe, et~al.]{cai2019human}
Carrie~J Cai, Emily Reif, Narayan Hegde, Jason Hipp, Been Kim, Daniel Smilkov,
  Martin Wattenberg, Fernanda Viegas, Greg~S Corrado, Martin~C Stumpe, et~al.
\newblock Human-centered tools for coping with imperfect algorithms during
  medical decision-making.
\newblock In \emph{Proceedings of the 2019 CHI Conference on Human Factors in
  Computing Systems}, pages 1--14, 2019.

\bibitem[Cole and Williamson(2019)]{Cole2019AvoidingRV}
Guy~W. Cole and Sinead~A. Williamson.
\newblock Avoiding resentment via monotonic fairness.
\newblock \emph{ArXiv}, abs/1909.01251, 2019.

\bibitem[Cook and Weisberg(1980)]{influence}
{R. Dennis} Cook and Sanford Weisberg.
\newblock Characterizations of an empirical influence function for detecting
  influential cases in regression.
\newblock \emph{Technometrics}, 22:\penalty0 495--508, 1980.

\bibitem[Dasgupta et~al.(2008)Dasgupta, Hsu, and
  Monteleoni]{dasgupta2008general}
Sanjoy Dasgupta, Daniel~J Hsu, and Claire Monteleoni.
\newblock A general agnostic active learning algorithm.
\newblock In \emph{Advances in neural information processing systems}, pages
  353--360, 2008.

\bibitem[Datta et~al.(2016)Datta, Sen, and Zick]{datta2016algorithmic}
Anupam Datta, Shayak Sen, and Yair Zick.
\newblock Algorithmic transparency via quantitative input influence: Theory and
  experiments with learning systems.
\newblock In \emph{2016 IEEE symposium on security and privacy (SP)}, pages
  598--617. IEEE, 2016.

\bibitem[Doshi-Velez and Kim(2017)]{doshi2017towards}
Finale Doshi-Velez and Been Kim.
\newblock Towards a rigorous science of interpretable machine learning.
\newblock \emph{arXiv preprint arXiv:1702.08608}, 2017.

\bibitem[Dua and Graff(2017)]{Dua:2019}
Dheeru Dua and Casey Graff.
\newblock {UCI} machine learning repository, 2017.

\bibitem[Fisher et~al.(2019)Fisher, Rudin, and Dominici]{rashomon}
Aaron Fisher, Cynthia Rudin, and Francesca Dominici.
\newblock All models are wrong, but many are useful: Learning a variable's
  importance by studying an entire class of prediction models simultaneously.
\newblock \emph{Journal of Machine Learning Research}, 20\penalty0
  (177):\penalty0 1--81, 2019.

\bibitem[Ghorbani and Zou(2019)]{ghorbani2019data}
Amirata Ghorbani and James Zou.
\newblock Data shapley: Equitable valuation of data for machine learning.
\newblock In \emph{International Conference on Machine Learning}, pages
  2242--2251, 2019.

\bibitem[Gretton et~al.(2012)Gretton, Borgwardt, Rasch, Sch{\"o}lkopf, and
  Smola]{gretton2012kernel}
Arthur Gretton, Karsten~M Borgwardt, Malte~J Rasch, Bernhard Sch{\"o}lkopf, and
  Alexander Smola.
\newblock A kernel two-sample test.
\newblock \emph{The Journal of Machine Learning Research}, 13\penalty0
  (1):\penalty0 723--773, 2012.

\bibitem[Gurumoorthy et~al.(2019)Gurumoorthy, Dhurandhar, Cecchi, and
  Aggarwal]{gurumoorthy2019efficient}
Karthik~S Gurumoorthy, Amit Dhurandhar, Guillermo Cecchi, and Charu Aggarwal.
\newblock Efficient data representation by selecting prototypes with importance
  weights.
\newblock In \emph{2019 IEEE International Conference on Data Mining (ICDM)},
  pages 260--269. IEEE, 2019.

\bibitem[Hampel(1974)]{hampel1974influence}
Frank~R Hampel.
\newblock The influence curve and its role in robust estimation.
\newblock \emph{Journal of the american statistical association}, 69\penalty0
  (346):\penalty0 383--393, 1974.

\bibitem[Hanneke(2009)]{hanneke2009theoretical}
Steve Hanneke.
\newblock Theoretical foundations of active learning.
\newblock 2009.

\bibitem[Hardt et~al.(2016{\natexlab{a}})Hardt, Price, and Srebro]{hardt2016}
Moritz Hardt, Eric Price, and Nati Srebro.
\newblock Equality of opportunity in supervised learning.
\newblock In \emph{Advances in Neural Information Processing Systems
  (NeurIPS)}, 2016{\natexlab{a}}.

\bibitem[Hardt et~al.(2016{\natexlab{b}})Hardt, Price, and
  Srebro]{hardt2016equality}
Moritz Hardt, Eric Price, and Nati Srebro.
\newblock Equality of opportunity in supervised learning.
\newblock In \emph{Advances in neural information processing systems}, pages
  3315--3323, 2016{\natexlab{b}}.

\bibitem[Hase and Bansal(2020)]{hase-bansal-2020-evaluating}
Peter Hase and Mohit Bansal.
\newblock Evaluating explainable {AI}: Which algorithmic explanations help
  users predict model behavior?
\newblock In \emph{Proceedings of the 58th Annual Meeting of the Association
  for Computational Linguistics}, pages 5540--5552, Online, July 2020.
  Association for Computational Linguistics.
\newblock \doi{10.18653/v1/2020.acl-main.491}.

\bibitem[Hastie et~al.(2009)Hastie, Tibshirani, and
  Friedman]{hastie2009elements}
Trevor Hastie, Robert Tibshirani, and Jerome Friedman.
\newblock \emph{The elements of statistical learning: data mining, inference,
  and prediction}.
\newblock Springer Science \& Business Media, 2009.

\bibitem[Hoffman et~al.(2018)Hoffman, Mueller, Klein, and
  Litman]{hoffman2018metrics}
Robert~R Hoffman, Shane~T Mueller, Gary Klein, and Jordan Litman.
\newblock Metrics for explainable ai: Challenges and prospects.
\newblock \emph{arXiv preprint arXiv:1812.04608}, 2018.

\bibitem[Jeyakumar et~al.(2020)Jeyakumar, Noor, Cheng, Garcia, and
  Srivastava]{jeyakumar2020can}
Jeya~Vikranth Jeyakumar, Joseph Noor, Yu-Hsi Cheng, Luis Garcia, and Mani
  Srivastava.
\newblock How can i explain this to you? an empirical study of deep neural
  network explanation methods.
\newblock \emph{Advances in Neural Information Processing Systems}, 33, 2020.

\bibitem[Khanna et~al.(2019)Khanna, Kim, Ghosh, and
  Koyejo]{khanna2019interpreting}
Rajiv Khanna, Been Kim, Joydeep Ghosh, and Sanmi Koyejo.
\newblock Interpreting black box predictions using fisher kernels.
\newblock In \emph{The 22nd International Conference on Artificial Intelligence
  and Statistics}, pages 3382--3390, 2019.

\bibitem[Kim et~al.(2016)Kim, Khanna, and Koyejo]{kim2016MMD}
Been Kim, Rajiv Khanna, and Sanmi Koyejo.
\newblock Examples are not enough, learn to criticize! criticism for
  interpretability.
\newblock In \emph{Advances in Neural Information Processing Systems}, 2016.

\bibitem[Koh and Liang(2017)]{koh2017understanding}
Pang~Wei Koh and Percy Liang.
\newblock Understanding black-box predictions via influence functions.
\newblock In \emph{Proceedings of the 34th International Conference on Machine
  Learning-Volume 70 ({ICML} 2017)}, pages 1885--1894. Journal of Machine
  Learning Research, 2017.

\bibitem[Koh et~al.(2019)Koh, Ang, Teo, and Liang]{koh2019accuracy}
Pang Wei~W Koh, Kai-Siang Ang, Hubert Teo, and Percy~S Liang.
\newblock On the accuracy of influence functions for measuring group effects.
\newblock In \emph{Advances in Neural Information Processing Systems}, pages
  5254--5264, 2019.

\bibitem[Krause and Golovin(2014)]{krause2014submodular}
Andreas Krause and Daniel Golovin.
\newblock Submodular function maximization., 2014.

\bibitem[Kusner et~al.(2017)Kusner, Loftus, Russell, and
  Silva]{kusner2017counterfactual}
Matt~J Kusner, Joshua Loftus, Chris Russell, and Ricardo Silva.
\newblock Counterfactual fairness.
\newblock In \emph{Advances in neural information processing systems}, pages
  4066--4076, 2017.

\bibitem[Kwon et~al.(2020)Kwon, Rivas, and Zou]{kwon2020efficient}
Yongchan Kwon, Manuel~A Rivas, and James Zou.
\newblock Efficient computation and analysis of distributional shapley values.
\newblock \emph{arXiv preprint arXiv:2007.01357}, 2020.

\bibitem[LeCun(1998)]{lecun1998mnist}
Yann LeCun.
\newblock The mnist database of handwritten digits.
\newblock \emph{http://yann. lecun. com/exdb/mnist/}, 1998.

\bibitem[Letham et~al.(2016)Letham, Letham, Rudin, and
  Browne]{letham2016prediction}
Benjamin Letham, Portia~A Letham, Cynthia Rudin, and Edward~P Browne.
\newblock Prediction uncertainty and optimal experimental design for learning
  dynamical systems.
\newblock \emph{Chaos: An Interdisciplinary Journal of Nonlinear Science},
  26\penalty0 (6):\penalty0 063110, 2016.

\bibitem[Libbrecht et~al.(2018)Libbrecht, Bilmes, and
  Noble]{libbrecht2018choosing}
Maxwell~W Libbrecht, Jeffrey~A Bilmes, and William~Stafford Noble.
\newblock Choosing non-redundant representative subsets of protein sequence
  data sets using submodular optimization.
\newblock \emph{Proteins: Structure, Function, and Bioinformatics}, 86\penalty0
  (4):\penalty0 454--466, 2018.

\bibitem[Lin and Bilmes(2011)]{lin-bilmes-2011-class}
Hui Lin and Jeff Bilmes.
\newblock A class of submodular functions for document summarization.
\newblock In \emph{Proceedings of the 49th Annual Meeting of the Association
  for Computational Linguistics: Human Language Technologies}, pages 510--520,
  Portland, Oregon, USA, June 2011. Association for Computational Linguistics.

\bibitem[Lipton(2018)]{lipton2018mythos}
Zachary~C Lipton.
\newblock The mythos of model interpretability.
\newblock \emph{Queue}, 16\penalty0 (3):\penalty0 31--57, 2018.

\bibitem[Lundberg(2020)]{lundberg2020explaining}
Scott~M Lundberg.
\newblock Explaining quantitative measures of fairness.
\newblock \emph{Fair \& Responsible AI Workshop @ CHI2020}, 2020.

\bibitem[Marx et~al.(2019)Marx, du~Pin~Calmon, and Ustun]{pred_mult}
Charles~T. Marx, Flavio du~Pin~Calmon, and Berk Ustun.
\newblock Predictive multiplicity in classification.
\newblock \emph{arXiv:1909.06677}, 2019.

\bibitem[Mirzasoleiman et~al.(2015)Mirzasoleiman, Badanidiyuru, Karbasi,
  Vondr{\'a}k, and Krause]{mirzasoleiman2015lazier}
Baharan Mirzasoleiman, Ashwinkumar Badanidiyuru, Amin Karbasi, Jan Vondr{\'a}k,
  and Andreas Krause.
\newblock Lazier than lazy greedy.
\newblock In \emph{Proceedings of the AAAI Conference on Artificial
  Intelligence}, volume~29, 2015.

\bibitem[Nemhauser et~al.(1978)Nemhauser, Wolsey, and
  Fisher]{nemhauser1978analysis}
George~L Nemhauser, Laurence~A Wolsey, and Marshall~L Fisher.
\newblock An analysis of approximations for maximizing submodular set
  functions—i.
\newblock \emph{Mathematical programming}, 14\penalty0 (1):\penalty0 265--294,
  1978.

\bibitem[Pearl(2009)]{pearl2009causality}
Judea Pearl.
\newblock \emph{Causality}.
\newblock Cambridge {U}niversity {P}ress, 2009.

\bibitem[Prasad et~al.(2014)Prasad, Jegelka, and Batra]{prasad2014submodular}
Adarsh Prasad, Stefanie Jegelka, and Dhruv Batra.
\newblock Submodular meets structured: Finding diverse subsets in
  exponentially-large structured item sets.
\newblock \emph{Advances in Neural Information Processing Systems}, 2014.

\bibitem[Shapley(1953)]{shapley52}
Lloyd~S Shapley.
\newblock A value for n-person games.
\newblock \emph{Contributions to the Theory of Games}, 2\penalty0
  (28):\penalty0 307--317, 1953.

\bibitem[Tschiatschek et~al.(2014)Tschiatschek, Iyer, Wei, and
  Bilmes]{tschiatschek2014learning}
Sebastian Tschiatschek, Rishabh~K Iyer, Haochen Wei, and Jeff~A Bilmes.
\newblock Learning mixtures of submodular functions for image collection
  summarization.
\newblock In \emph{Advances in neural information processing systems}, 2014.

\bibitem[Van Der~Maaten(2014)]{van2014accelerating}
Laurens Van Der~Maaten.
\newblock Accelerating t-sne using tree-based algorithms.
\newblock \emph{The Journal of Machine Learning Research}, 15\penalty0
  (1):\penalty0 3221--3245, 2014.

\bibitem[Wiener and El-Yaniv(2011)]{agnostic}
Yair Wiener and Ran El-Yaniv.
\newblock Agnostic selective classification.
\newblock In J.~Shawe-Taylor, R.~S. Zemel, P.~L. Bartlett, F.~Pereira, and
  K.~Q. Weinberger, editors, \emph{Advances in Neural Information Processing
  Systems 24}, pages 1665--1673. Curran Associates, Inc., 2011.

\bibitem[Xiao et~al.(2017)Xiao, Rasul, and Vollgraf]{xiao2017fashion}
Han Xiao, Kashif Rasul, and Roland Vollgraf.
\newblock Fashion-mnist: a novel image dataset for benchmarking machine
  learning algorithms.
\newblock \emph{arXiv preprint arXiv:1708.07747}, 2017.

\bibitem[Yeh et~al.(2018)Yeh, Kim, Yen, and Ravikumar]{yeh2018representer}
Chih-Kuan Yeh, Joon Kim, Ian En-Hsu Yen, and Pradeep~K Ravikumar.
\newblock Representer point selection for explaining deep neural networks.
\newblock In \emph{Advances in neural information processing systems}, pages
  9291--9301, 2018.

\bibitem[Zafar et~al.(2017)Zafar, Valera, Rogriguez, and
  Gummadi]{zafar2017fairness}
Muhammad~Bilal Zafar, Isabel Valera, Manuel~Gomez Rogriguez, and Krishna~P
  Gummadi.
\newblock Fairness constraints: Mechanisms for fair classification.
\newblock In \emph{Artificial Intelligence and Statistics}, pages 962--970,
  2017.

\bibitem[Zhou et~al.(2019)Zhou, Li, Hu, Yu, Chen, Li, and
  Wang]{zhou2019effects}
Jianlong Zhou, Zhidong Li, Huaiwen Hu, Kun Yu, Fang Chen, Zelin Li, and Yang
  Wang.
\newblock Effects of influence on user trust in predictive decision making.
\newblock In \emph{Extended Abstracts of the 2019 CHI Conference on Human
  Factors in Computing Systems}, pages 1--6, 2019.

\end{thebibliography}
\bibliographystyle{plainnat}


\clearpage

\appendix
\section*{Appendix}
This appendix is formatted as follows.
\begin{enumerate}
    \item We present a \textbf{practitioner guide} in Appendix~\ref{guide}. We discuss how one would go about selecting the various parameters used to find DIVINE points. 
    \item We discuss extensions of \textbf{counterfactual prediction} for training point importance in Appendix~\ref{cfa_app}.
    \item We provide additional details about our \textbf{experimental setup} in Appendix~\ref{app:setup}.
    \item We report \textbf{additional experimental results} in Appendix~\ref{add_experiments}.
    \item We discuss details of our \textbf{user studies} in Appendix~\ref{hse}.
\end{enumerate}

\section{Practitioner Guide}
\label{guide}
Throughout the paper, we use the word ``practitioners'' to refer to data scientists who can use DIVINE in practical ML settings where explainability is valued, or those who hope to refine their models by better understanding their training data.
In this guide, we explain how practitioners can select the parameters used in DIVINE: importance measure $\mathcal{I}$, evaluation function $f$, diversity function $\mathcal{R}(\mathcal{S})$, influence-diversity tradeoff $\gamma$, and the number of DIVINE points, $m$. 
Our code can be extended to support additional influence measures $\mathcal{I}$, submodular diversity functions $\mathcal{R}$, evaluation functions $f$, and $\gamma$ selection strategies.

\subsection{Influence Measure $\mathcal{I}$}
Within our work, the influence measure $\mathcal{I}$ assigns importance to individual data points and to groups of data points. We aim to find an importance score $I_i$ for the $i$-th training point. Under our additivity assumption per~\citet{koh2019accuracy}, we let the importance of a set of points $\mathcal{S}$ be $\mathcal{I}(\mathcal{S}) = \sum_{x_i \in \mathcal{S}} I_i$. We can obtain importance scores from various methods. In our main paper and in our additional experiments (Appendix~\ref{add_experiments}), we let $\mathcal{I}$ be influence functions~\citep{koh2017understanding}, Data Shapley~\citep{ghorbani2019data}, counterfactual prediction~\citep{bhatt2020counterfactual}, or leave-one-out (LOO)~\citep{hastie2009elements}.
In Table~\ref{tab:practitioner}, we compare various methods for finding valuable training points with respect to a model $\theta$ or a diversity function $\mathcal{R}$.
\begin{table*}[htb]
\centering
\begin{tabular}{l|l|l|l}
\toprule
                Method    & $\theta$ & $\mathcal{R}(\mathcal{S})$ &  $f(\theta)$ \\ 
\midrule
Prototypes/Criticisms~\citep{kim2016MMD} & \xmark & \cmark  &  N/A \\
\cline{1-4} 
ProtoDash~\citep{gurumoorthy2019efficient} & \xmark & \cmark  &  N/A \\
\hline
Influential Points~\citep{koh2017understanding} & \cmark & \xmark  &  Loss \\ 
\cline{1-4} 
Representer Points~\citep{yeh2018representer} & \cmark & \xmark  &  Loss \\ 
\cline{1-4} 
Seq. Bayesian Quadrature~\citep{khanna2019interpreting} & \cmark & \xmark  &  Loss \\ 
\cline{1-4} 
Data Shapley~\citep{ghorbani2019data} & \cmark & \xmark  &  Accuracy and AUC \\ 
\cline{1-4} 
RelatIF~\citep{barshan2020relatif} & \cmark & \xmark  &  Loss \\ 
\midrule
DIVINE (Ours) & \cmark & \cmark  &  Any $f$ (Loss, Unfairness, etc.) \\ 
\bottomrule
\end{tabular}
\caption{Practitioners can leverage DIVINE to value data points based on their contributions to model-specific evaluation metrics and then select a diverse, influential subset of points as a summary. We list other methods based on (1) their dependence on model parameters (e.g., prototypes are model independent), (2) the diversity in the points to which they assign high importance, and (3) the metric with respect to which the data points are valued.}
\label{tab:practitioner}
\end{table*}

One would select influence functions if they prefer a fast computation (assuming the Hessian computation is done once for a reasonably sized set of parameters). 
One could use Data Shapley if they want importance scores that adhere to the game-theoretic guarantees of the Shapley value.
Counterfactual prediction is a training point importance method that is robust to label noise: we discuss this  at length in Appendix~\ref{cfa_app}.
LOO scores are also possible to compute, but note these may be computationally expensive as retraining is required for every point that is dropped. We show how all methods behave in comparison to each other on our synthetic data in Appendix~\ref{add_experiments}. In our package, influence functions, Data Shapley, LOO, and counterfactual prediction are supported.

\label{eval_funcs}
Moreover, when applying an importance scoring measure, a practitioner may have to select an evaluation function $f$ with respect to which one wants to value datapoints.
While we primarily let $f$ be $f_\text{loss}$ or $f_\text{unf}$ in our experiments, we report additional evaluation functions $f$ that can be used to score data points in Table~\ref{tab:eval}. For every evaluation function, we write the function such that a large $f(\theta)$ is undesirable: a negative importance score (if our score $I$ is taken to be the difference between $f$ evaluated at the new and old parameters -- trained with and without the point respectively) implies that a point is harmful to $f$. 
Moreover, we can also consider functions that do not simply take a difference between the evaluation function's value at the new and old parameters. These functions might be independent of $f$, e.g.,
$||\theta_{\text{new}} - \theta_{\text{old}}||_p$.
However, from a model debugging perspective, these definitions may not be relevant when the dimensionality of $\theta$ is even moderately large. 
We can also define importance scores that jointly consider unfairness and loss: $f(\theta) = f_{\text{train}}(\theta) + f_{\text{unf}}(\theta)$ would tell us how important a point is for unfairness and loss. 
The evaluation functions in Table~\ref{tab:eval} represent other popular group fairness metrics. We hope that future work consider finding DIVINE points with respect to robustness and privacy, which requires devising a new $f$.

\begin{table*}[htb]
\centering
\begin{tabular}{l|l}
\toprule
                Metric    &  Evaluation Function ($f$) \\ 
\midrule

Loss (e.g., wrt $\mathcal{D}_\text{train}$) & $f_{\text{train}}(\theta) = \mathcal{L}(\mathcal{D}_\text{train}) = \sum_{i}^{n} l(x_i,y_i; \theta)$ \\
\midrule
Equal Accuracy~\citep{berk2018fairness} & \small $f_\text{ea}(\theta) = f_\text{unf}(\theta) = \sum_{j \in \{-1,1\}} |P_{j,a,j} - P_{j,b,j}|$ \\
\midrule
Equal Opportunity~\citep{hardt2016} & $f_\text{eq}(\theta) =  P_{1,a,1} - P_{1,b,1}$ \\
\midrule
Equalized Odds~\citep{hardt2016} & $f_\text{eo}(\theta) =  \sum_{j \in \{-1,1\}}  |P_{1,a,j} - P_{1,b,j}|$ \\
\bottomrule
\end{tabular}
\caption{Various candidate evaluation functions $f$; note $P_{j,a,j} = P(\hat y = j |A = a, y = j)$}
\label{tab:eval}
\end{table*}
\textit{Practitioners should select the $f$ that captures the property, for which they wish to test their model.} If one wants to see the impact of datapoints on performance, $f_\text{loss}$ would be a good option. If one wants to see the impact of datapoints on fairness, $f_\text{unf}$ would be a good option.

\subsection{Diversity Function $\mathcal{R}$}
One main ingredient in our DIVINE point selection is a submodular function $\mathcal{R}$. This allows us to perform greedy selection when adding points to our DIVINE set. While other non-submodular diversity functions $\mathcal{R}$ are possible, they would to benefit from the ease of using greedy selection. Future work might benefit from more clever set selection methodologies. Through out the paper, we mostly let our diversity function be the sum-redundancy function, $\mathcal{R}_\text{SR}$. This function ensures that our selected points differ from each other, i.e., have low similarity~\citep{lin-bilmes-2011-class,libbrecht2018choosing}. In Appendix~\ref{add_experiments}, we demonstrate how the facility location function, $\mathcal{R}_\text{FL}$, and maximum mean discrepancy, $\mathcal{R}_\text{MMD}$, perform. While the equations for each $\mathcal{R}$ appears in Section~\ref{submod}, $\mathcal{R}_\text{FL}$ is the submodular facility location function~\citep{krause2014submodular} from the sensor-placement literature, selects points that minimize are similar to the most number of points in the entire dataset, and does not explicitly prohibit redundancy between the points selected.  When $\gamma$ is large and $\mathcal{R}_\text{MMD}$ is maximized, the prototypes of \citet{kim2016MMD} are recovered. If a practitioner does not mind some potential redundancy in the points selected and wants a set of points representative of the dataset, then $\mathcal{R}_\text{FL}$ may be suitable. On the other hand, $\mathcal{R}_\text{MMD}$ from~\citet{kim2016MMD} selects a set of points that summarize the entire dataset and penalizes similarity between the chosen points. When $\gamma$ is large and $\mathcal{R}_\text{MMD}$ is maximized, the prototypes of \citet{kim2016MMD} are recovered. If a practitioner wants  representativeness without much redundancy, $\mathcal{R}_\text{MMD}$ might suffice. All three submodular diversity functions are implemented in our package.

\subsection{Influence-Diversity Tradeoff $\gamma$}
In Section~\ref{experiments}, we introduce influence-diversity tradeoff curve. This illustrates how $\gamma$ controls how much influence to forgo in favor of diversity with respect to $\mathcal{R}$. While practitioners can select any $\gamma$ along the curve. We suggest two ways to pick $\gamma$. A practitioner could specify the maximum $\%$ of influence they are willing to sacrifice. A practitioner could also find the $\gamma$ that optimizes average pairwise distance between the $m$ selected points. By default our divine package favors the latter. Practitioners could also implement other $\gamma$ selection strategies as they see fit.

\subsection{Number of DIVINE points $m$}
Selecting the number of DIVINE points to find and visualize will be use case dependent. If the goal of finding DIVINE points is to display them as an explanation of model behavior, we suggest displaying at most $5$ points, which aligns with the number of cognitive chunks a user can handle at any given moment~\citep{doshi2017towards}. We suggest curating the size of the explanation to the needs of the stakeholders who will be analyzing the DIVINE points~\citep{bhatt2020explainable}. When selecting points to remove, a practitioner may consider checking our additivity assumption, i.e., see if removing a large number of low-value points at once does not affect other metrics of interest. We discuss how one would go about doing this analysis in Appendix~\ref{app:setup}. We find that, for small $m$, our additivity assumption is valid. Therefore, one might consider recalculating importance scores after removing a few batches of $m$. We hope future work develops additional methodology for choosing $m$.

\subsection{Using Our Code}
Our code is publicly available at \url{https://github.com/umangsbhatt/divine-release}, with a comprehensive README describing our implementation of DIVINE. We intend our code to be usable out of the box. We describe typical use-cases for our DIVINE codebase in the README.
Practitioners can use our codebase by importing necessary files (as shown in \textbf{tutorial.py}) into their own code. All use-cases are runnable from \textbf{tutorial.py}. More details are available in our README.

\section{Counterfactual Prediction}
\label{cfa_app}
We extend the work of \citet{bhatt2020counterfactual} to be compatible with our approach to finding training point importance. We restrict ourselves to standard binary classification tasks, where our goal is to find a parameter $\theta \in \Theta$ such that $\theta$ learns a mapping between inputs $x \in \mathbb{R}^d$ and labels $y \in \{-1,1\}$. 
Given a training dataset $\mathcal{D}^{n} = \{ (x_{i},y_{i})\}^{n}_{i=1}$ from some underlying, unknown distribution $\mathcal{D}$ and a non-negative loss function $l$, our goal is to learn $\theta$ that minimizes training error yet performs well on \textit{unseen} test data. The expected loss of $\theta$ is given by:
$R(\theta) = \mathbb{E}_\mathcal{D} \left[ l\left(x, y; \theta \right)\right]$.
Since we do not know $\mathcal{D}$, we calculate the average loss $\hat R$ over the training dataset, $\hat R(\theta) = \frac{1}{n}\sum_{i=1}^n l\left(x_i, y_i; \theta \right)$.
In an ERM setup, we find the optimal parameter $\hat \theta$ parameterized by $\theta$ as follows:
$\hat \theta = \argmin_{\theta \in \Theta} \sum_{i=1}^n l\left(x_i, y_i; \theta \right)$.

Given a point $z$ and its predicted label $\rho(\hat \theta,z)$,
we want to find  alternate parameters $\theta^\prime$ such that we minimize empirical risk \textit{with the condition} that the predicted label of $z$ if flipped: $\rho(\hat \theta,z) \neq \rho(\theta^\prime,z)$.
We find an alternative classifier via $\theta^\prime_z = \argmin_{\theta \in \Theta}   \sum_{i=1}^n l\left(x_i, y_i; \theta \right) \; \;
        \textrm{s.t.} \; \rho(\hat \theta,z) \neq \rho( \theta, z) $.
We can view this problem as optimizing over $\Theta$ with an added constraint, or as optimizing over a subspace, $\Theta^\prime_{z} = \{\theta \in \Theta : \rho(\hat \theta,z) \neq \rho( \theta, z)\}$; note $\Theta^\prime_{z} \subseteq \Theta$. 
If $f$ is training loss, this quantity tells us how much our loss suffers when we introduce a constraint to conflict the predictions of $\hat \theta$ and $\theta^\prime_z$ on a point $z$. The importance of the $i$-th training point $z$ is given by: $I_i^{\text{CFP}} = f(\theta^\prime_{z}) - f(\hat \theta)$. 

The expected loss of $\theta^\prime$ is given by:
$R(\theta^\prime) = \mathbb{E}_\mathcal{D} \left[ l\left(x, y; \theta^\prime \right)\right]$. 
The average loss of $\theta^\prime$ over the training data is given by 
$\hat R(\theta^\prime) = \frac{1}{n}\sum_{i=1}^n l\left(x_i, y_i; \theta^\prime \right)$.
Let $C(\theta^\prime,\hat \theta) = R(\theta^\prime) - R(\hat \theta)$ tell us how much our loss suffers when we introduce a constraint to conflict the predictions of $\theta^\prime$ and $\hat \theta$ on a point $z$. For ease of reading, we let $C_{z} = C(\theta^\prime,\hat \theta)$. Since we do not know $\mathcal{D}$, we calculate an empirical variant over our training dataset: $\hat C(\theta^\prime,\hat \theta) = \hat R(\theta^\prime) - \hat R(\hat \theta)$.
We write the \textit{empirical extra loss} $\hat C_{z}$ as the difference in training loss between $\hat \theta$ and $\theta^\prime$: we call this the \textit{counterfactual prediction} (CFP), denoted by ${C}_{z}$.
For any point $z$ and a given parameter space $\Theta$, we can find its corresponding counterfactual prediction. ``Counterfactual'' is not used in the causal sense of \citet{pearl2009causality} but captures what would happen to loss (or any $f$) if we were to constrain our objective to alter the prediction of $z$.

\subsection{CFP in Prior Work}
\citet{breiman2001statistical} noted that there can exist multiple hypotheses that fit a training dataset equally well, leading to different stories about the relationship between the input features and output response. There are a few recent works that relate to our formulation. 
Firstly, \citet{rashomon} defines the empirical $\epsilon$-Rashomon set is defined as:
$S_\epsilon = \{\theta \in \Theta: \hat R(\theta) \leq \hat R(\hat \theta) + \epsilon \}$.
$S_\epsilon$ can be seen as the set of all classifiers in $\Theta$ that have an average loss no more than $\epsilon$ greater than the average loss of $\hat \theta$. \citet{pred_mult} study predictive multiplicity within the Rashomon set (calling it the $\epsilon$-level set): they define a metric called \textit{ambiguity}, which is the proportion of training data points whose $\theta^\prime \in S_\epsilon$. To calculate ambiguity, they find $\theta^\prime$ via a mixed integer program (MIP) for each $z$.
While these works study how to deal with varying predictions in $S_\epsilon$, we essentially solve a dual problem where we want to find the minimum $\epsilon$ such that the empirical $S_\epsilon$ contains at least \textbf{one} model with different predictions for $z$. More concretely, we want to find $\epsilon_i > 0$, where 
$\epsilon_i = \text{min} \;  \epsilon \; \; \textrm{s.t.}\; \exists \theta \in S_\epsilon: \rho(\theta,z_i) \neq \rho(\hat \theta,z_i)$.
\citet{letham2016prediction} looks across $S_\epsilon$ to identify classifiers which have maximally different predictions (similar to \textit{discrepancy} defined in \cite{pred_mult}).
We do something similar but different: we ask how much does your average loss need to suffer in order to change the prediction of a test point.

\citet{agnostic} used $C_{z}$ to create a \textit{selection function} for rejection option classification: they call $C_{z}$ the disbelief index. They consider  Specifically, if the $C_{z} > \Delta$, then they proceed to predict on $z$ using $\hat \theta$. If $C_{z} \leq \Delta$, then they abstain from providing a prediction for $z$. They choose $\Delta = \sigma(n, \delta/4, d) = 4  \sqrt{\frac{2d\left(\ln \frac{2ne}{d}\right) + \ln \frac{2}{\delta}}{n}}$, where $\sigma$ is the slack of a uniform deviation bound in terms of the training set size $n$, a confidence parameter $\delta$, and the VC-dimension $d$ of $\Theta$.
Using a weighted SVM, they find $\theta^\prime$ by adding $z^\prime = (z, -\rho(\hat \theta,z))$ to the training data and then upweight the penalty of misclassifying $z^\prime$ (penalty is set to ten times the weight of all other training points combined). Since $C_{z}$ is a noisy statistic and depends heavily on the $n$ samples chosen from $\mathcal{D}$, they use bootstrap sampling and then take the median of all measurements as the final value of the disbelief index, which is closely related to the disagreement coefficient from \citet{hanneke2009theoretical}.

Earlier, \citet{dasgupta2008general} used an ERM oracle with an example-based constraint in the context of active learning. In a setup similar to \citet{agnostic} and ours, \citet{dasgupta2008general} decides to request a label for $z$ if $C_{z} \leq \Delta$. They also select $\Delta$ to be in terms of the empirical errors of both classifiers, $n$, $\delta$, and $d$.

\begin{table*}[htb]
    \centering
    \begin{tabular}{c|c|c}
    \toprule
         & High Density & Low Density \\
         \midrule
        Low $C_{z}$  & $\rho(\hat \theta,z)$ is uncertain in its prediction for $z$ & $\rho(\hat \theta,z)$ is uncertain in its prediction for $z$ \\
        High $C_{z}$  & $\rho(\hat \theta,z)$ is certain in its prediction for $z$ & $z$ is a potential outlier\\
    \bottomrule
    \end{tabular}
    \caption{The Interplay between CFP $C_{z}$ and the data density around $z$}
    \label{tab:density}
\end{table*}

\subsection{Intuition}
\begin{figure*}[htb]
\centering
    \begin{subfigure}[b]{0.245\linewidth}
        \includegraphics[width=\textwidth]{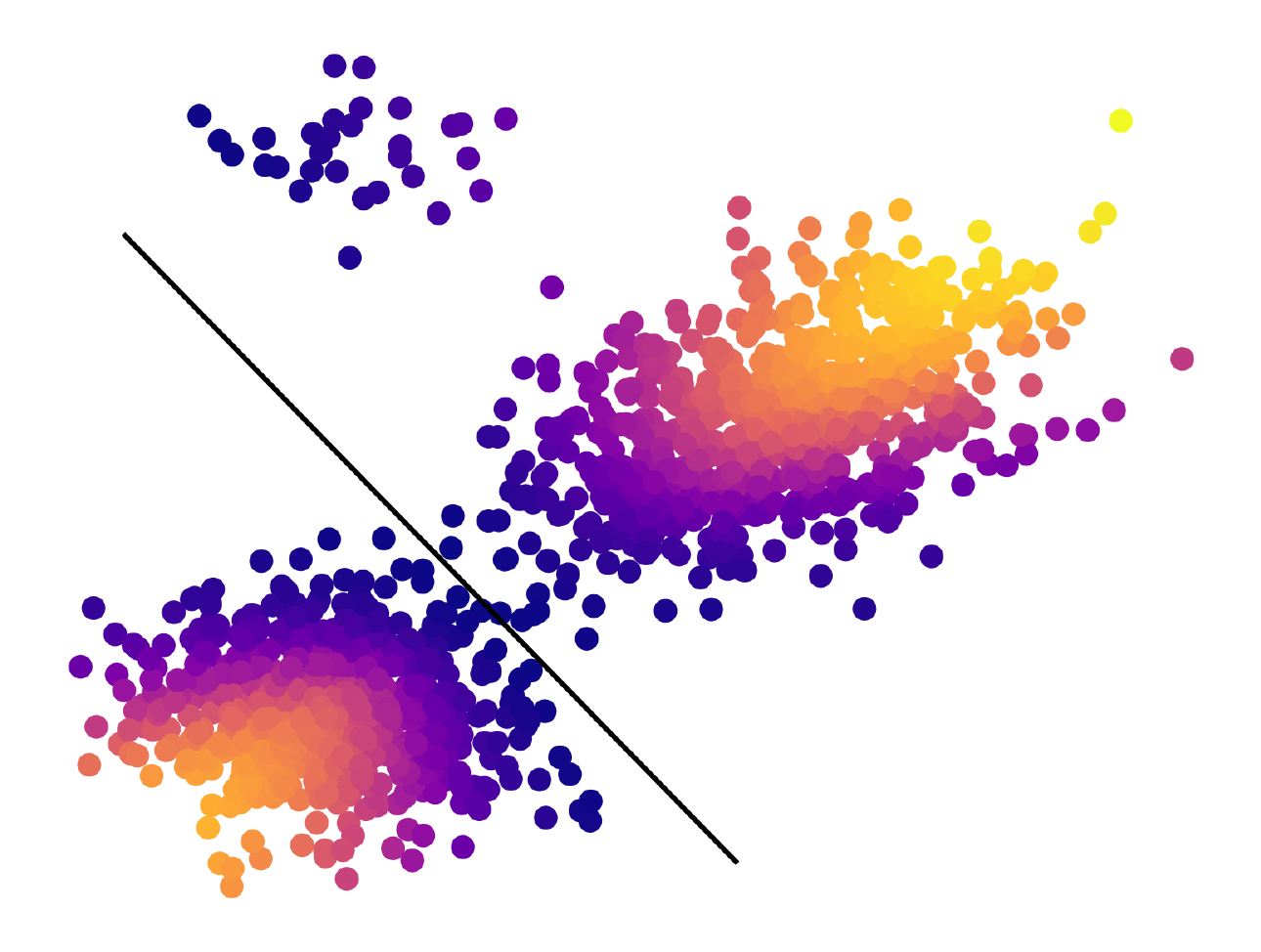}
        \caption{CFP with $f_\text{loss}$}
    \end{subfigure}
    \begin{subfigure}[b]{0.245\linewidth}    
        \includegraphics[width=\textwidth]{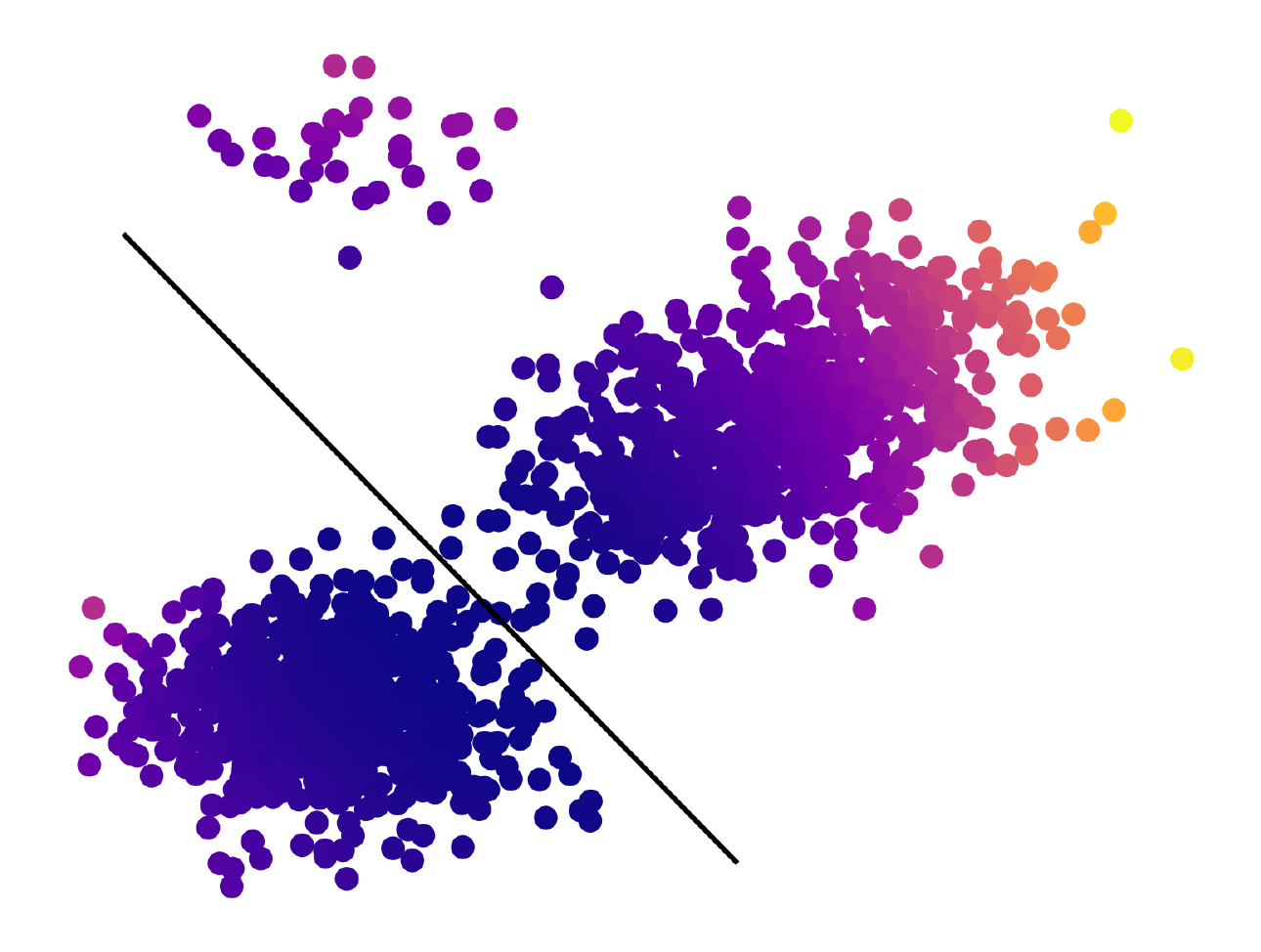}
        \caption{IF with $f_\text{loss}$}
    \end{subfigure}
    \begin{subfigure}[b]{0.245\linewidth}
          \includegraphics[width=\textwidth]{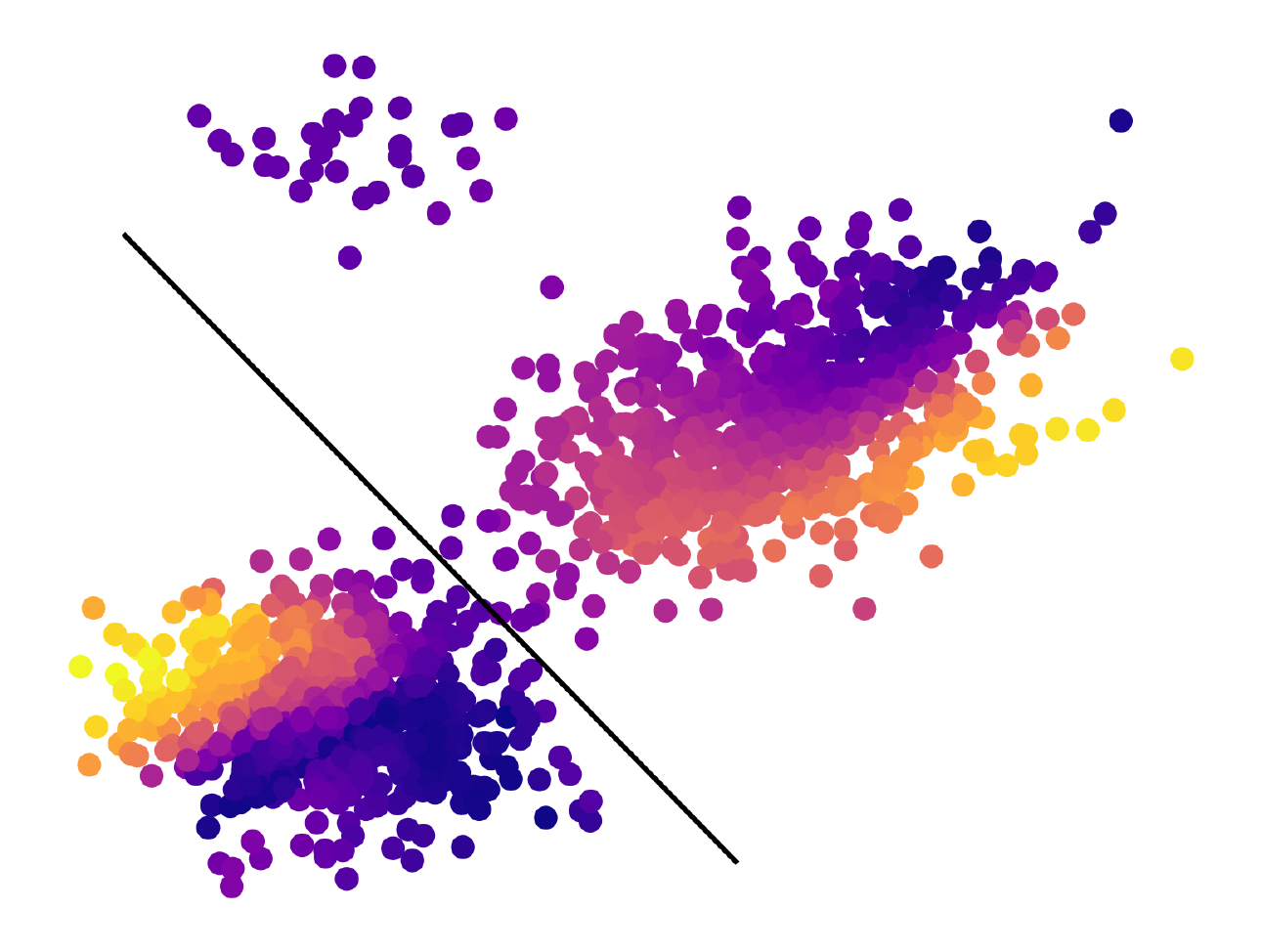}
            \caption{CFP with $f_\text{unf}$}
    \end{subfigure}%
    \begin{subfigure}[b]{0.245\linewidth}        
            \includegraphics[width=\textwidth]{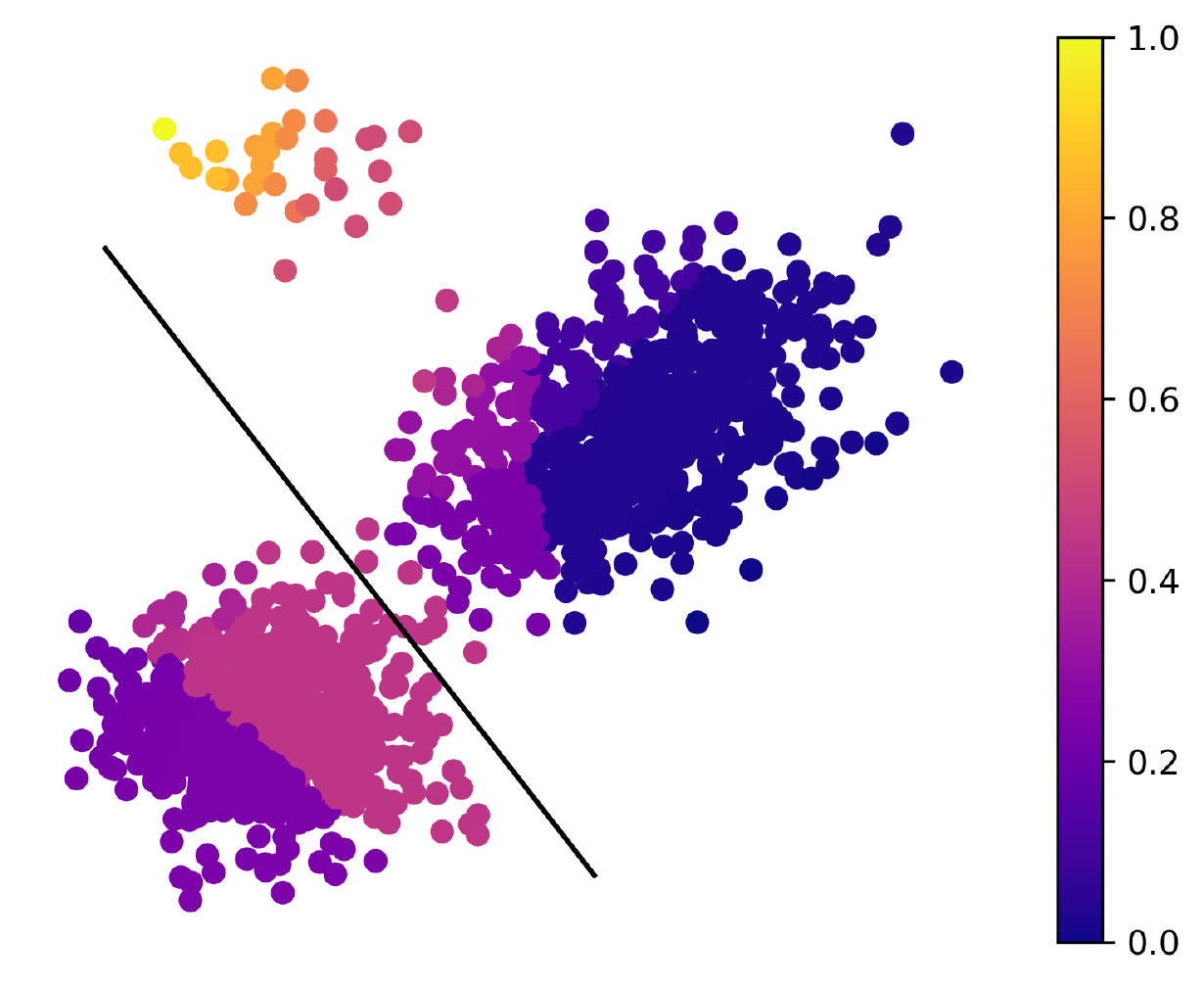}
            \caption{IF with $f_\text{unf}$}
    \end{subfigure}
    \caption{Comparing various training point importance methods based on how they assign influence; we normalize influence $I_i \in [0,1]$ when we visualize.}\label{fig:all_sub_heat}
\end{figure*}

In Figure~\ref{fig:all_sub_heat}, we compare CFP to IF for $f_\text{loss}$ and $f_\text{unf}$: CFP respects the data density more than IF, which correlates heavily with the distance to decision boundary. When visualizing, we normalize importance such that $I_i \in [0,1]$. We now provide intuition behind CFP.  Recall we simply care about the absolute difference in loss between the training loss of the ERM parameter $\hat R(\hat \theta)$ and the training loss of the constrained parameter $\hat R(\theta^\prime)$, where the constraint mandates that $\rho(\hat \theta,z) \neq \rho(\theta^\prime,z)$. We denote this difference by $C_{z}$.

If $C_{z}$ is large, the parameters must change a lot in order to fit an opposite label for $z$; therefore, we see model performance drop and can be confident that $\hat \theta$ correctly classified $z$, given $\mathcal{D}^n$ and $\Theta$: \citet{dasgupta2008general} would not request a label for $z$ and \citet{agnostic} would accept $\rho(\hat \theta,z)$. 
If $C_{z}$ is small, the parameters learned are similar and it was easy to fit an opposite label for $z$, so we cannot be sure of the label for $z$: \citet{dasgupta2008general} would request a label for $z$ and \citet{agnostic} would reject $\rho(\hat \theta,z)$. 

We expect $C_{z}$ to change rapidly based on the data density. We can expect $C_{z}$ to be high in dense regions. In Table~\ref{tab:density}, we discuss the interplay between $C_{z}$ and data density.
If $C_{z}$ is high and we are in a high density region, then we know $\rho(\hat \theta,z)$ is correct and certain. 
If $C_{z}$ is low and we are in a high density region, then we know $\rho(\hat \theta,z)$ is uncertain: we may have the incorrect label or noise in our covariates. This happens in regions of high class overlap (high aleatoric uncertainty). 
If $C_{z}$ is low and we are in a low density region, then we know $\rho(\hat \theta,z)$ is uncertain: we may have an outlier or simply high epistemic uncertainty. 
If $C_{z}$ is high and we are in a low density region, $z$ is a potential outlier, since $\rho(\hat \theta,z)$ changed considerably to alter the prediction of $z$.


\subsection{Connecting CFP to Neural Networks}
\label{infConn}

For many complex models (i.e., large neural networks), re-computing the model with an added constraint is computationally expensive. As such, we propose an approximate version of  using influence functions to find the set of perturbed parameters. We want to estimate the effect of training with a flipped label on the model parameters. 
In a weighted ERM setup, we find $\hat \theta$ that minimizes empirical risk:
$$
R(\theta) = \sum_{i=1}^{n} w_i l(x_i, y_i; \theta)
$$
To weight every training data point equally, we set $w_i = \frac{1}{n}$ for all $i$.  Like~\citet{koh2017understanding}, we assume that $R$ is twice differentiable and convex in $\theta$. As such, we assume the Hessian exists and is given by:
$$
H_\theta = \nabla^2 R(\theta) = \sum_{i=1}^{n} w_i \nabla_{\theta}^2  l(x_i, y_i; \theta)
$$
We assume $H_\theta$ is positive definite to guarantee the existence of $H_\theta^{-1}$. \citet{koh2017understanding} define the perturbed parameters obtained when upweighting a single training point, $z = \{x, y\}$ as follows:
$$
\hat \theta_{\epsilon, z} = \argmin_{\theta \in \Theta} \{R(\theta) + \epsilon l(x, y; \theta)\}
$$
Assuming $y \in \{-1,1\}$, let $z^\prime = \{x, -y\}$ be the label flipped point. We can analogously get the following:
$$
\hat \theta_{\epsilon, z^\prime, z} = \argmin_{\theta \in \Theta} \{R(\theta) + \epsilon l(x^\prime, y^\prime; \theta) - \epsilon l(x, y; \theta)\}
$$
\citet{koh2017understanding} study the affect of an input perturbation on the models parameters. They define this as follows:
$$
\mathcal{I}_{\mathrm{up}, \operatorname{param}}\left(\vec{z}\right) = \frac{d\hat \theta_{\epsilon, z}}{d\epsilon} \bigg |_{\epsilon = 0} = - H_{\hat \theta}^{-1} \nabla_{\theta} l\left(x, y; \hat \theta \right)
$$

We can define the effect of the label flipped point as:
$$
\frac{d\hat \theta_{\epsilon, z^\prime, z}}{d\epsilon} \bigg |_{\epsilon = 0} = 
- H_{\hat \theta}^{-1} \left ( \nabla_{\theta} l\left(x, -y; \hat \theta \right) - \nabla_{\theta} l\left(x, y; \hat \theta \right)\right)
$$

Thus, the new parameters are approximately given by:
\begin{align*}
\hat \theta_{\epsilon, z^\prime, z} &\approx \hat \theta + \frac{d\hat \theta_{\epsilon, z^\prime, z}}{d\epsilon} |_{\epsilon = 0} \; \epsilon  \\
& \approx
\hat \theta - H_{\hat \theta}^{-1} \left ( \nabla_{\theta} l\left(x, -y; \hat \theta \right) - \nabla_{\theta} l\left(x, y; \hat \theta \right)\right) \epsilon
\end{align*}
In the case of logistic regression, we can approximate the CFP parameters in closed form. We know that for logistic regression $p(y|x) = \sigma(y\theta^\intercal x)$ where $\sigma(a) = \frac{1}{1 + \exp(a)}$. The loss is given by $l(x, y; \theta) = \log(1 + \exp(y\theta^\intercal x))$ and its derivative is given by $\nabla_{\theta} l(x, y; \theta) = -\sigma(-y\theta^\intercal x)yx$. 

The difference in the derivatives with respect to the loss for the flipped-label point and the original point can be re-written as $yx(\sigma(y\theta^\intercal x))+ \sigma(-y\theta^\intercal x))$. Since we know $\sigma(a) + \sigma(-a) = 1$, we can write the updated parameters after flipping the label of $z$ as:
$$
\hat \theta_{\epsilon, z^\prime, z} \approx 
\hat \theta - H_{\hat \theta}^{-1}yx \epsilon
$$

\clearpage
\section{Experimental Setup}
\label{app:setup}

\subsection{Dataset Metadata}
\label{metadata}

We employ 6 datasets in our experiments, 4 tabular and 2 image. All are publicly available, with details given in Table~\ref{tab:appendix_datasets}.  For all datasets, we use a $70\%$ train, $20\%$ validation, and $10\%$ test split.
\begin{table}[h]
\centering
\caption{Summary of datasets used in our experiments. (*)We use a 7 feature version of COMPAS; however, other versions exist.}
\label{tab:appendix_datasets}
\begin{tabular}{@{}cccccc@{}}
\toprule
Name       & Targets     & Input Type                & \# Features      & \# Total Samples\\ \midrule
LSAT       & Continuous  & Continuous \& Categorical & $3$            & $21791$   \\
COMPAS     & Binary      & Continuous \& Categorical & $7^{*}$        & $5278$    \\
Adult & Binary  & Continuous \& Categorical              & $11$           & $45222$    \\
Bank   & Binary      & Continuous \& Categorical & $21$           & $41188$ \\
MNIST      & Categorical & Image (greyscale)         & $28{\times}28$ & $70000$   \\ \
FashionMNIST      & Categorical & Image (greyscale)         & $28{\times}28$ & $70000$  \\ \bottomrule
\end{tabular}
\end{table}

We use the LSAT loading script from \citet{Cole2019AvoidingRV}'s github page. The raw data can be downloaded from \url{https://raw.githubusercontent.com/throwaway20190523/MonotonicFairness/master/data/law_school_cf_test.csv} and \url{https://raw.githubusercontent.com/throwaway20190523/MonotonicFairness/master/data/law_school_cf_train.csv}. We let ``sex'' be our protected attribute and drop ``race'' from the dataset when running our experiments. Features used are undergraduate grade point average, LSAT score, and sex. The predicted label is first year law school performance.

For the COMPAS criminal recidivism prediction dataset we use a modified version of \citet{zafar2017fairness}'s loading and pre-processing script. It can be found at \url{https://github.com/mbilalzafar/fair-classification/blob/master/disparate_mistreatment/propublica_compas_data_demo/load_compas_data.py}. We add an additional feature: ``days served'' which we compute as the difference, measured in days, between the ``c\_jail\_in'' and ``c\_jail\_out'' variables. The raw data is found at \url{https://github.com/propublica/compas-analysis/blob/master/compas-scores-two-years.csv}.  We let ``race'' be our protected attribute. Other features used are age, sex, charge degree (felony or misdemeanor), and priors count. The predicted label is recidivism within 2 years.

The adult dataset~\citep{Dua:2019} can be obtained from and is described in detail at \url{https://archive.ics.uci.edu/ml/datasets/adult}. The features we used are age, work class, education, education number, marital status, occupation, relationship, capital gain, capital loss, hours per week, and native country. More details are available at the link. We let ``sex'' be our protected attribute. The predicted label is whether the person makes more than 50K a year.

The bank marketing dataset~\citep{Dua:2019} can be obtained from and is described in detail at \url{https://archive.ics.uci.edu/ml/datasets/Bank+Marketing}. The features we used are described in detail at the link. We let ``age'' be our protected attribute. The predicted label is whether a client will subscribe to a term deposit. 

The MNIST handwritten digit image dataset~\citep{lecun1998mnist} can be obtained from \url{http://yann.lecun.com/exdb/mnist/}. 

The FashionMNIST image dataset~\citep{xiao2017fashion} can be obtained from \url{https://github.com/zalandoresearch/fashion-mnist}.

\begin{figure}[]
\centering
    \begin{subfigure}[b]{0.49\linewidth}
            \centering
            \includegraphics[width=\linewidth]{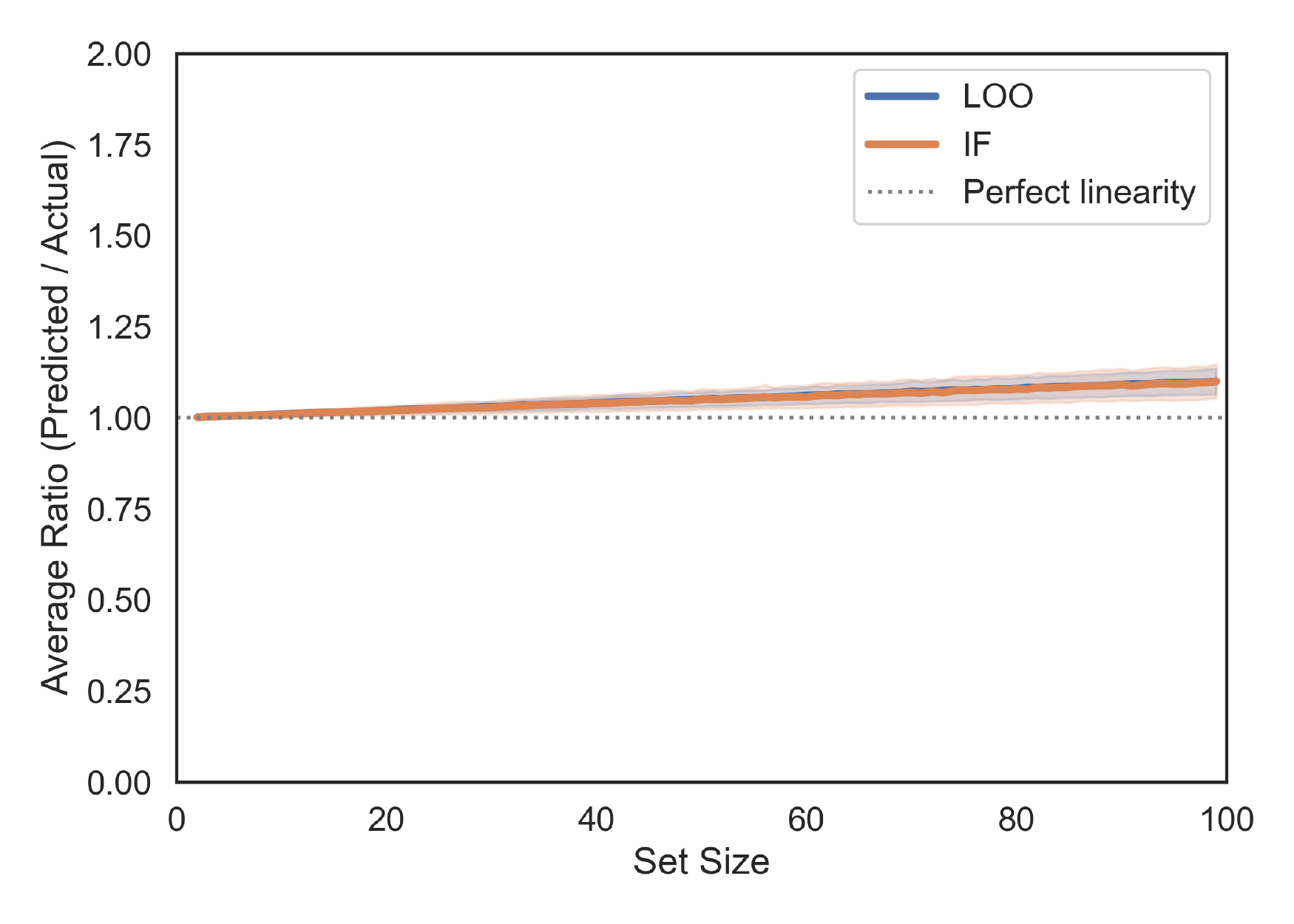}
            \caption{LSAT}
            \label{fig:lsat_lin}
    \end{subfigure}
    \begin{subfigure}[b]{0.49\linewidth}    
            \includegraphics[width=\linewidth]{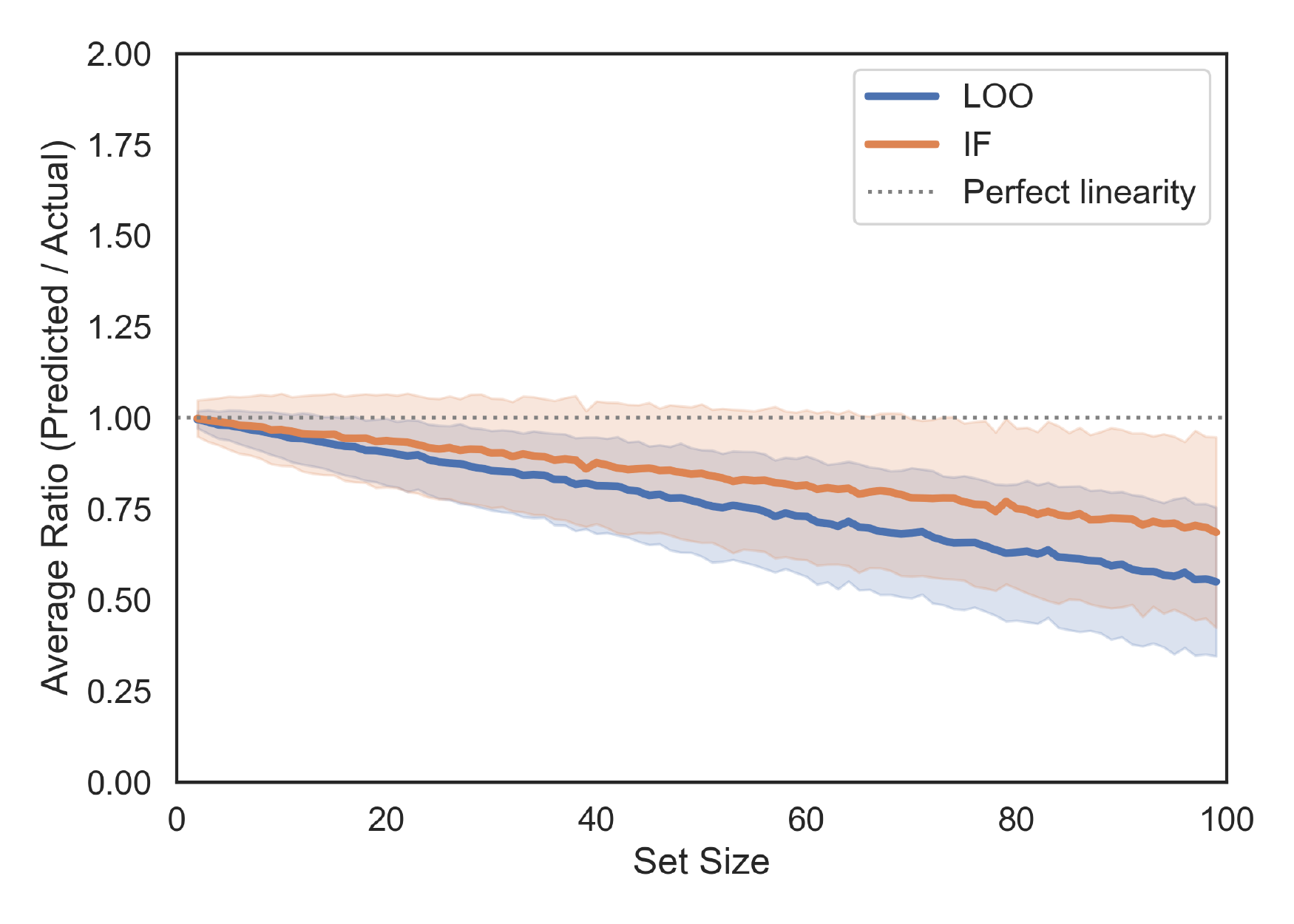}
            \caption{COMPAS}
            \label{fig:adult_lin}
    \end{subfigure}%
    \caption{We show $\frac{I_i + I_j}{I_{\{i,j\}}}$ when $f$ is taken to be equal accuracy.}\label{fig:linearity}
\end{figure}

\subsection{Models}
\label{tradeoff}
In Section~\ref{experiments}, we primarily use logistic regression for our tabular data experiments. For all tabular datasets, we append an intercept to the input features before learning our parameters, $\theta \in \mathbb{R}^d$: this is customary in such settings, i.e.,~\citet{zafar2017fairness} has a similar set up. We learn classifier's parameters using \textbf{scipy.optimize}, using the SLSQP (Sequential Least SQuares Programming) solver. For image datasets, we use \textbf{tensorflow} to learn a three-layered multilayer perceptron (for MNIST) and a three-layered convolutional neural network (for FashionMNIST). We then leverage our DIVINE codebase (Section~\ref{guide}) to calculate the DIVINE points for each model.

\subsection{Additivity implies Modularity}
\label{linearity}
We next comment on the modularity of $\mathcal{I}(\mathcal{S})$. If $I_i > 0$ holds $\forall i$, then $\mathcal{I}(\mathcal{S})$ is monotone. To show the modularity of $\mathcal{I}(\mathcal{S})$, it suffices to show each importance measure with a selected $f$, is additive, which implies importance scores are linear. 
When $f$ is loss, \citet{koh2019accuracy} find that influence functions are approximately linear, $I_{ij} \approx I_{i} + I_{j}$. Since we can recast counterfactual prediction in terms of influence functions in Appendix~\ref{cfa_app}, we know importance scores from counterfactual prediction are approximately linear. 
Furthermore, irrespective of $f$, Shapley values, by construct, satisfy linearity: see~\citep{shapley52,ghorbani2019data} for a thorough treatment. 
While we know that $\mathcal{I}(\mathcal{S})$ will be modular when $f$ is a function of loss, we consider linearity using influence functions with $f_\text{unf}$.
In Figure~\ref{fig:linearity}, we plot how a linearity approximation of importance (calculated using $f_\text{unf}$) performs as we increase the number of points we remove from the dataset. The average difference between the predicted importance score $I_i + I_j$, and the actual importance score
$I_{\{i,j\}}$, over 1000 samples of $m$ sized sets. On LSAT, shown in Figure~\ref{fig:lsat_lin}, all three scoring maintain linearity as set sizes increases, implying that $\mathcal{I}(\mathcal{S})$ is modular. Linearity is also satisfied on Adult (Figure~\ref{fig:adult_lin}) with LOO scores. For IF, importance score is linear for small values of $m$. As $m$ increases, our approximation no longer maintains linearity. If we keep $m$ relatively small, we can use our linearity approximation, as we desire simple explanations with few cognitive chunks~\citep{doshi2017towards}. However, as we can assume additivity with larger set sizes when we use LOO, we expect it to perform better overall in identifying large sets of important points on high-dimensional data.
Practitioners might find using these graphs when deciding how large to make $m$; if $m$ is too large, then DIVINE points for unfairness might not be trustworthy.


\clearpage
\section{Additional Experiments}
\label{add_experiments}

\subsection{Analyzing Different Diversity Functions}
\begin{figure*}[htb]
\centering
    \begin{subfigure}[b]{0.245\linewidth}        
            \includegraphics[width=\textwidth]{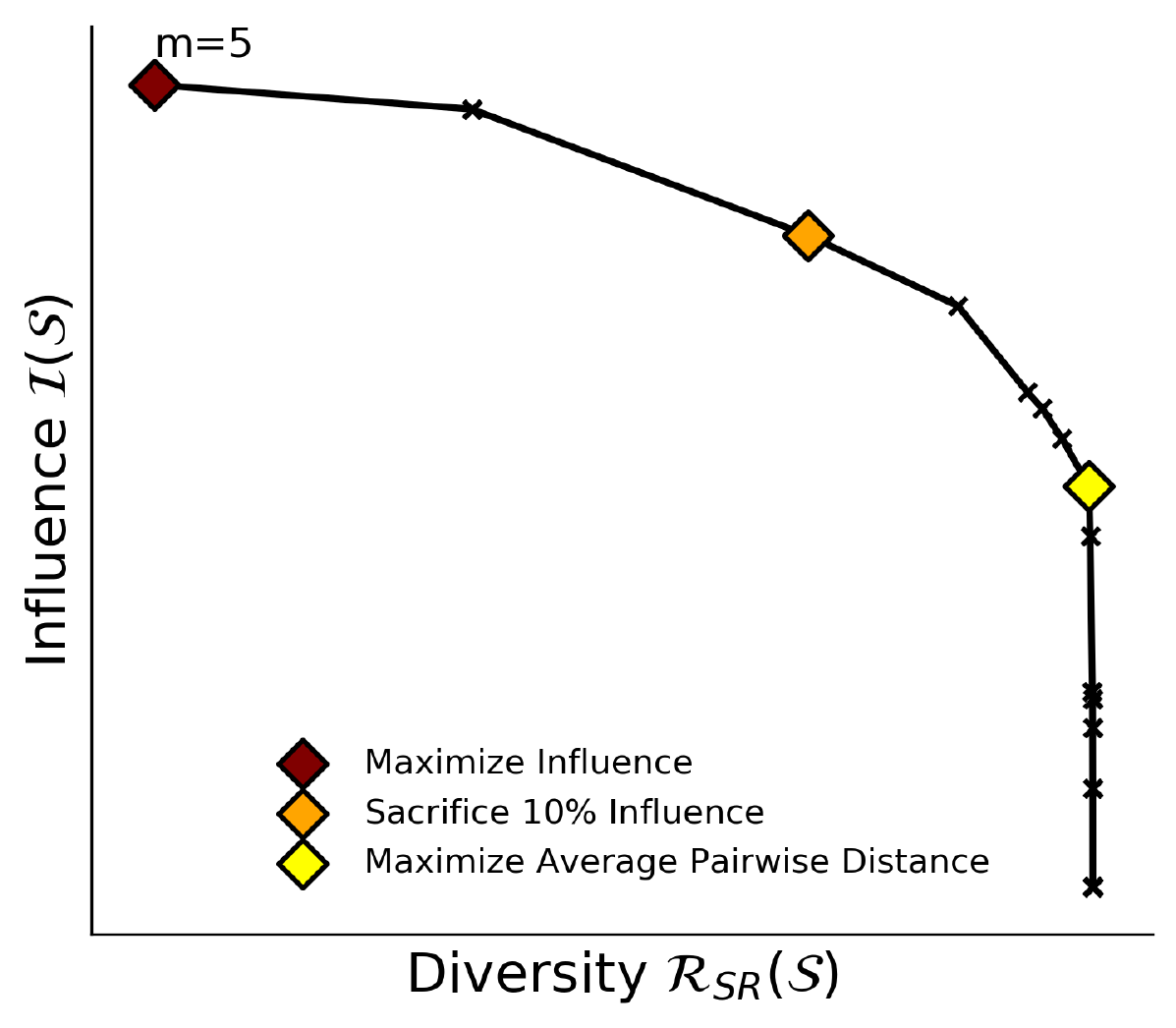}
            \caption{$\mathcal{I}$-$\mathcal{R}$ Tradeoff}

    \end{subfigure}
    \begin{subfigure}[b]{0.245\linewidth}
        \includegraphics[width=\textwidth]{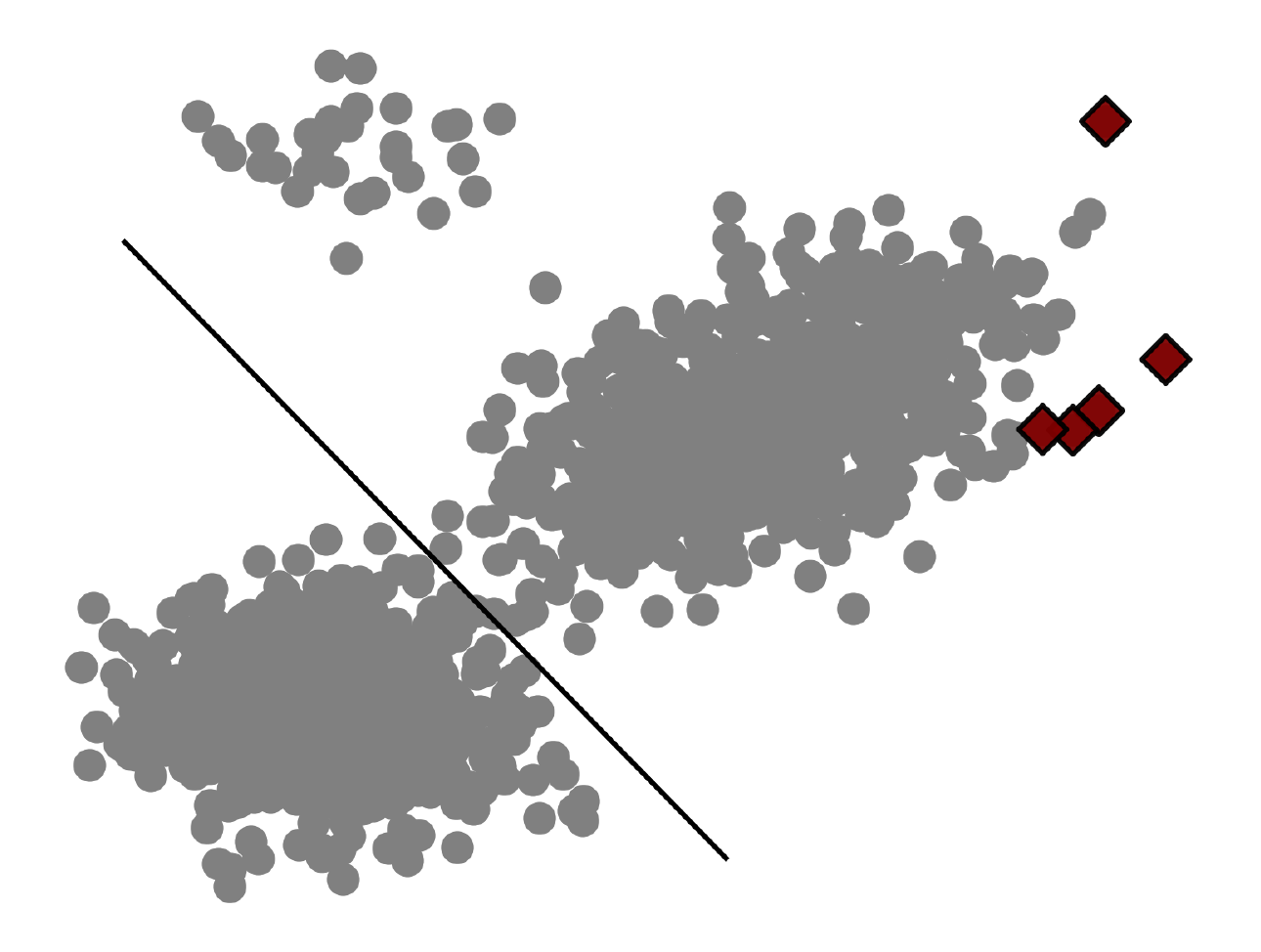}
        \caption{DIVINE $\gamma = 0$}
  
    \end{subfigure}
    \begin{subfigure}[b]{0.245\linewidth}    
        \includegraphics[width=\textwidth]{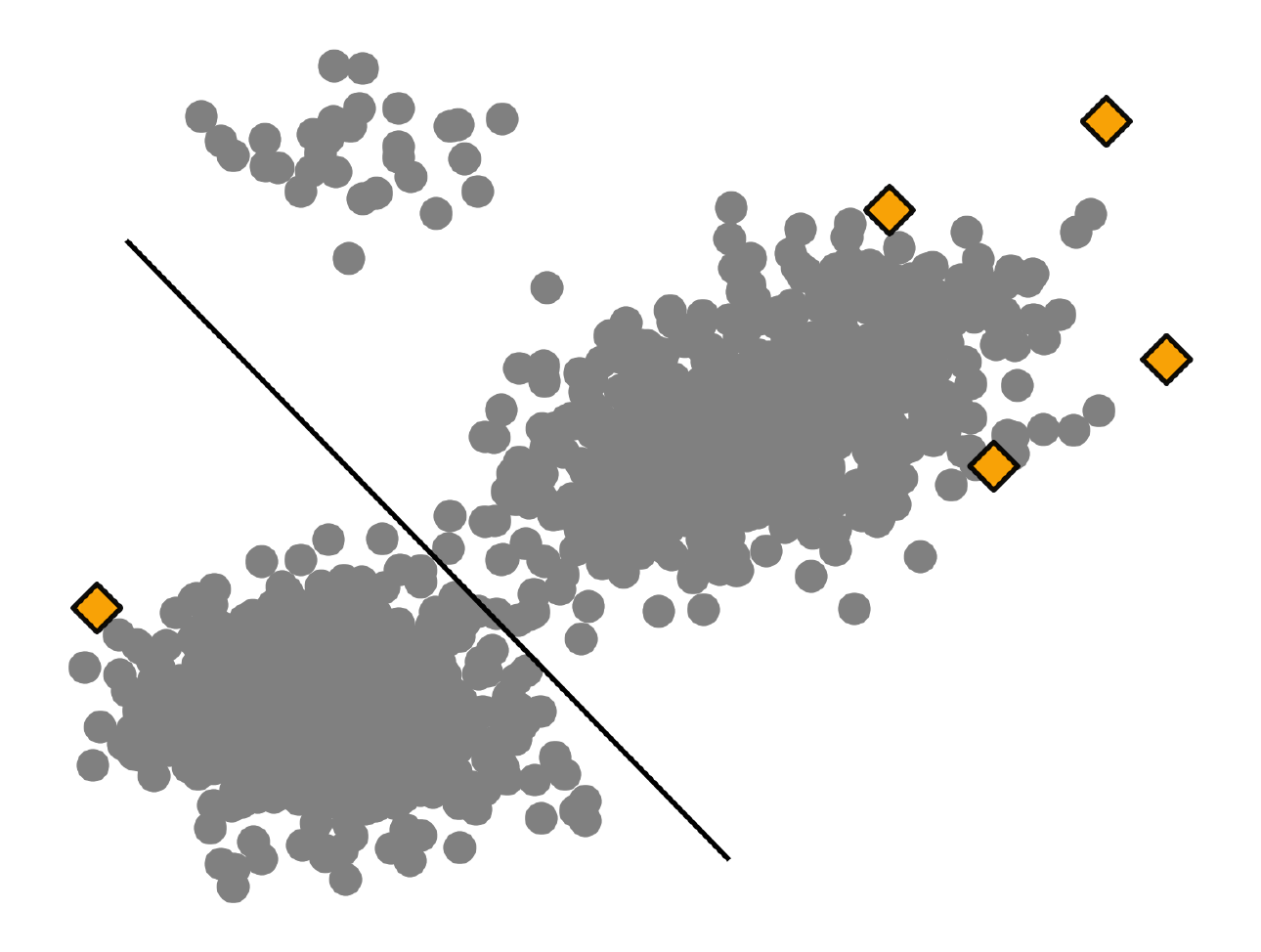}
        \caption{DIVINE $\gamma = 161$}
    
    \end{subfigure}
    \begin{subfigure}[b]{0.245\linewidth}
        \includegraphics[width=\textwidth]{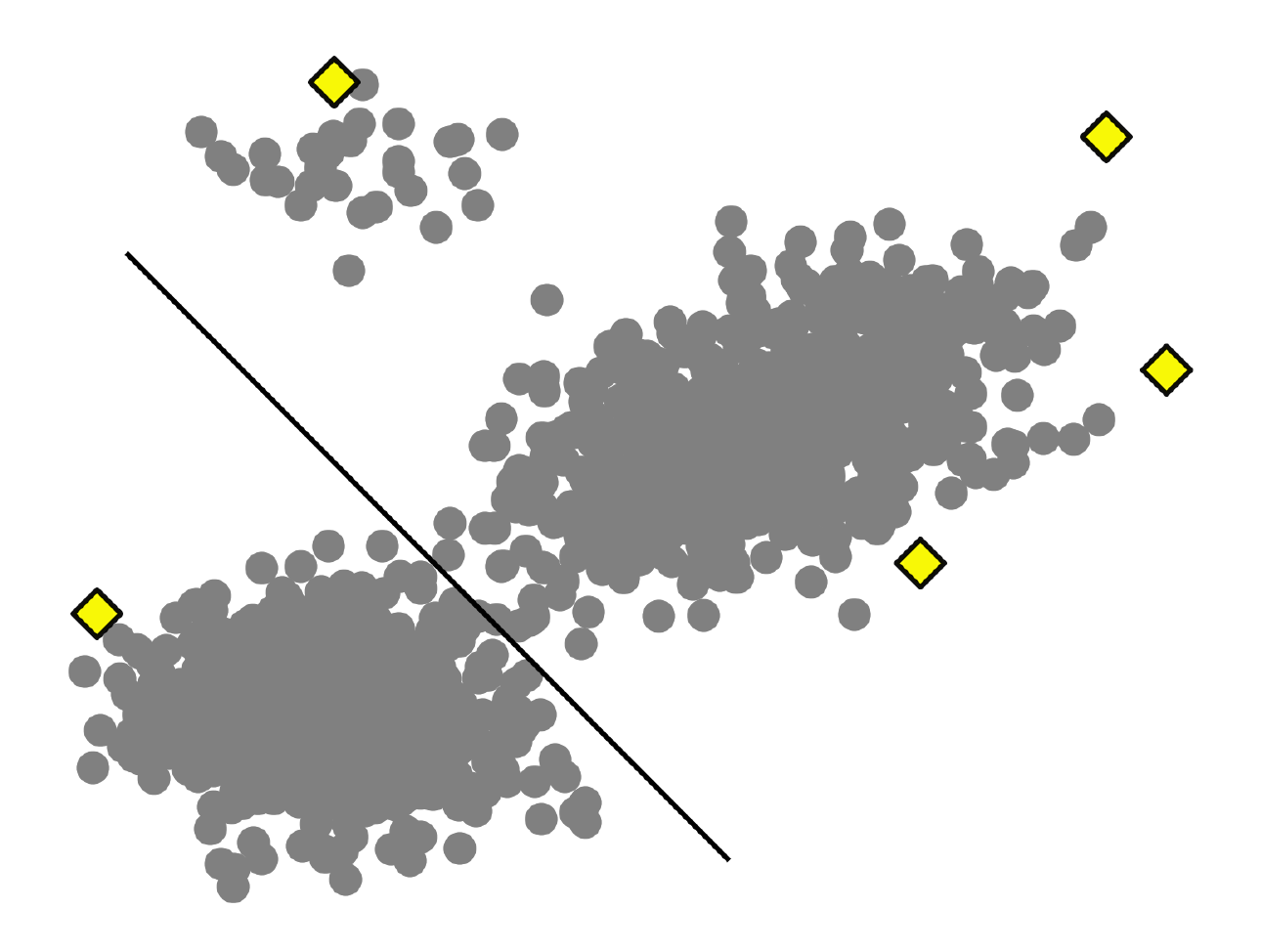}
            \caption{DIVINE $\gamma = 434$}
       
    \end{subfigure}%
    \caption{\small $\mathcal{R}_\text{SR}$ -- note this figure is the same as Figure~\ref{fig:all_subplots} from the main paper
     }\label{fig:all_subplots_copy}
\end{figure*}

\begin{figure*}[htb]
\vspace{-0.5cm}
\centering
    \begin{subfigure}[b]{0.245\linewidth}        
            \includegraphics[width=\textwidth]{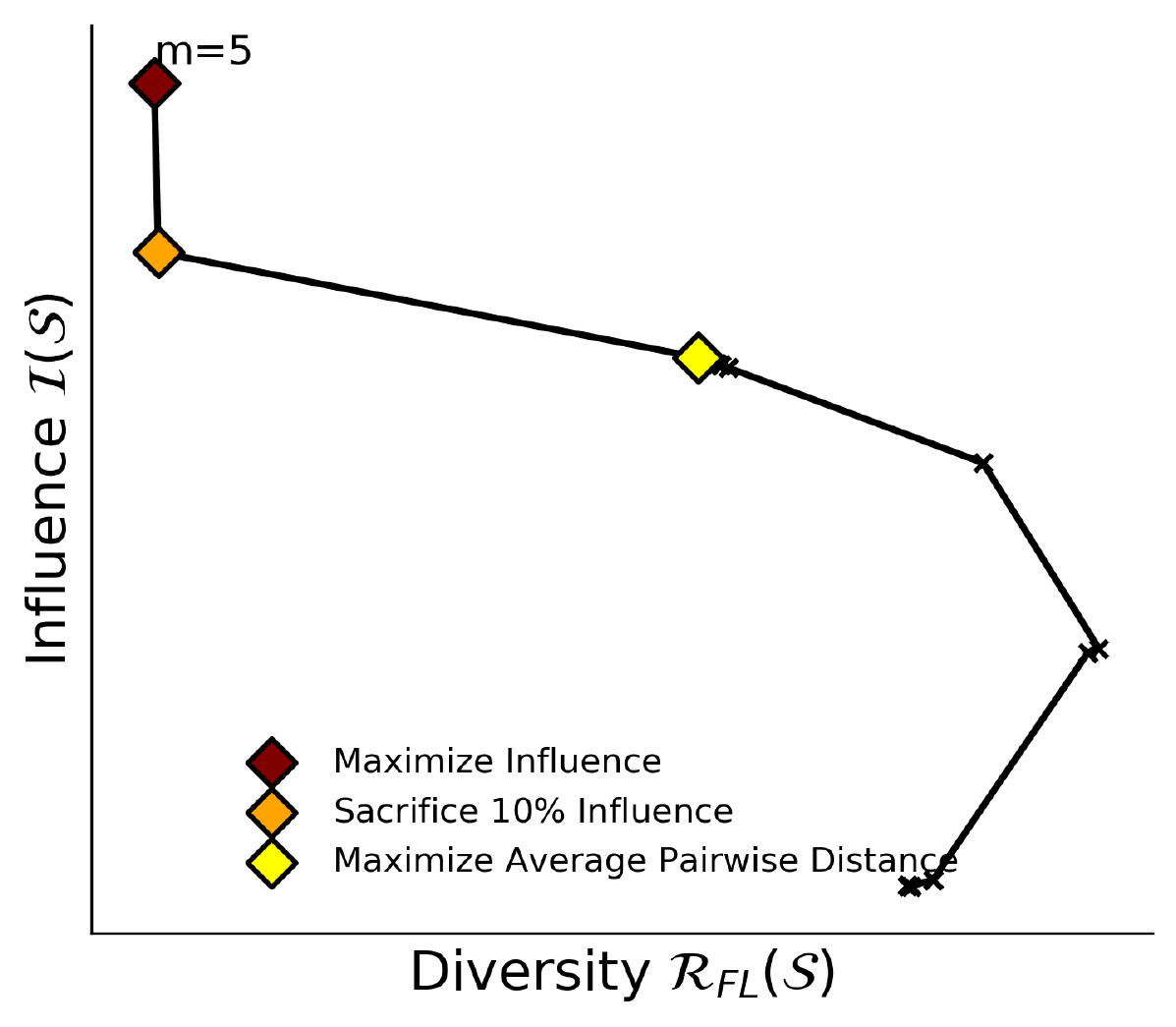}
            \caption{$\mathcal{I}$-$\mathcal{R}$ Tradeoff}
    \end{subfigure}
    \begin{subfigure}[b]{0.245\linewidth}
        \includegraphics[width=\textwidth]{figures/pdfs_app/fig_two_if_train_gam0.pdf}
        \caption{DIVINE $\gamma = 0$}
    \end{subfigure}
    \begin{subfigure}[b]{0.245\linewidth}    
        \includegraphics[width=\textwidth]{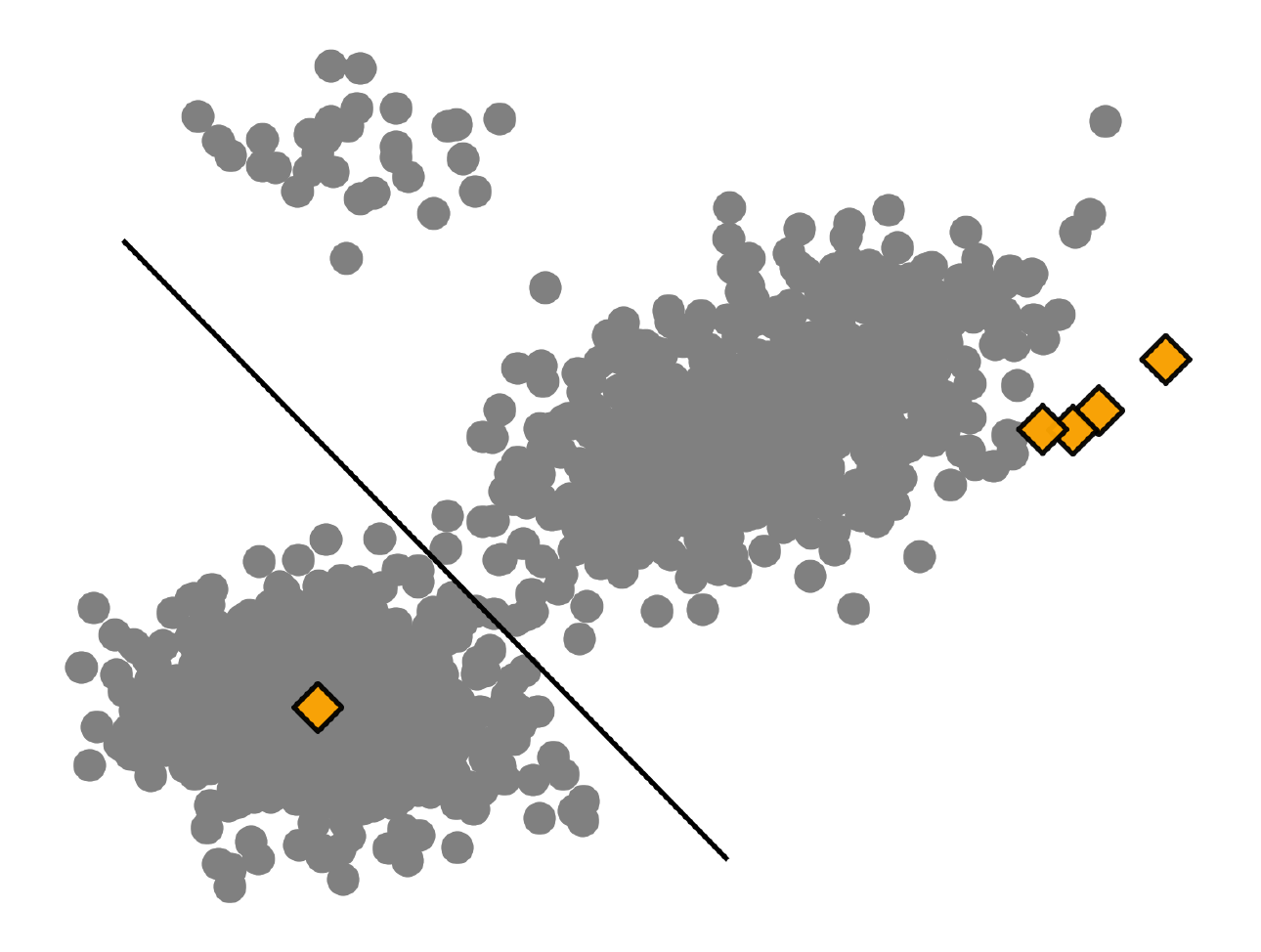}
        \caption{DIVINE $\gamma = 0.526$}
    \end{subfigure}
    \begin{subfigure}[b]{0.245\linewidth}
        \includegraphics[width=\textwidth]{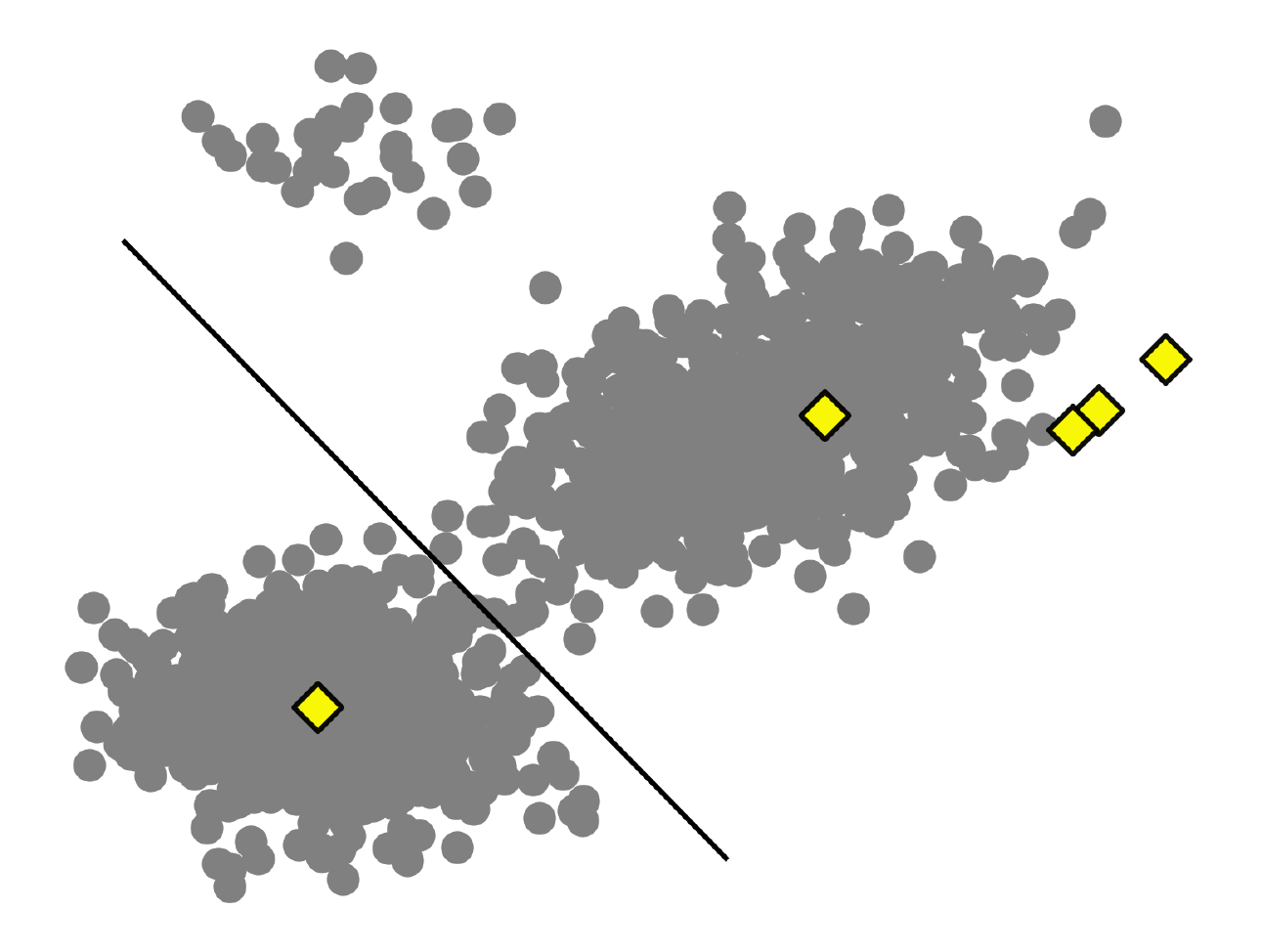}
            \caption{DIVINE $\gamma = 0.628$}
    \end{subfigure}%
    \caption{\small $\mathcal{R}_\text{FL}$. Some points are representative (near cluster center), but others are redundant (top right).
     }\label{fig:all_subplots_copy_fl}
\end{figure*}

\begin{figure*}[htb]
\vspace{-0.5cm}
\centering
    \begin{subfigure}[b]{0.245\linewidth}        
            \includegraphics[width=\textwidth]{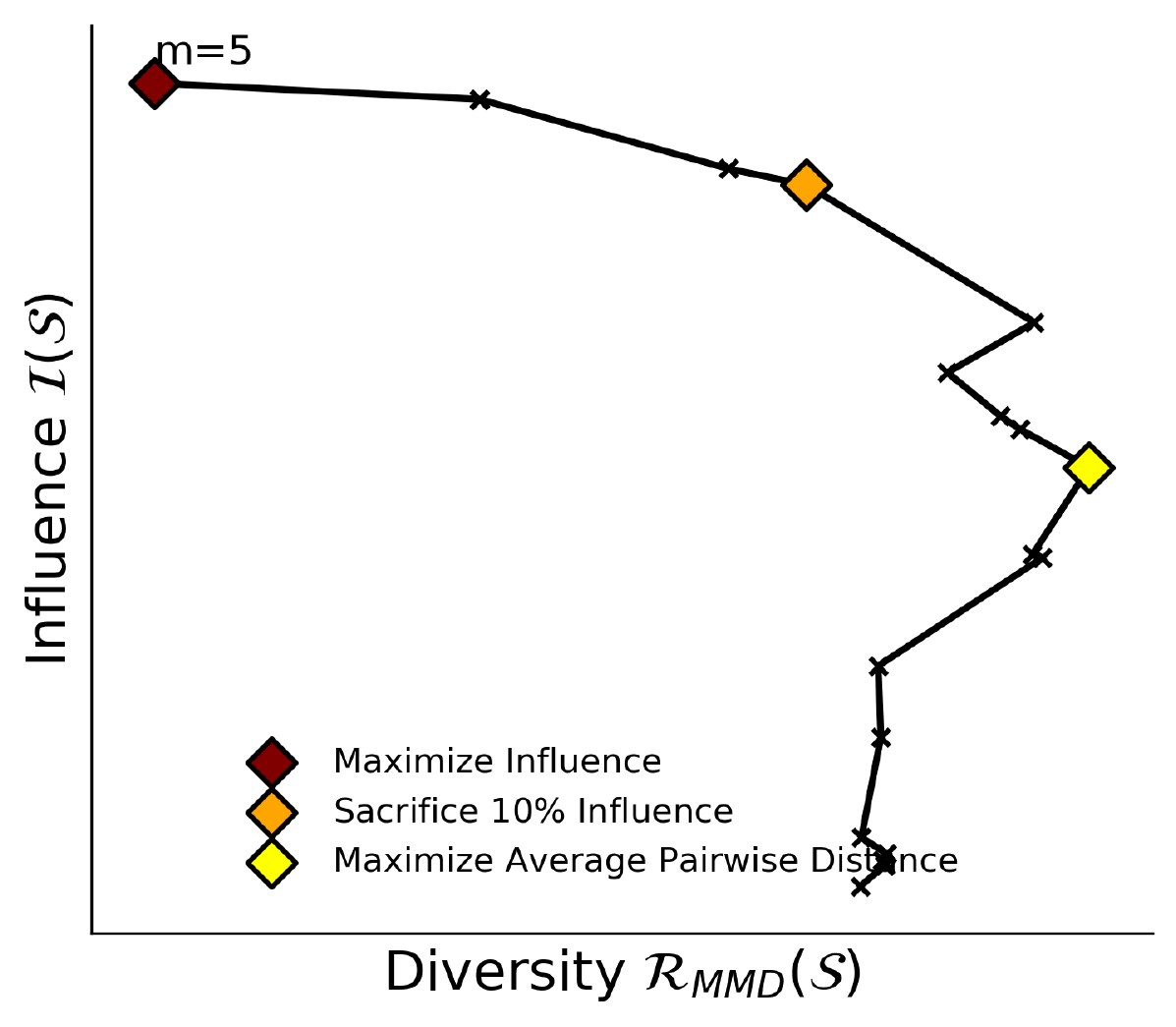}
            \caption{$\mathcal{I}$-$\mathcal{R}$ Tradeoff}
    \end{subfigure}
    \begin{subfigure}[b]{0.245\linewidth}
        \includegraphics[width=\textwidth]{figures/pdfs_app/fig_two_if_train_gam0.pdf}
        \caption{DIVINE $\gamma = 0$}
    \end{subfigure}
    \begin{subfigure}[b]{0.245\linewidth}    
        \includegraphics[width=\textwidth]{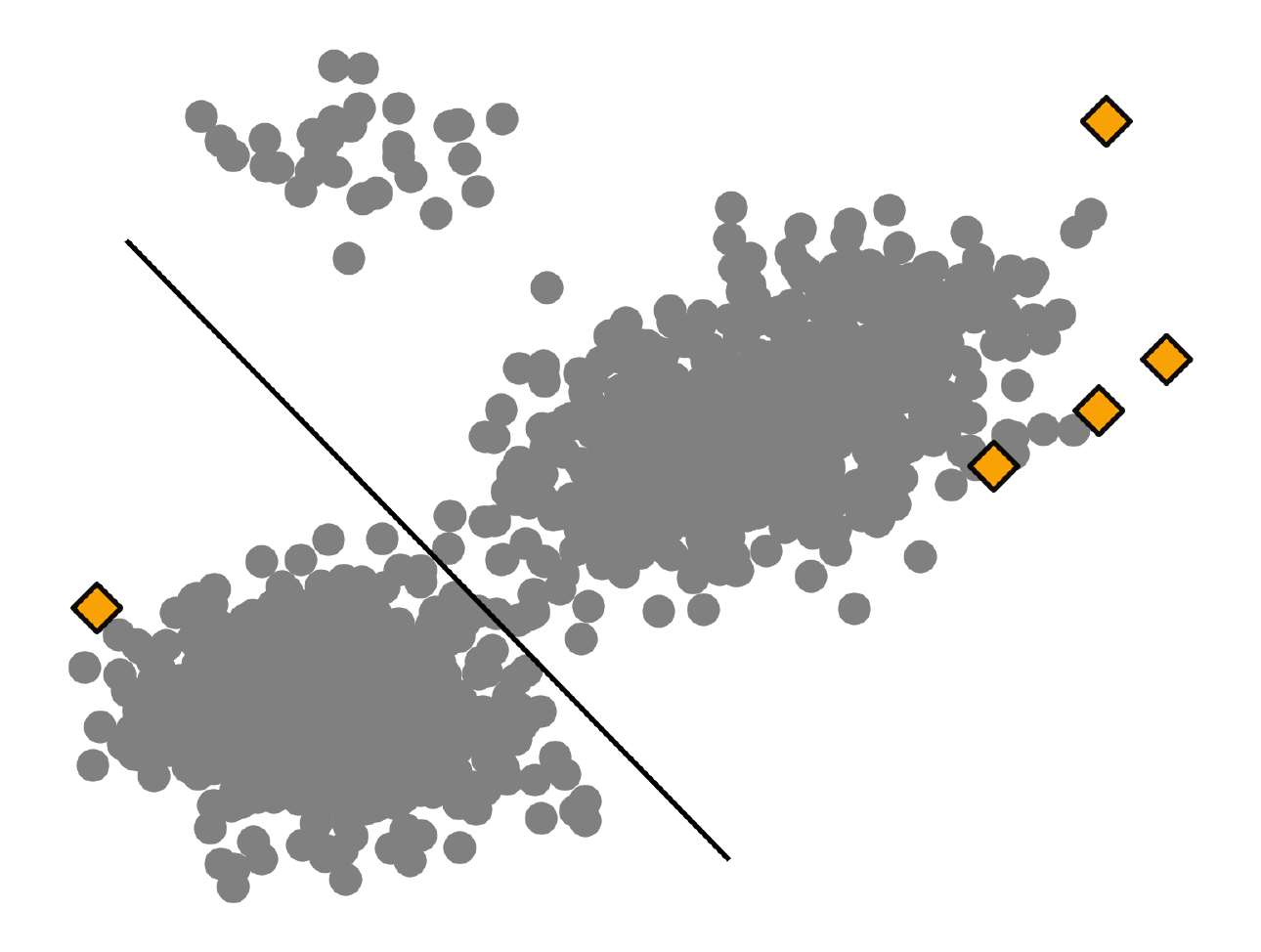}
        \caption{DIVINE $\gamma = 201$}
    \end{subfigure}
    \begin{subfigure}[b]{0.245\linewidth}
        \includegraphics[width=\textwidth]{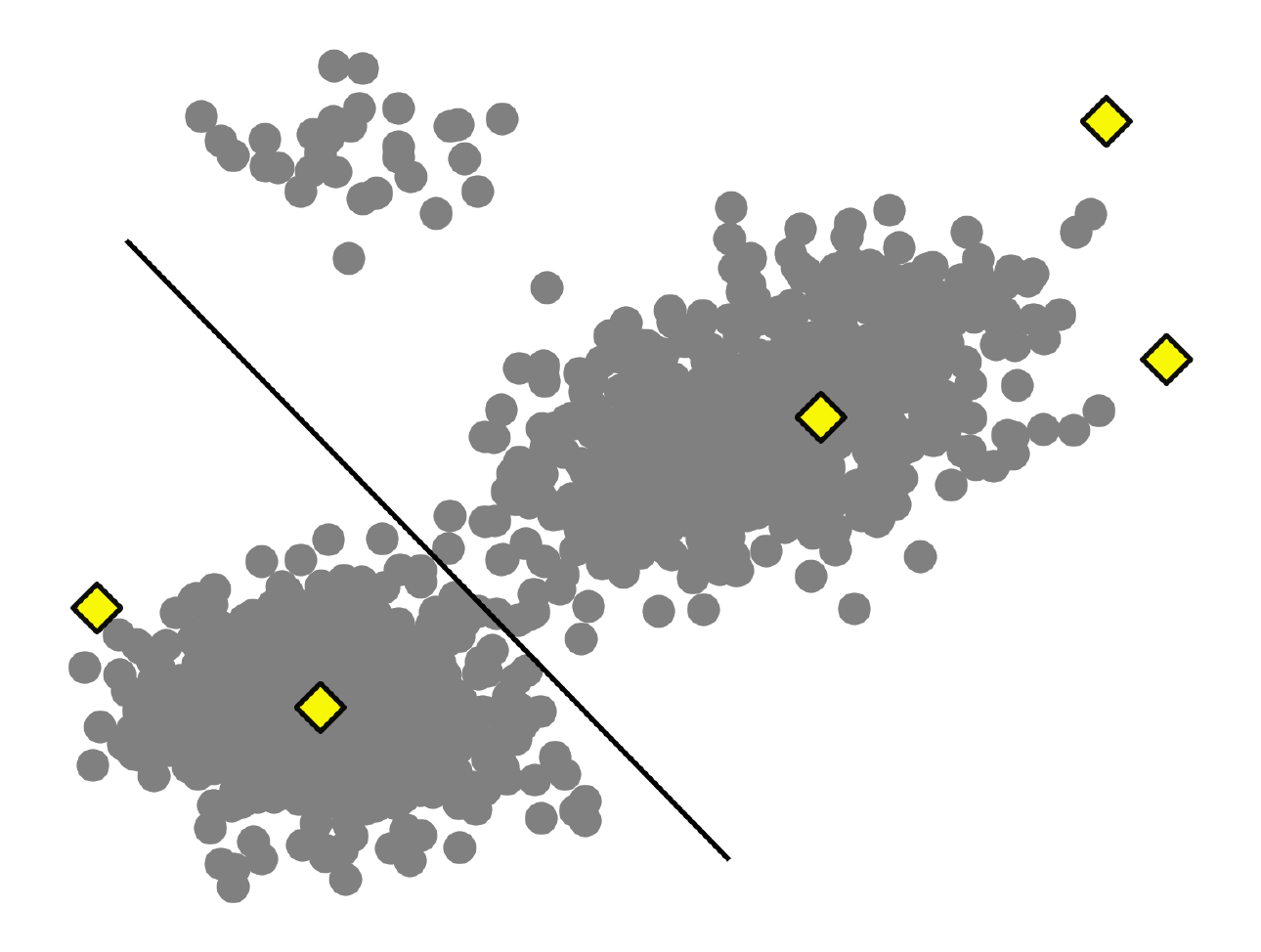}
            \caption{DIVINE $\gamma = 614$}
    \end{subfigure}%
    \caption{\small $\mathcal{R}_\text{MMD}$. Our selected points are representative, though the top cluster is missed. There are no redundant points, in contrast to points selected with $\mathcal{R}_\text{FL}$.
     }\label{fig:all_subplots_copy_mmd}
\end{figure*}

\begin{figure*}[htb]
\centering
\vspace{-0.5cm}
    \begin{subfigure}[b]{0.3\linewidth}
        \includegraphics[width=\textwidth]{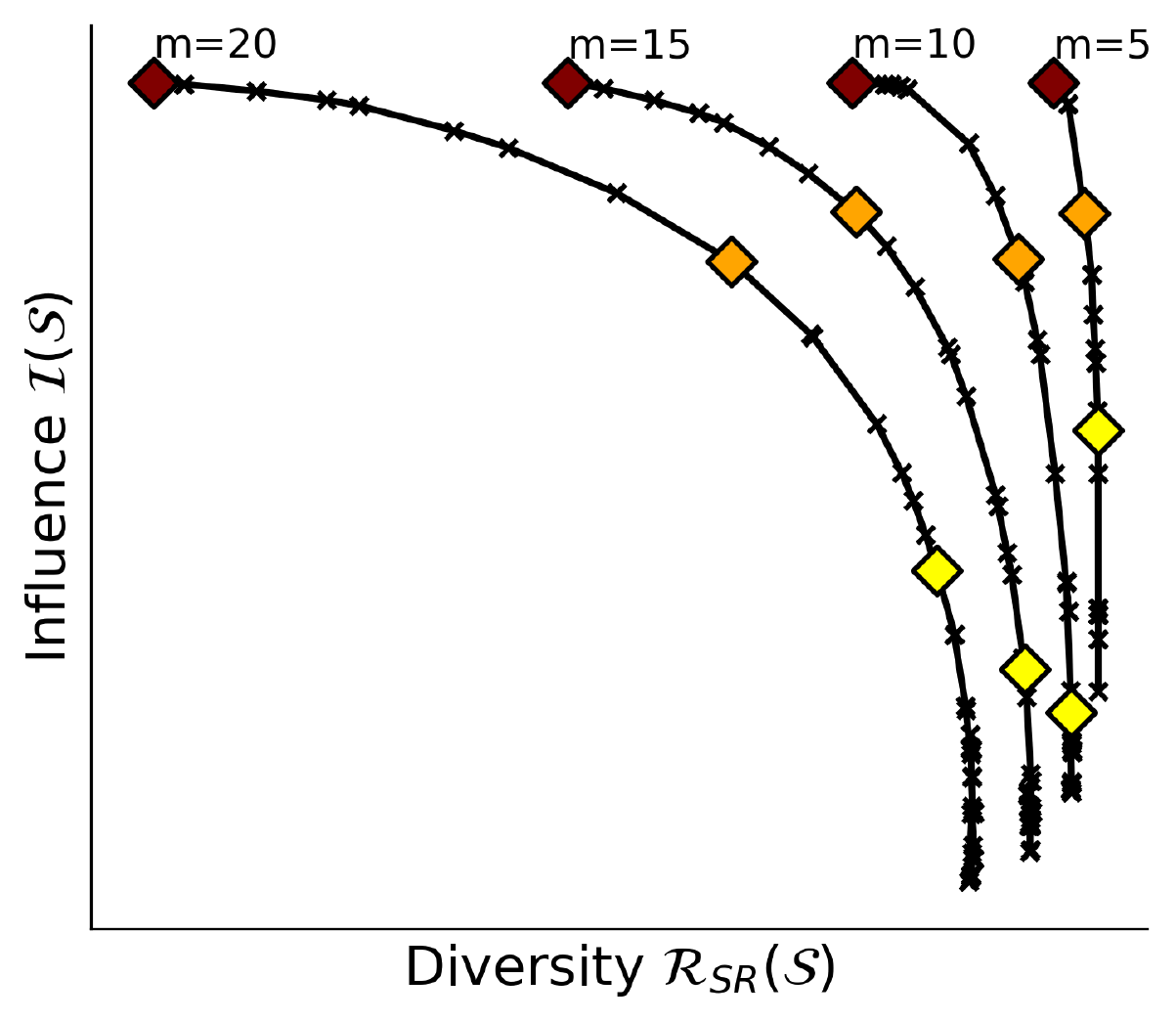}
        \caption{$\mathcal{R}_\text{SR}$}
    \end{subfigure}
        \begin{subfigure}[b]{0.3\linewidth}
        \includegraphics[width=\textwidth]{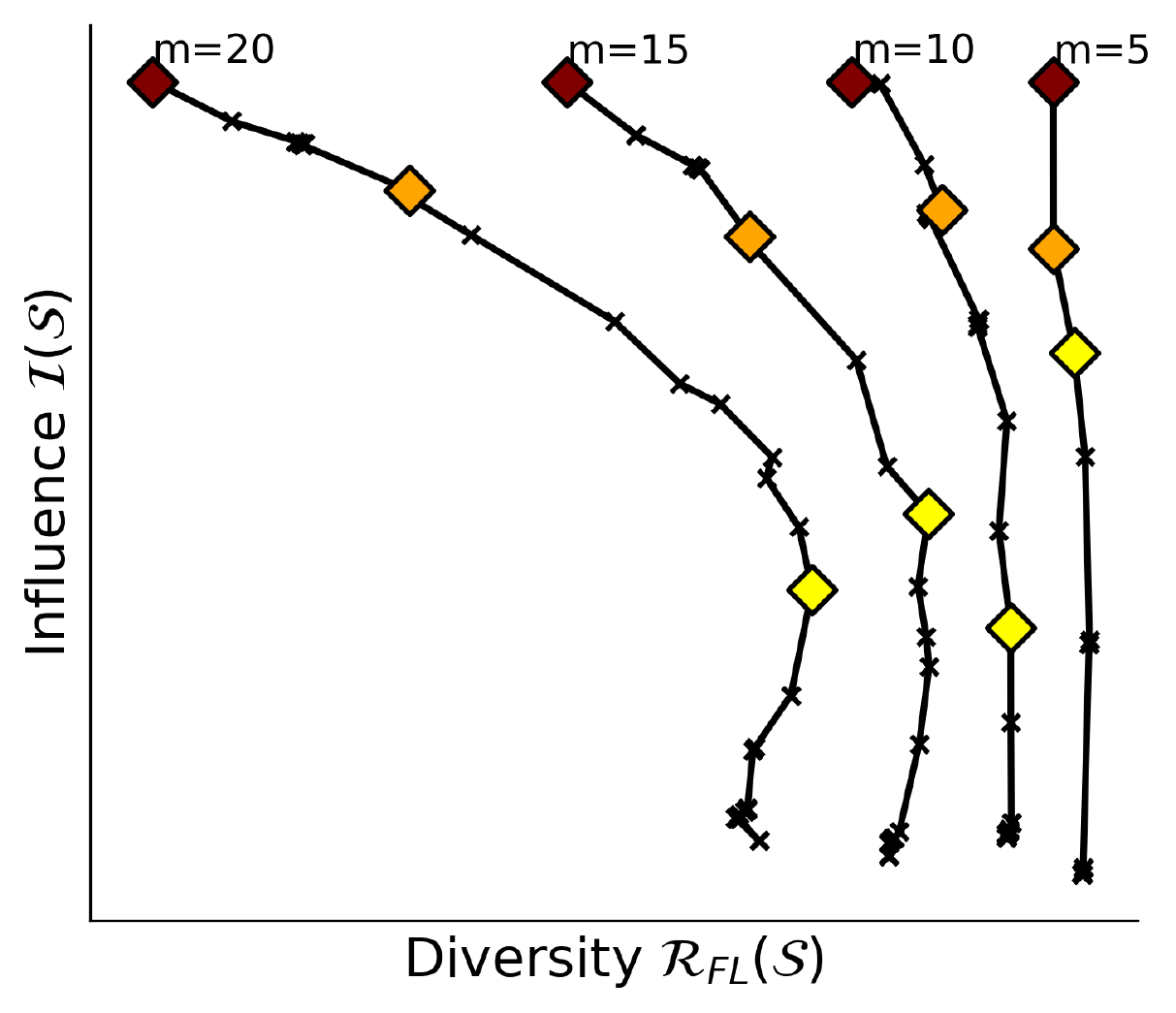}
            \caption{$\mathcal{R}_\text{FL}$}
    \end{subfigure}%
    \begin{subfigure}[b]{0.3\linewidth}    
        \includegraphics[width=\textwidth]{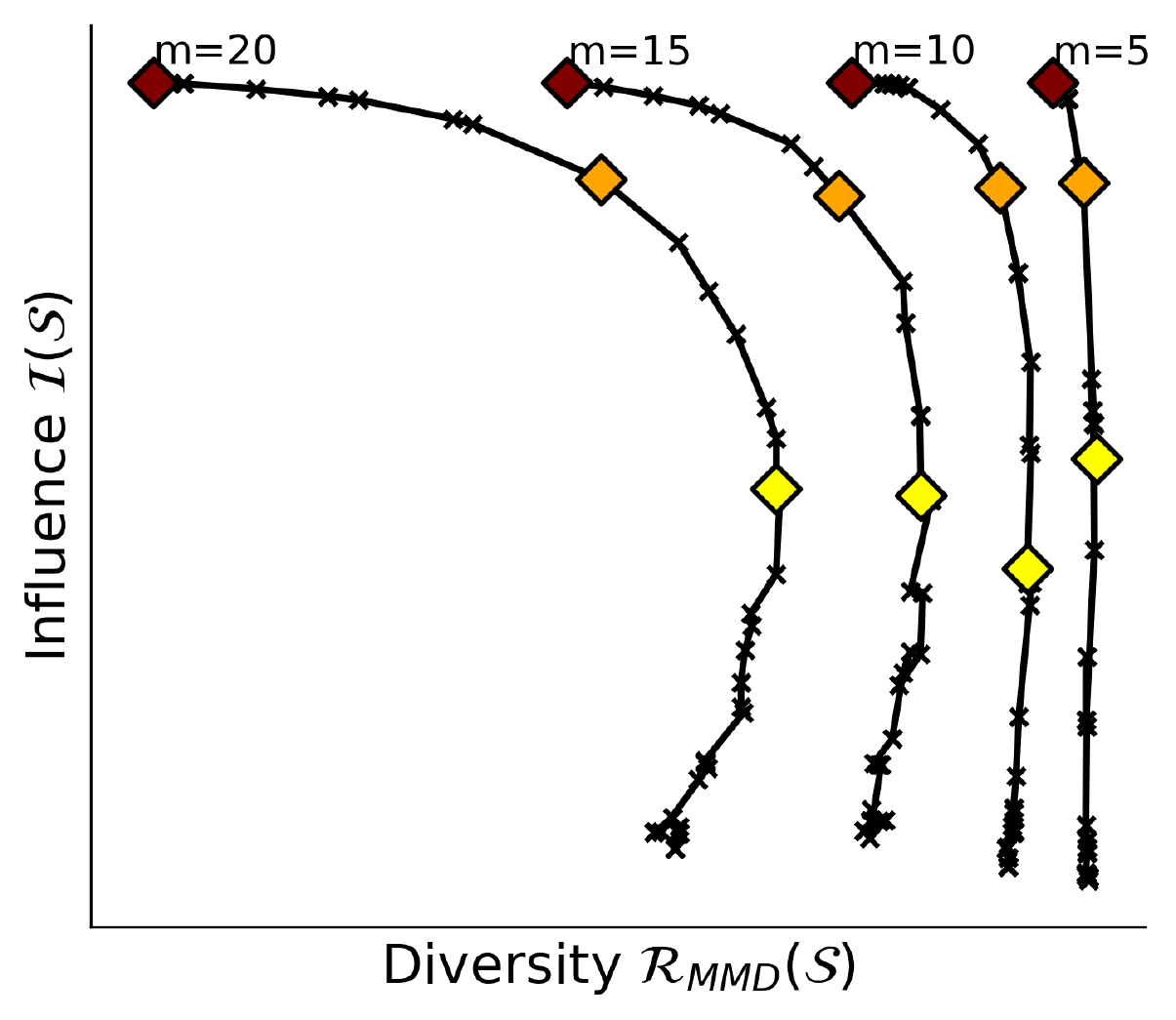}
        \caption{$\mathcal{R}_\text{MMD}$}
    \end{subfigure}
    \caption{We report the DIVINE trade-off as a function of $m$ for various $\mathcal{R}$ with our synthetic data.}\label{fig:tradem_appm}
\end{figure*}

First, we replicate Figure~\ref{fig:all_subplots} with various diversity functions on our synthetic data. Here $m=5$ and $\mathcal{I}$ is taken to be influence functions~\citep{koh2017understanding} with $f_\text{loss}$.  
We notice similar trends for $\mathcal{R}_\text{MMD}$ and $\mathcal{R}_\text{FL}$ as we did for $\mathcal{R}_\text{SR}$ in Figures~\ref{fig:all_subplots_copy},~\ref{fig:all_subplots_copy_fl}, and~\ref{fig:all_subplots_copy_mmd}. We also visualize the top-5 DIVINE points for select values of $\gamma$. The red diamonds are $\gamma = 0$, which recovers the top points from IF alone. This is the same, irrespective of $\mathcal{R}$. The orange diamonds are the $\gamma$ we find such that our DIVINE points have $10\%$ less influence than IF points. The yellow diamonds are the $\gamma$ we find such that we maximize the average pairwise distance between our DIVINE points.  Notice how both $\mathcal{R}_\text{MMD}$ and $\mathcal{R}_\text{FL}$ encourage representativeness by selecting a DIVINE point that is near the center of the Gaussians. However, since $\mathcal{R}_\text{FL}$ does not penalize redundancy between points, it selects three points near to each other in the top right. As $\gamma$ approaches $\infty$, $\mathcal{R}_\text{MMD}$ will recover the prototypes of~\citet{kim2016MMD}. In Figure~\ref{fig:tradem_appm}, we show how varying $m$ affects our trade-off curves for various $\mathcal{R}$.

\subsection{Analyzing Different Influence Measures}
\begin{figure*}[h]

\centering
    \begin{subfigure}[b]{0.245\linewidth}        
            \includegraphics[width=\textwidth]{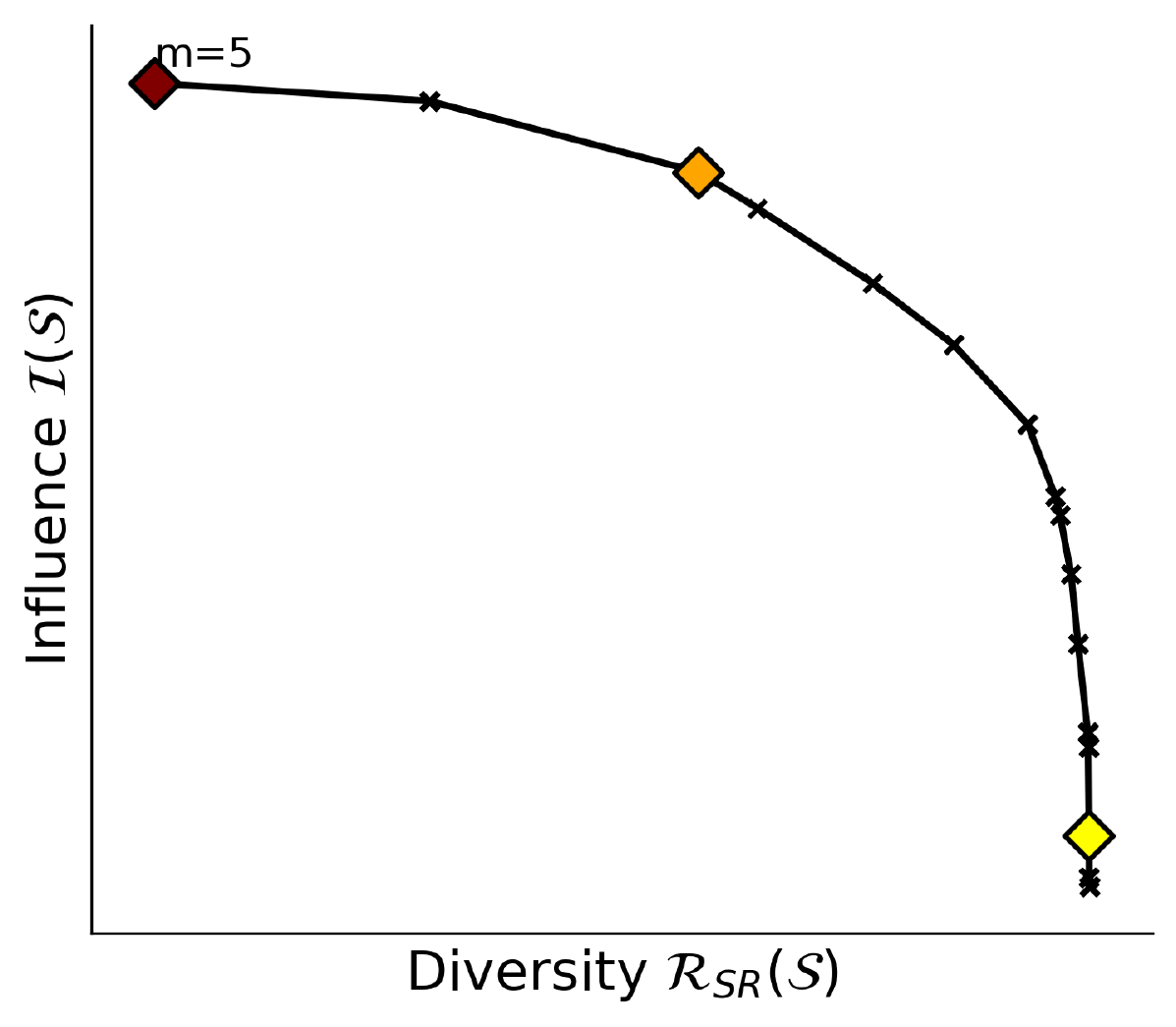}
            \caption{$\mathcal{I}$-$\mathcal{R}$ Tradeoff}
    \end{subfigure}
    \begin{subfigure}[b]{0.245\linewidth}
        \includegraphics[width=\textwidth]{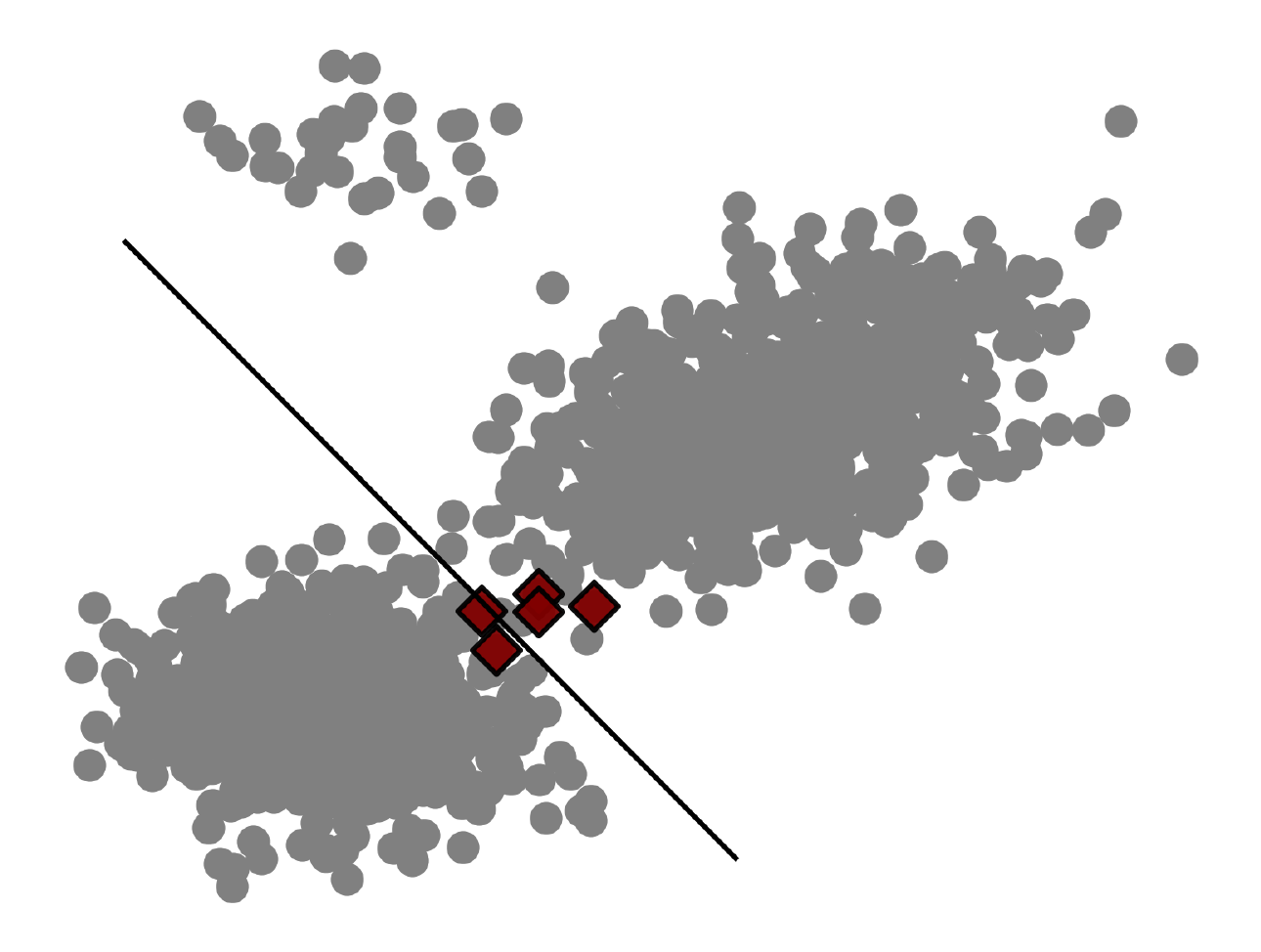}
        \caption{DIVINE $\gamma = 0$}
    \end{subfigure}
    \begin{subfigure}[b]{0.245\linewidth}    
        \includegraphics[width=\textwidth]{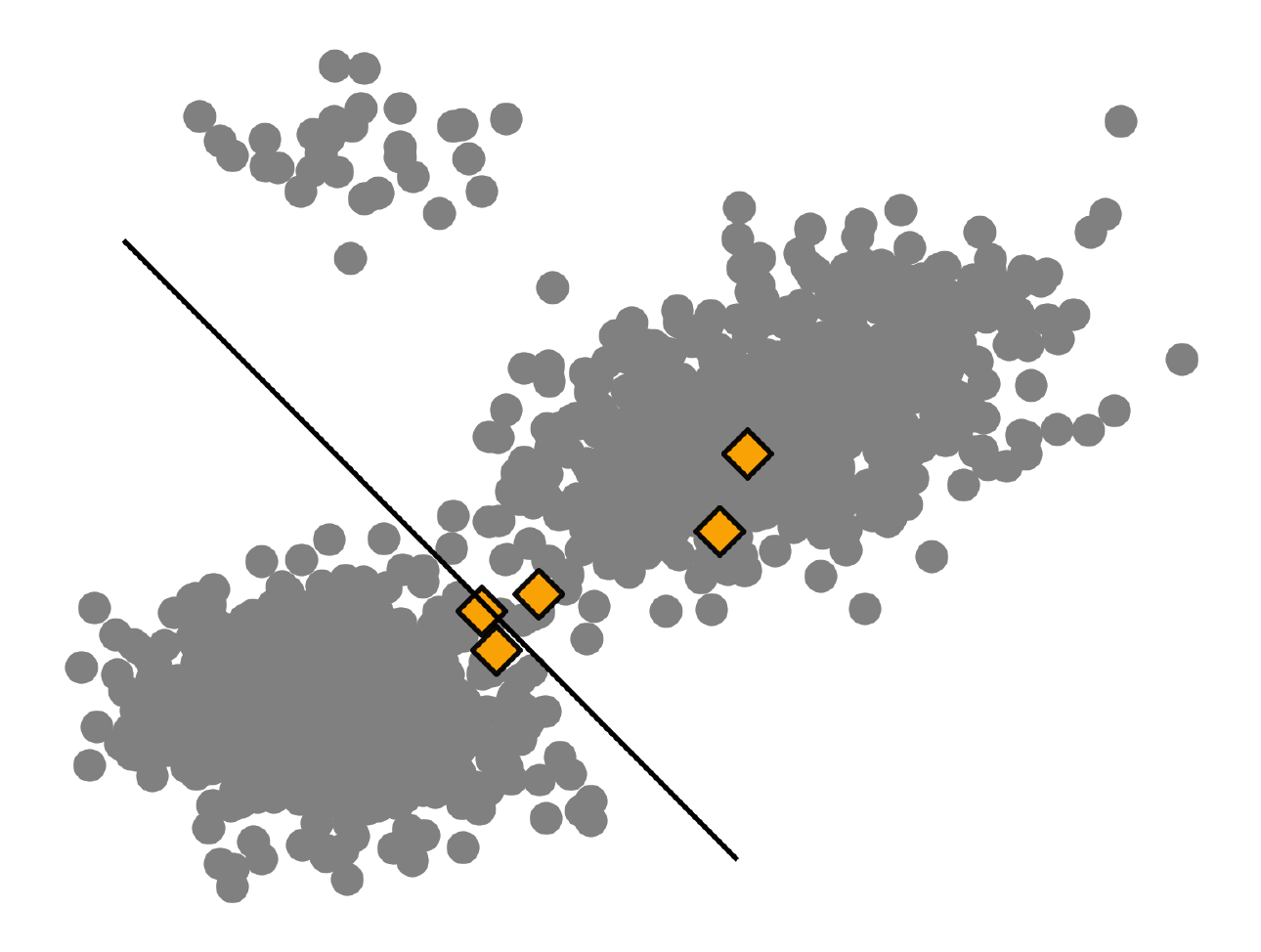}
        \caption{DIVINE $\gamma = 0.041$}
    \end{subfigure}
    \begin{subfigure}[b]{0.245\linewidth}
        \includegraphics[width=\textwidth]{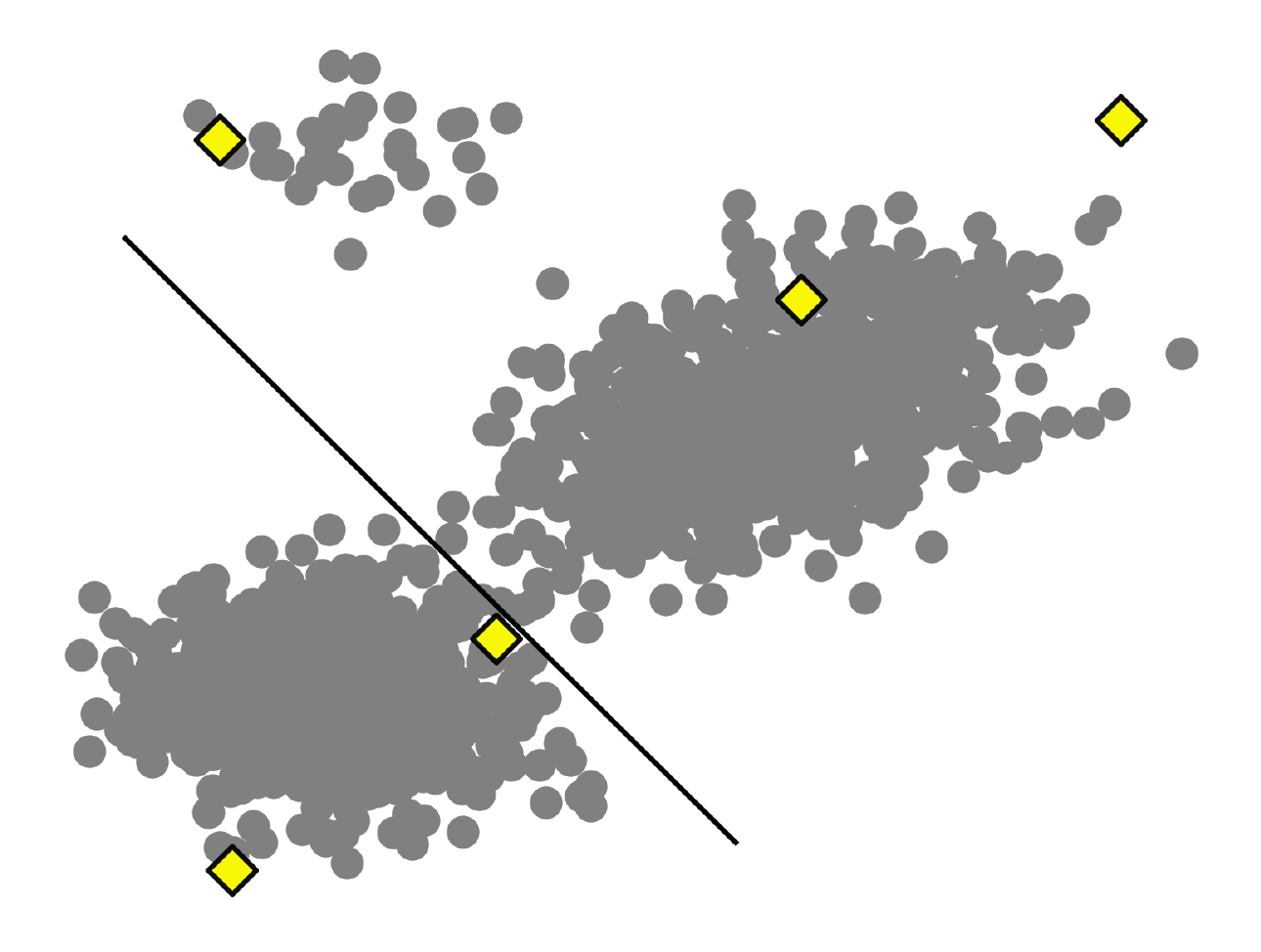}
            \caption{DIVINE $\gamma = 12.42$}
    \end{subfigure}%
    \caption{\small Data Shapley
     }\label{fig:all_subplots_copy_ds}
\end{figure*}

\begin{figure*}[h]
\vspace{-0.5cm}
\centering
    \begin{subfigure}[b]{0.245\linewidth}        
            \includegraphics[width=\textwidth]{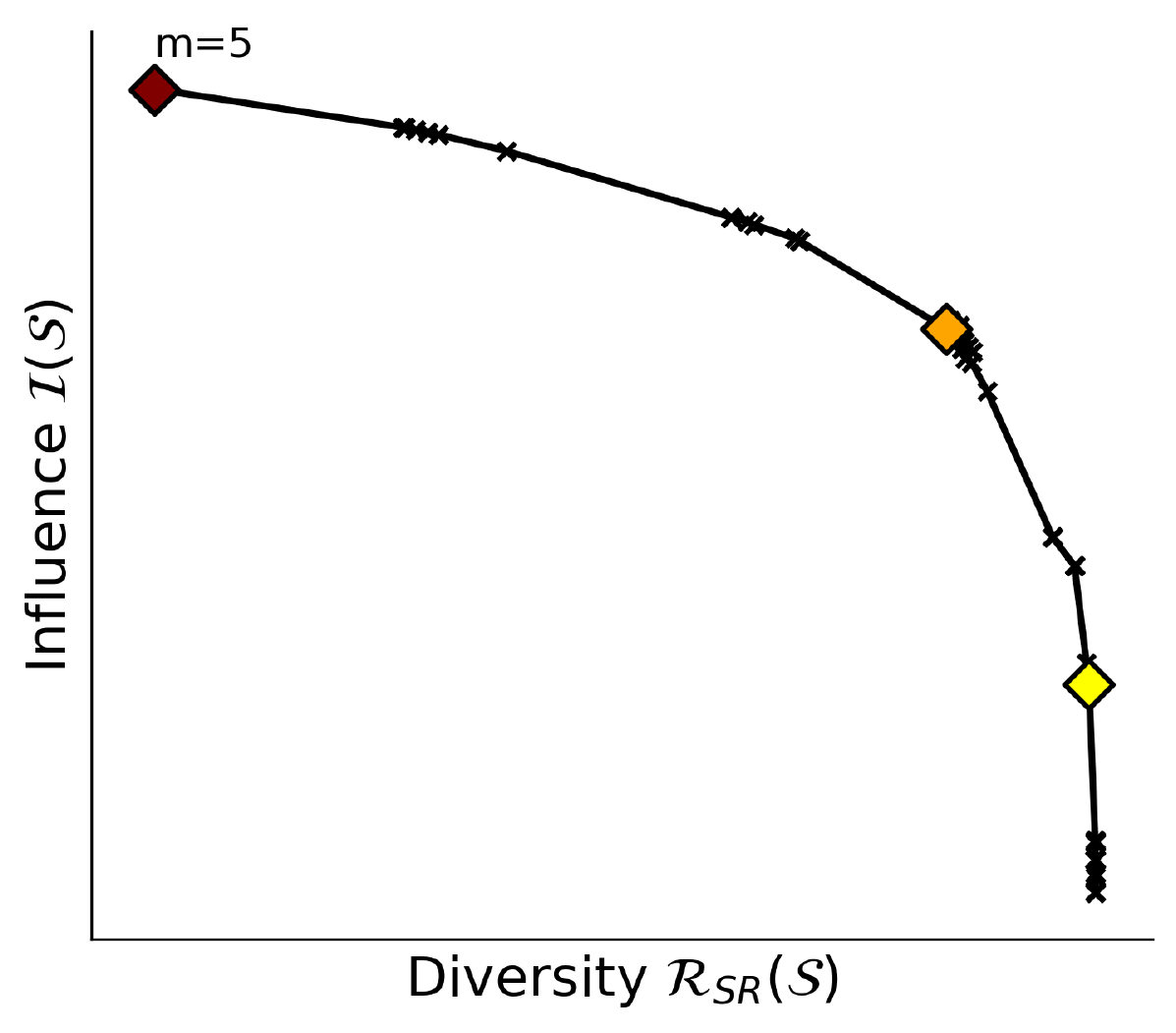}
            \caption{$\mathcal{I}$-$\mathcal{R}$ Tradeoff}
    \end{subfigure}
    \begin{subfigure}[b]{0.245\linewidth}
        \includegraphics[width=\textwidth]{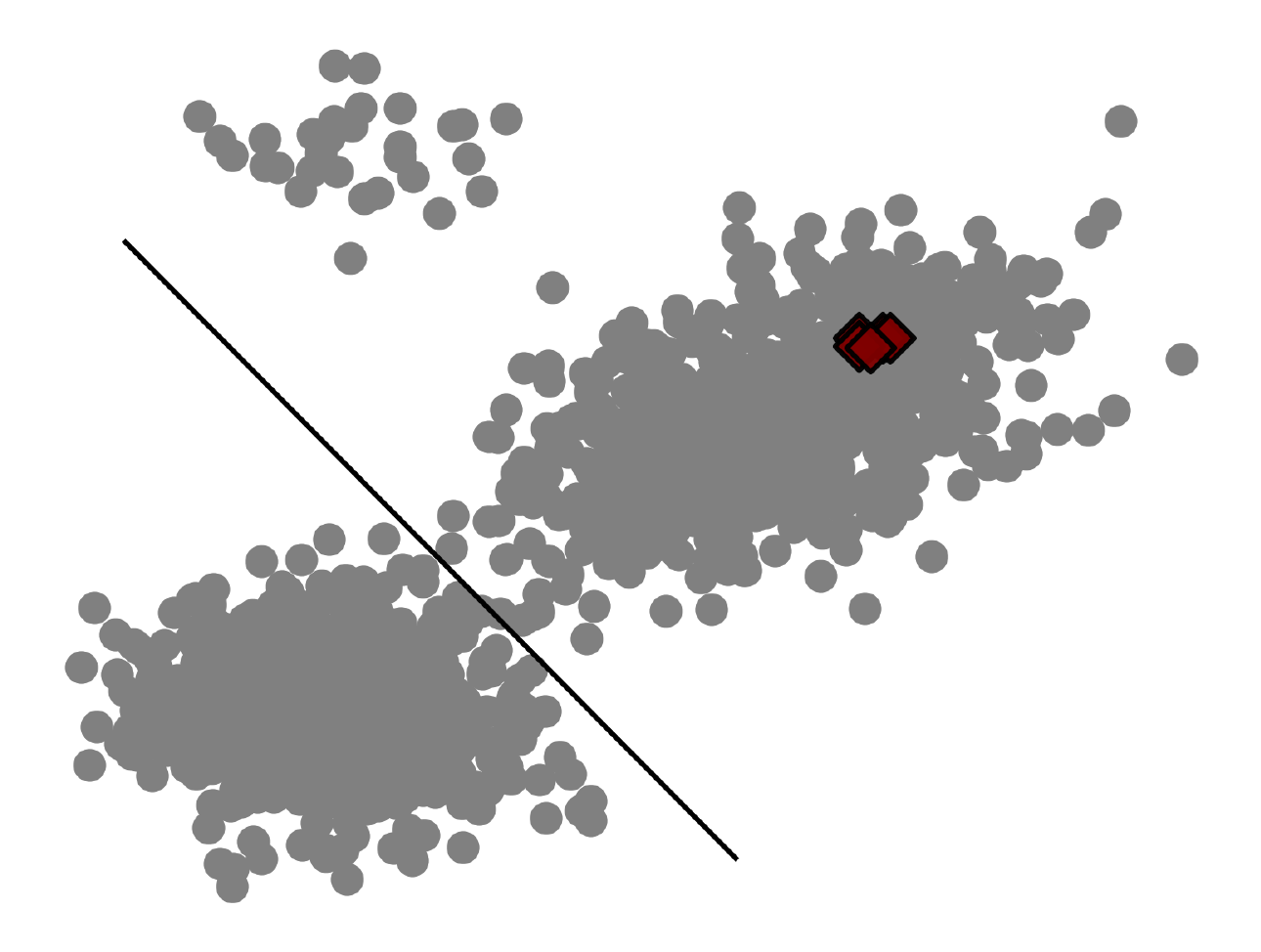}
        \caption{DIVINE $\gamma = 0$}
    \end{subfigure}
    \begin{subfigure}[b]{0.245\linewidth}    
        \includegraphics[width=\textwidth]{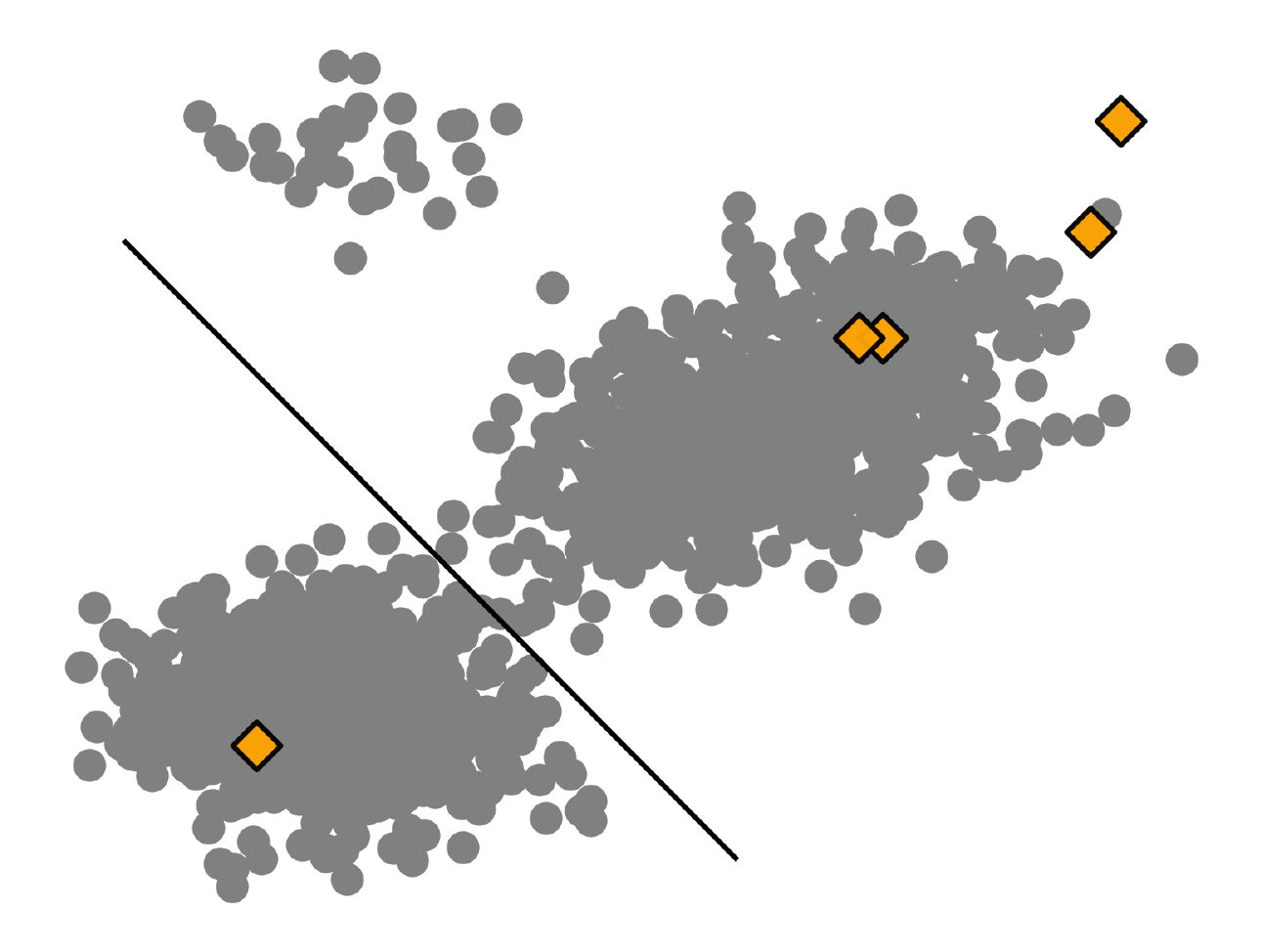}
        \caption{DIVINE $\gamma = 127$}
    \end{subfigure}
    \begin{subfigure}[b]{0.245\linewidth}
        \includegraphics[width=\textwidth]{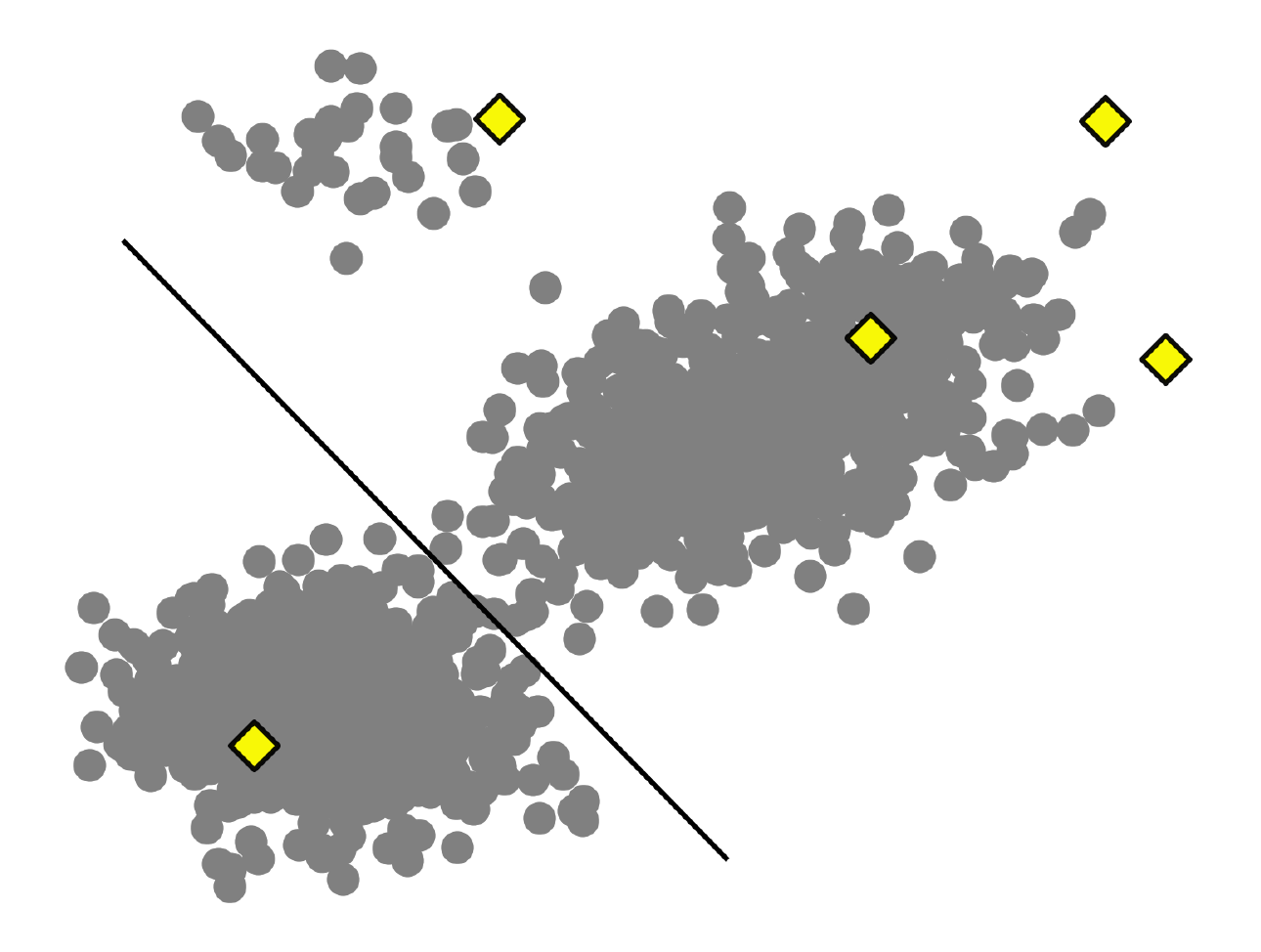}
            \caption{DIVINE $\gamma = 1813$}
    \end{subfigure}%
    \caption{\small Counterfactual Prediction
     }\label{fig:all_subplots_copy_cfp}
\end{figure*}

\begin{figure*}[h]
\vspace{-0.5cm}
\centering
    \begin{subfigure}[b]{0.245\linewidth}        
            \includegraphics[width=\textwidth]{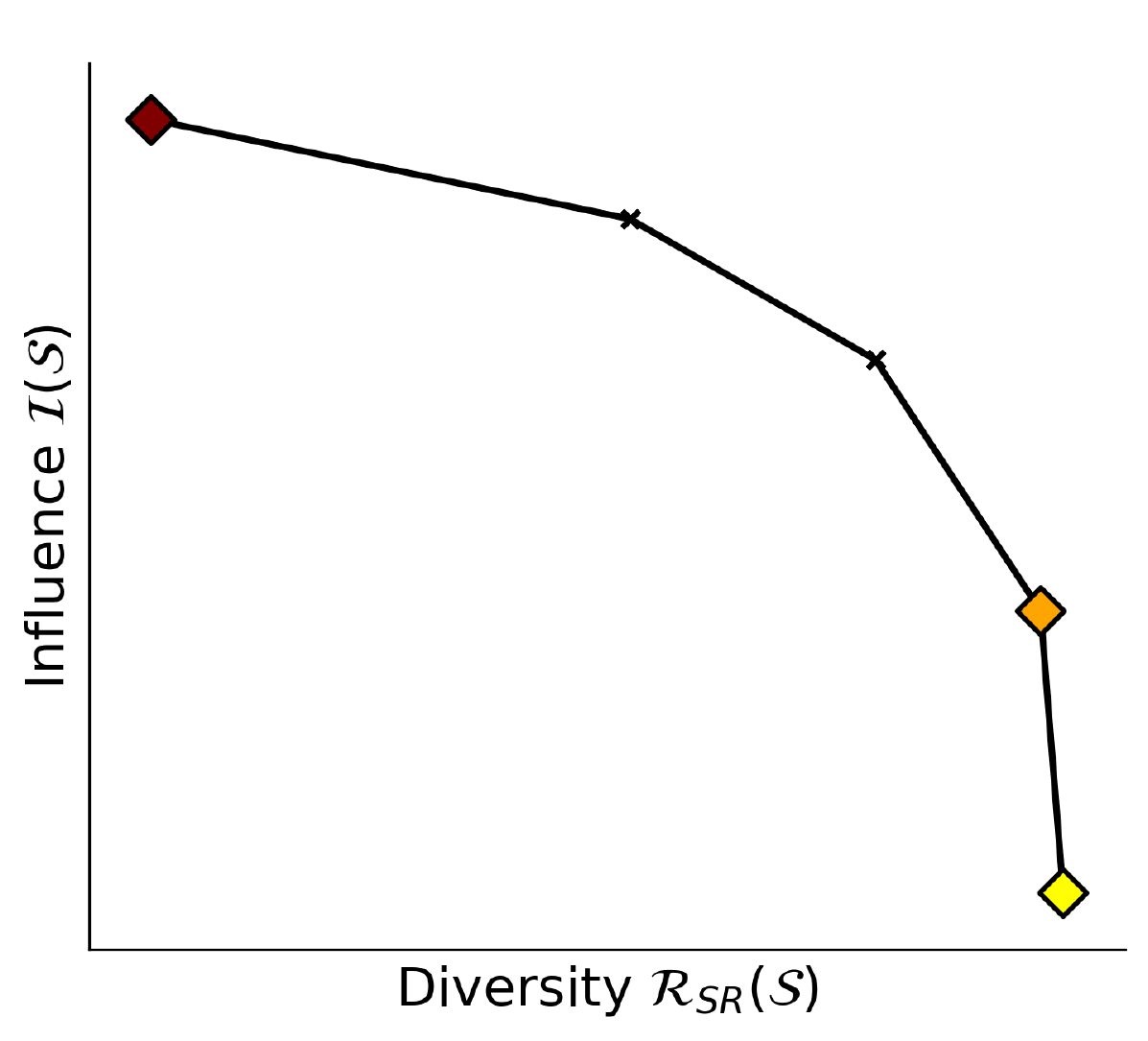}
            \caption{$\mathcal{I}$-$\mathcal{R}$ Tradeoff}
    \end{subfigure}
    \begin{subfigure}[b]{0.245\linewidth}
        \includegraphics[width=\textwidth]{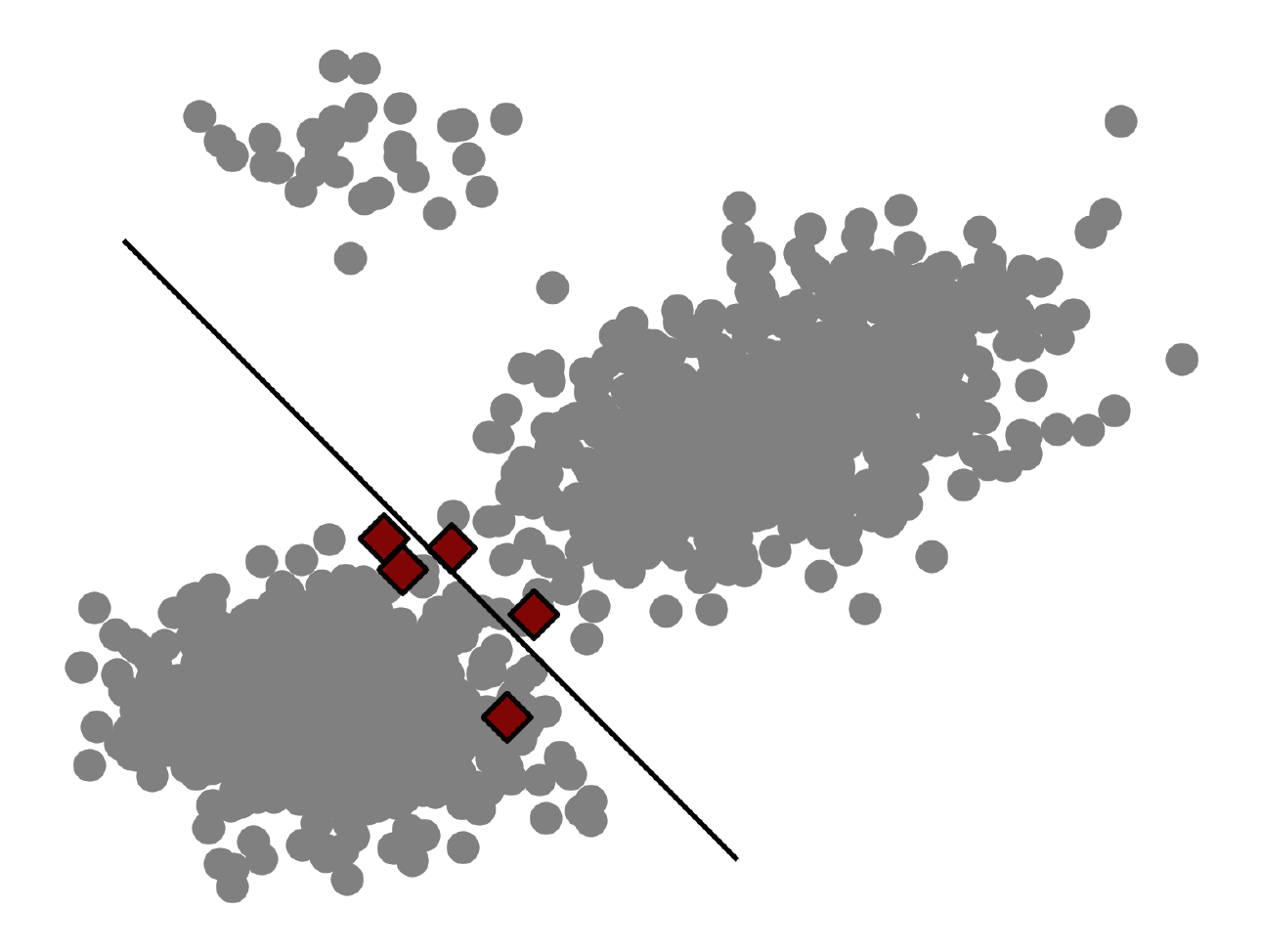}
        \caption{DIVINE $\gamma = 0$}
    \end{subfigure}
    \begin{subfigure}[b]{0.245\linewidth}    
        \includegraphics[width=\textwidth]{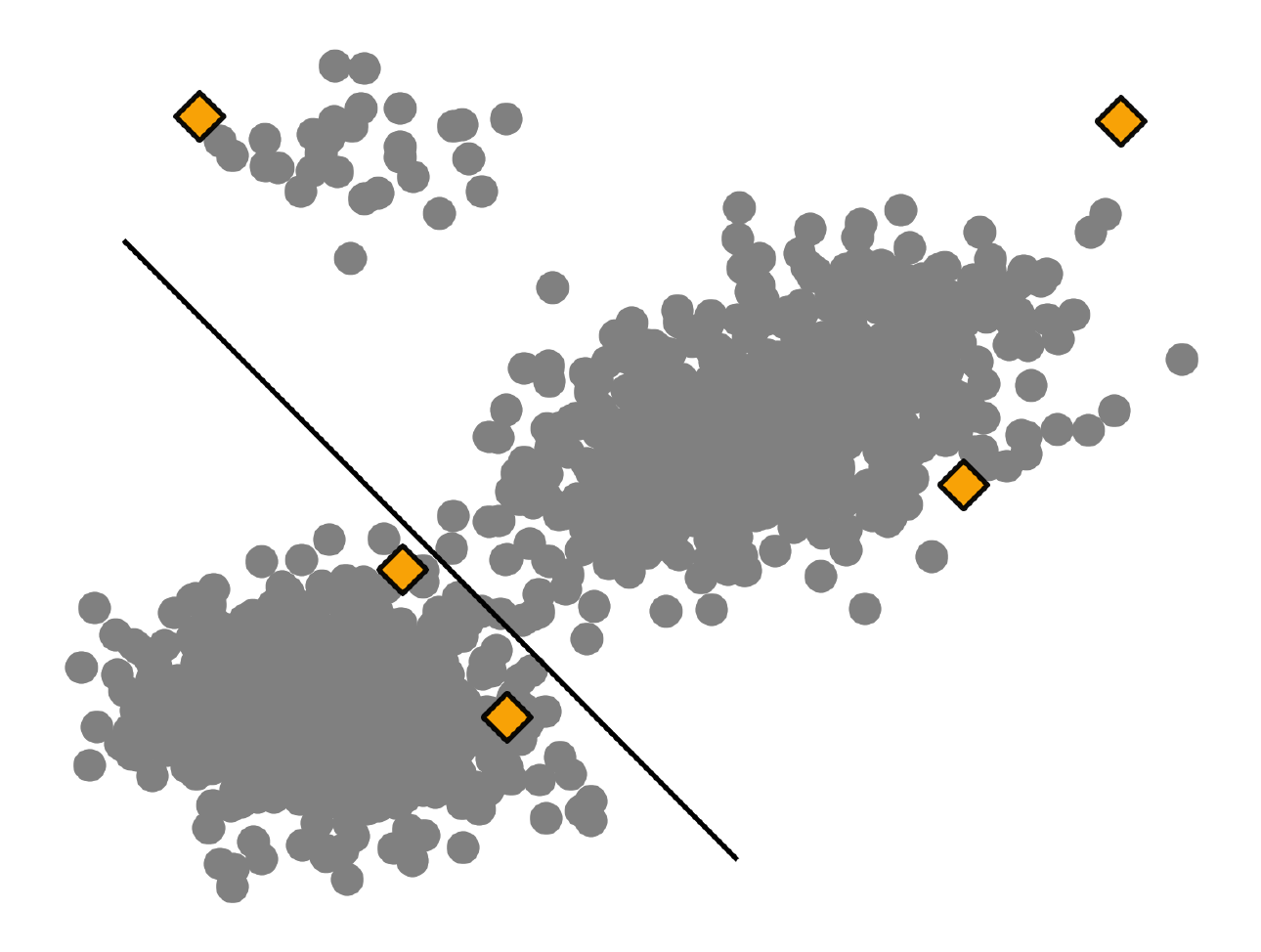}
        \caption{DIVINE $\gamma = 0.97$}
    \end{subfigure}
    \begin{subfigure}[b]{0.245\linewidth}
        \includegraphics[width=\textwidth]{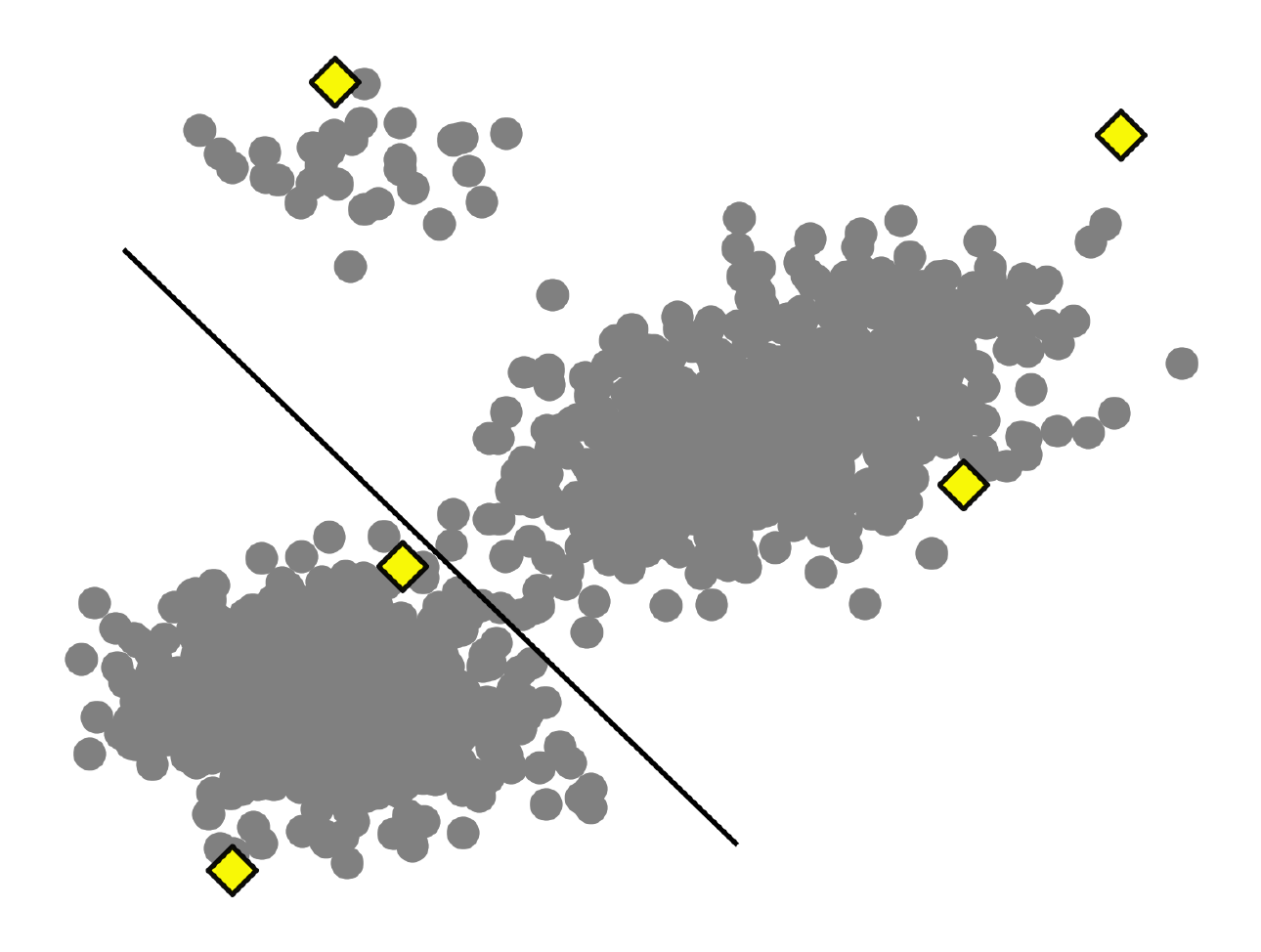}
            \caption{DIVINE $\gamma = 7.97$}
    \end{subfigure}%
    \caption{\small Leave-one-out
     }\label{fig:all_subplots_copy_loo}
\end{figure*}

\begin{figure*}[htb]
\centering
\vspace{-0.5cm}
    \begin{subfigure}[b]{0.245\linewidth}
        \includegraphics[width=\textwidth]{figures/pdfs_app/app_sr_all.pdf}
        \caption{\small Influence Functions}
    \end{subfigure}
    \begin{subfigure}[b]{0.245\linewidth}    
        \includegraphics[width=\textwidth]{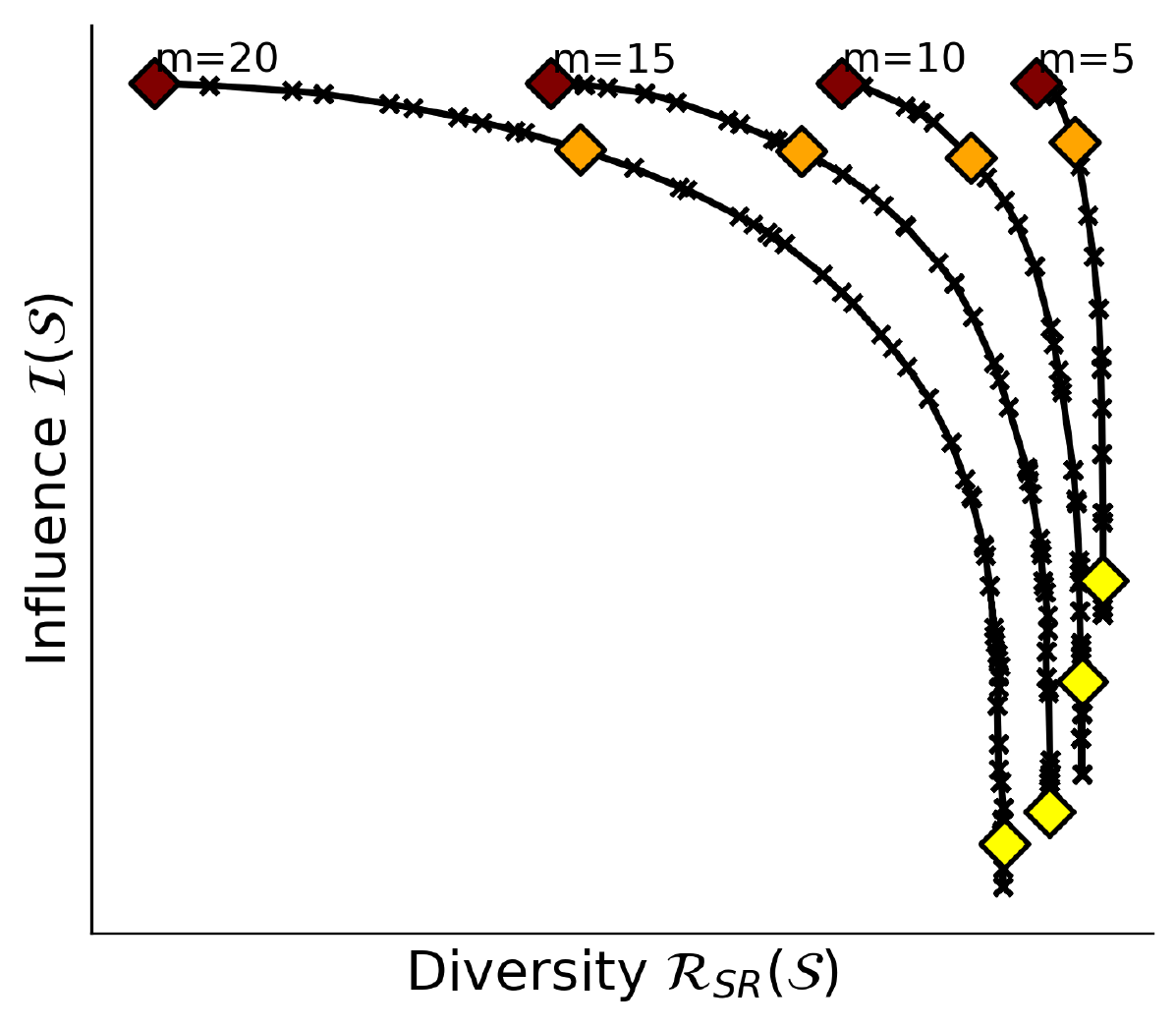}
        \caption{DS}
    \end{subfigure}
    \begin{subfigure}[b]{0.245\linewidth}
        \includegraphics[width=\textwidth]{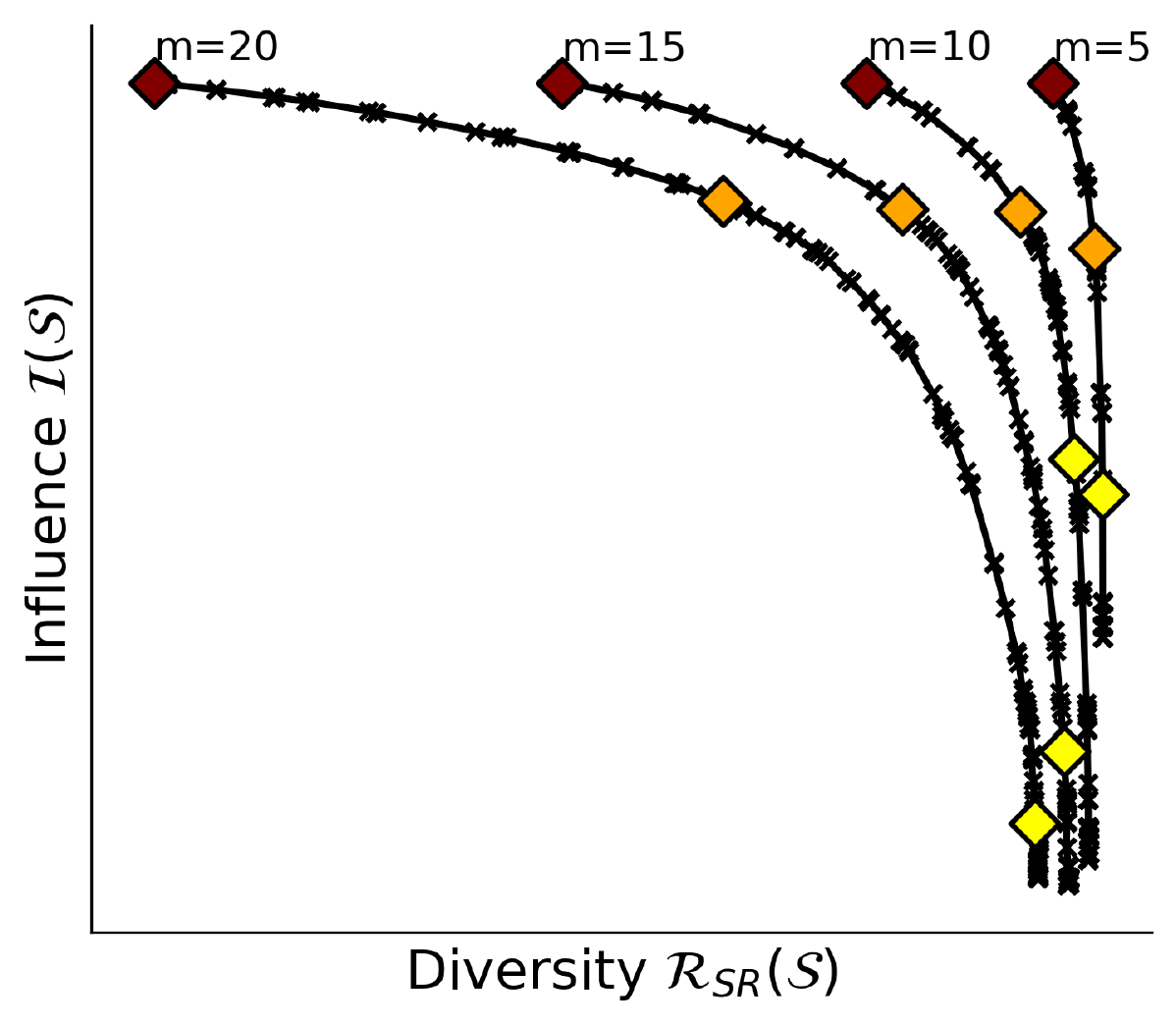}
            \caption{CFP}
    \end{subfigure}%
    \begin{subfigure}[b]{0.245\linewidth}        
            \includegraphics[width=\textwidth]{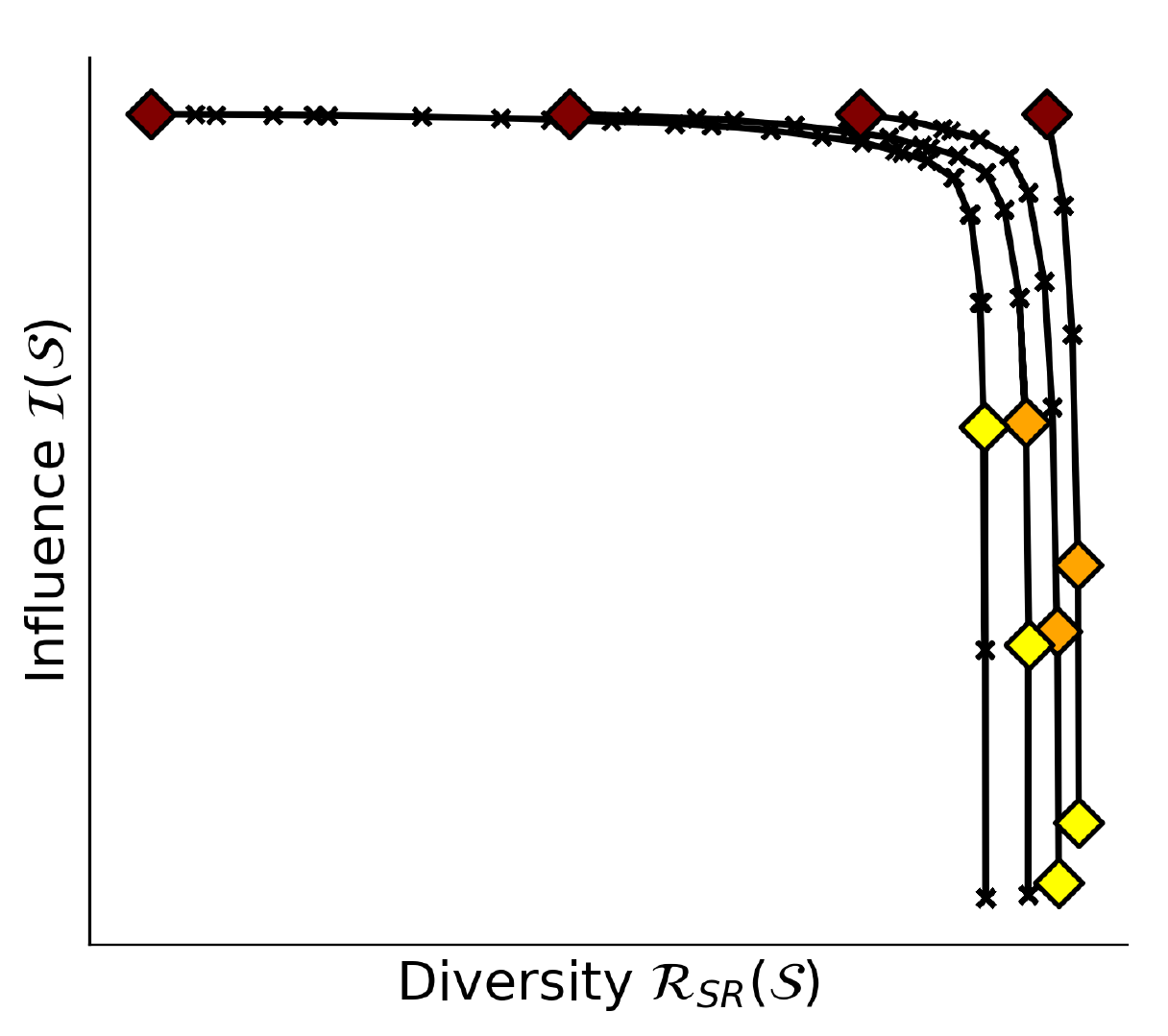}
            \caption{LOO}
    \end{subfigure}
    \caption{We report the DIVINE trade-off as a function of $m$ for various $\mathcal{I}$ with our synthetic data.}\label{fig:tradem_appmI}
\end{figure*}

We next consider the effects of varying the underlying influence measure $\mathcal{I}$ but use $f_\text{loss}$ for all experiments herein. In Figure~\ref{fig:all_subplots_copy_ds},~\ref{fig:all_subplots_copy_cfp}, and~\ref{fig:all_subplots_copy_loo}, we show how DIVINE points are selected for DataShapley, Counterfactual Prediction, and Leave-one-out, respectively. Note that when $\gamma = 0$, the DIVINE points are simply the highest scoring points from each method alone. Every method selects similar points (all high importance are located in a small cluster) when no diversity is considered. Then, we trade-off $\mathcal{R}_\text{SR}$ with influence, and obtain similar trade-off plots. In Figure~\ref{fig:tradem_appmI}, we find that as we increase $m$ we get similar behavior for other influence measures that we obtained for influence functions before. For all influence measures, we use $f_\text{loss}$ as our evaluation function. 

  
    
       
\clearpage
\subsection{Analyzing Different Datasets}
\begin{figure*}[htb]
\centering
    \begin{subfigure}[b]{0.245\linewidth}
        \includegraphics[width=\textwidth]{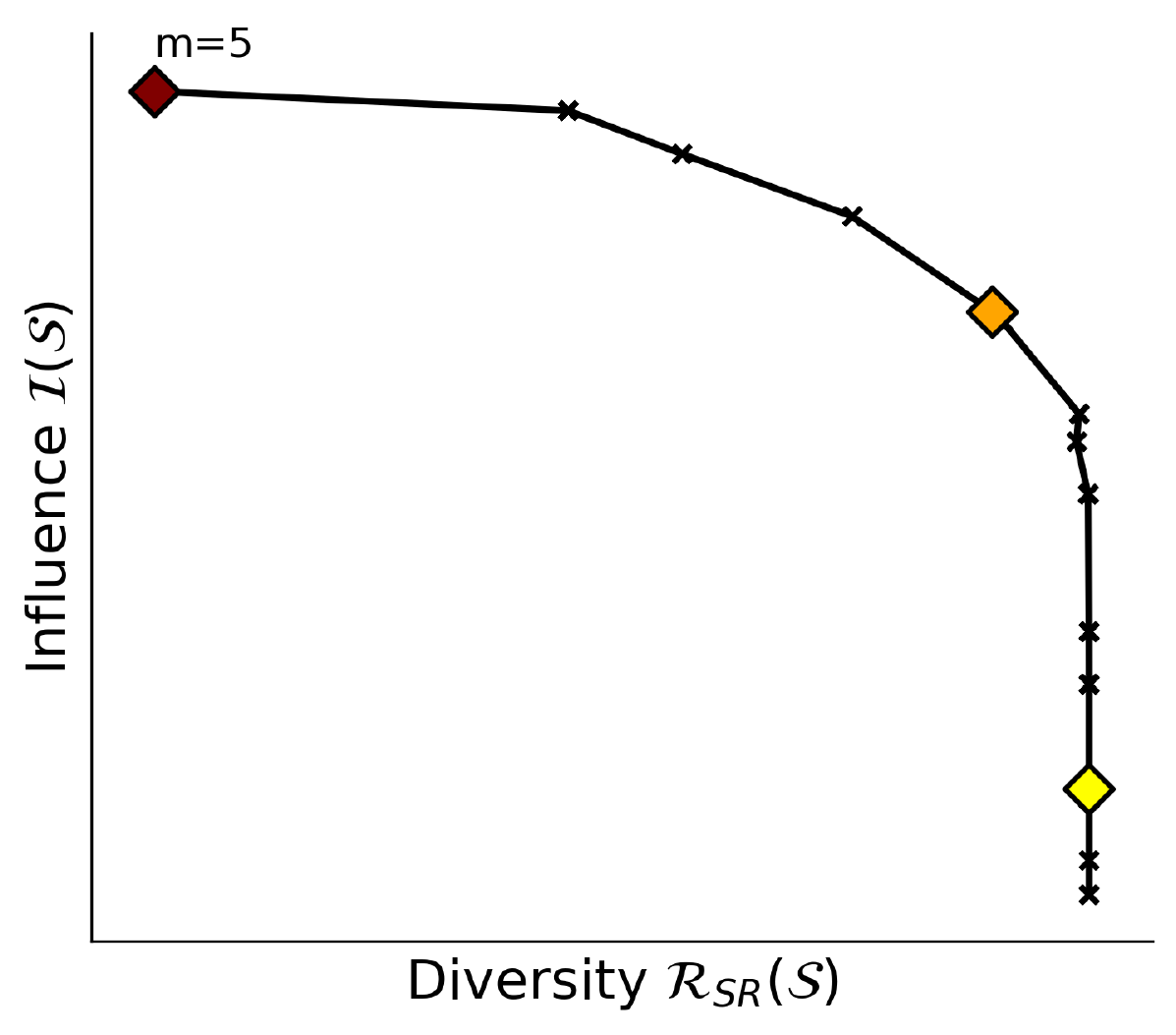}
        \caption{LSAT}
    \end{subfigure}
    \begin{subfigure}[b]{0.245\linewidth}    
        \includegraphics[width=\textwidth]{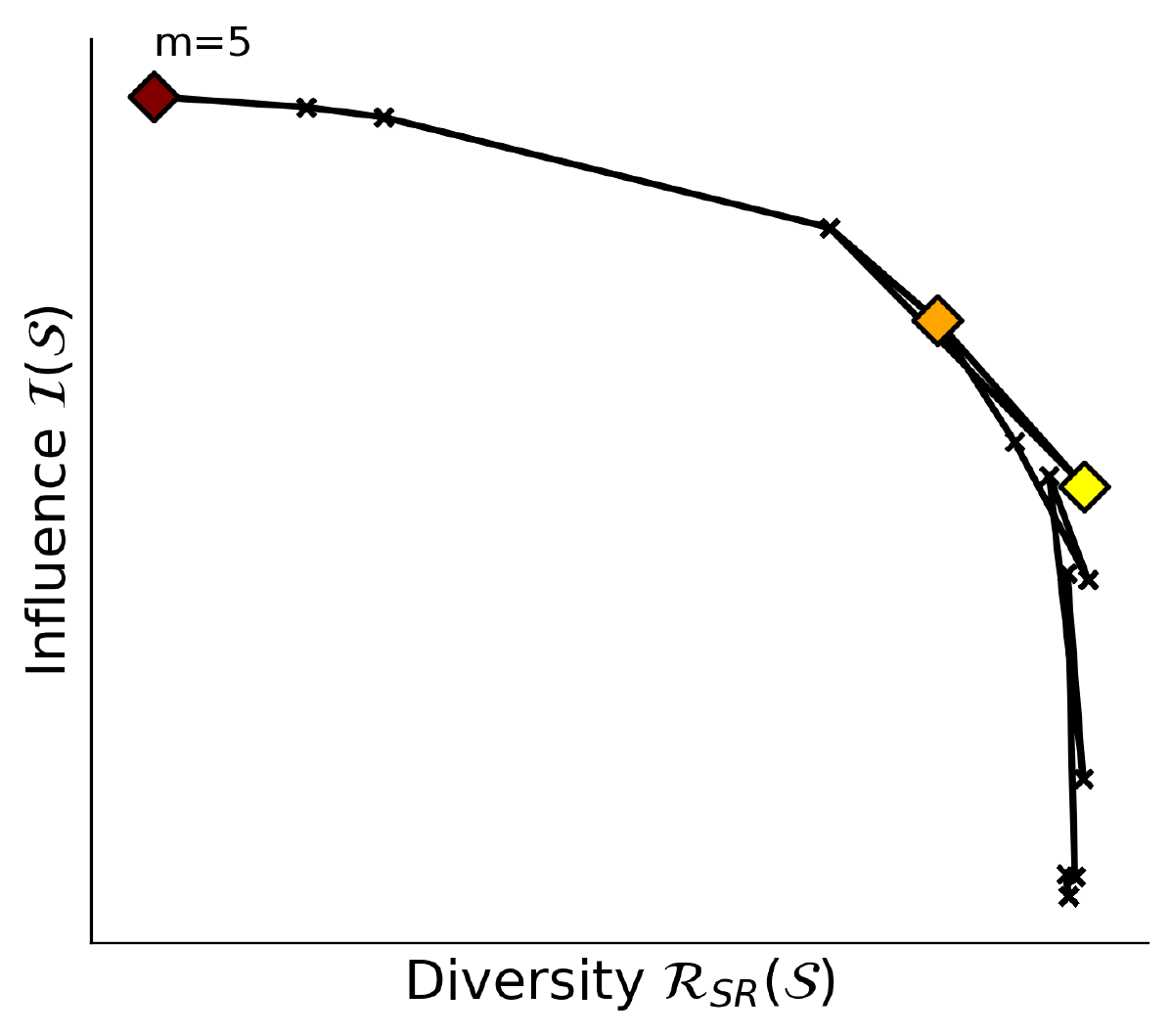}
        \caption{COMPAS}
    \end{subfigure}
    \begin{subfigure}[b]{0.245\linewidth}
        \includegraphics[width=\textwidth]{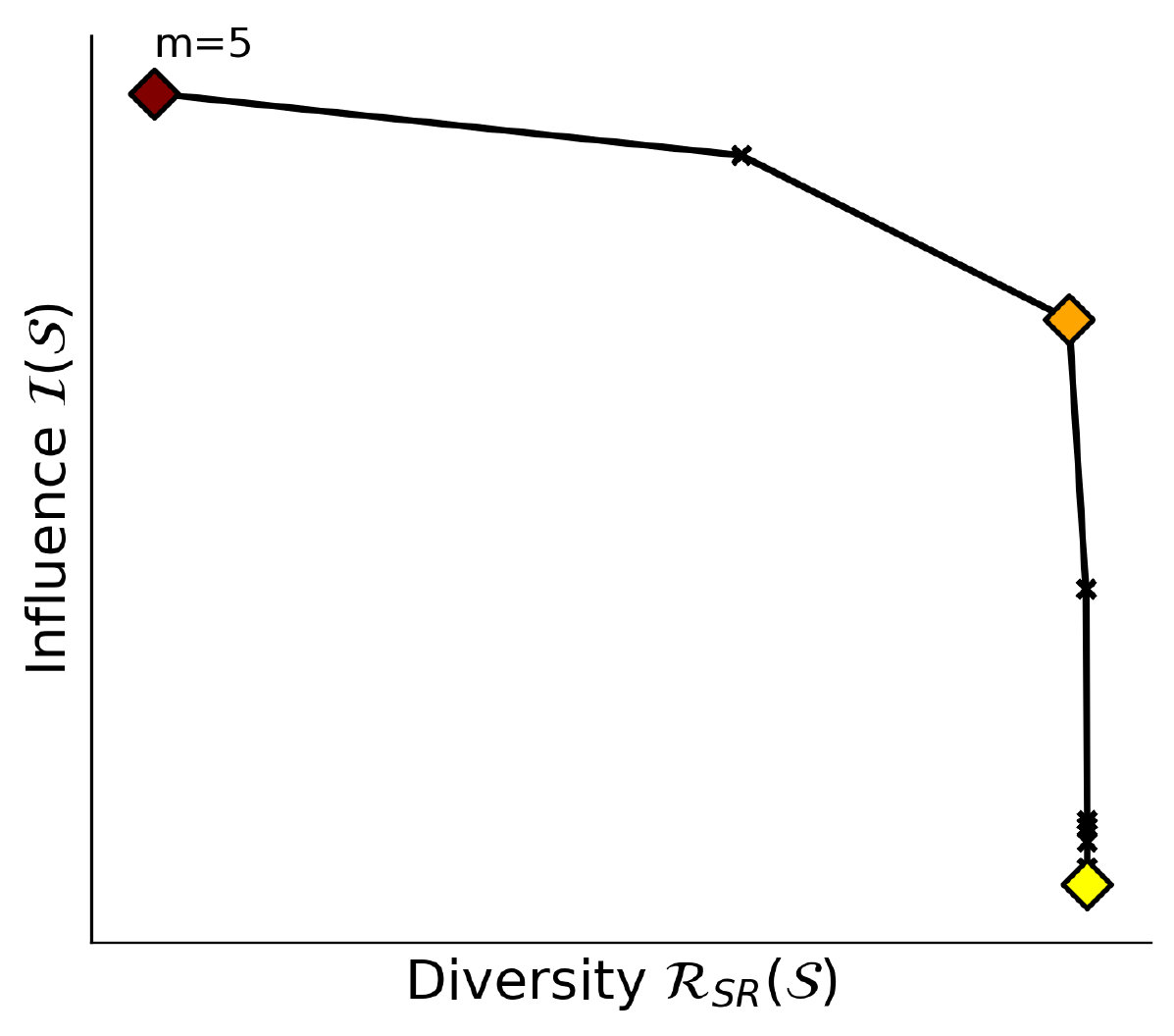}
            \caption{Adult}
    \end{subfigure}%
    \begin{subfigure}[b]{0.245\linewidth}        
            \includegraphics[width=\textwidth]{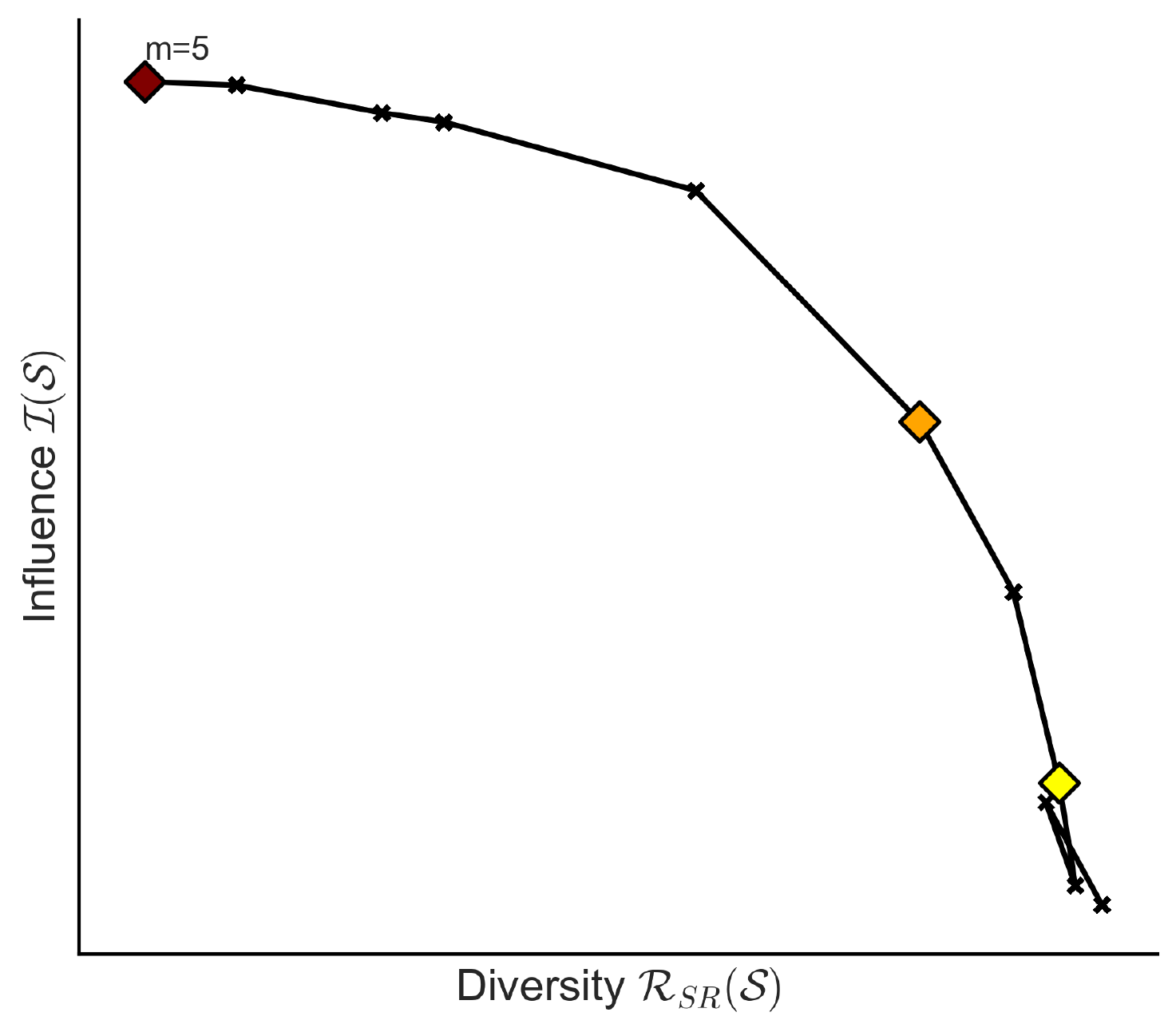}
            \caption{FashionMNIST}
            \label{5loc}
    \end{subfigure}
    \caption{For four datasets and $m = 5$, we characterize the influence-diversity trade-off. In~\ref{5loc}, we show a trade-off curve for FashionMNIST. Notice that it has a similar shape to~\ref{trade}, even though the model type is a CNN not LR and the data type is image not tabular.}\label{fig:other_data_5}
\end{figure*}

\begin{figure*}[htb]
\centering
    \begin{subfigure}[b]{0.245\linewidth}
        \includegraphics[width=\textwidth]{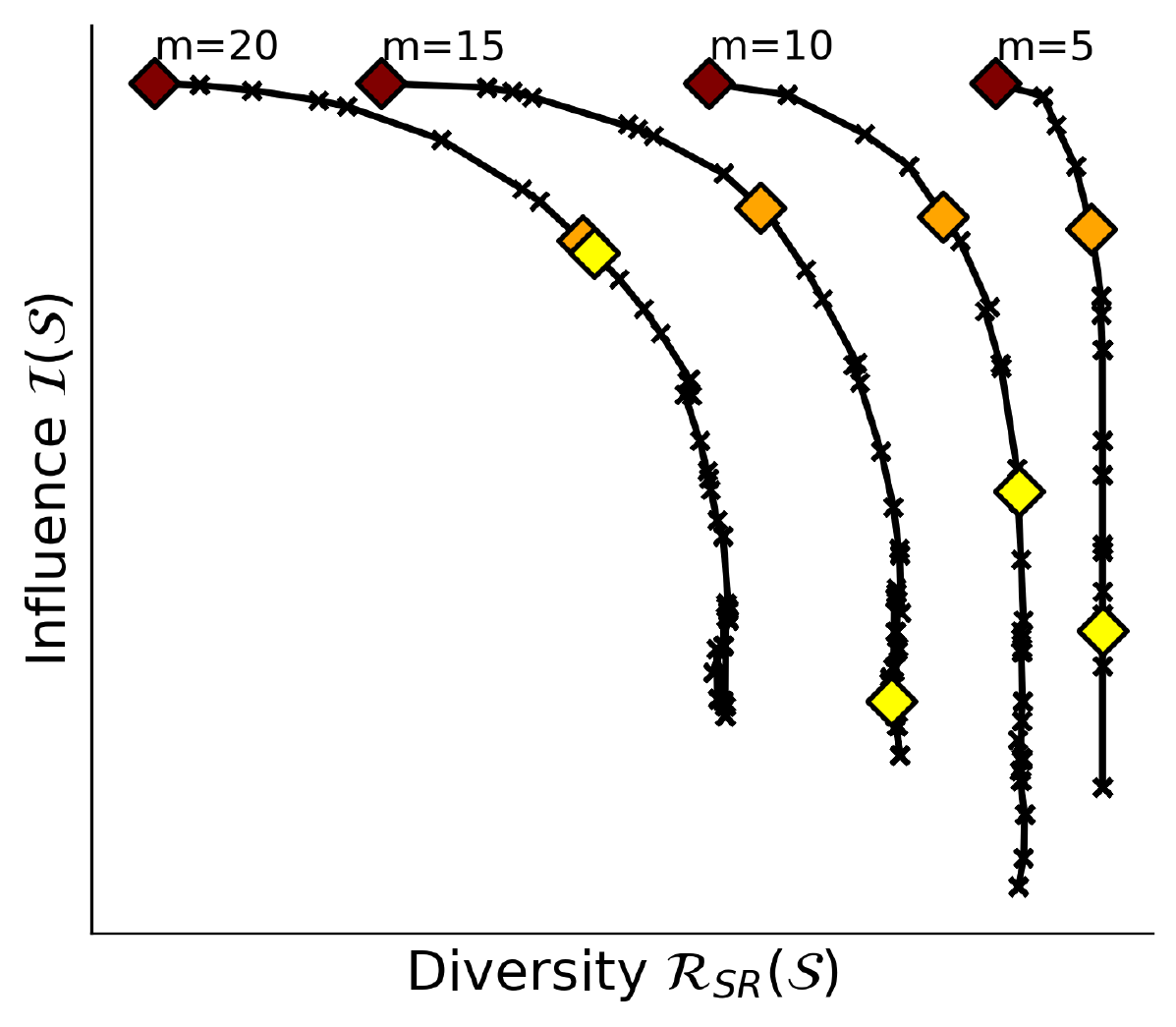}
        \caption{LSAT}
    \end{subfigure}
    \begin{subfigure}[b]{0.245\linewidth}    
        \includegraphics[width=\textwidth]{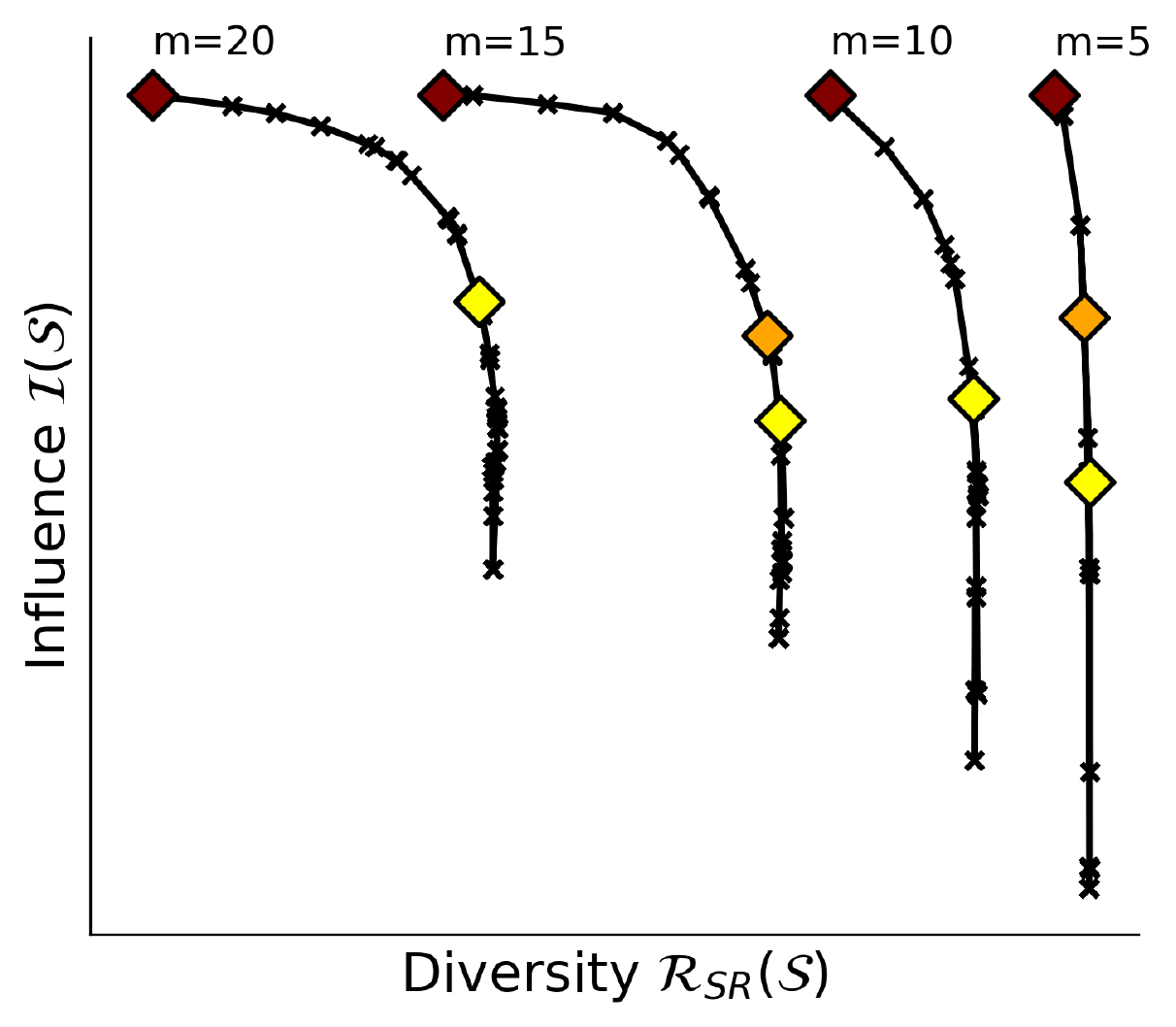}
        \caption{COMPAS}
    \end{subfigure}
    \begin{subfigure}[b]{0.245\linewidth}
        \includegraphics[width=\textwidth]{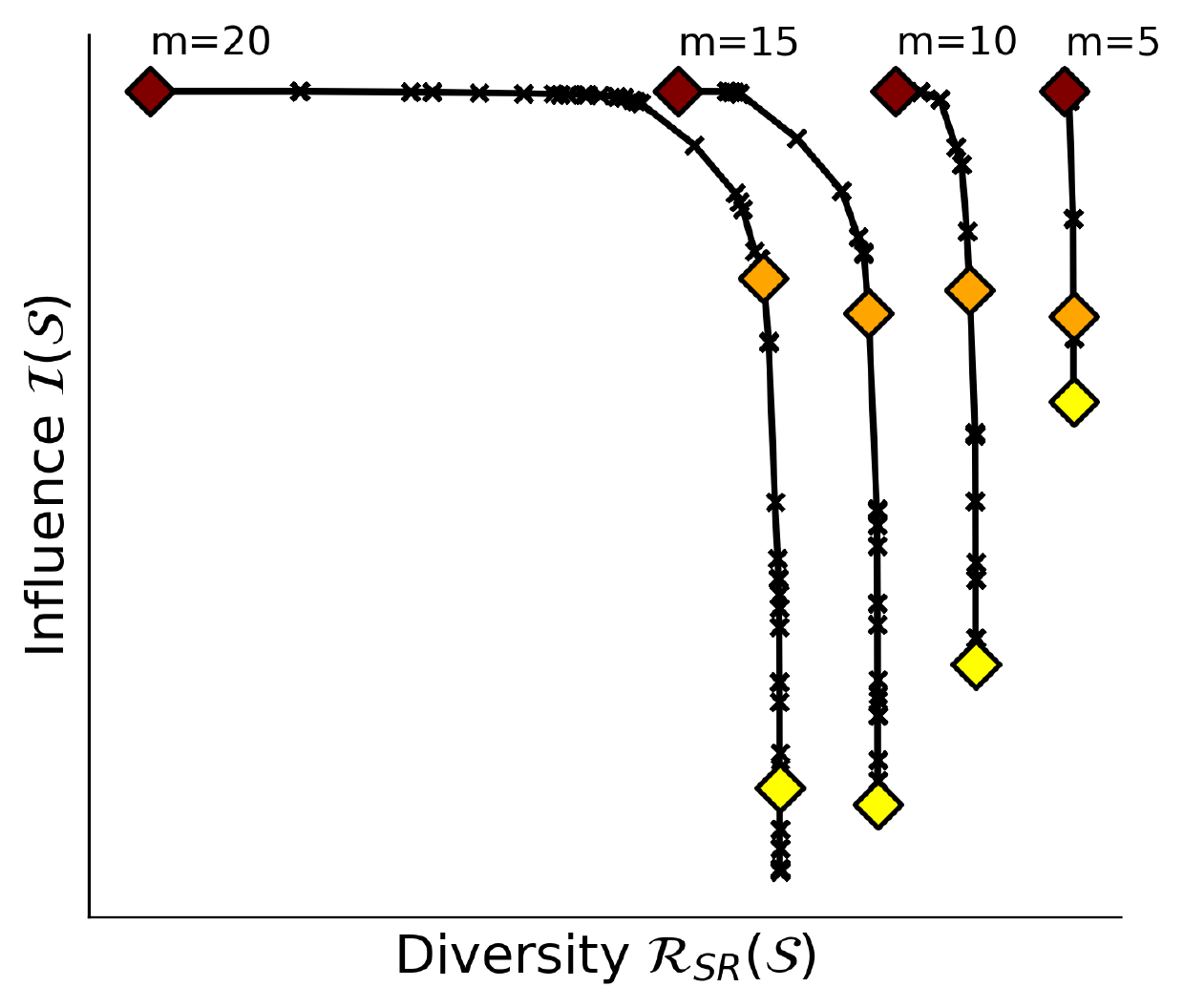}
            \caption{Adult}
    \end{subfigure}%
    \begin{subfigure}[b]{0.245\linewidth}        
            \includegraphics[width=\textwidth]{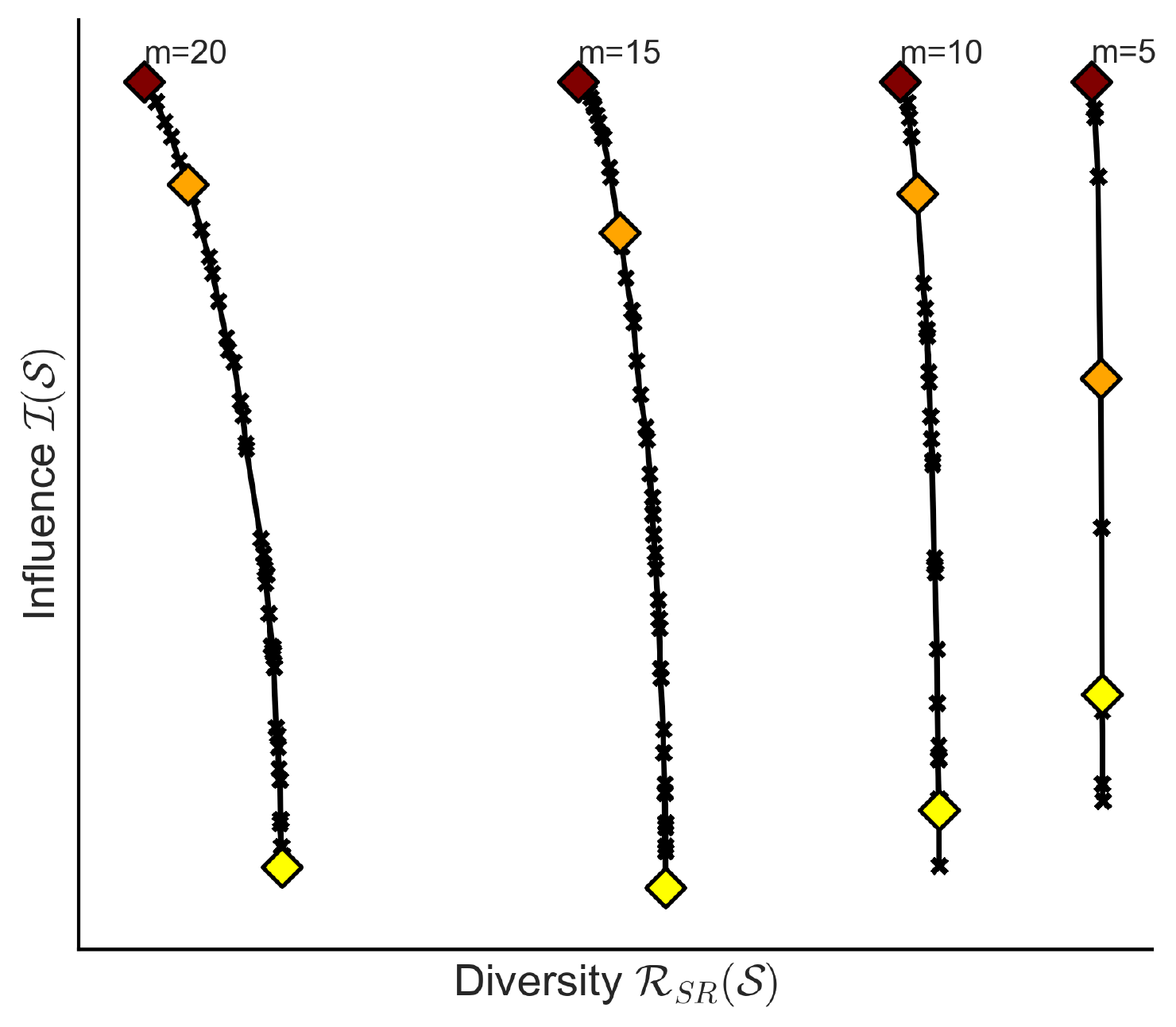}
            \caption{FashionMNIST}
    \end{subfigure}
    \caption{For four datasets and four values of $m = \{5,10,15,20\}$, we characterize the influence-diversity trade-off. The red diamond indicates where top IF points lie. The orange diamond is where $10\%$ of the influence has been foregone for diversity. The yellow diamond is where the average pairwise distance between DIVINE points is maximized. With COMPAS, we find that the orange and yellow diamonds coincide for multiple values of $m$. Recall that just because the lines look linear does not mean that they do not resemble the curve shown in Figure~\ref{fig:other_data_5}. Even with FashionMNIST where average pairwise distance in input space might not be meaningful, we find that our curves hold, as a gamma selection strategy.}\label{fig:other_data}
\end{figure*}

We show how the trade-off curves look for various $m$ from various datasets: LSAT, COMPAS, Adult, and FashionMNIST. We use IF as our influence measure, $\mathcal{R}_\text{SR}$ as our diversity function, and$f_\text{loss}$ as our evaluation function.  In Figure~\ref{fig:other_data_5}, we report the trade-off curves for when $m=5$.
In Figure~\ref{fig:other_data}, we further illustrate the flexibility of our approach to obtain DIVINE points under multiple changes: in model type, in input dimensions, and in explanation size $m$. 

\subsection{DIVINE for Image Classifiers}
\begin{figure}
    \centering
    \includegraphics[width=0.9\textwidth]{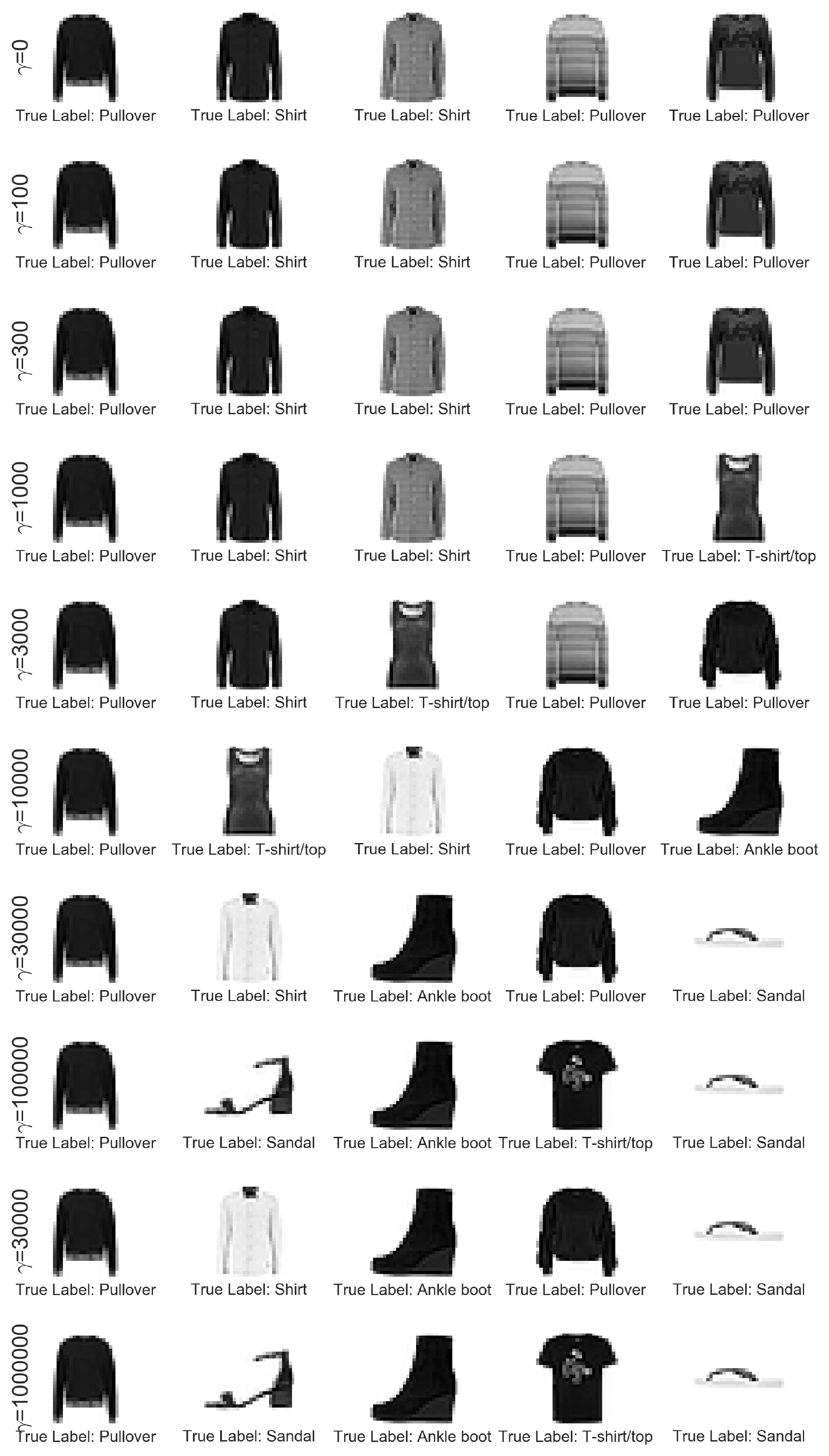}
    \caption{We show the top-$5$ points when trading off influence (IF with $f_\text{loss}$) with $\mathcal{R}_\text{SR}$.}
    \label{fig:fmnist_sweep}
\end{figure}

In the main paper, we discuss how to find DIVINE points for a CNN trained on FashionMNIST. In Figure~\ref{fig:fmnist_sweep}, we show the top-$5$ DIVINE points as we sweep over $\gamma$.  When finding the $\gamma$ that maximizes average pairwise distance, we select $30000$. These DIVINE points are used in one of our user studies.

\begin{figure*}[ht]
\centering
    \begin{subfigure}[b]{0.49\linewidth}
    \centering
    \includegraphics[width=0.7\linewidth]{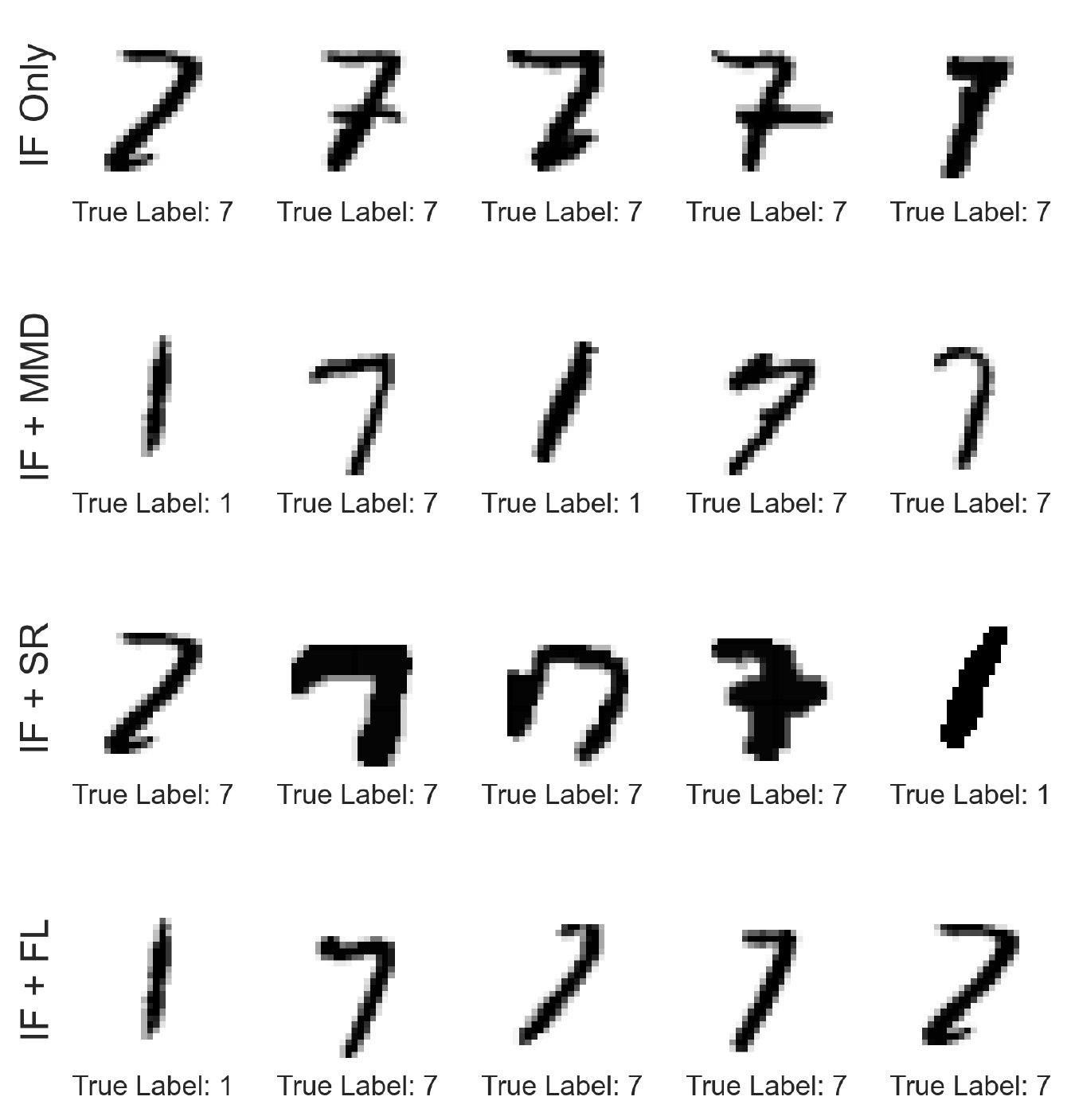}
            \caption{Logistic Regression}
            \label{fig:global_mnist_lr}
    \end{subfigure}
    \begin{subfigure}[b]{0.49\linewidth}
            \centering
          \includegraphics[width=0.7\linewidth]{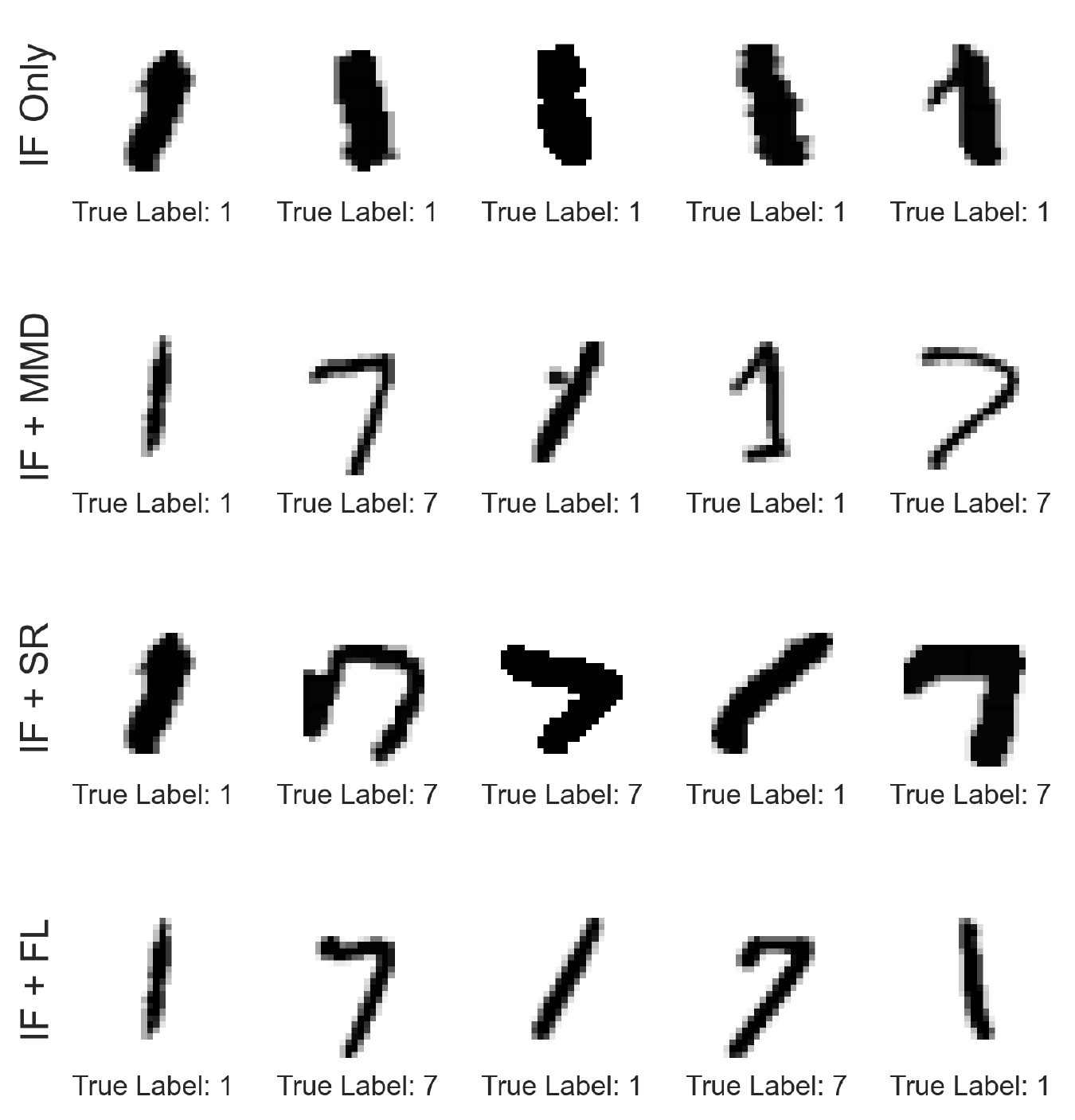}
            \caption{MLP}
            \label{fig:local_mnist_lr}
    \end{subfigure}
    \caption{We show that top DIVINE points for LR and a MLP differ. Most important point for each method in leftmost position. Note that the last three rows are three potential sets of DIVINE points all with IF as the influence measure but with the diversity function varying.}\label{fig:g_mnist_lr}
\end{figure*}

\begin{figure*}[ht]
\centering
    \begin{subfigure}[b]{0.49\linewidth}
    \centering
    \includegraphics[width=.6\linewidth]{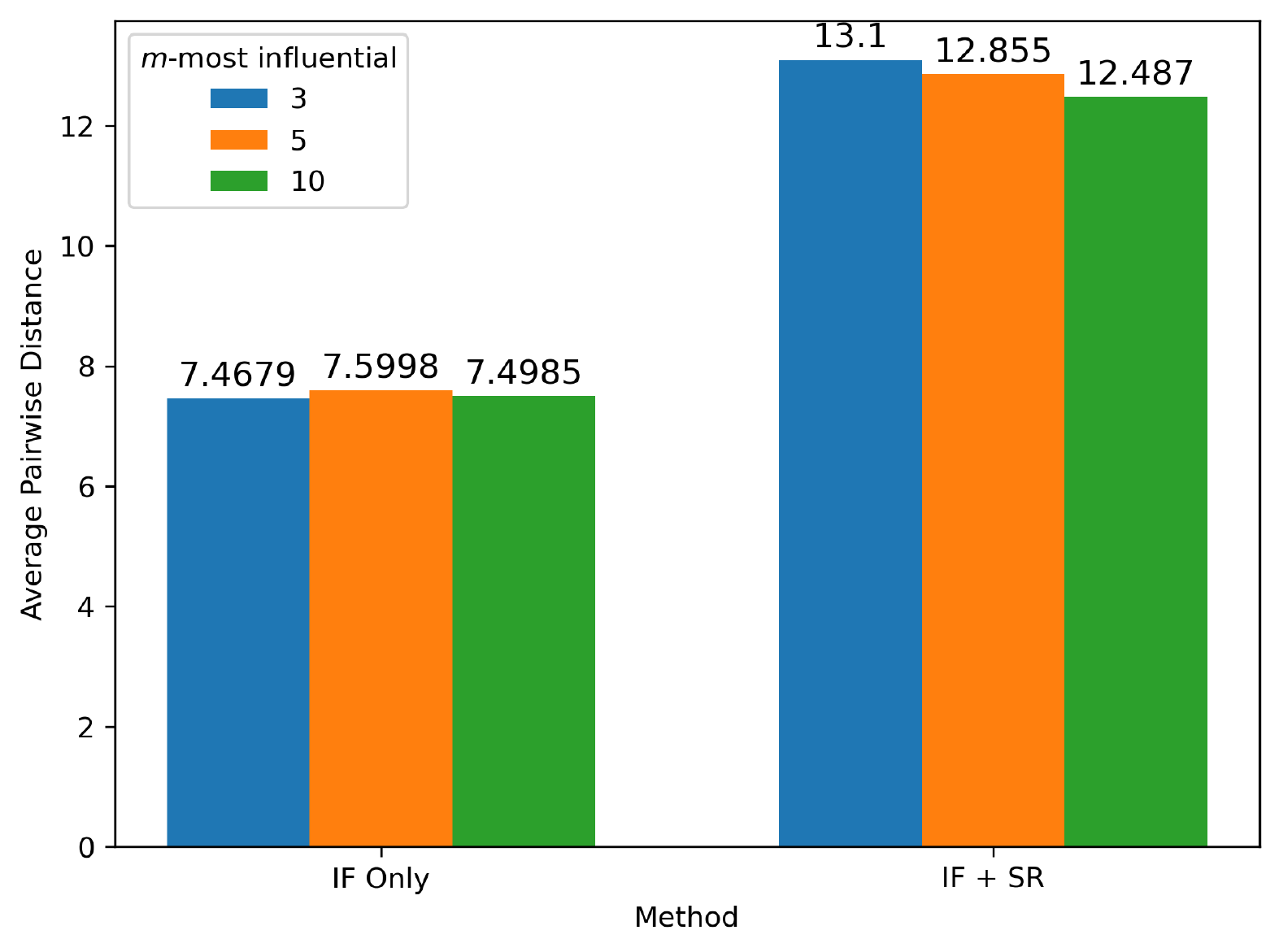}
            \caption{Logistic Regression}
    \end{subfigure}
    \begin{subfigure}[b]{0.49\linewidth}
            \centering
          \includegraphics[width=.6\linewidth]{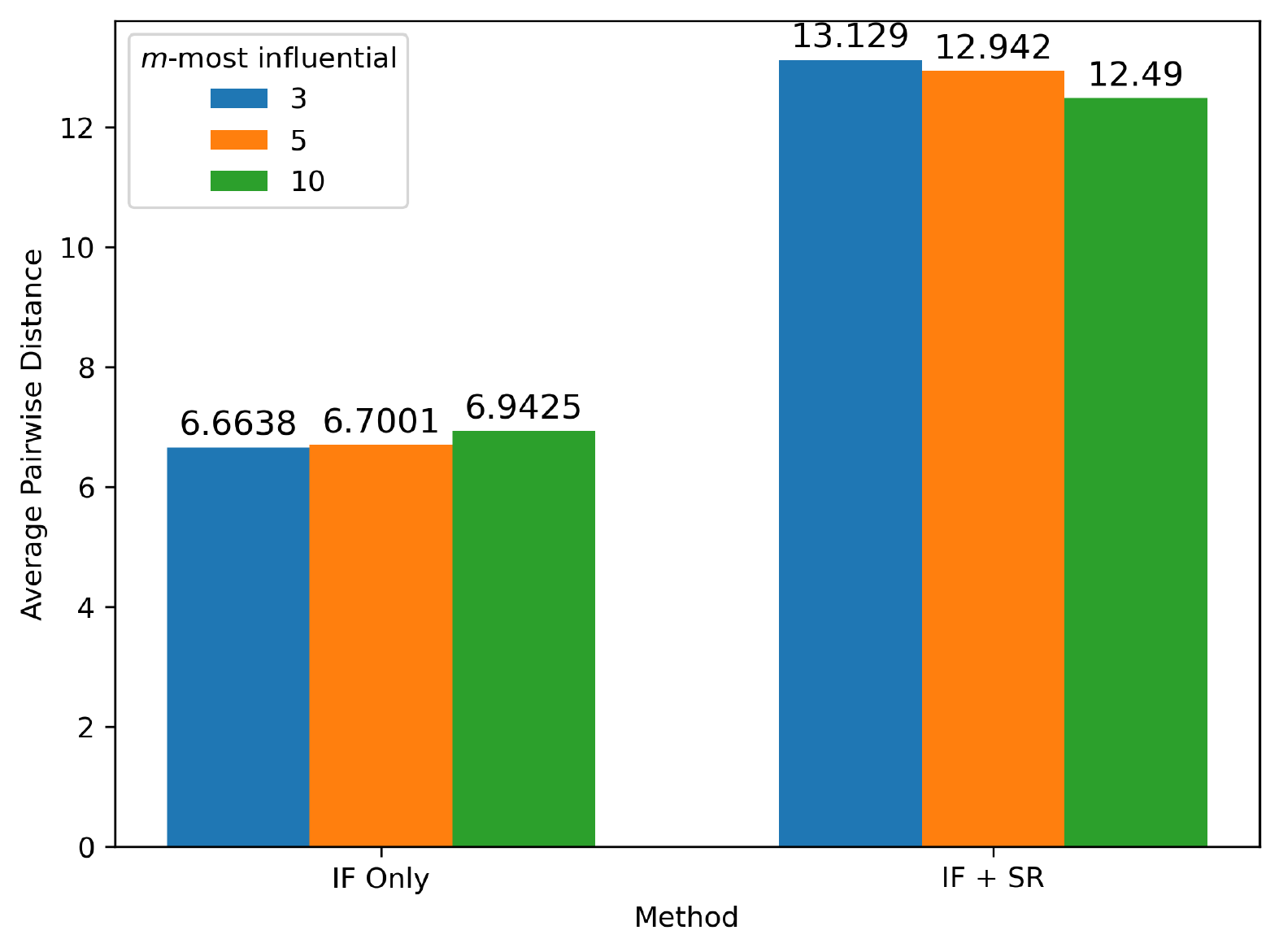}
            \caption{MLP}
    \end{subfigure}
    \caption{Average pairwise distance between DIVINE points exceeds that of IF alone for both models. The $m$-sized explanations are found $100$ random test points and then averaged. }\label{fig:barplot}
\end{figure*}

\begin{figure*}[ht]
\centering
    \begin{subfigure}[b]{0.49\linewidth}
    \centering
    \includegraphics[width=0.7\linewidth]{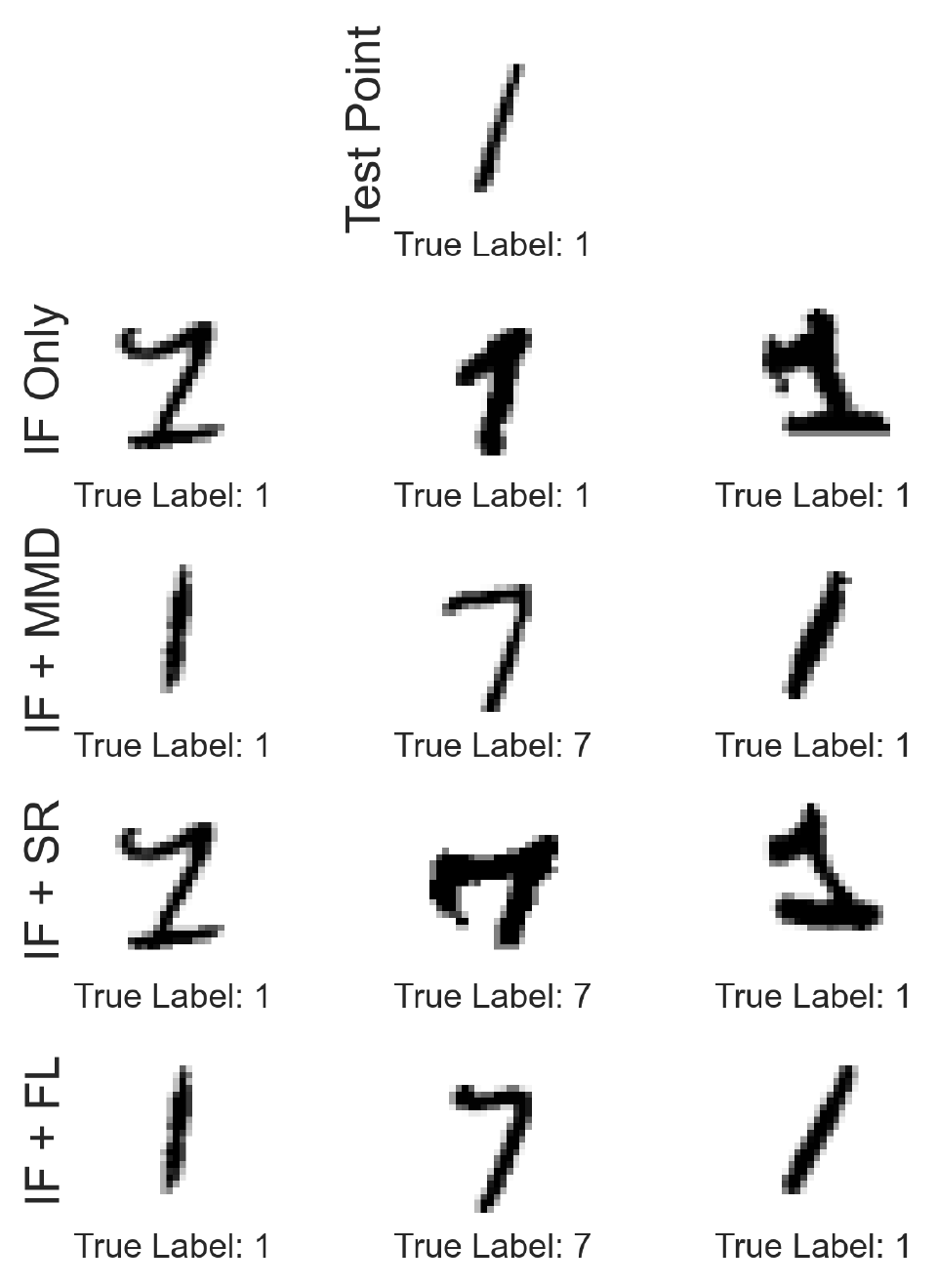}
            \caption{Logistic Regression}
    \end{subfigure}
    \begin{subfigure}[b]{0.49\linewidth}
            \centering
          \includegraphics[width=0.7\linewidth]{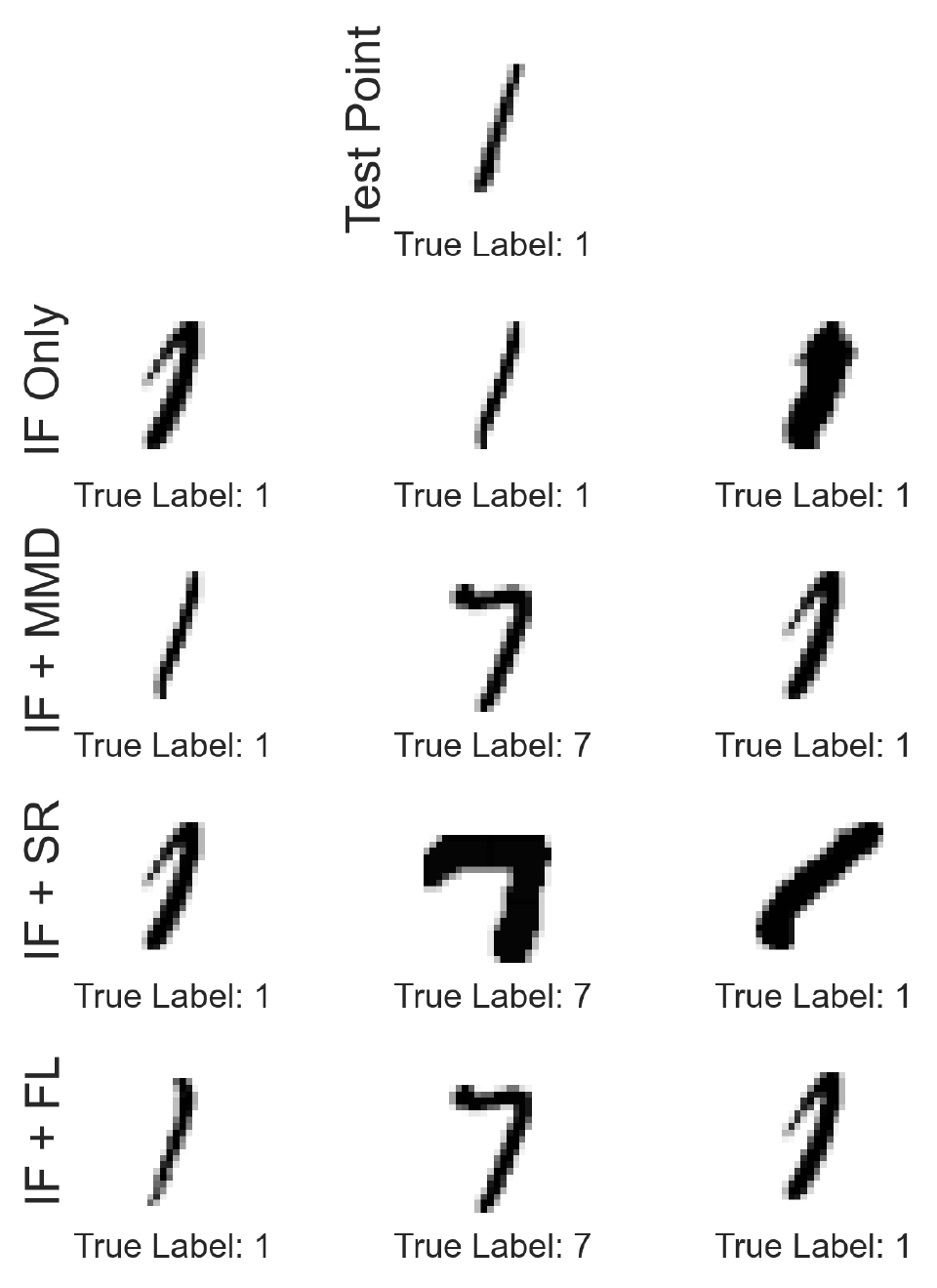}
            \caption{MLP}
    \end{subfigure}
    \caption{We show that the influential samples selected by IF are less diverse than the ones we select, when locally explaining a test point, for both logistic regression and a MLP. Note both of the test points are correctly classified. Note that the last three rows are three potential sets of DIVINE points all with IF as the influence measure but with the diversity function varying.}\label{fig:mnist_lr}
\end{figure*}

While we report results for a CNN trained on FashionMNIST in the main text, we compare the top influential and DIVINE points from a MLP and from a simpler logistic regression classifier trained on MNIST. In Figure~\ref{fig:g_mnist_lr}, we show the most influential points to an entire test set for our logistic regression classifier. Herein we value points with rest to $f_\text{loss}$. Note how IF alone does contain label diversity: it simply captures two canonical sevens (ones with lines in the middle and ones without). As sanity check, we train a logistic regression classifier on the top $m$ selected points by each method and report accuracies. The LR classifier trained on the MNIST task used $13007$ training points and achieved a test accuracy of $99.72\%$. We use a validation set and $f_\text{loss}$ to select $100$ important points via IF and IF+ diversity methods. We find that the model trained on points selected by IF has a test accuracy of $48.19\%$. With DIVINE points, we find that IF+MMD gets $65.00\%$, IF+FL gets $48.75\%$, and IF+SR gets $99.44\%$. Ergo, our method allows us to select a subset of points important for model performance.
In Figure~\ref{fig:barplot}, we find that the average pairwise distance between DIVINE points exceeds that of IF for MNIST. In Figure~\ref{fig:mnist_lr}, we show how explanations for a specific test point differ by model (LR and MLP) and by method (IF and IF + Diversity). Notice that we achieve label diversity with all three of our $\mathcal{R}$, and achieve mode diversity with $\mathcal{R}_\text{SR}$.

\clearpage
\subsection{Additional Fairness Experiments}


\subsubsection{Generalization}
\label{generalization}
\begin{figure*}[htb]
\centering
    \begin{subfigure}[b]{0.245\linewidth}
            \centering
            \includegraphics[width=\linewidth]{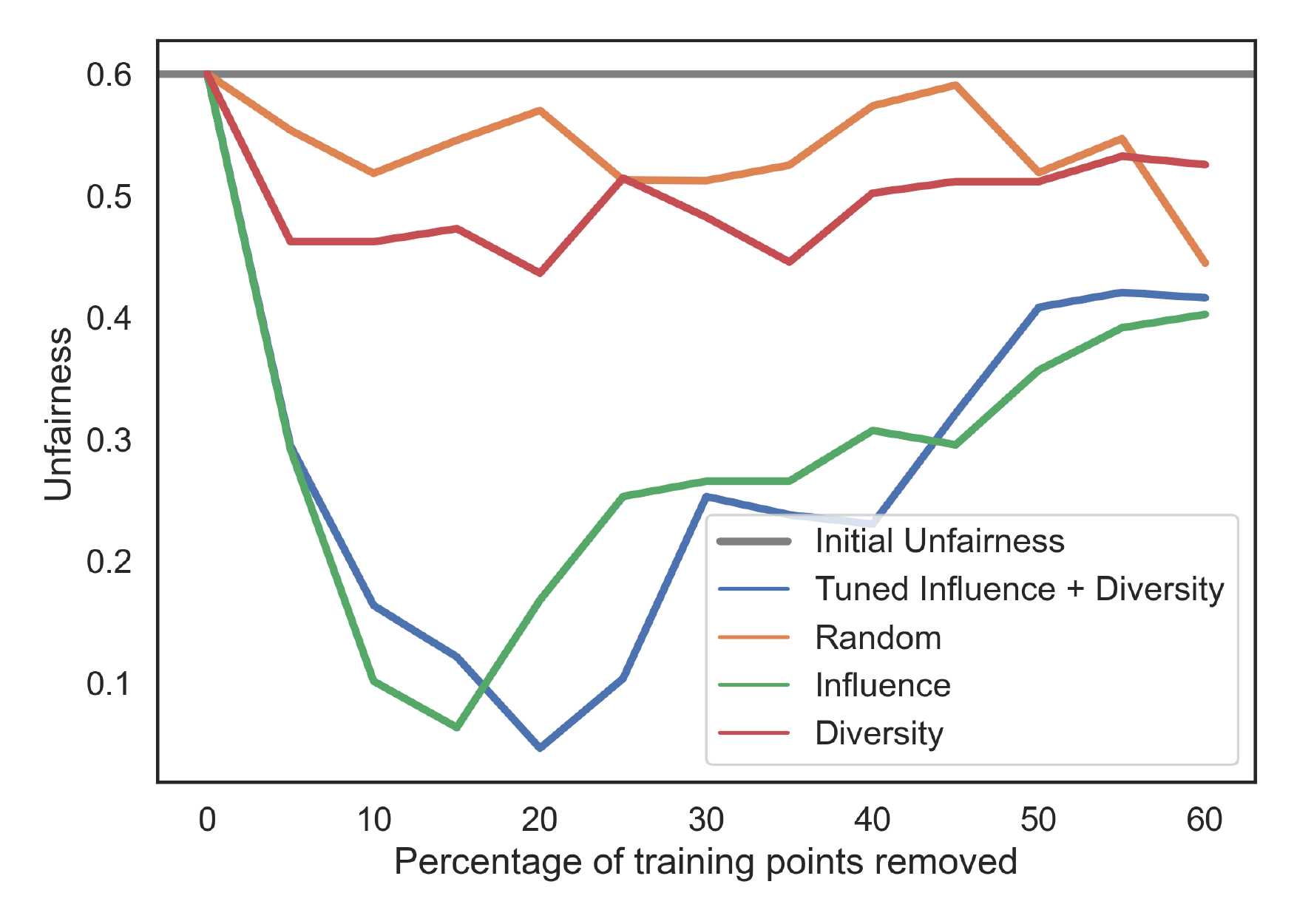}
    \end{subfigure}
    \begin{subfigure}[b]{0.245\linewidth}
        \centering
        \includegraphics[width=\linewidth]{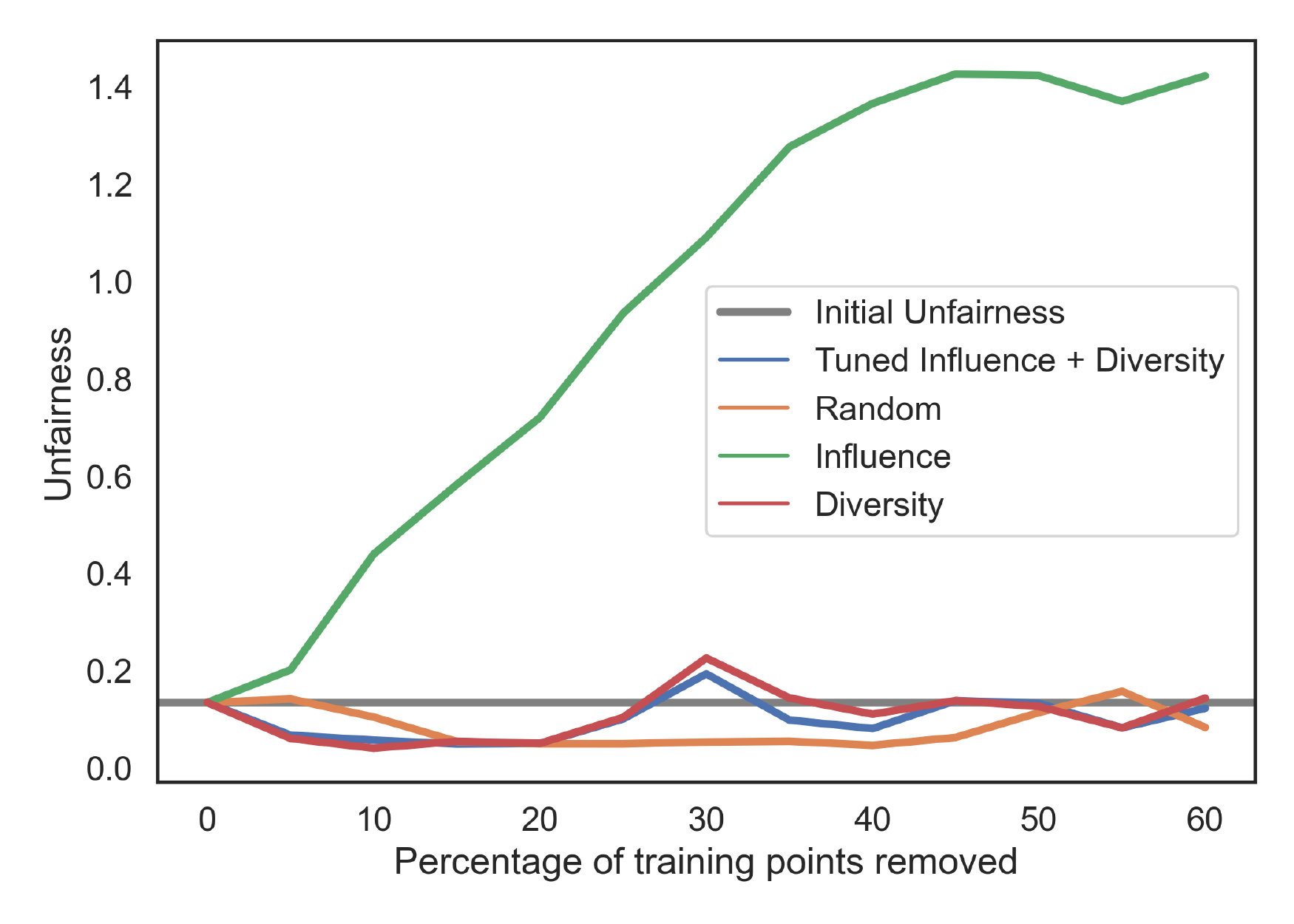}
    \end{subfigure}
    \begin{subfigure}[b]{0.245\linewidth}
            \centering
            \includegraphics[width=\linewidth]{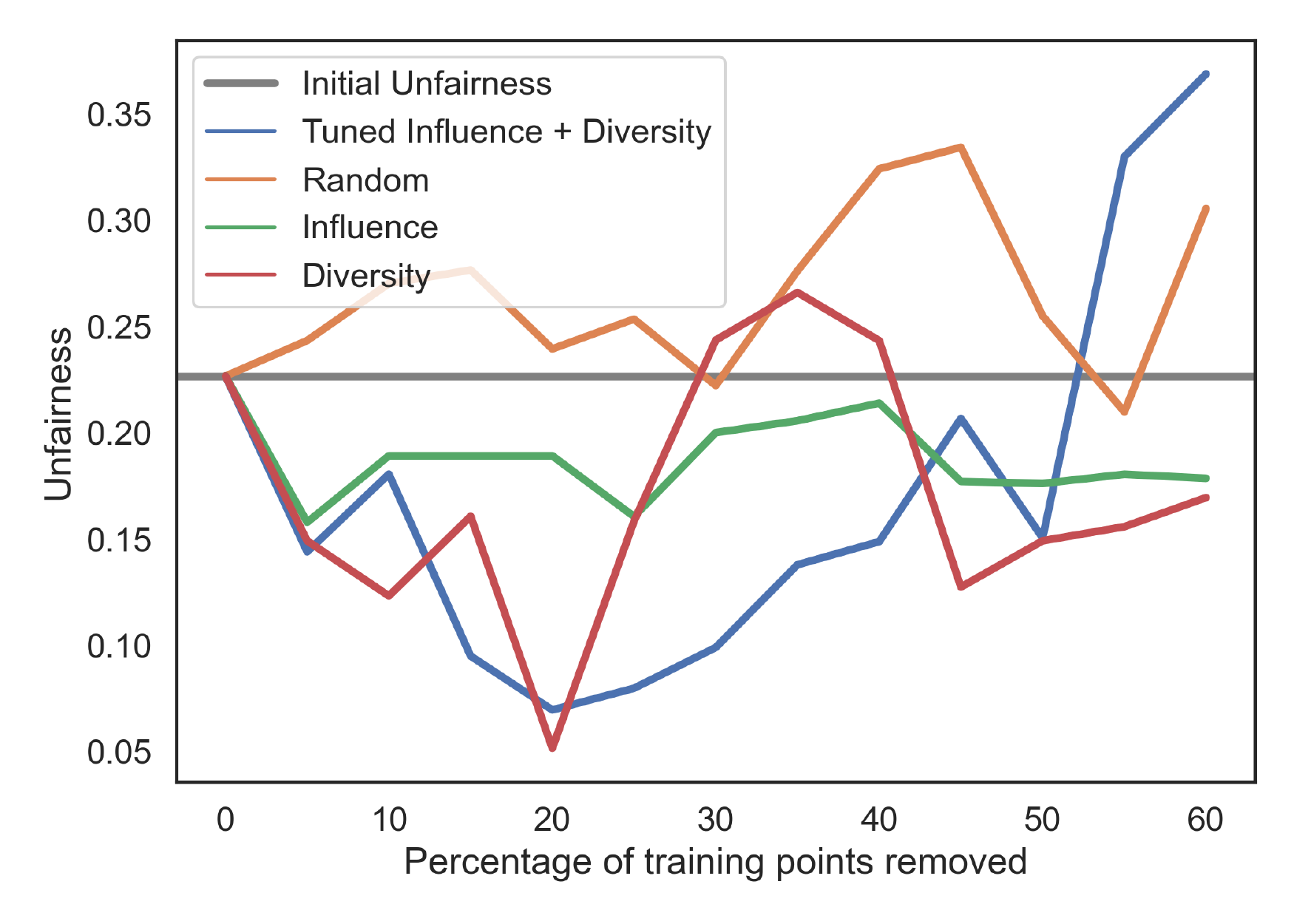}
    \end{subfigure}
    \begin{subfigure}[b]{0.245\linewidth}
            \centering
            \includegraphics[width=\linewidth]{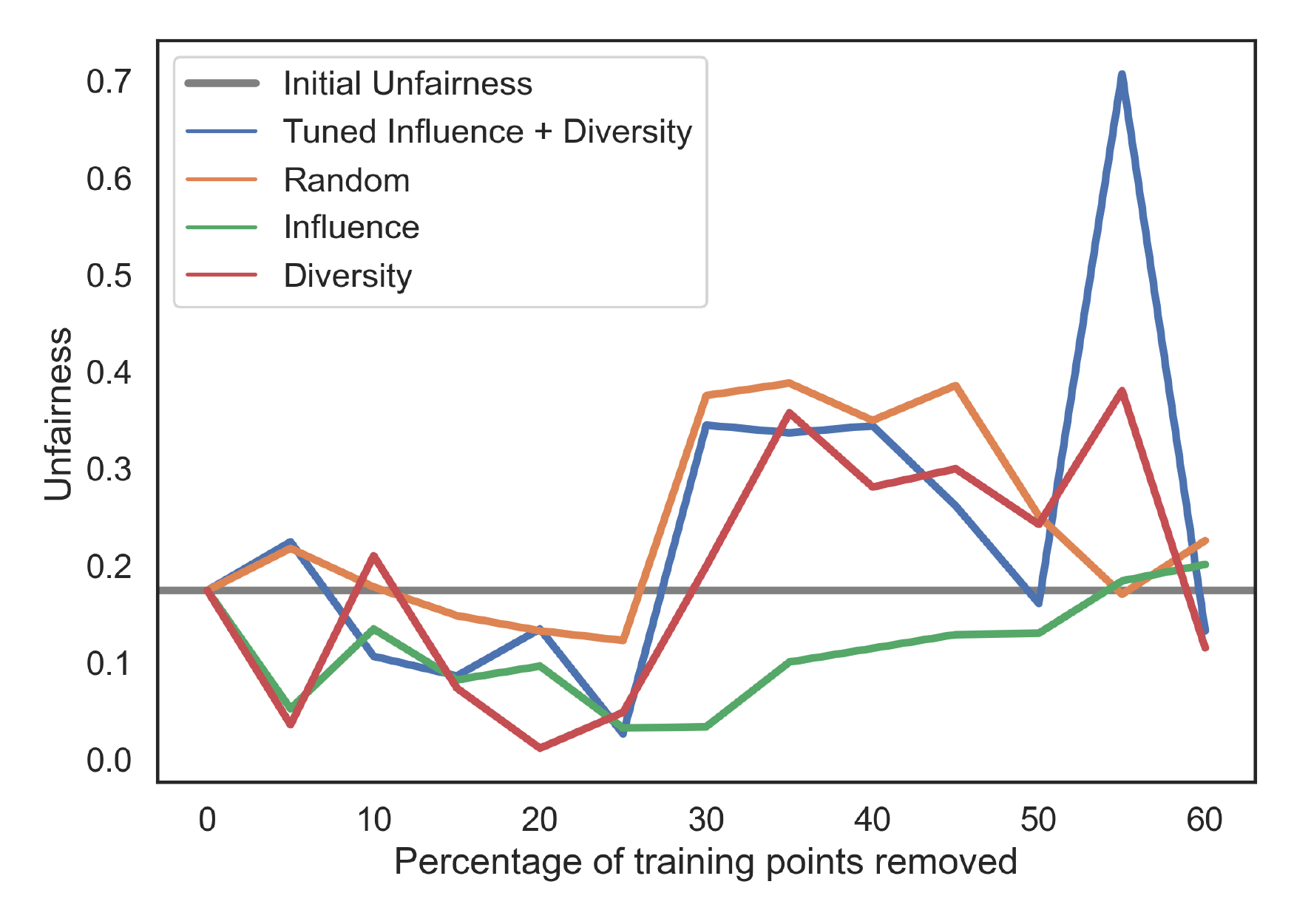}
    \end{subfigure}\\
    \begin{subfigure}[b]{0.245\linewidth}    
            \includegraphics[width=\linewidth]{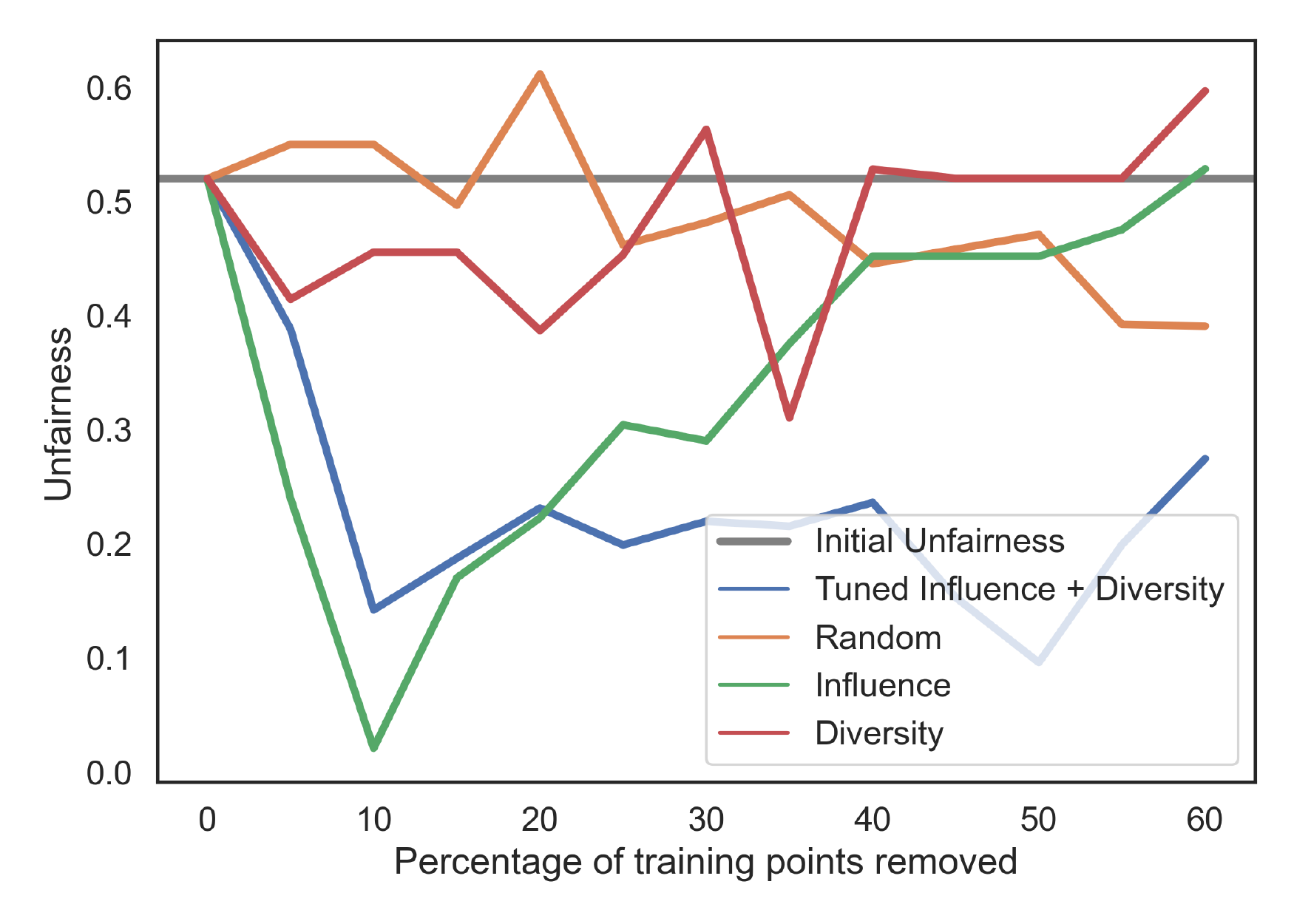}
            \caption{COMPAS}
    \end{subfigure}%
    \begin{subfigure}[b]{0.245\linewidth}    
            \includegraphics[width=\linewidth]{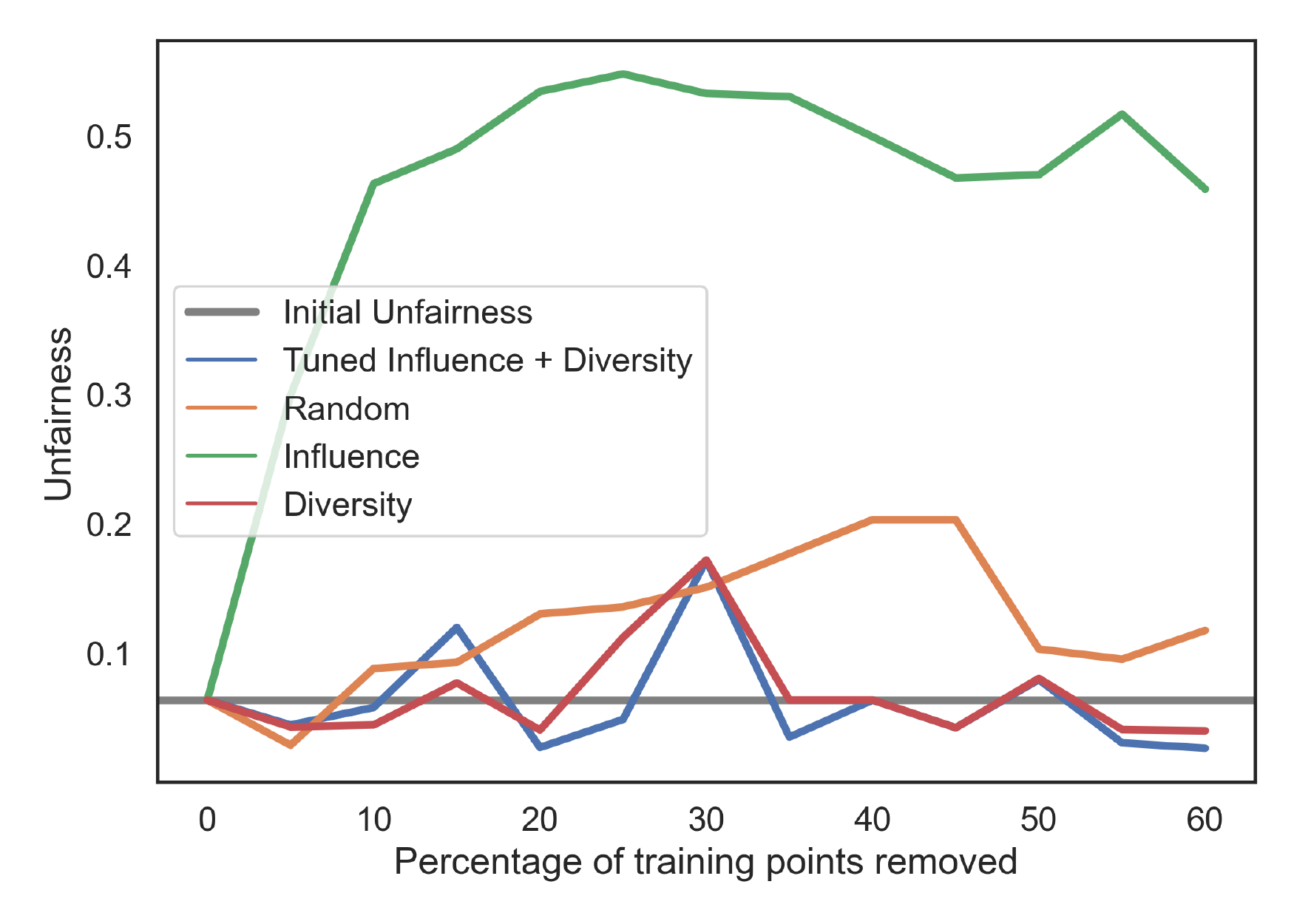}
            \caption{LSAT}
    \end{subfigure}%
    \begin{subfigure}[b]{0.245\linewidth}    
            \includegraphics[width=\linewidth]{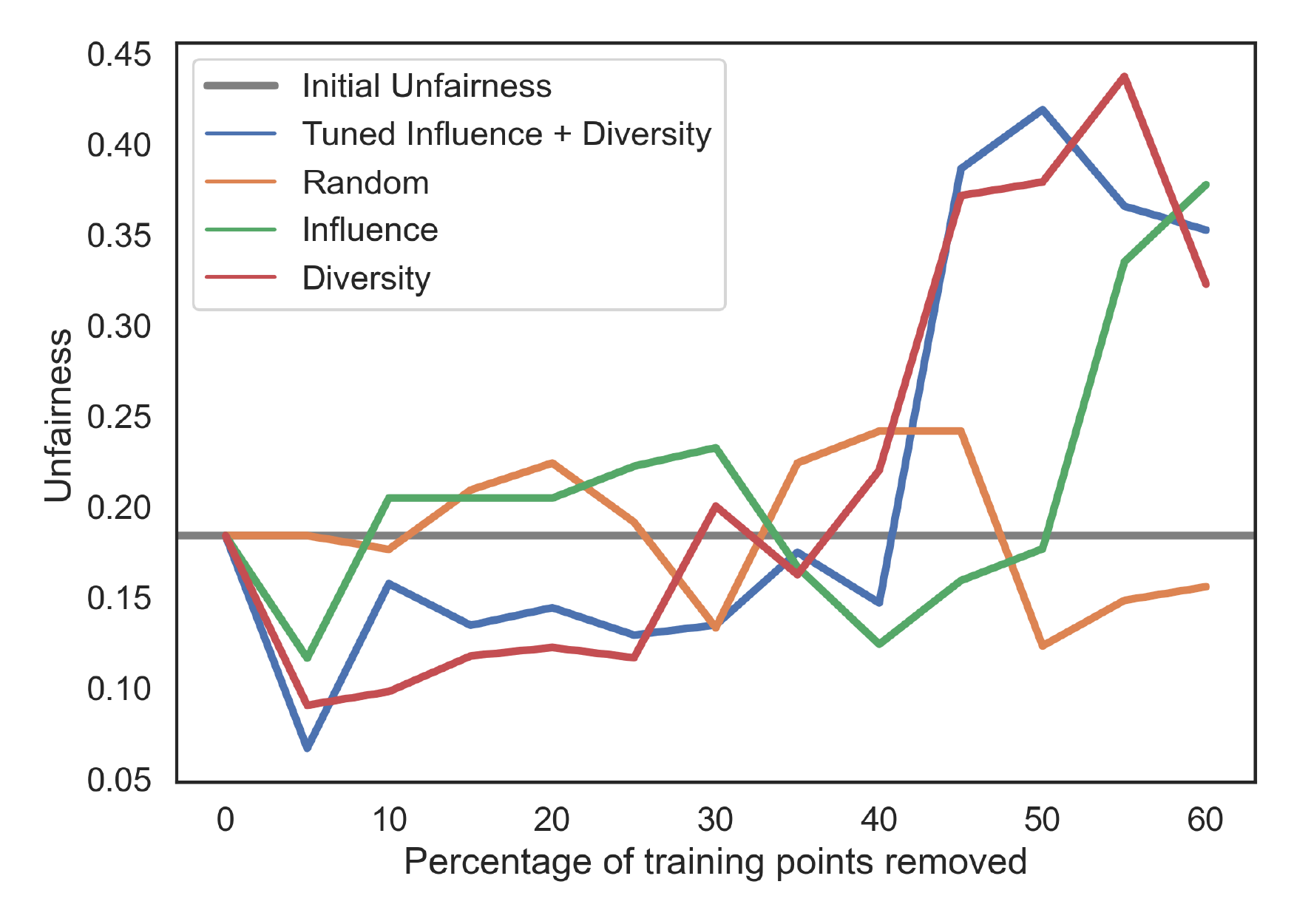}
            \caption{Adult}
    \end{subfigure}%
    \begin{subfigure}[b]{0.245\linewidth}    
            \includegraphics[width=\linewidth]{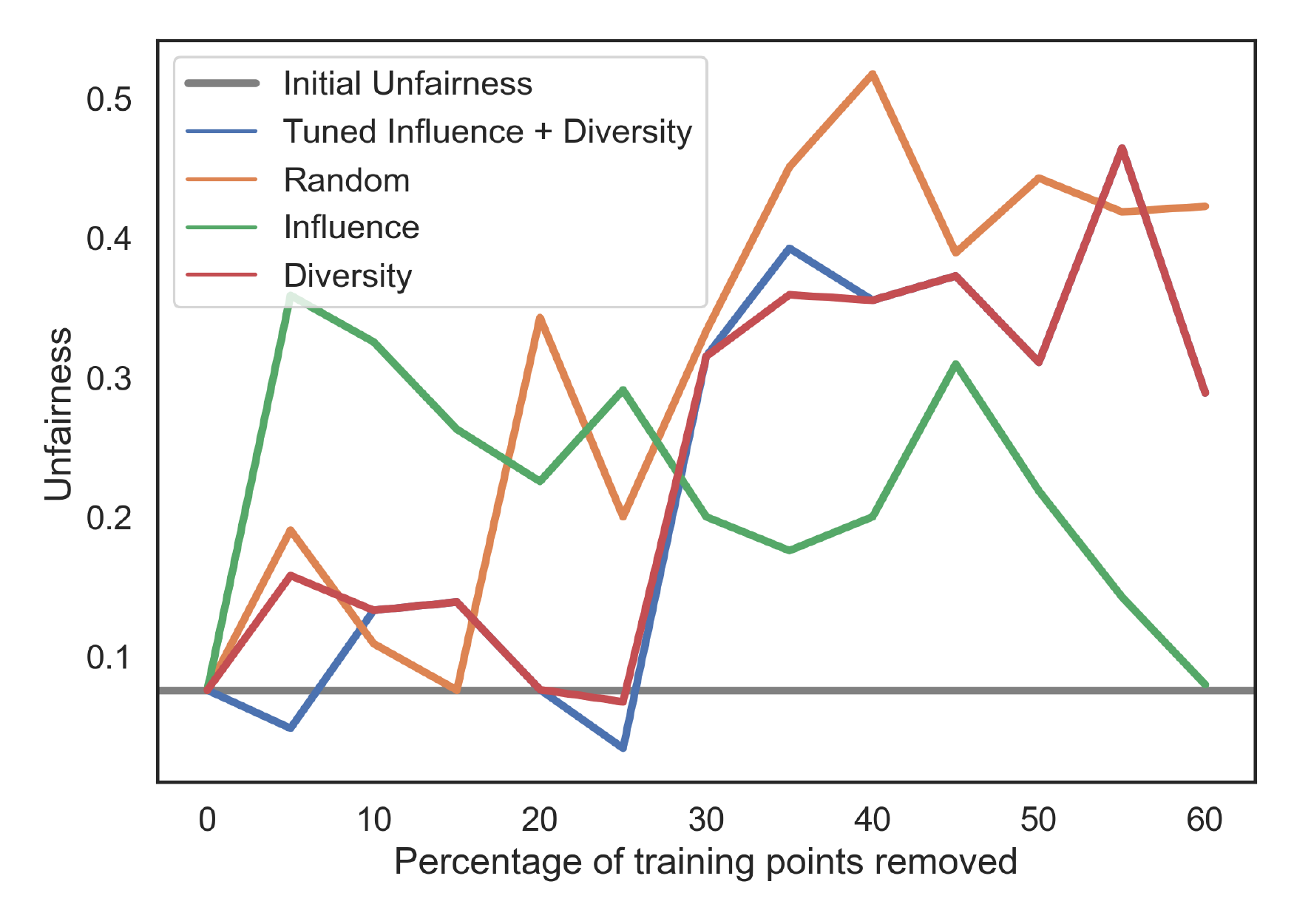}
            \caption{Bank}
    \end{subfigure}%
    \caption{Impact of removing unfairness-inducing points on generalization. In the top row, we score each training point based the change in $f_\text{unf}$ on the training dataset upon removing the point. We report unfairness on a test dataset, after removing $5\%$ of the most unfairness inducing points (selected based on methods differing by color) at a time. In the bottom row, we score each training point based the change in $f_\text{unf}$ on a separate validation dataset upon removing the point. We then report unfairness on a held-out test dataset.}
    \label{fig:gen_all}
\end{figure*}

While practitioners might be interested in understanding the effect of training data points with respect to metrics evaluated on the training data itself, practitioners may also like to achieve better generalization. We report the effect of removing points scored with respect to training unfairness on the test data in the top row of Figure~\ref{fig:gen_all}. In this section, we use $f_\text{unf}$ as our evaluation function and use influence functions as our influence measure.
Notice that we are able to achieve a lower unfairness in generalization on all datasets with either points selected based on importance scores alone or with DIVINE-selected points, which incorporate a diversity term. We observe unexpected results with LSAT removal and the influence score alone: this is likely due to the low-dimensionality and high redundancy in the dataset. We posit that there only exist a handful of unfairness inducing points that need to be removed.

Moreover, practitioners may be interested in scoring points specifically for generalization. We can score training data points based on their impact on the unfairness of a held-out, validation dataset (i.e., $f_\text{unf}$ is calculated on validation data) and then measure the impact removing unfairness-inducing points on a separate test set. 
We demonstrate this approach in the bottom row of Figure~\ref{fig:gen_all}. As expected, the first few percentages of points removed lead to a decrease in model unfairness. For every dataset, removing DIVINE points (blue line) outperforms removing points at random, by only their importance score ($\mathcal{I}(\mathcal{S})$), or by only their diversity ($\mathcal{R}_\text{SR}$ in this case).

A practitioner can define a stopping criterion for the removal of unfairness-inducing points. 
In Table~\ref{num_miss}, we report the number of unfairness-inducing points highlighted by various methods. We include a column denoting the number of correctly classified points our method has identified as unfair to show that it does not simply recommend misclassified points for removal. Unfairness-inducing points can also be correctly classified points that change the decision boundary significantly upon removal, such that unfairness is reduced while accuracy is maintained.

\subsubsection{Removal with Recalculation}
Instead of simply calculating the importance scores once (with respect to the entire training dataset), we can recalculate importance scores after the removal of each set of $m$ points. Although computationally expensive, a practitioner may elect to do this to avoid divergence from additivity for large $m$ (as described in Appendix~\ref{linearity}) or to improve generalization. To improve generalization, we may also calculate importance scores with respect to a validation data set and report performance on a test data set (as described in Appendix~\ref{generalization}). In Figure~\ref{fig:all_recalculate}, we use both of these approaches to improve generalization, and report the removal with recalculation results for the first $10$ training points removed, with importance scores recalculated after every single point removed ($m=1$). For LSAT, Adult, and Bank, we notice that unfairness decreases steadily as we remove the most harmful data point based on newly calculated influence scores. However, for COMPAS, after $2$ iterations, our algorithm no longer identifies unfair points (importance scores are all 0 or greater) and so accuracy and unfairness both remain constant. 

\begin{table}[htb]
\centering
\begin{tabular}{c|c|c|cHHHH}
\toprule
         Dataset & \makecell*{Importance\\Method} &\makecell*{$\#$ Unfairness\\Inducing\\ Points} &  \makecell*{$\#$ Correct} & Train Acc. & Train Unf.  & Test Acc. & Test Unf. \\ 
 \midrule
\multirow{3}{*}{Adult}  
&LOO & $87$  &  $26$ & $0.85$ & $0.05$ & $0.86$ & $0.12$  \\ \cline{2-8} 
&  IF & $428$  &  $418$ & $0.84$ & $0.27$ & $0.85$ & $0.29$ \\ \cline{2-8} 
&  CFP & $432$  &  $379$ & $0.85$ & $0.23$ & $0.86$ & $0.22$ \\  \midrule
\multirow{3}{*}{Bank}  
&LOO & $66$  &  $38$ & $0.91$ & $0.11$ & $0.90$ & $0.12$  \\ \cline{2-8} 
&  IF & $305$  &  $294$ & $0.92$ & $0.49$ & $0.91$ & $0.36$ \\ \cline{2-8} 
&  CFP & $375$  &  $325$ & $0.92$ & $0.30$ & $0.90$ & $0.26$ \\  \midrule
\multirow{3}{*}{COMPAS}
&LOO & 614  &  $325$ & $0.59$ & $0.27$ & $0.59$ & $0.27$  \\ \cline{2-8} 
&  IF & $598$  &  $519$ & $0.54$ & $1.07$ & $0.52$ & $1.05$ \\ \cline{2-8} 
&  CFP & $1116$  &  $775$ & $0.66$ & $0.54$ & $0.63$ & $0.62$ \\  \midrule
\multirow{3}{*}{LSAT}
&LOO & $158$  &  $50$ & $0.59$ & $0.38$ & $0.59$ & $0.38$  \\ \cline{2-8} 
&  IF & $72$  &  $70$ & $0.60$ & $0.45$& $0.60$ & $0.35$ \\ \cline{2-8} 
&  CFP & $706$  &  $391$ & $0.60$ & $0.21$ & $0.60$ & $0.09$ \\ 
\bottomrule
\end{tabular}
\vspace{1ex}
\caption{We report the number of unfairness inducing points by each importance method. Recall unfairness inducing points are points with negative importance scores (i.e., $I_i < 0$). We also note the number of unfairness inducing points that were correctly classified by the original model.}
\label{num_miss}
\end{table}

\begin{figure*}[htb]
\centering

    \begin{subfigure}[b]{0.245\linewidth}
            \centering
            \includegraphics[width=\linewidth]{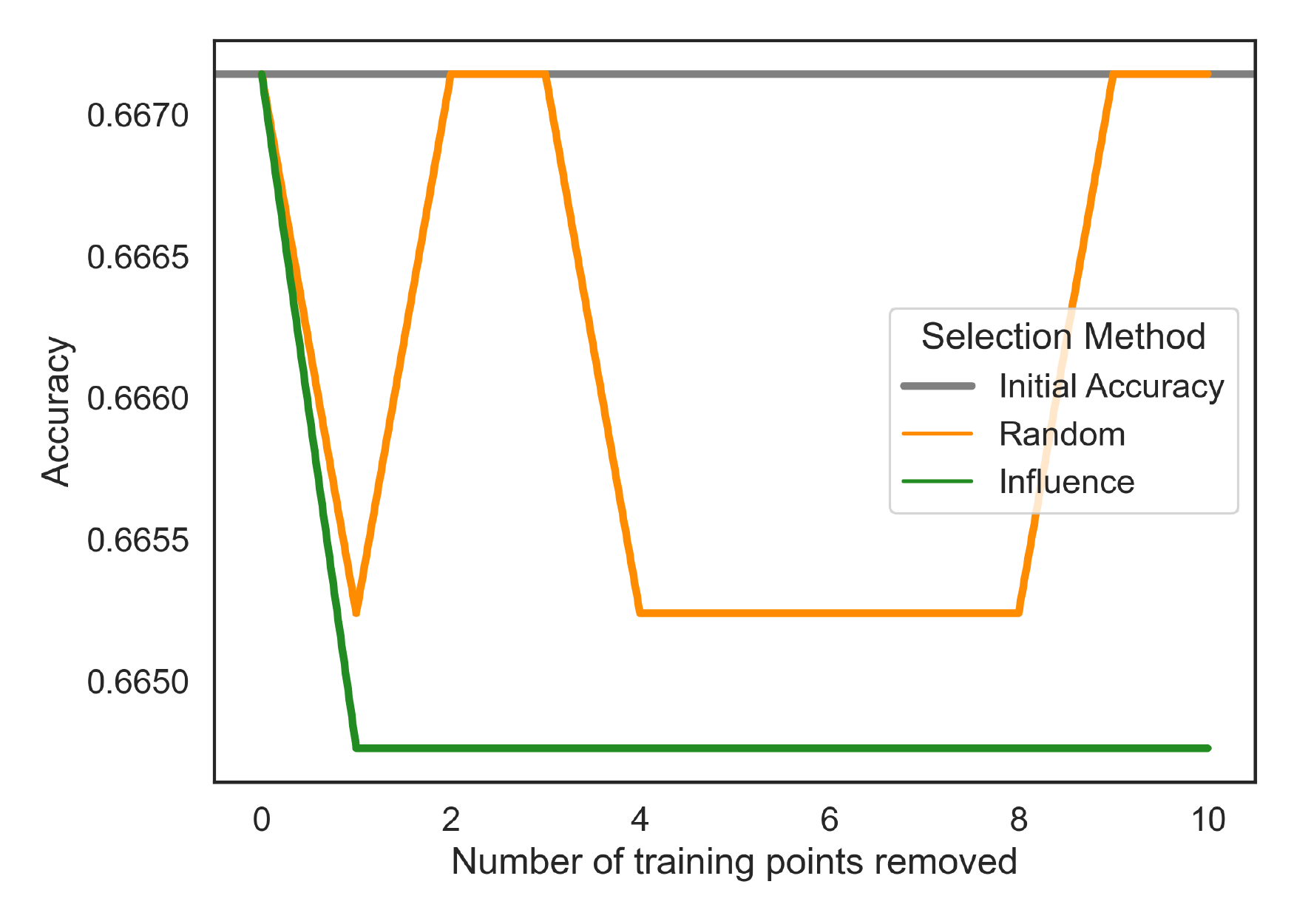}
    \end{subfigure}
    \begin{subfigure}[b]{0.245\linewidth}
            \centering
            \includegraphics[width=\linewidth]{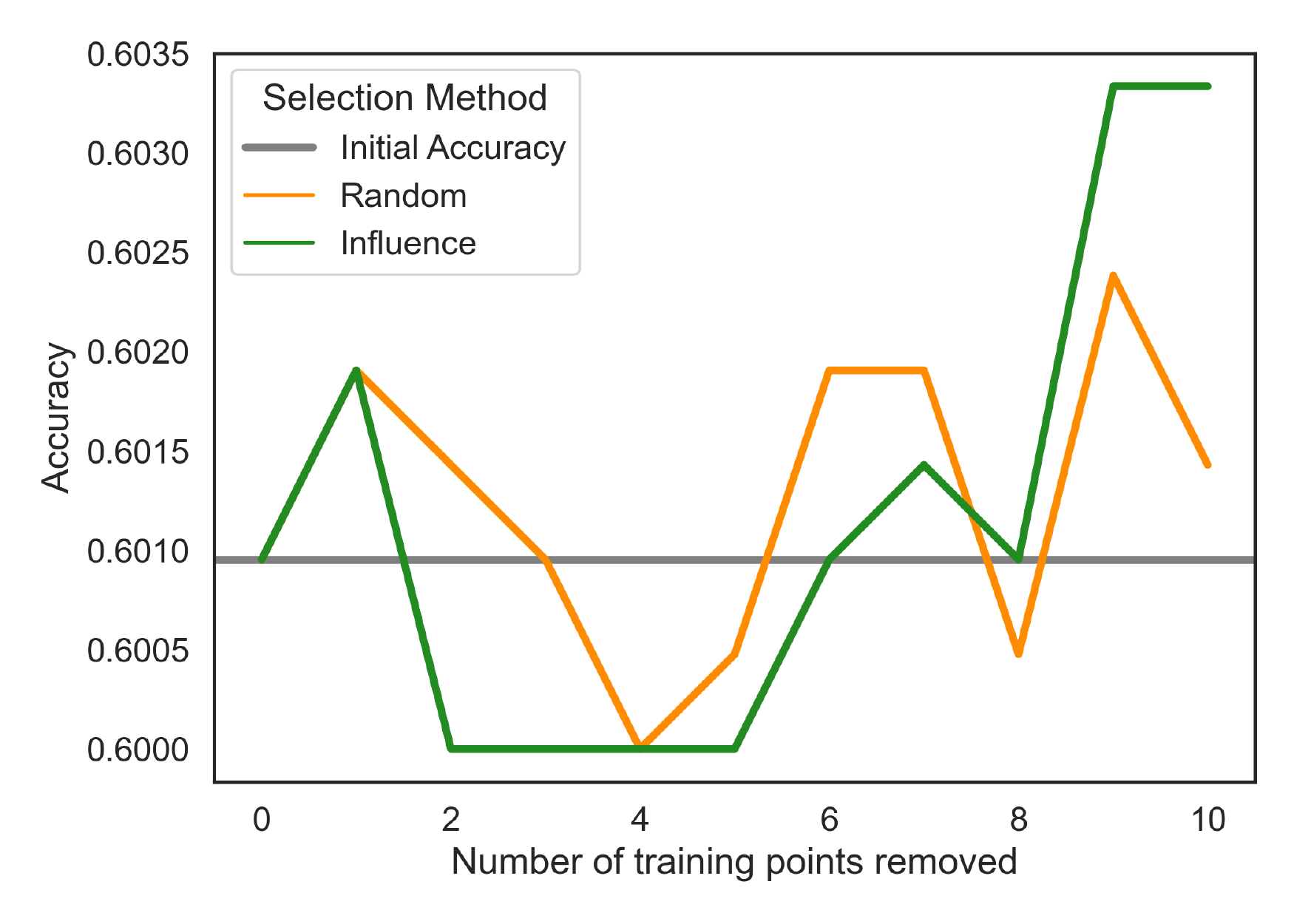}
    \end{subfigure}
    \begin{subfigure}[b]{0.245\linewidth}
            \centering
            \includegraphics[width=\linewidth]{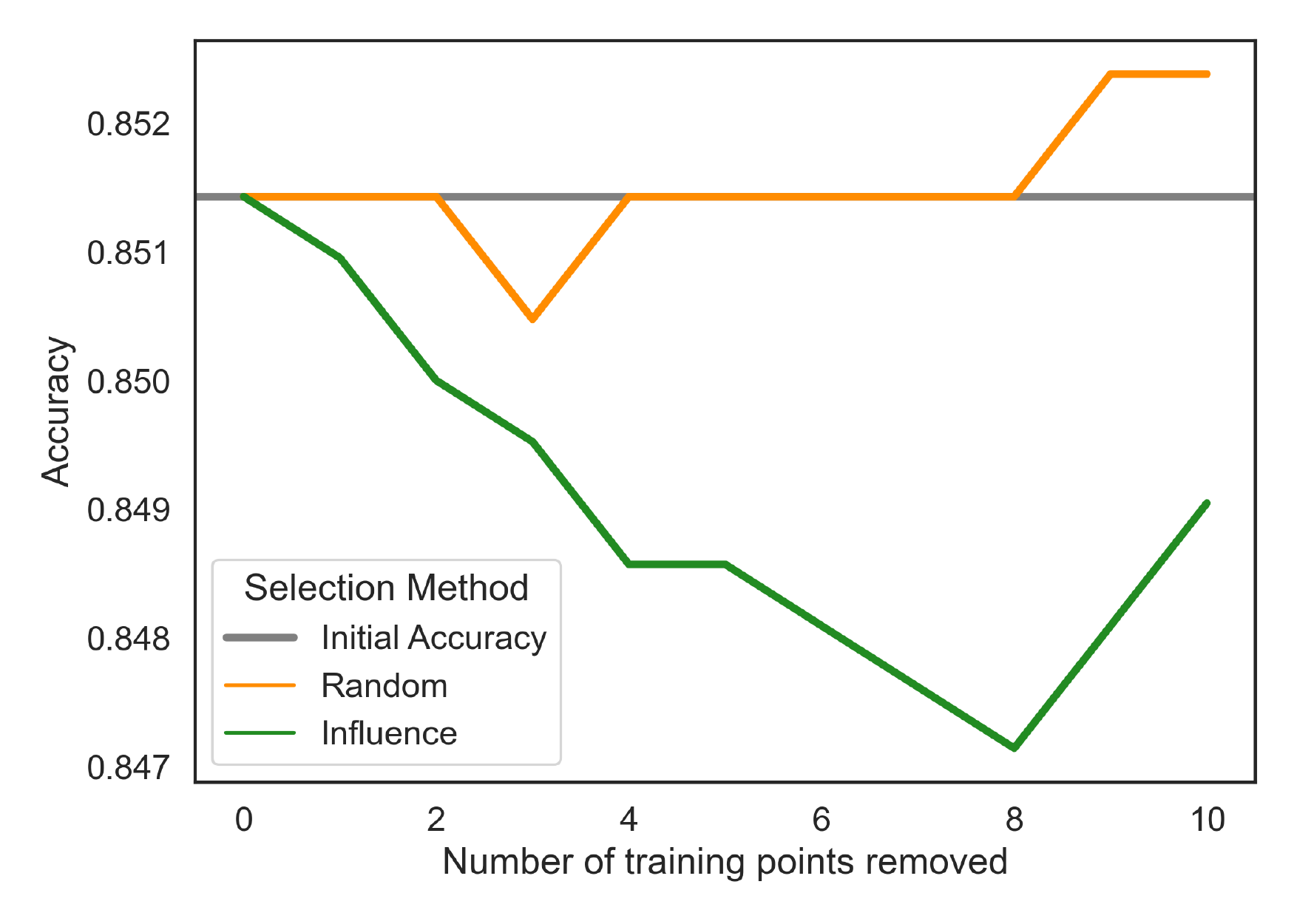}
    \end{subfigure}
    \begin{subfigure}[b]{0.245\linewidth}
            \centering
            \includegraphics[width=\linewidth]{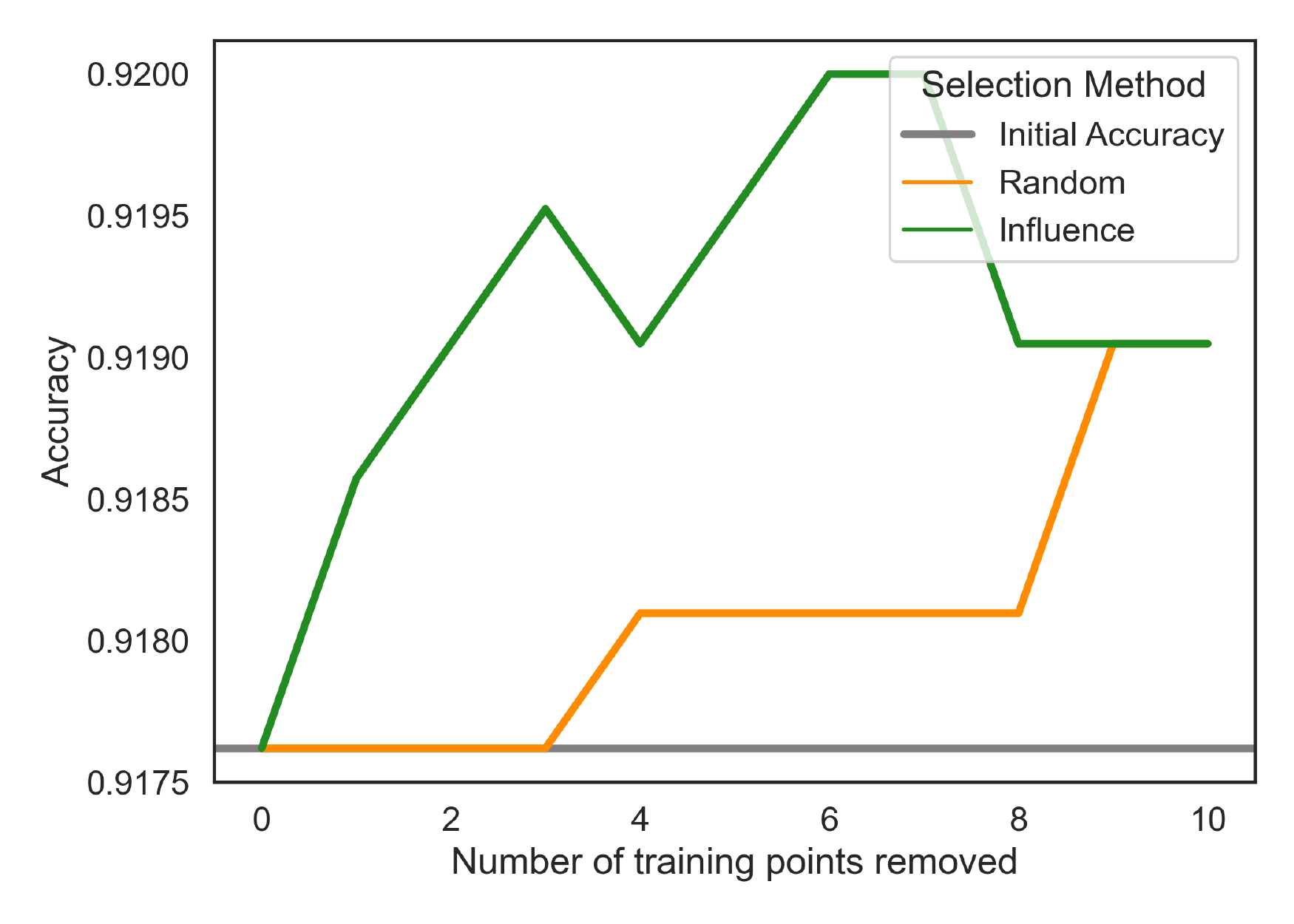}
    \end{subfigure}
    \\
    \begin{subfigure}[b]{0.245\linewidth}    
    \centering
        \includegraphics[width=\linewidth]{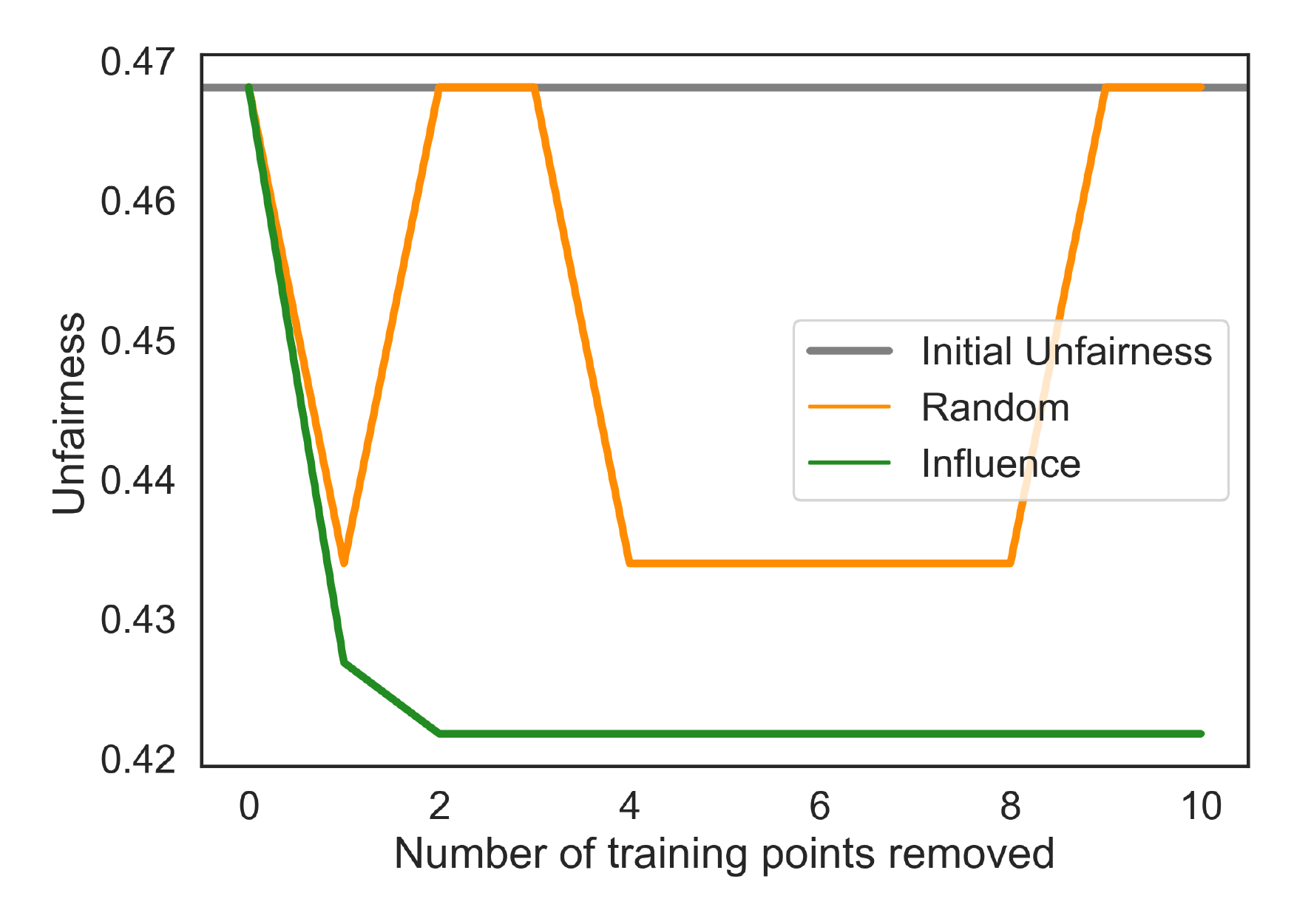}
        \caption{COMPAS}
    \end{subfigure}%
    \begin{subfigure}[b]{0.245\linewidth}  
    \centering
            \includegraphics[width=\linewidth]{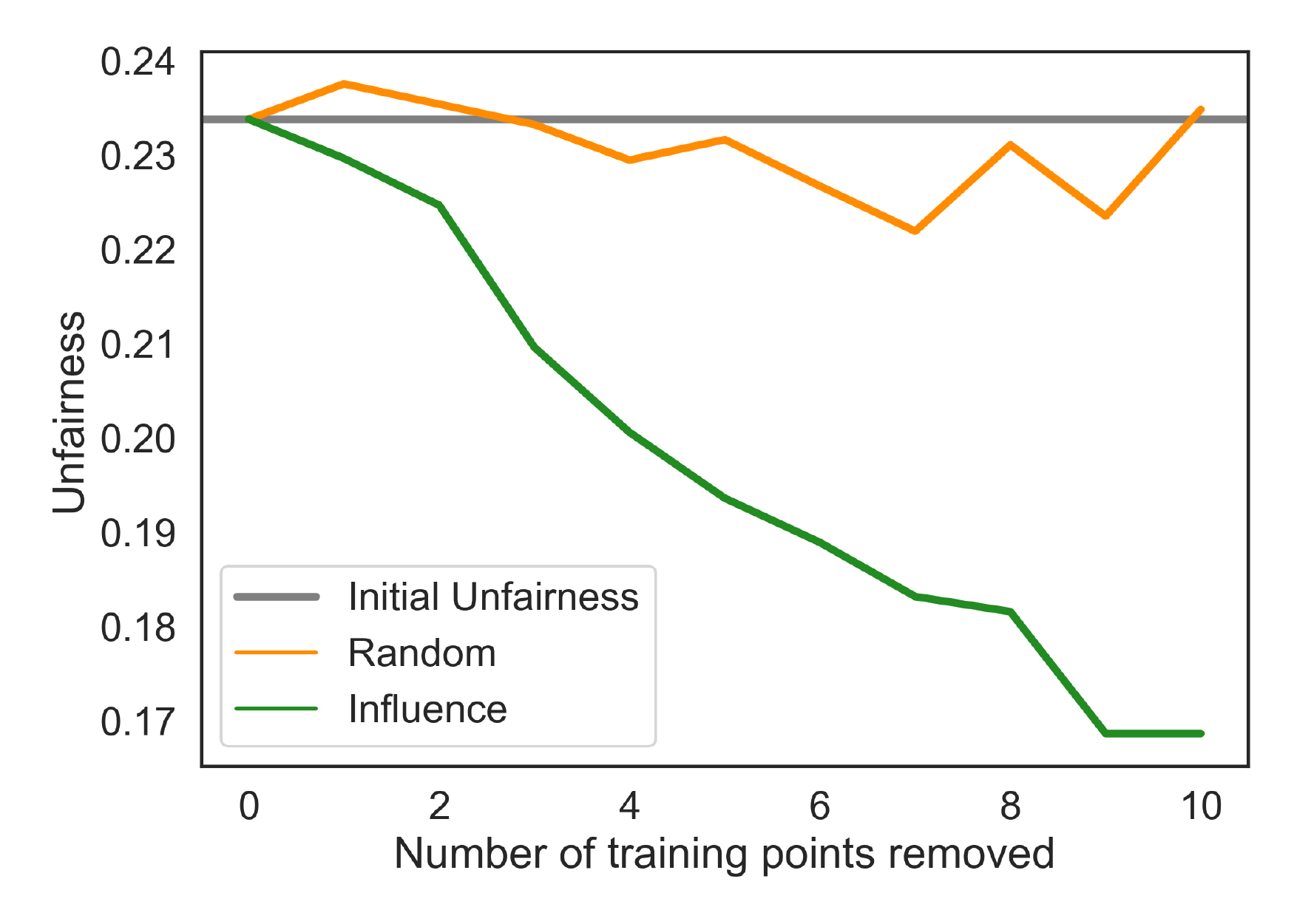}
            \caption{LSAT}
    \end{subfigure}%
    \begin{subfigure}[b]{0.245\linewidth}  
    \centering
            \includegraphics[width=\linewidth]{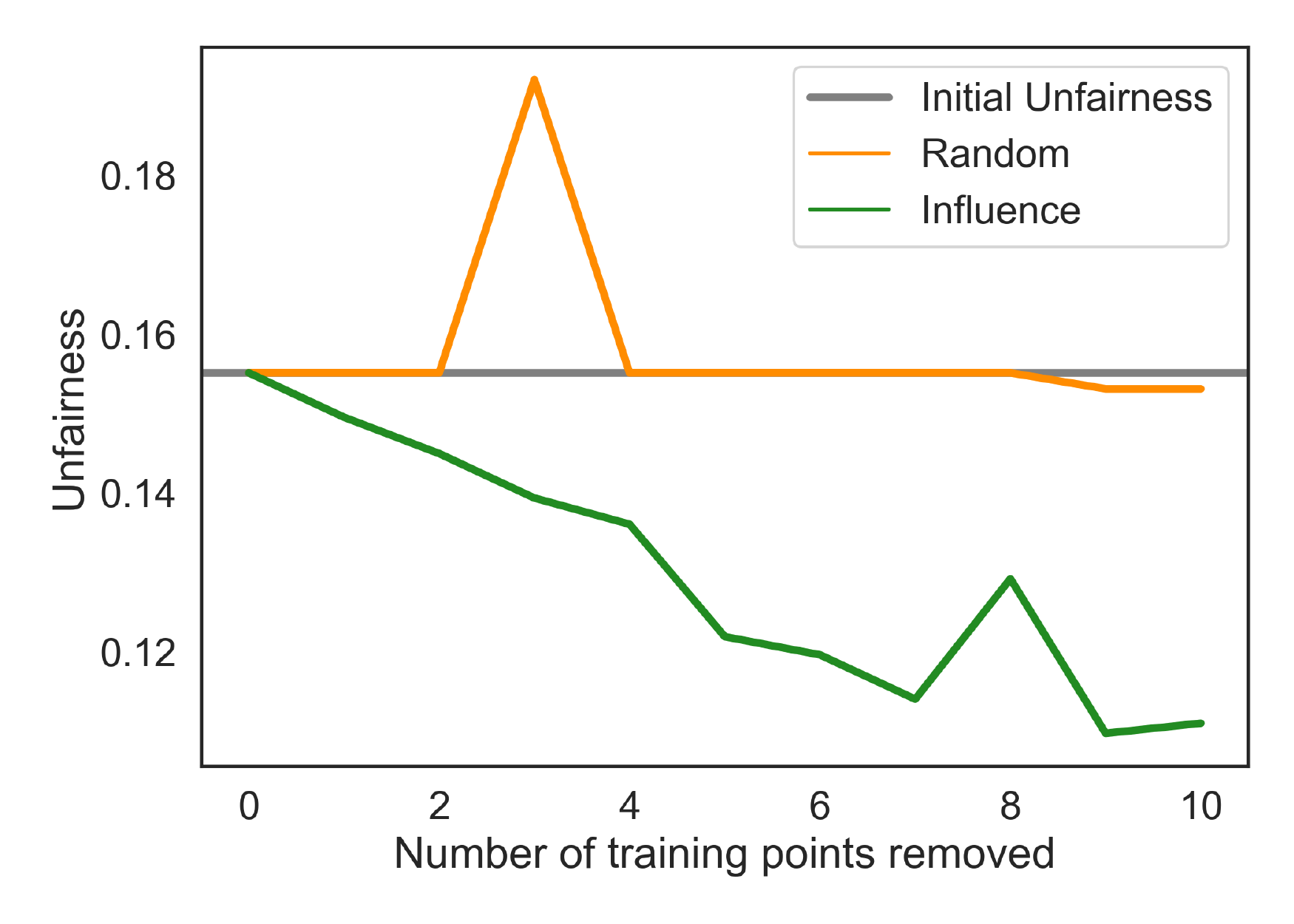}
            \caption{Adult}
    \end{subfigure}%
    \begin{subfigure}[b]{0.245\linewidth}    
    \centering
            \includegraphics[width=\linewidth]{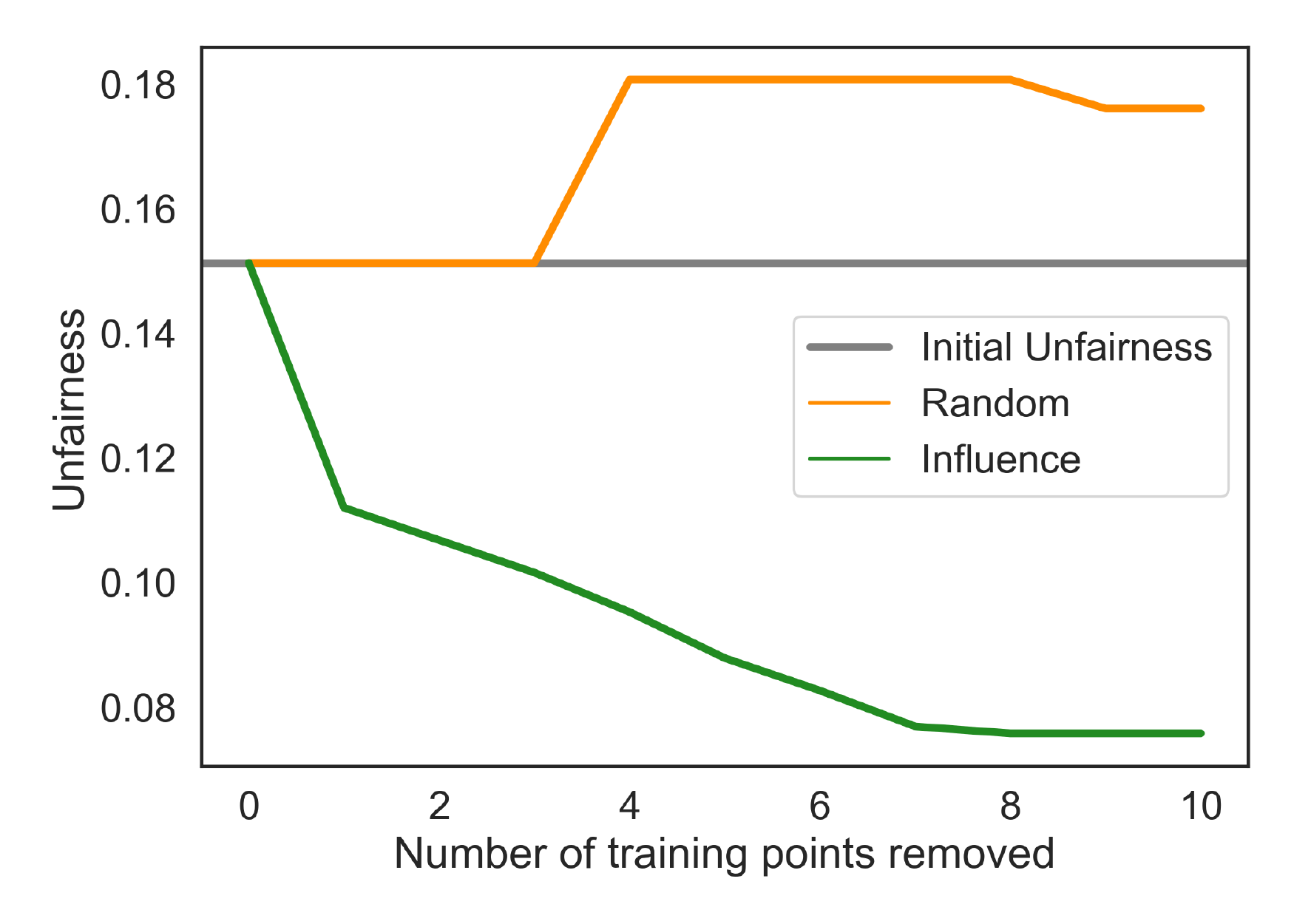}
            \caption{Bank}
    \end{subfigure}%
    \caption{Impact on training data performance (accuracy and unfairness) of removing DIVINE points with recalculation after the most harmful unfairness-inducing point is dropped. Points are selected using $\mathcal{I}(\mathcal{S})$ alone.}
    \label{fig:all_recalculate}
\end{figure*}

\clearpage
\section{User Studies}
\label{hse}

\subsection{Experiment Details}
\begin{figure*}[htb]
\vspace{-0.05in}
    \centering
    \begin{subfigure}{0.5\textwidth}
        \centering
        \includegraphics[width=.98\linewidth]{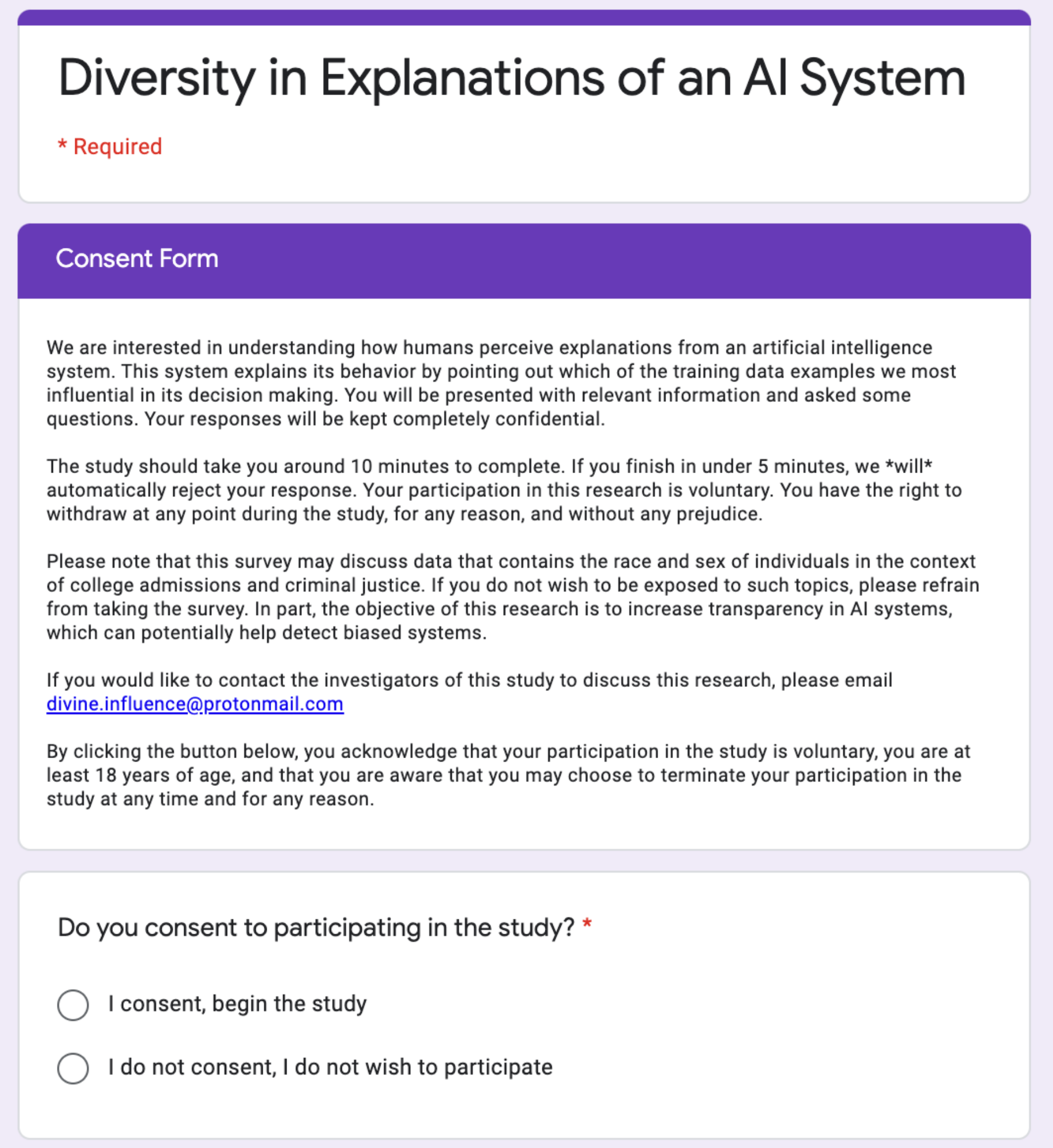}
        \caption{Consent Form for the Survey}
        \label{fig:consent_form}
    \end{subfigure}%
    ~
    \begin{subfigure}{0.5\textwidth}
        \centering
        \includegraphics[width=\linewidth]{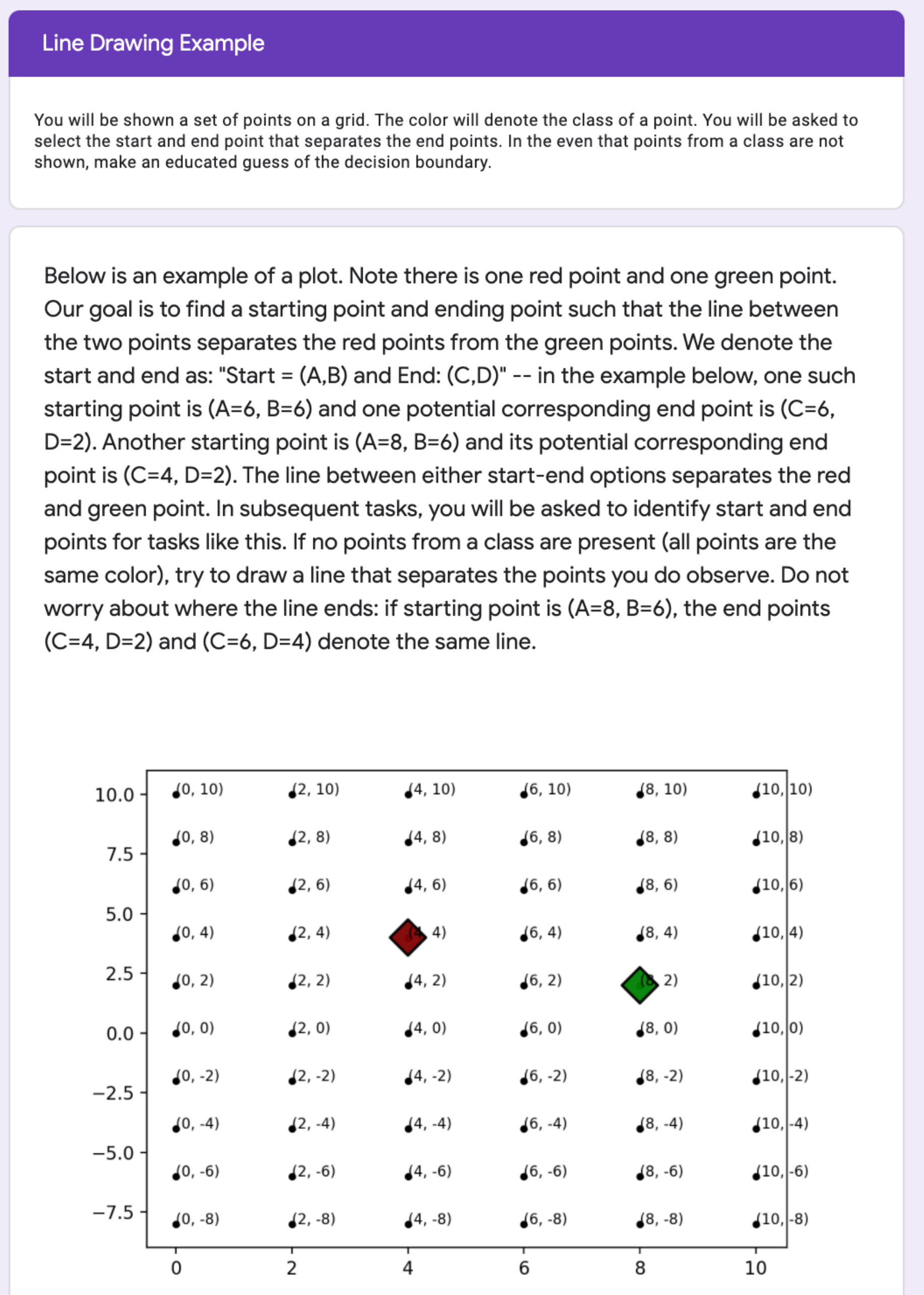}
        \caption{Attention Check for Simulatability}
        \label{fig:attention_check}
    \end{subfigure}
    \caption{Setup of user studies.}
    \label{fig:hs_example_side}
\end{figure*}

We conduct all our user studies on the Prolific platform. We use Google Forms to record our answers. Each study variant was taken by 10 participants, who had a high approval rating on the platform, had computer science experience, and completed a Bachelors degree. We ran 2 different user studies. 
One study asked about the DIVINE points for FashionMNIST (Section~\ref{examples}) and one study contained the simulatability experiment (Section~\ref{linedrawing}). 
In~\Cref{fig:consent_form}, we include the consent form used in our user studies. This user study was performed with the approval of 
the University of Cambridge's Department of Engineering Research Ethics Committee.
Only two participants who were asked to take the survey did not provide consent and thus exited the form. We still ensured at least 10 participants took each survey variant.
The maximum allocated time for either study was 44 minutes. In order to rule-out bogus answers, users were instructed that their responses would be discarded if they finished the task before 5 minutes. The average time spent by a participant over all studies was 9 minutes and 53 seconds. In all, we had 40 participants over all studies and all variants. Each participant was paid approximately \pounds 11.93 per hour. We spent approximately \pounds 86 in total.

We included an example question for the line drawing task, called an ``attention check.'' An example is shown in~\Cref{fig:attention_check}. Note that the answer to this example question is provided in line. Later in the study, we ask participants this exact same question. If participants get the attention check wrong, we void their results. We only had to void three results. This did not affect our criteria of ten completed surveys per variant. 
While consent form was the same for all study variants, the attention check only appeared in the simulatability user study.

\subsection{FashionMNIST}
\begin{figure*}[htb]
\vspace{-0.05in}
    \centering
    \begin{subfigure}{0.5\textwidth}
        \centering
        \includegraphics[width=.98\linewidth]{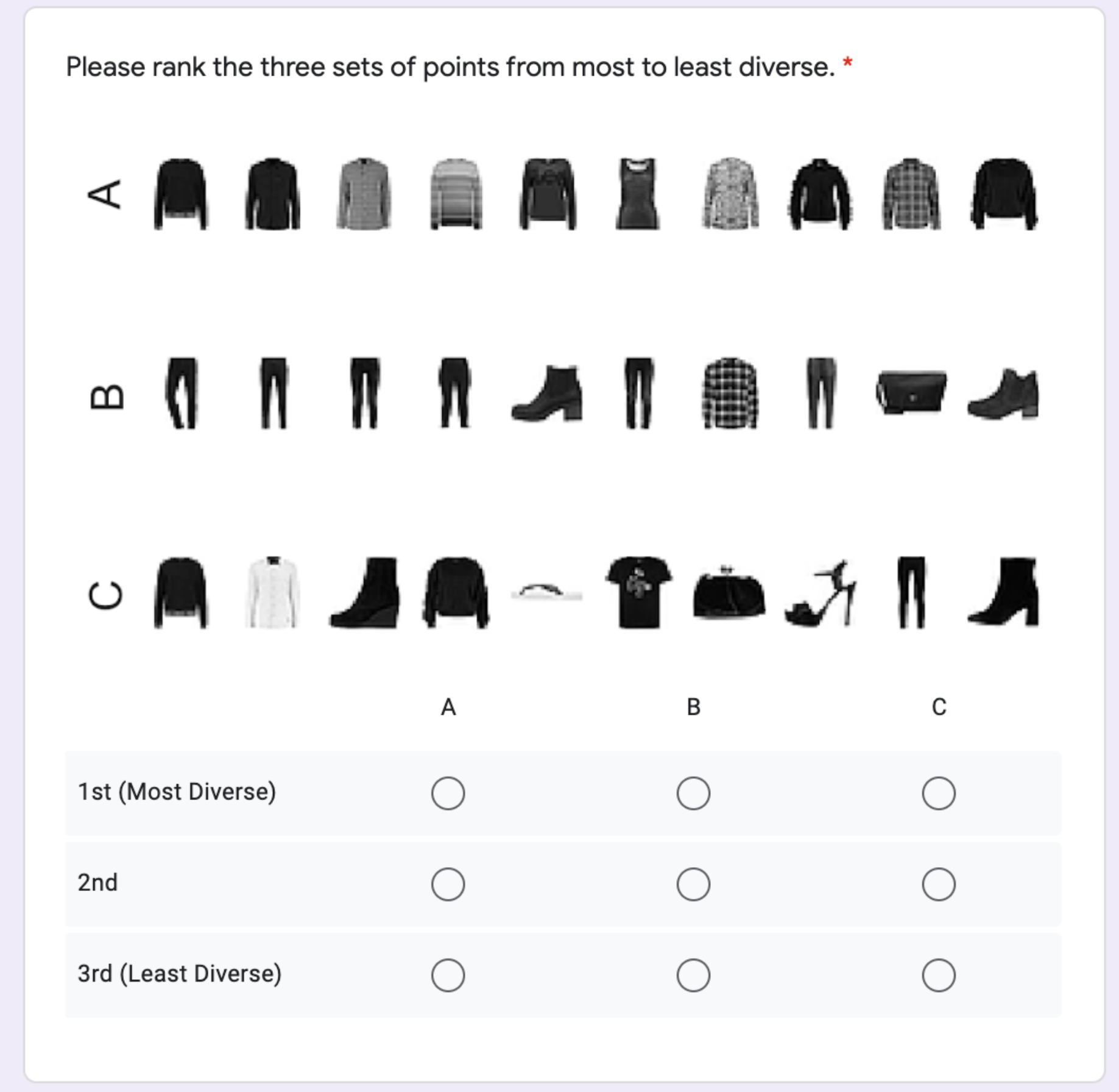}
        \caption{Diversity in top-$10$ points}
        \label{fig:diverse}
    \end{subfigure}%
    ~
    \begin{subfigure}{0.5\textwidth}
        \centering
        \includegraphics[width=\linewidth]{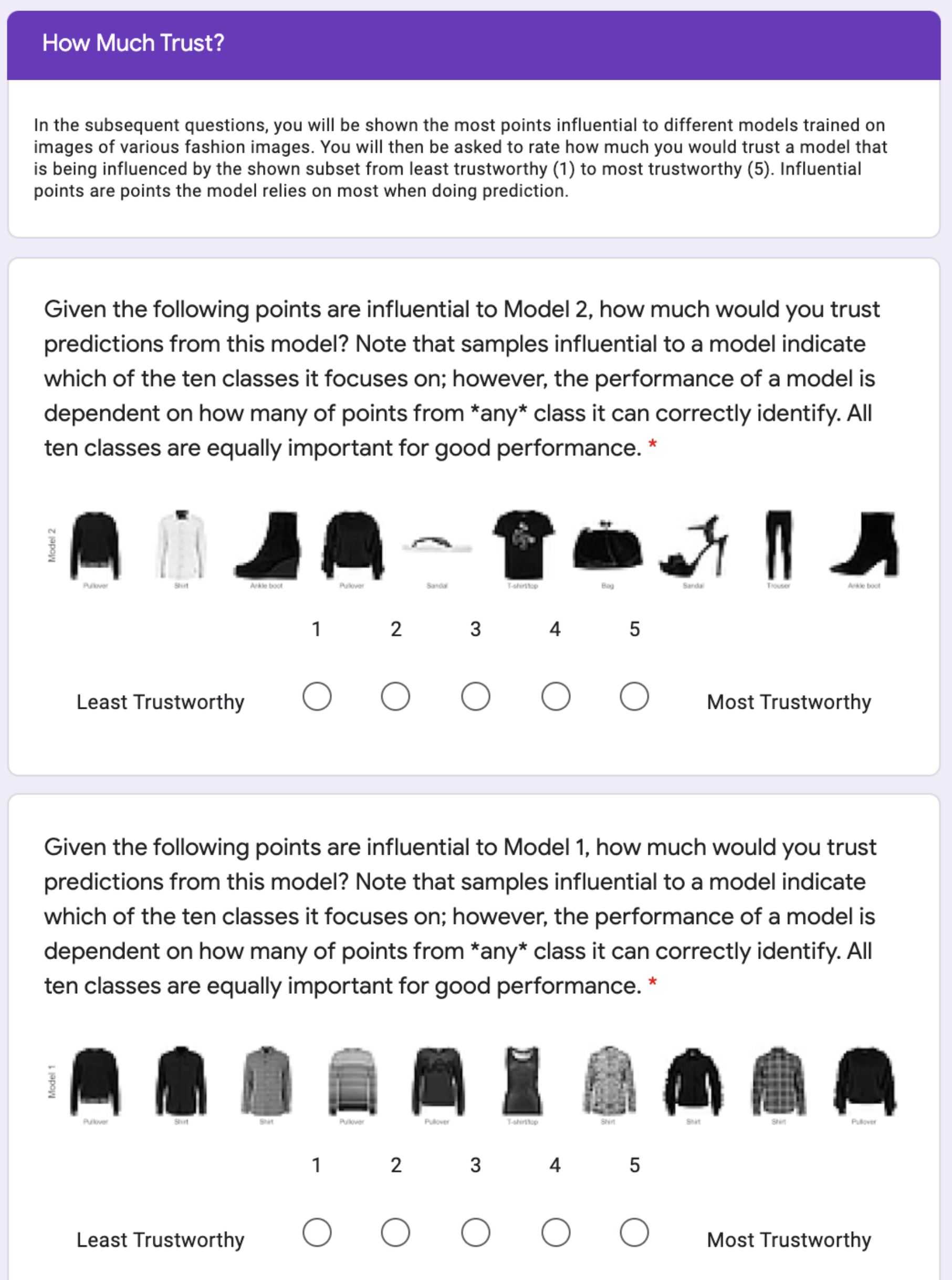}
        \caption{Trustworthiness Question}
        \label{fig:trust}
    \end{subfigure}
    \caption{FashionMNIST User Study}
    \label{fig:hse_fashion}
\end{figure*}

Before running any study, we run a 10 person study with one question. We show participants the top-$5$ points from Influence Functions, DataShapley, and DIVINE. We ask them to pick which set of points is more diverse. $100\%$ selected DIVINE to be more diverse.
In our first main user study, we asked 10 participants to do four tasks.
\begin{enumerate}
    \item Choose which of top-$10$ points from Influence Functions, DataShapley, or DIVINE are more diverse.  Question shown in Figure~\ref{fig:diverse}.
\item Rank the trustworthiness of the top-$10$ points from Influence Functions and DIVINE from 1 to 5. Answer if and why they deemed one set of points more trustworthy than the other (minimum 10 words). Question shown in Figure~\ref{fig:trust}.
    \item Select which set of top-$3$ points from Influence Functions, RelatIF, or DIVINE were useful for understanding a misclassified test point. Answer if and why they deemed one set of points more useful than the other (minimum 10 words).
\end{enumerate}
Before running any tasks, we prime them with one randomly selected example from each of the 10 classes: this ensures that participants understand the extent of the model's intended behavior. The results and select quotes from participants are included in the main text (Section~\ref{examples}).

\subsection{Simulatability}
\begin{figure*}[htb]
\centering
    \begin{subfigure}[b]{0.32\linewidth}
        \includegraphics[width=\textwidth]{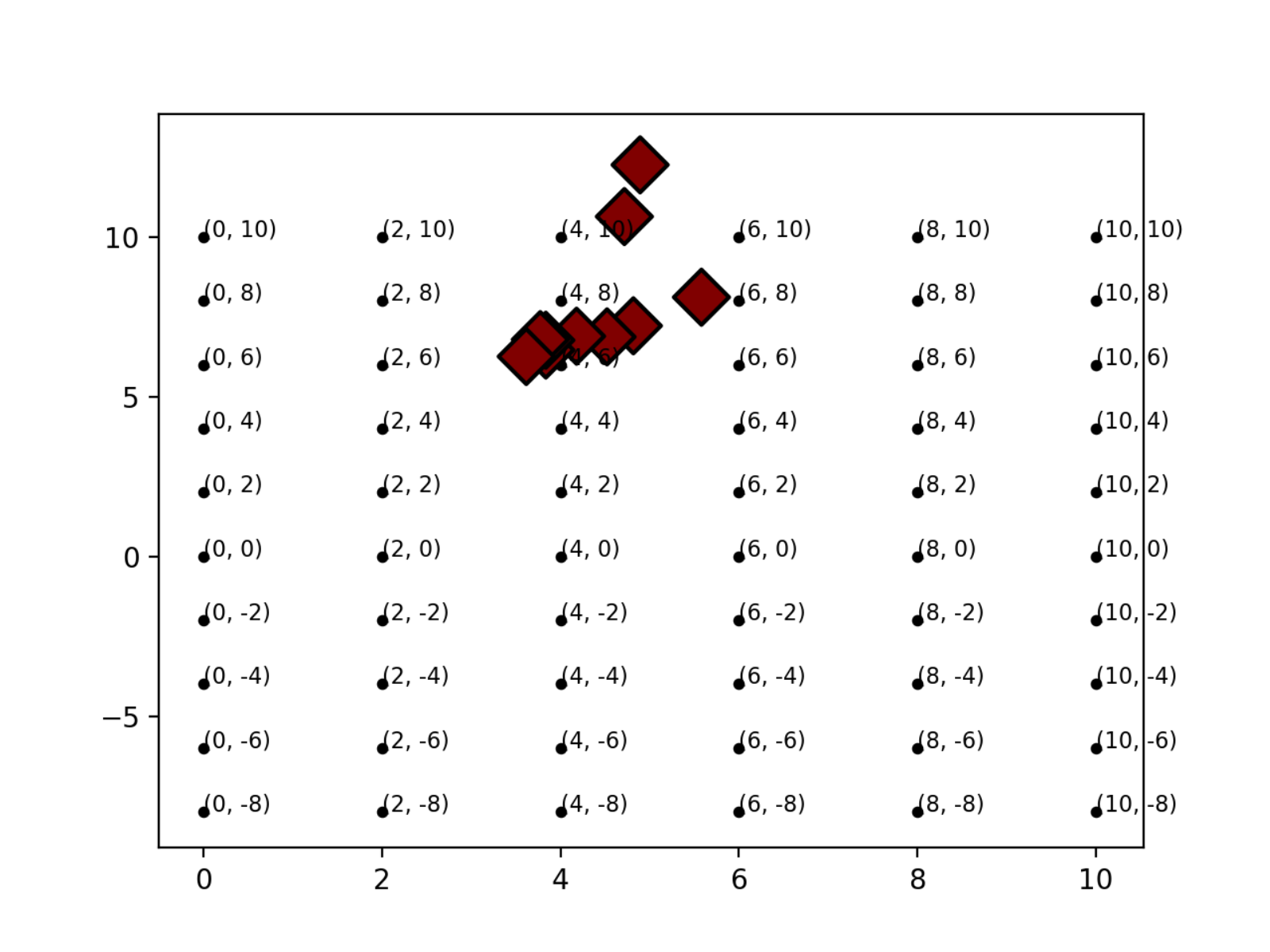}
        \caption{Top IF Points}
    \end{subfigure}
    \begin{subfigure}[b]{0.32\linewidth}    
        \includegraphics[width=\textwidth]{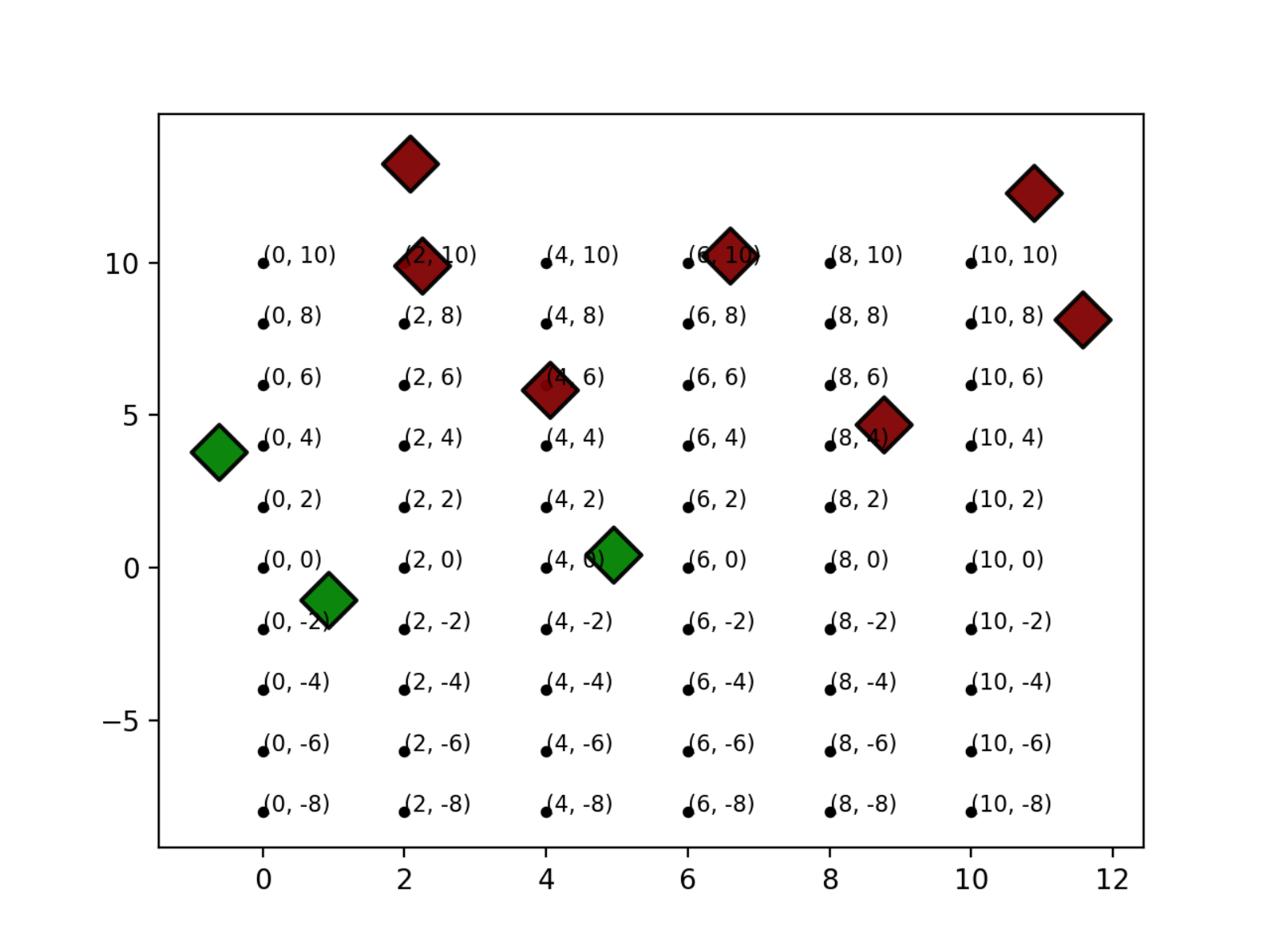}
        \caption{Top DIVINE Points}
    \end{subfigure}
    \begin{subfigure}[b]{0.32\linewidth}
        \includegraphics[width=\textwidth]{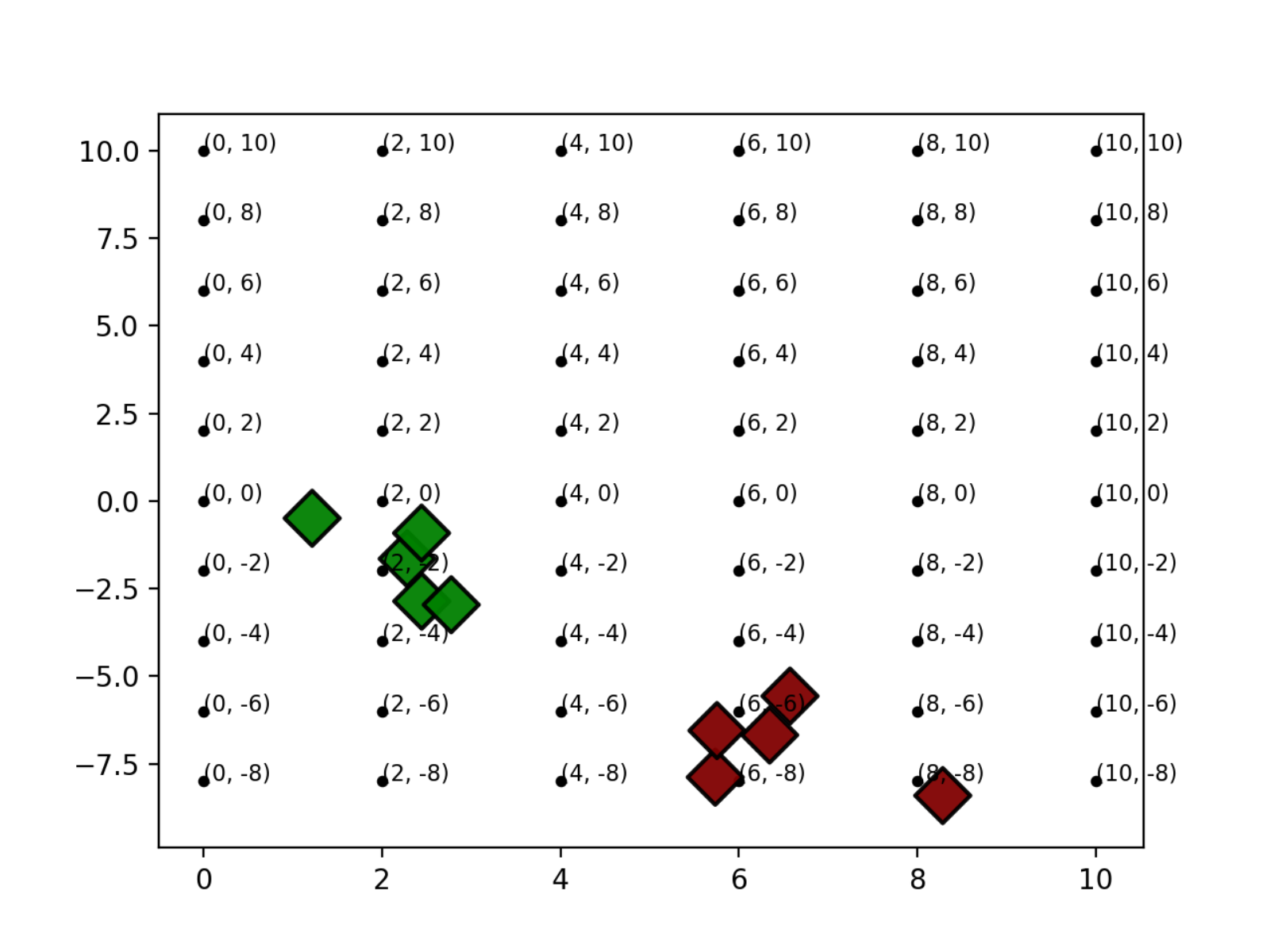}
            \caption{Random Points}
    \end{subfigure}%
    \caption{Simulatability task. Points shown to users on coordinate grid for $m=10$.}
    \label{fig:line_drawing}
\end{figure*}

\begin{figure*}[htb]
\centering
    \begin{subfigure}[b]{0.32\linewidth}
        \includegraphics[width=\textwidth]{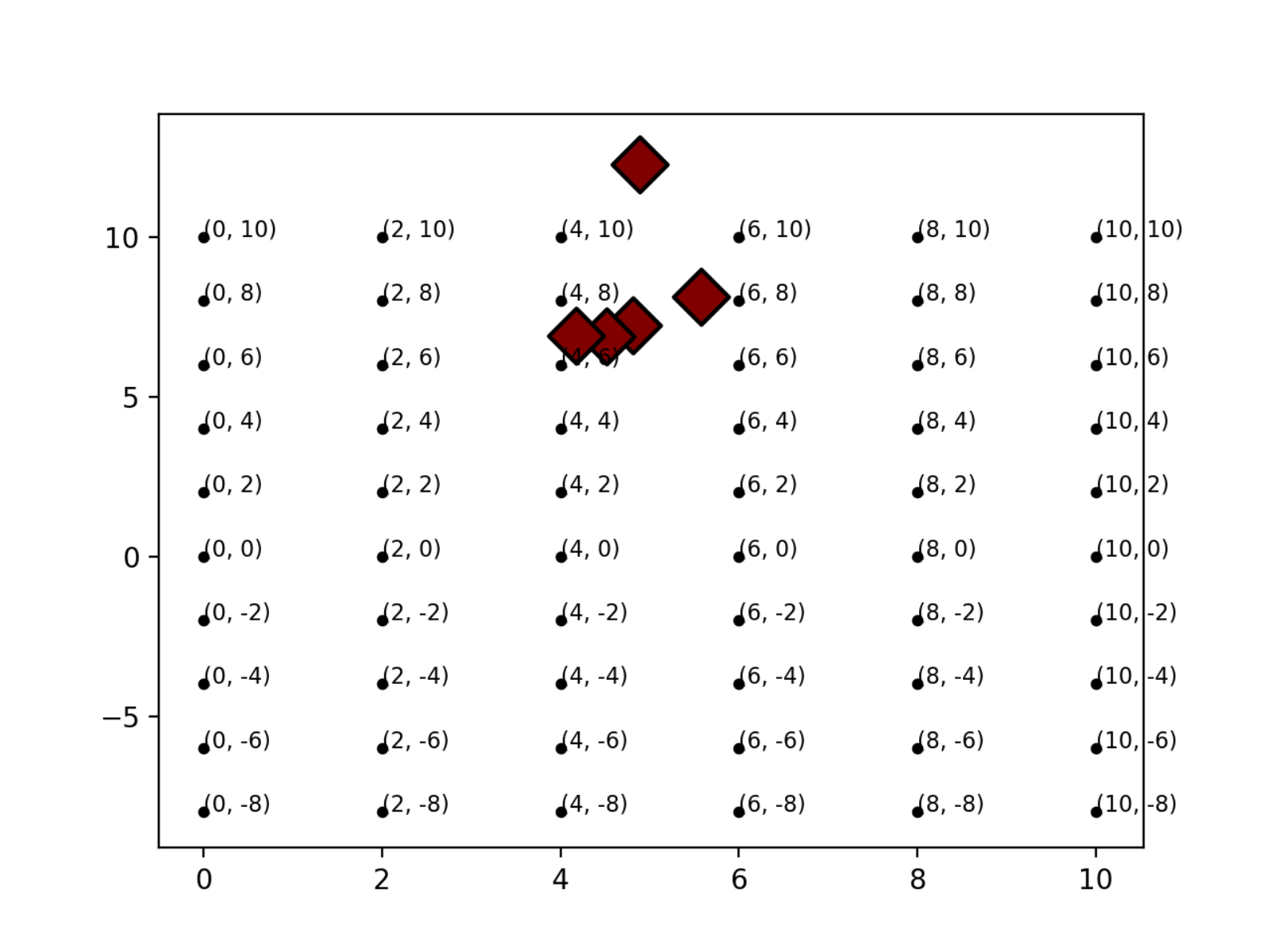}
        \caption{Top IF Points}
    \end{subfigure}
    \begin{subfigure}[b]{0.32\linewidth}    
        \includegraphics[width=\textwidth]{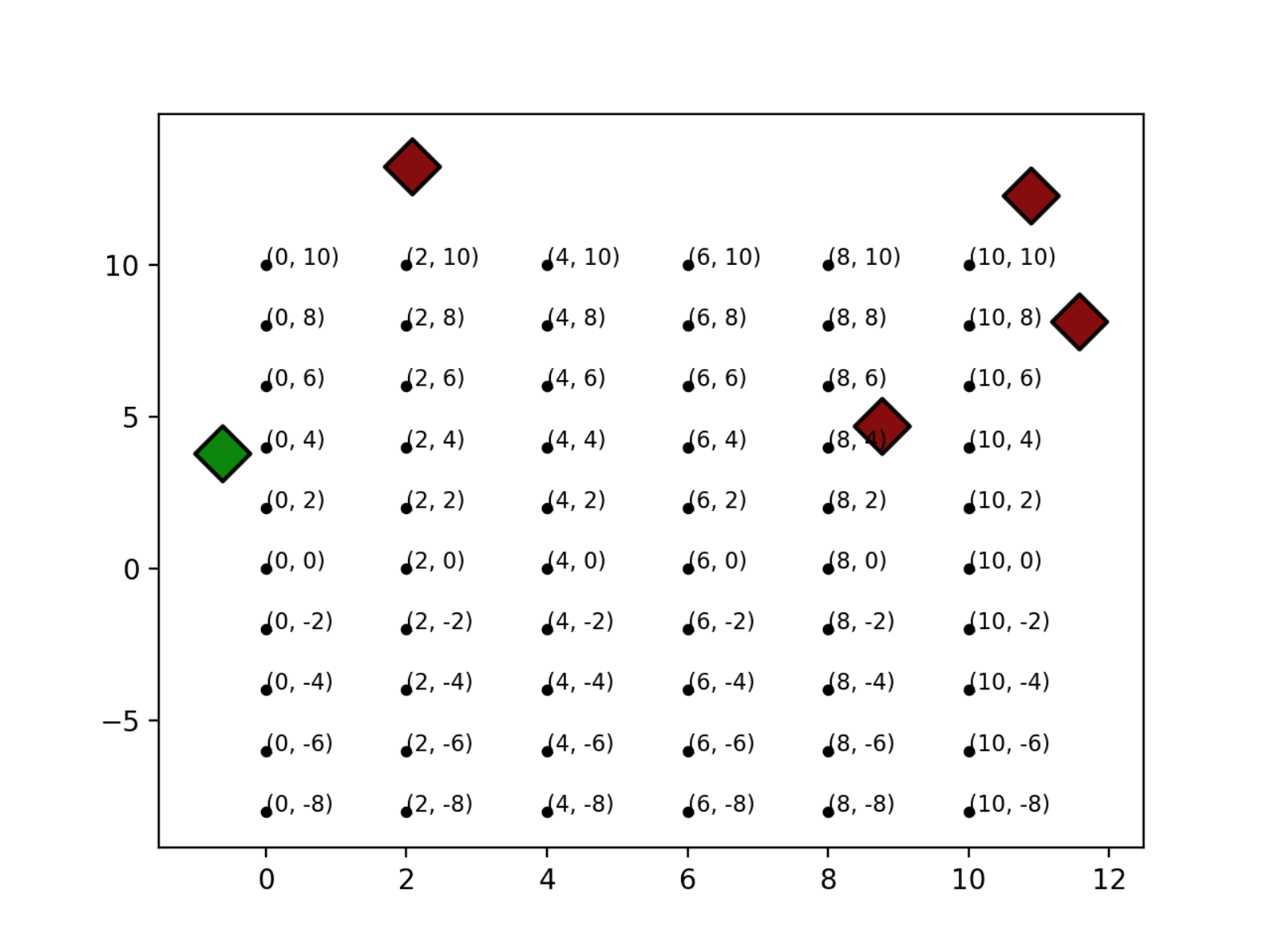}
        \caption{Top DIVINE Points}
    \end{subfigure}
    \begin{subfigure}[b]{0.32\linewidth}
        \includegraphics[width=\textwidth]{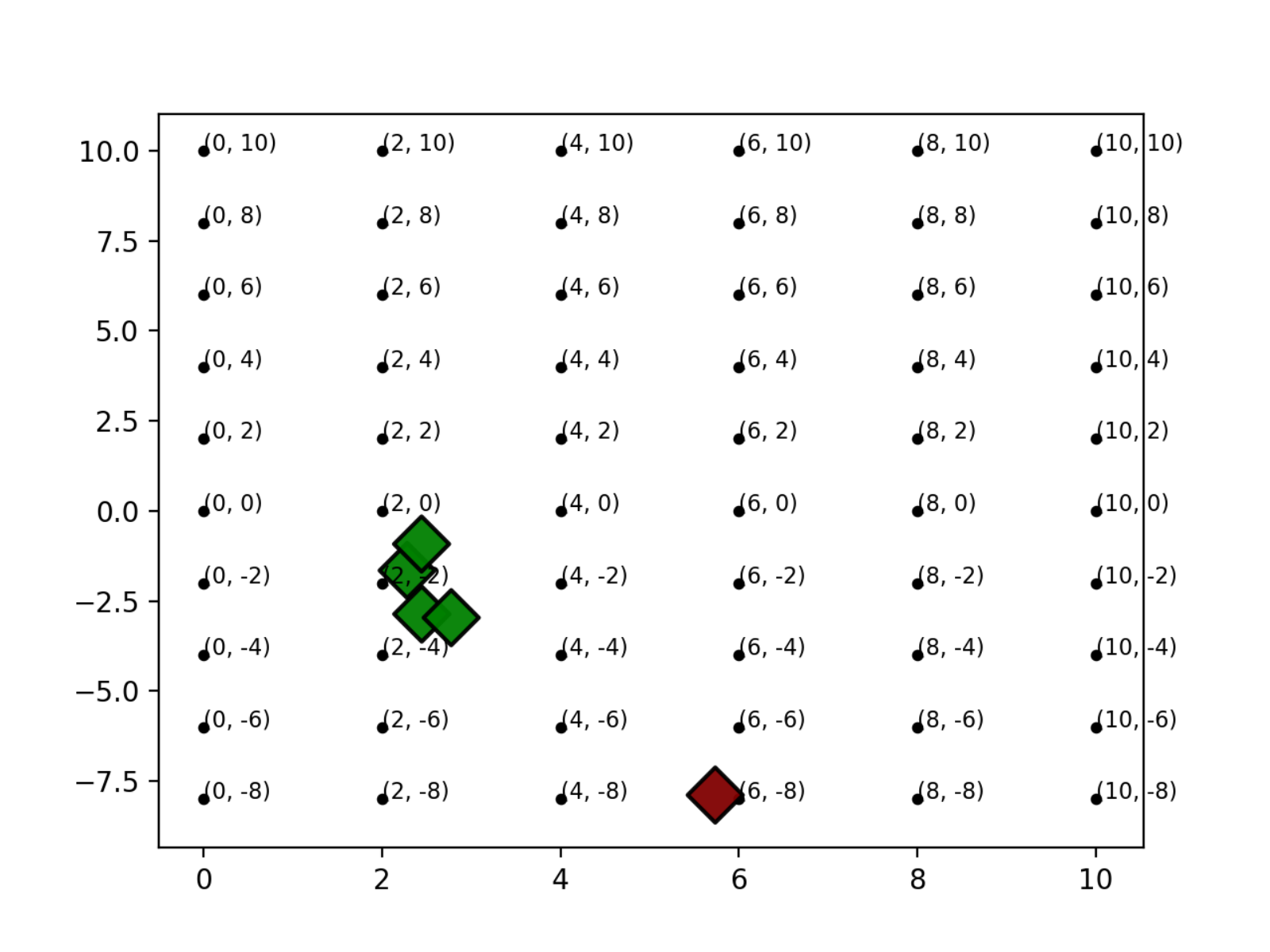}
            \caption{Random Points}
    \end{subfigure}%
    \caption{Simulatability task. Points shown to users on coordinate grid for $m=5$.}
    \label{fig:line_drawing5}
\end{figure*}

For a different set of 20 participants, we ask how well participants can draw a decision boundary given points colored by their predicted class. We generate points in 3 ways: randomly, top-$m$ via influence functions, and top-$m$ via DIVINE (IF+SR). We run one variant with $m=5$ and one with $m=10$. 10 individuals take each variant.
In Figure~\ref{fig:line_drawing}, we show the setup for the task where the users were asked to guess the classifier boundary given $10$ points. Similar task details shown in Figure~\ref{fig:line_drawing5} when $m=5$. The users were shown the coordinate labels and were asked to enter the coordinates of the start and end point of the inferred line in their response. We then calculate the parameters for the drawn line and compare it to the true decision boundary by calculating the cosine similarity between the drawn and actual boundaries.
Note we randomly translate or reflect the points before line drawing to ensure the boundary is not in the same place for all variants. We revert the line drawn between the selected points before calculating the cosine similarity.

\begin{figure*}[htb]
\centering
    \begin{subfigure}[b]{0.49\linewidth}
        \includegraphics[width=\textwidth]{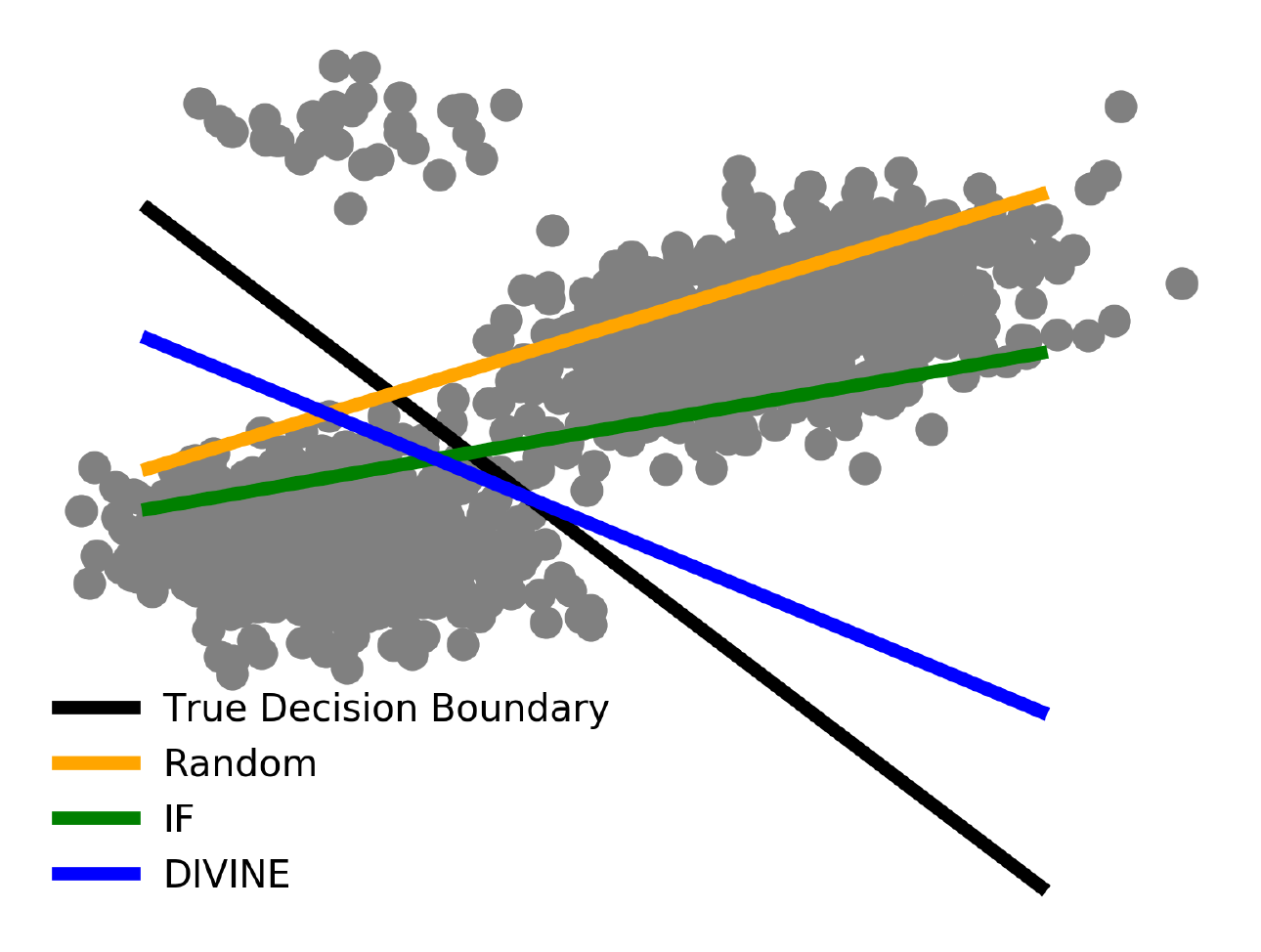}
        \caption{Average}
    \end{subfigure}
    \begin{subfigure}[b]{0.49\linewidth}    
        \includegraphics[width=\textwidth]{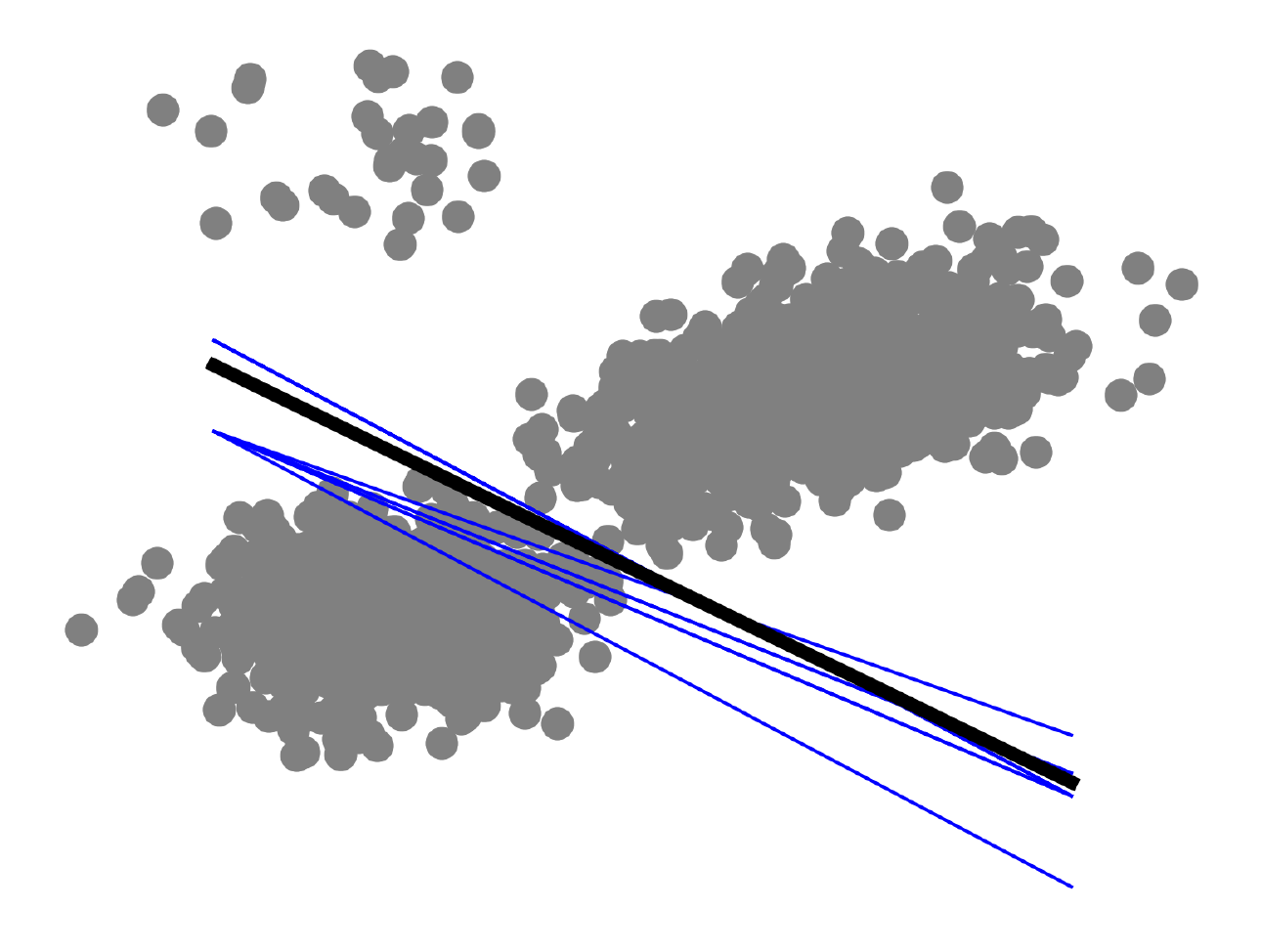}
        \caption{DIVINE}
    \end{subfigure}
    \begin{subfigure}[b]{0.49\linewidth}
        \includegraphics[width=\textwidth]{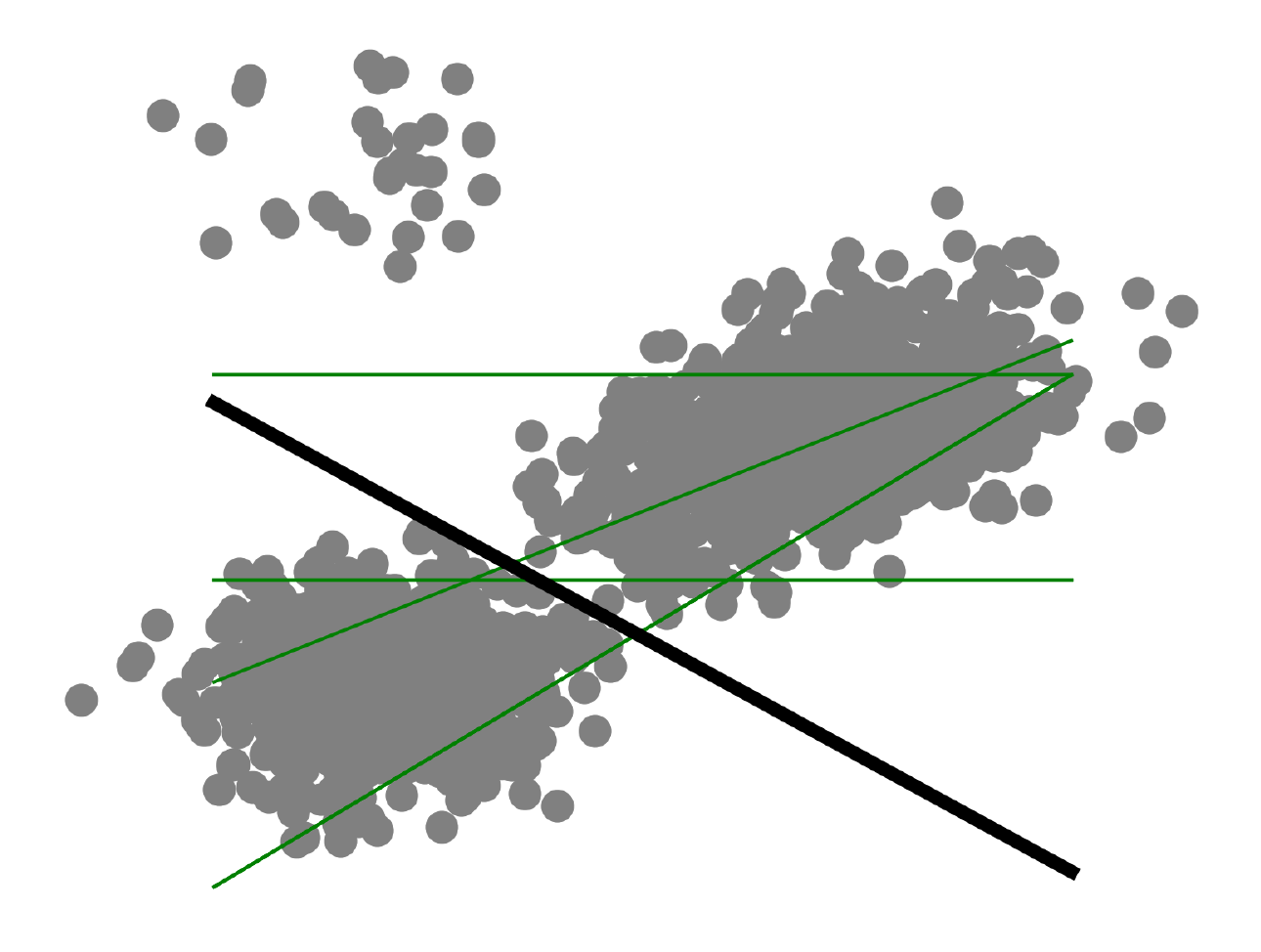}
            \caption{IF}
    \end{subfigure}%
    \begin{subfigure}[b]{0.49\linewidth}        
            \includegraphics[width=\textwidth]{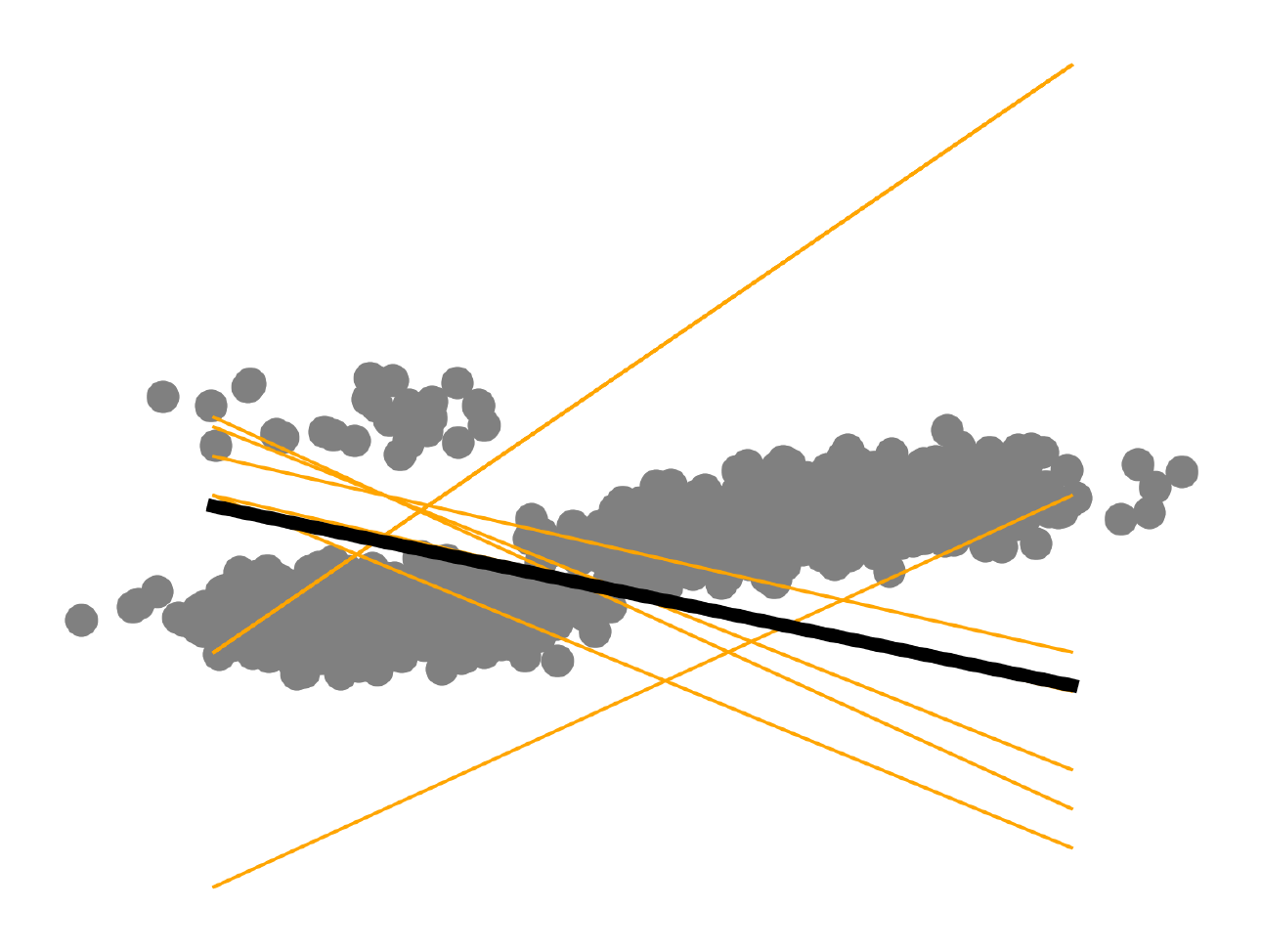}
            \caption{Random}
    \end{subfigure}
    \caption{User drawn decision boundaries. $m=10$}\label{fig:hse10}
\end{figure*}

\begin{figure*}[htb]
\centering
    \begin{subfigure}[b]{0.49\linewidth}
        \includegraphics[width=\textwidth]{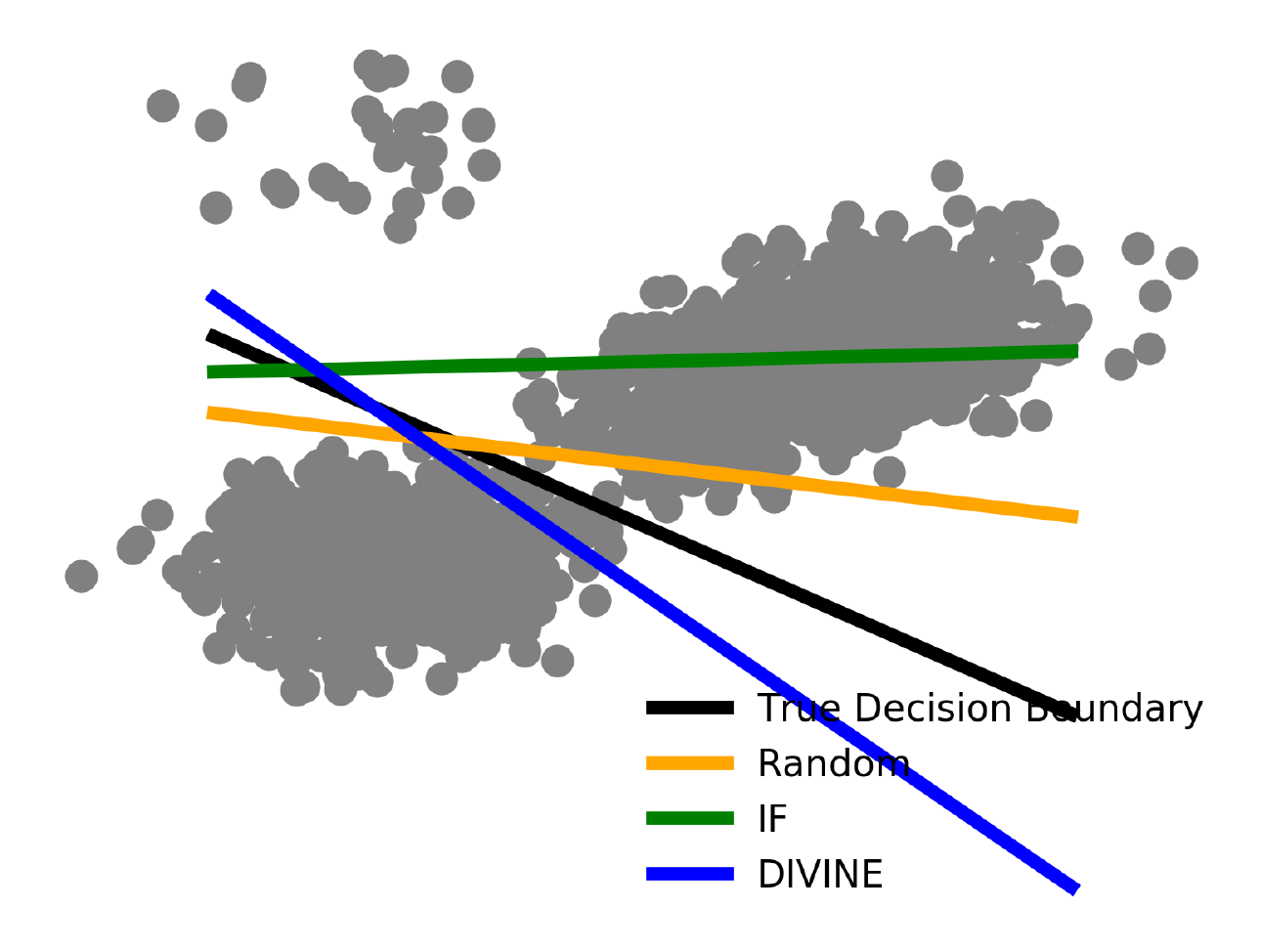}
        \caption{Average}
    \end{subfigure}
    \begin{subfigure}[b]{0.49\linewidth}    
        \includegraphics[width=\textwidth]{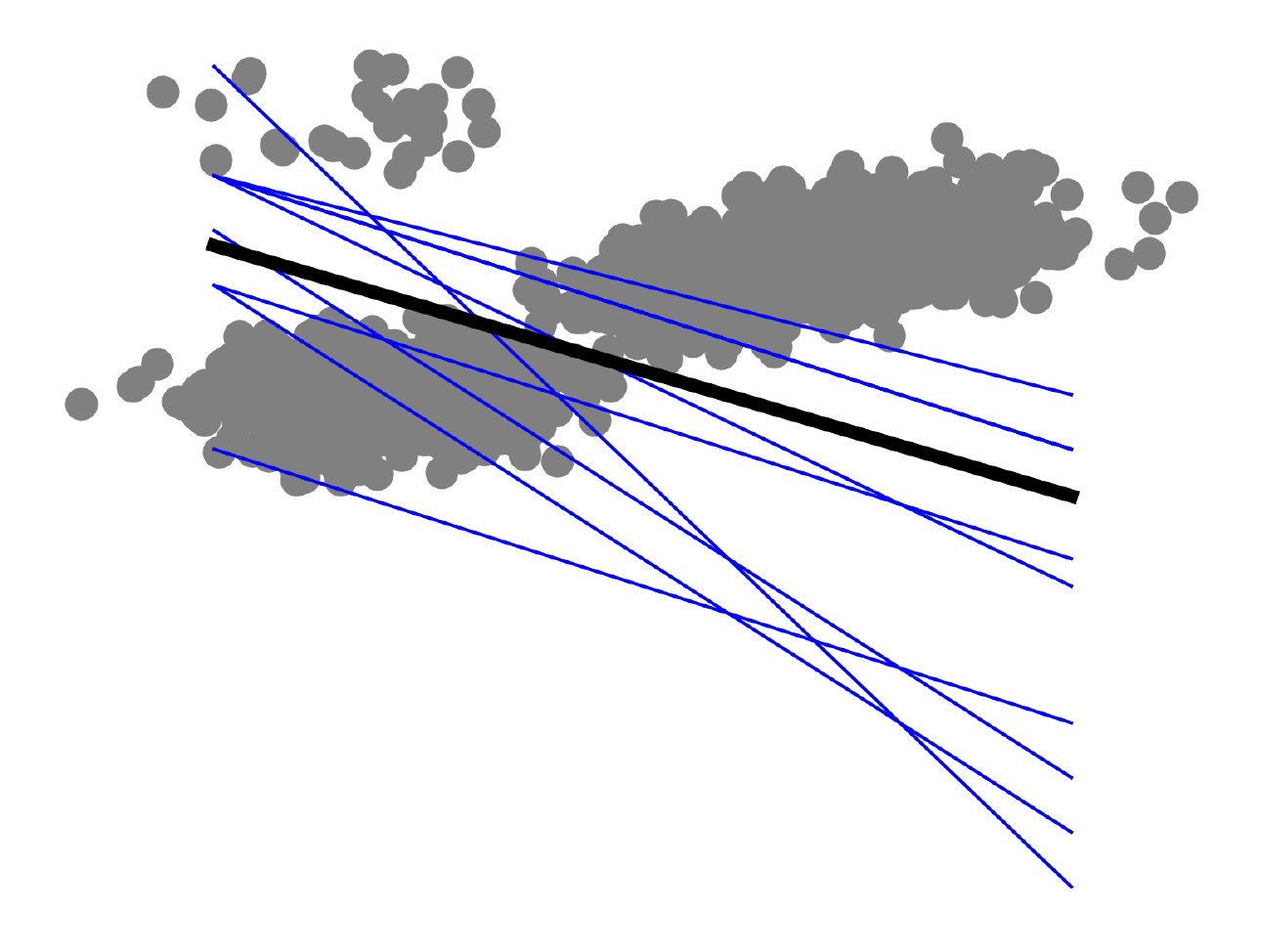}
        \caption{DIVINE}
    \end{subfigure}
    \begin{subfigure}[b]{0.49\linewidth}
        \includegraphics[width=\textwidth]{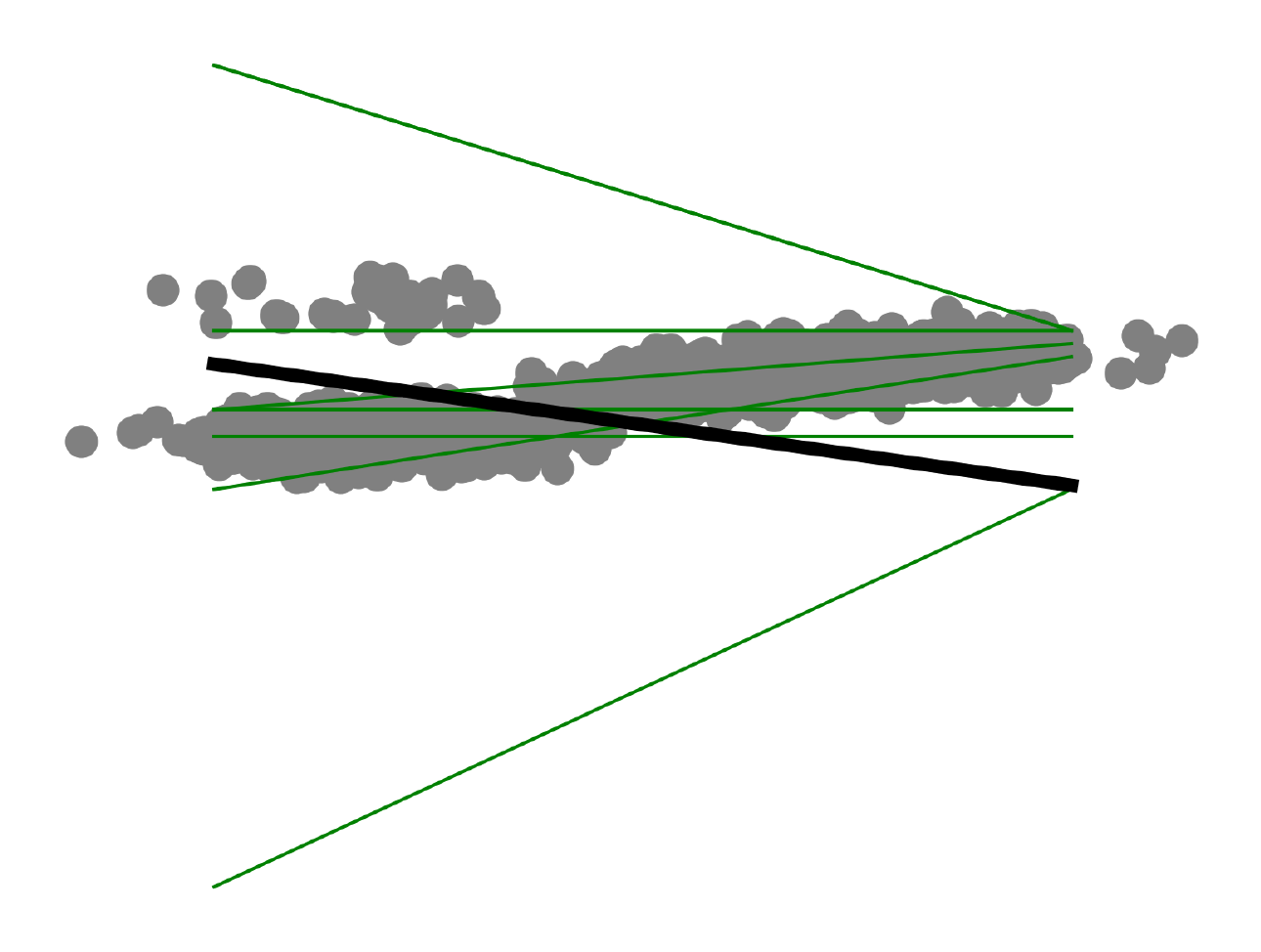}
            \caption{IF}
    \end{subfigure}%
    \begin{subfigure}[b]{0.49\linewidth}        
            \includegraphics[width=\textwidth]{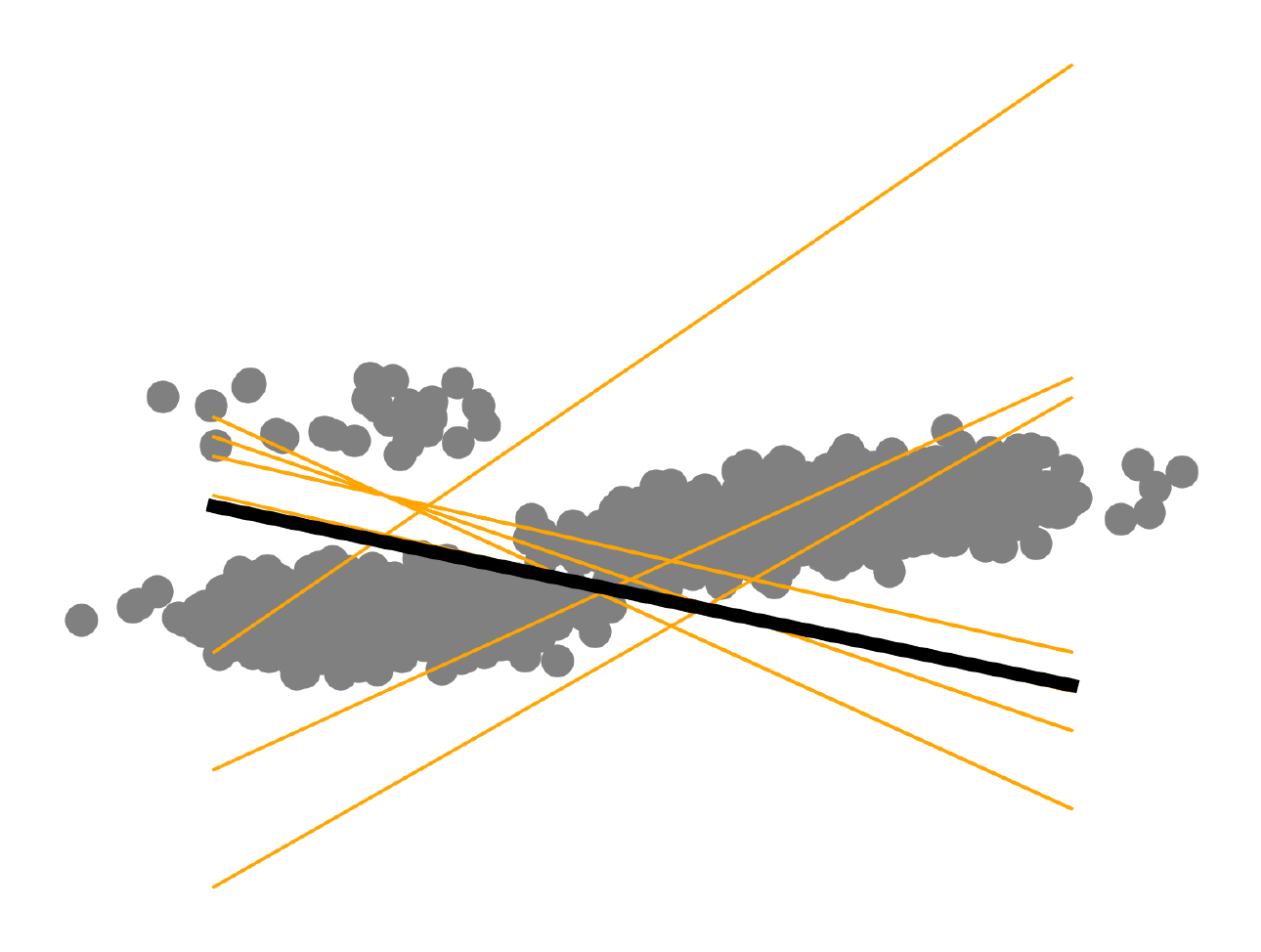}
            \caption{Random}
    \end{subfigure}
    \caption{User drawn decision boundaries when $m=5$. Notice that the user drawn decision boundaries are most similar to the true decision boundary upon seeing DIVINE points.}\label{fig:hse5}
\end{figure*}

In Figure~\ref{fig:hse10}, we show the user drawn decision boundaries when shown $10$ points from each method. Notice how the boundaries drawn when shown DIVINE points is more similar to the true decision boundary. We represent each user drawn decision boundary with a slope and a bias term. We then calculate the cosine similarity between the user drawn decision boundary parameters and the true decision boundary parameters. In the main text (Section~\ref{linedrawing}), we report how the cosine similarity of true decision boundary and the user drawn decision boundary after seeing DIVINE points exceeds that the decision boundaries drawn after seeing random points or the top points from influence functions. Similar results hold in Figure~\ref{fig:hse5} for the 10 participants who only saw $5$ points from each method. However, note that the cosine similarity was a bit lower implying that it might become easier to simulate the model after seeing more points. However, future work can explore how large $m$ needs to be for participants to recover the decision boundary well. More broadly, simulatability is an important direction to pursue in the explainability community. We hope to further study how simulatability can be used to evaluate the efficacy of explanations for the behavior of machine learning models.


\end{document}